\pdfoutput=1
\RequirePackage{snapshot}

\def\plain{true}

\ifdefined\medphys      % ---------- AIP Medical Physics
  
  \ifdefined\review
    \documentclass[aapm,preprint,nofootinbib]{revtex4-1}
  \else
    \documentclass[aapm,reprint,nofootinbib]{revtex4-1}
  \fi

\fi\ifdefined\plain     % ---------- plain (default)

  \ifdefined\review
    \documentclass[12pt,draft]{article}
  \else
    \documentclass{article}
  \fi
  
\fi

\ifdefined\medphys              % ----- AIP Medical Physics
  \ifdefined\review
    \usepackage{lineno}
    \modulolinenumbers[2]
    \linenumbers\relax
  \fi
\fi\ifdefined\plain             % ----- plain
  \usepackage[left=0.85in,right=0.85in,top=0.7in,bottom=0.7in]{geometry}
  \usepackage{authblk}
  \usepackage{setspace}
\fi

\usepackage[cmex10]{amsmath}
\usepackage{amssymb}
\usepackage{array}
\usepackage{dcolumn}
\usepackage{accents}
\usepackage{bm}

\usepackage[thinspace,thinqspace,squaren]{SIunits}

\usepackage[export]{adjustbox}
\usepackage{graphicx}
\usepackage[hyperref,x11names,cmyk,table,pdftex]{xcolor}
\usepackage[caption=false]{subfig}
\usepackage[many]{tcolorbox}
\usepackage{tikz}
\ifdefined\review
  \usepackage[section]{placeins}
\else
  \usepackage{placeins}
\fi

\usepackage[font=footnotesize,hypcap=true]{caption}

\usepackage{booktabs}
\usepackage{multirow}

\usepackage{enumerate}
\usepackage{paralist}

\usepackage[resetfonts]{cmap}
\usepackage[T1]{fontenc}
\usepackage[utf8]{inputenc}

\usepackage{algorithm}
\usepackage{algpseudocode}
\usepackage{etoolbox}
\usepackage{calc}

\usepackage{wrapfig}

\usepackage[pdftex,breaklinks=true,bookmarks=true]{hyperref}
\usepackage{natbib}
\usepackage{cleveref}

\usepackage{rotating}

\ifdefined\medphys
  \title{Iterative Inversion of Deformation Vector Fields with Feedback Control}
\fi\ifdefined\plain
  \title{Iterative Inversion of Deformation Vector Fields with\\Feedback Control}
\fi

\makeatletter
\let\titletext\@title
\makeatother

\ifdefined\review
  \newcommand{\pdfauthors}{}
\else
  \newcommand{\pdfauthors}{%
    A.K. Dubey, A.S. Iliopoulos, X. Sun, F. Yin, L. Ren}
\fi

\graphicspath{%
  {whiteboards/}%
  {../../figs/dvf-mappings/pdfs/}%
  {../../figs/dvf-mappings/jpg/}%
  {../../figs/dvf-mappings/png/}%
  {../../figs/dvf-mappings/ipe/}%
}

\crefname{figure}{Fig.}{Figs.}
\crefname{table}{Table}{Tables}
\crefname{algorithm}{Alg.}{Algs.}
\crefname{section}{Section}{Sections}
\crefname{paragraph}{Section}{Sections}
\crefname{appendix}{Appendix}{Appendices}
\crefname{equation}{}{}
\crefrangeformat{equation}{#3(#1)#4--#5(#2)#6}
\crefrangeformat{figure}{Figs.~#3#1#4--#5#2#6}
\crefrangeformat{table}{Tables~#3#1#4--#5#2#6}
\crefrangeformat{algorithm}{Algs.~#3#1#4--#5#2#6}

\ifdefined\plain\ifdefined\review\else
\fi\fi

\hypersetup{%
  pdffitwindow=false,%
  pdfstartview={FitH},%
  colorlinks,%
  pdfauthor={\pdfauthors},%
  pdftitle={\titletext},%
  pagebackref=false,%
  citecolor=Blue1,%
  filecolor=Blue1,%
  linkcolor=Blue1,%
  urlcolor=Blue1%
}

\setkeys{Gin}{draft=false}

\pdfpageattr{/Group <</S /Transparency /I true /CS /DeviceRGB>>}

\let\leftorig\left
\let\rightorig\right
\renewcommand{\left}{\mathopen{}\mathclose\bgroup\leftorig}
\renewcommand{\right}{\aftergroup\egroup\rightorig}

\ifdefined\review
  
\else
  
\fi

\hypersetup{final}

\providecommand{\email}[1]{\href{mailto:#1}{#1}}

\makeatletter
\DeclareRobustCommand\ttfamily{%
\not@math@alphabet\ttfamily\mathtt
\fontfamily\ttdefault\selectfont\hyphenchar\font=-1\relax}
\makeatother
\DeclareTextFontCommand{\tturl}{\ttfamily\hyphenchar\font=`/}

\algrenewcommand{\alglinenumber}[1]{%
  \sf\footnotesize \textcolor{gray}{#1}}

\algrenewcommand{\algorithmicrequire}{\textbf{Input:}}

\algrenewcomment[1]{\hfill \(\triangleright\) \textit{#1}}

\makeatletter
\newcommand{\etal}{%
  et al\@ifnextchar.{}{.}}
\makeatother

\ifdefined\medphys\ifdefined\review
  \makeatletter
  \patchcmd\linenumberpar{\@LN@parpgbrk}{\penalty\@LN@parpgpen\relax}{}{}
  \makeatother
\fi\fi

\ifdefined\medphys\ifdefined\review
  \newcommand*\patchAmsMathEnvironmentForLineno[1]{%
    \expandafter\let\csname old#1\expandafter\endcsname\csname #1\endcsname
    \expandafter\let\csname oldend#1\expandafter\endcsname\csname end#1\endcsname
    \renewenvironment{#1}%
    {\linenomath\csname old#1\endcsname}%
    {\csname oldend#1\endcsname\endlinenomath}}% 
  \newcommand*\patchBothAmsMathEnvironmentsForLineno[1]{%
    \patchAmsMathEnvironmentForLineno{#1}%
    \patchAmsMathEnvironmentForLineno{#1*}}%
  \AtBeginDocument{%
    \patchBothAmsMathEnvironmentsForLineno{equation}%
    \patchBothAmsMathEnvironmentsForLineno{align}%
    \patchBothAmsMathEnvironmentsForLineno{flalign}%
    \patchBothAmsMathEnvironmentsForLineno{alignat}%
    \patchBothAmsMathEnvironmentsForLineno{gather}%
    \patchBothAmsMathEnvironmentsForLineno{multline}%
  }
\fi\fi

\makeatletter
\newdimen\SOUL@dimen %new
\def\SOUL@ulunderline#1{{%
    \setbox\z@\hbox{#1}%
    \SOUL@dimen=\wd\z@ %new
    \dimen@i=\SOUL@uloverlap
    \advance\SOUL@dimen2\dimen@i %\dimen@ exchanged too
    \rlap{%
        \null
        \kern-\dimen@i
        \SOUL@ulcolor{\SOUL@ulleaders\hskip\SOUL@dimen}% new
    }%
    \unhcopy\z@
}}

\makeatletter
\newcolumntype{B}[3]{>{\boldmath\DC@{#1}{#2}{#3}}c<{\DC@end}}
\makeatother
\newcolumntype{d}[1]{D{.}{.}{#1}}
\newcolumntype{e}[1]{D{e}{\text{\textsc{e}}\!\!}{#1}}
\newcolumntype{b}[1]{B{e}{\text{\bf\textsc{e}}\!\!}{#1}}
\newcolumntype{S}{>{\centering\arraybackslash}m{3em}}
\newcolumntype{M}{>{\centering\arraybackslash}m{4em}}
\newcolumntype{L}{>{\centering\arraybackslash}m{6em}}

\ifdefined\medphys
  \AtBeginDocument{
    \heavyrulewidth=.08em
    \lightrulewidth=.05em
    \cmidrulewidth=.03em
    \belowrulesep=.65ex
    \belowbottomsep=0pt
    \aboverulesep=.4ex
    \abovetopsep=0pt
    \cmidrulesep=\doublerulesep
    \cmidrulekern=.5em
    \defaultaddspace=.5em
  }
\fi

\tcbset{
  highlight math style={
    colback=yellow,
    arc=0pt,
    outer arc=0pt,
    boxrule=0pt,
    top=2pt,
    bottom=2pt,
    left=2pt,
    right=2pt,
  }
}
\ifdefined\review
  
\else
  
\fi

\makeatletter
\newsavebox{\@tabnotebox}
\providecommand\tmark{} % so having ctable or not is irrelevant
\providecommand\tnote{}
\newenvironment{tabularwithnotes}[3][c]
  {\long\def\@tabnotes{#3}%
   \renewcommand\tmark[1][a]{\makebox[0pt][l]{\textsuperscript{\itshape##1}}}%
   \renewcommand\tnote[2][a]{\textsuperscript{\itshape##1}\,##2\par}
   \begin{lrbox}{\@tabnotebox}
   \begin{tabular}{#2}}
  {\end{tabular}\end{lrbox}%
   \parbox{\wd\@tabnotebox}{
     \usebox{\@tabnotebox}\par
     \smallskip\raggedright\scriptsize\@tabnotes
   }%
  }
\makeatother

\ifdefined\review               % ----- review mode
  \newcommand{\inlineMathDisplay}[1]{\displaystyle #1}
\else
  \newcommand{\inlineMathDisplay}[1]{#1}
\fi

\newcommand{\mbs}[1]{\mathbf{#1}}

\newcommand{\RF}{\mbs{r_u}}
\newcommand{\R}{\mbs{r}}

\newcommand{\I}{\mbs{I}}
\newcommand{\F}{\mbs{u}}
\newcommand{\G}{\mbs{v}}
\newcommand{\x}{\mbs{x}}
\newcommand{\xxi}{\boldsymbol{\mbs{\xi}}}
\newcommand{\Ghat}{\mbs{\hat{v}}}

\newcommand{\f}{\mbs{f}}
\newcommand{\g}{\mbs{g}}

\newcommand{\B}{\mbs{B}}
\newcommand{\E}{\mbs{e}}
\newcommand{\EPM}{\mbs{P}}
\newcommand{\ICM}{\mbs{Q}}
\newcommand{\J}{\mbs{J}}

\newcommand{\imRef}{I_{\text{ref}}}
\newcommand{\imTgt}{I_{\text{tgt}}}

\newcommand{\idRef}{\imRef}        % backwards compatibility
\newcommand{\idTgt}{\imTgt}        % backwards compatibility

\newcommand*{\ud}{\mathop{}\!\mathrm{d}}
\newcommand{\setcond}[2]{\left\{#1 \;\middle|\; #2 \right\}}

\newcommand{\smsp}{\mspace{2mu}}
\DeclareMathOperator*{\argmin}{arg\smsp min}

\newcommand{\abs}[1]{\left\lvert #1 \right\rvert}
\newcommand{\norm}[1]{\left\lVert #1 \right\rVert}
\newcommand{\prctile}{\beta}

\newcommand{\tpose}[1]{#1^{\mathsf{T}}}
\DeclareMathOperator{\real}{Re}
\DeclareMathOperator{\imag}{Im}

\renewcommand{\det}[1]{\abs{#1}}

\newcommand{\reciprocalGap}{\gamma}
\newcommand{\scalarfield}{\phi}

\newcommand{\nhood}{\mathcal{N}}

\colorlet{clrMidSep}{black!30!white}
\newlength{\tabMidSepWidth}
\newlength{\tabvskip}
\newlength{\valignImgSkip}
\newlength{\vskipColorbar}
\newlength{\subcaptionSkip}
\newlength{\trimCrule}
\usetikzlibrary{fit,backgrounds}

\ifdefined\medphys              % -------------------- Medical Physics
  \ifdefined\review             % ----- review mode
    \newcommand{\widthConvergenceFig}{0.80\linewidth}
    \newcommand{\wideonecol}{0.475\linewidth}

    \newcommand{\thirdonecol}{0.30\linewidth}
    
    \newcommand{\widetwocol}{0.98\linewidth}
    
    \newcommand{\halftwocol}{0.48\linewidth}
    
    \newcommand{\ichalftwocolleft}{0.50000001\linewidth}
    \newcommand{\ichalftwocolright}{0.44999999\linewidth}
    
    \newcommand{\ictextmarginleft}{0.08\linewidth}
    \newcommand{\icwidthImgleft}{0.9\linewidth}
    \newcommand{\icwidthImgright}{\linewidth}
    \newcommand{\rhotextmargin}{0.08\linewidth}
    \newcommand{\rhoThirdAxial}{0.31517444304\linewidth}
    \newcommand{\rhoThirdCoronal}{0.30241277847\linewidth}
    \newcommand{\rhoThirdSagittal}{0.30241277847\linewidth}
    \newcommand{\dctextmargin}{0.08\linewidth}
    \newcommand{\dcwidthImg}{0.78\linewidth}
    \newcommand{\dcThirdAxial}{0.34258091635\linewidth}
    \newcommand{\dcThirdCoronal}{0.32870954182\linewidth}
    \newcommand{\dcThirdSagittal}{0.32870954182\linewidth}
    \newcommand{\dcwidthCbar}{0.1\linewidth}
    \newcommand{\dvfThirdAxial}{0.33908768745\linewidth}
    \newcommand{\dvfThirdCoronal}{0.32545615627\linewidth}
    \newcommand{\dvfThirdSagittal}{0.32545615627\linewidth}

    \newcommand{\rwidthImgP}{0.875075393552367\linewidth}
    \newcommand{\rwidthCbarP}{0.109924606447633\linewidth}
    
    \newcommand{\thirdAxialPR}{0.32544\linewidth}
    \newcommand{\thirdCoronalPR}{0.31654\linewidth}
    \newcommand{\thirdSagittalPR}{0.31654\linewidth}

  \else                         % ----- journal mode
    \newcommand{\dvfThirdAxial}{0.33908768745\linewidth}
    \newcommand{\dvfThirdCoronal}{0.32545615627\linewidth}
    \newcommand{\dvfThirdSagittal}{0.32545615627\linewidth}
    \newcommand{\dctextmargin}{0.08\linewidth}
    \newcommand{\dcwidthImg}{0.78\linewidth}
    \newcommand{\dcThirdAxial}{0.34258091635\linewidth}
    \newcommand{\dcThirdCoronal}{0.32870954182\linewidth}
    \newcommand{\dcThirdSagittal}{0.32870954182\linewidth}
    \newcommand{\dcwidthCbar}{0.1\linewidth}
    \newcommand{\rhotextmargin}{0.08\linewidth}
    \newcommand{\rhoThirdAxial}{0.31517444304\linewidth}
    \newcommand{\rhoThirdCoronal}{0.30241277847\linewidth}
    \newcommand{\rhoThirdSagittal}{0.30241277847\linewidth}
    \newcommand{\ichalftwocolleft}{0.50000001\linewidth}
    \newcommand{\ichalftwocolright}{0.44999999\linewidth}
    
    \newcommand{\ictextmarginleft}{0.08\linewidth}
    \newcommand{\icwidthImgleft}{0.9\linewidth}
    \newcommand{\icwidthImgright}{\linewidth}
    \newcommand{\widthConvergenceFig}{1.0\linewidth}
    \newcommand{\wideonecol}{1.0\linewidth}

    \newcommand{\thirdonecol}{0.31\linewidth}
    
    \newcommand{\widetwocol}{0.98\linewidth}
    
    \newcommand{\halftwocol}{0.48\linewidth}

    \newcommand{\thirdAxialPR}{0.33\linewidth}
    \newcommand{\thirdCoronalPR}{0.32\linewidth}
    \newcommand{\thirdSagittalPR}{0.32\linewidth}

    \newcommand{\rwidthImgP}{0.875075393552367\linewidth}
    \newcommand{\rwidthCbarP}{0.109924606447633\linewidth}

  \fi
\fi\ifdefined\plain             % -------------------- plain
  \ifdefined\review             % ----- draft mode
    \newcommand{\dvfThirdAxial}{0.33908768745\linewidth}
    \newcommand{\dvfThirdCoronal}{0.32545615627\linewidth}
    \newcommand{\dvfThirdSagittal}{0.32545615627\linewidth}
    \newcommand{\dctextmargin}{0.08\linewidth}
    \newcommand{\dcwidthImg}{0.78\linewidth}
    \newcommand{\dcThirdAxial}{0.34258091635\linewidth}
    \newcommand{\dcThirdCoronal}{0.32870954182\linewidth}
    \newcommand{\dcThirdSagittal}{0.32870954182\linewidth}
    \newcommand{\dcwidthCbar}{0.1\linewidth}
    \newcommand{\rhotextmargin}{0.08\linewidth}
    \newcommand{\rhoThirdAxial}{0.31517444304\linewidth}
    \newcommand{\rhoThirdCoronal}{0.30241277847\linewidth}
    \newcommand{\rhoThirdSagittal}{0.30241277847\linewidth}
    \newcommand{\ichalftwocolleft}{0.50000001\linewidth}
    \newcommand{\ichalftwocolright}{0.44999999\linewidth}
    \newcommand{\ictextmarginleft}{0.08\linewidth}
    \newcommand{\icwidthImgleft}{0.9\linewidth}
    \newcommand{\icwidthImgright}{\linewidth}
    \newcommand{\widthConvergenceFig}{0.80\linewidth}
    \newcommand{\wideonecol}{0.475\linewidth}

    \newcommand{\thirdonecol}{0.32\linewidth}
    
    \newcommand{\widetwocol}{0.98\linewidth}
    
    \newcommand{\halftwocol}{0.49\linewidth}

    \newcommand{\thirdAxialPR}{0.33\linewidth}
    \newcommand{\thirdCoronalPR}{0.32\linewidth}
    \newcommand{\thirdSagittalPR}{0.32\linewidth}

    \newcommand{\rwidthImgP}{0.875075393552367\linewidth}
    \newcommand{\rwidthCbarP}{0.109924606447633\linewidth}

  \else                         % ----- plain mode
    \newcommand{\dvfThirdAxial}{0.33908768745\linewidth}
    \newcommand{\dvfThirdCoronal}{0.32545615627\linewidth}
    \newcommand{\dvfThirdSagittal}{0.32545615627\linewidth}
    \newcommand{\dctextmargin}{0.08\linewidth}
    \newcommand{\dcwidthImg}{0.78\linewidth}
    \newcommand{\dcThirdAxial}{0.34258091635\linewidth}
    \newcommand{\dcThirdCoronal}{0.32870954182\linewidth}
    \newcommand{\dcThirdSagittal}{0.32870954182\linewidth}
    \newcommand{\dcwidthCbar}{0.1\linewidth}
    \newcommand{\rhotextmargin}{0.08\linewidth}
    \newcommand{\rhoThirdAxial}{0.31517444304\linewidth}
    \newcommand{\rhoThirdCoronal}{0.30241277847\linewidth}
    \newcommand{\rhoThirdSagittal}{0.30241277847\linewidth}
    \newcommand{\ichalftwocolleft}{0.50000001\linewidth}
    \newcommand{\ichalftwocolright}{0.44999999\linewidth}
    
    \newcommand{\ictextmarginleft}{0.08\linewidth}
    \newcommand{\icwidthImgleft}{0.9\linewidth}
    \newcommand{\icwidthImgright}{\linewidth}
    \newcommand{\widthConvergenceFig}{1.0\linewidth}
    \newcommand{\wideonecol}{1.0\linewidth}

    \newcommand{\thirdonecol}{0.33\linewidth}
    
    \newcommand{\widetwocol}{0.98\linewidth}
    
    \newcommand{\halftwocol}{0.49\linewidth}

    \newcommand{\thirdAxialPR}{0.33\linewidth}
    \newcommand{\thirdCoronalPR}{0.32\linewidth}
    \newcommand{\thirdSagittalPR}{0.32\linewidth}

    \newcommand{\rwidthImgP}{0.875075393552367\linewidth}
    \newcommand{\rwidthCbarP}{0.109924606447633\linewidth}

  \fi
\fi

\begin{document}

% double line spacing for draft mode
\ifdefined\plain\ifdefined\review
  \doublespacing
\fi\fi

%%%%%%%%%%%%%%%%%%%%%%%%%%%%%%%%%%%%%%%%%%%%%%%%%%
%%% FRONT MATTER

\ifdefined\medphys              % ---------- Medical Physics

  % ----- title
  \title{\titletext}

\ifdefined\medphys              % ---------- AIP Medical Physics

  \author{Abhishek Kumar Dubey}%
  \thanks{A.K.\ Dubey and A.S.\ Iliopoulos have contributed equally to this
    study, and are considered as co-first authors.}

  \author{Alexandros-Stavros Iliopoulos}%
  \thanks{A.K.\ Dubey and A.S.\ Iliopoulos have contributed equally to this
    study, and are considered as co-first authors.}
  
  \author{Xiaobai Sun}

  \affiliation{Department of Computer Science, Duke University, Durham,
    NC 27708, USA}

  \author{Fang-Fang Yin}

  \affiliation{Department of Radiation Oncology, Duke University School of
    Medicine, Durham, NC 27710, USA
    \newline
    Medical Physics Program, Duke Kunshan University, Kunshan, Jiangsu,
    China 215316}

  \author{Lei Ren}
  \email[Author to whom correspondence should be addressed.
    Electronic mail: ]{lei.ren@duke.edu}

  \affiliation{Department of Radiation Oncoology, Duke University School of
    Medicine, Durham, NC 27710, USA}
  
  \ifdefined\review             % ----- review

    \begin{minipage}[t][0.96\textheight][b]{1.0\linewidth}
      \footnotesize \textsuperscript{a)}%
      A.K.\ Dubey and A.S.\ Iliopoulos have contributed equally to this
      study, and are considered as co-first authors.
      \\
      \footnotesize \textsuperscript{b)}%
      Author to whom correspondence should be addressed.
      Electronic mail: \href{mailto:lei.ren@duke.edu}{lei.ren@duke.edu}
    \end{minipage}
    \par
    \vspace*{-\textheight}
    \resetlinenumber[1]

  \fi
  
\fi\ifdefined\plain             % ---------- plain

  \author[1]{Abhishek~Kumar~Dubey%
             \footnote{A.K.\ Dubey and A.S.\ Iliopoulos have contributed
               equally to this study, and are considered as co-first
               authors.}}
  % [https://tex.stackexchange.com/a/173331/11788]
  \newcommand\CoAuthorMark{\footnotemark[\arabic{footnote}]}
  \author[1]{Alexandros-Stavros~Iliopoulos%
             \protect\CoAuthorMark}
  \author[1]{Xiaobai~Sun}
  \author[2]{Fang-Fang~Yin}
  \author[2]{Lei~Ren}

  \affil[1]{%
    Department~of~Computer~Science,
    Duke~University,
    Durham,~NC~27705,~USA}
  \affil[2]{%
    Department~of~Radiation~Oncology,
    Duke~University~School~of~Medicine,
    Durham,~NC~27710,~USA}

\fi

  % ----- date [review mode only]
  \ifdefined\review
    \date{March 12, 2018}
  \fi
  
  % ----- abstract
  \begin{abstract}
    \label{sec:abstract}

\noindent
\textbf{Purpose:}
Often, the inverse deformation vector field~(DVF) is needed together with
the corresponding forward DVF in 4D reconstruction and dose calculation,
adaptive radiation therapy, and simultaneous deformable registration.
This study aims at improving both accuracy and efficiency of iterative
algorithms for DVF inversion, and advancing our understanding of
divergence and latency conditions.

% --------------------------------------------------

\noindent
\textbf{Method:}
We introduce a framework of fixed-point iteration algorithms with active
feedback control for DVF inversion.
Based on rigorous convergence analysis, we design control mechanisms for
modulating the inverse consistency (IC) residual of the current iterate, to
be used as feedback into the next iterate.  
The control is designed adaptively to the input DVF with the objective to 
enlarge the convergence area and expedite convergence.
Three particular settings of feedback control are introduced: constant
value over the domain throughout the iteration; alternating values between
iteration steps; and spatially variant values.
We also introduce three spectral measures of the displacement Jacobian for
characterizing a DVF.
These measures reveal the critical role of what we term the
non-translational displacement component (NTDC) of the DVF.
% in our analysis and adaptive iteration design.
%
We carry out inversion experiments with an analytical DVF pair, and with
DVFs associated with thoracic CT images of $6$ patients at end of
expiration and end of inspiration.
%
% --------------------------------------------------

\noindent
\textbf{Results:}
NTDC-adaptive iterations are shown to attain a larger convergence region at
a faster pace compared to previous non-adaptive DVF inversion iteration
algorithms.
By our numerical experiments, alternating control yields smaller IC
residuals and inversion errors than constant control.  Spatially variant
control renders smaller residuals and errors by at least an order of
magnitude, compared to other schemes, in no more than $10$ steps.
Inversion results also show remarkable quantitative agreement with
analysis-based predictions.

% --------------------------------------------------

\noindent
\textbf{Conclusion:}
Our analysis captures properties of DVF data associated with clinical CT
images, and provides new understanding of iterative DVF inversion
algorithms with a simple residual feedback control.  Adaptive control is
necessary and highly effective in the presence of non-small NTDCs.  The
adaptive iterations or the spectral measures, or both, may potentially be
incorporated into deformable image registration methods.

% ----- abstract
  \end{abstract}

  % ----- typeset!
  \maketitle

\fi\ifdefined\plain             % ---------- plain

\ifdefined\medphys              % ---------- AIP Medical Physics

  \author{Abhishek Kumar Dubey}%
  \thanks{A.K.\ Dubey and A.S.\ Iliopoulos have contributed equally to this
    study, and are considered as co-first authors.}

  \author{Alexandros-Stavros Iliopoulos}%
  \thanks{A.K.\ Dubey and A.S.\ Iliopoulos have contributed equally to this
    study, and are considered as co-first authors.}
  
  \author{Xiaobai Sun}

  \affiliation{Department of Computer Science, Duke University, Durham,
    NC 27708, USA}

  \author{Fang-Fang Yin}

  \affiliation{Department of Radiation Oncology, Duke University School of
    Medicine, Durham, NC 27710, USA
    \newline
    Medical Physics Program, Duke Kunshan University, Kunshan, Jiangsu,
    China 215316}

  \author{Lei Ren}
  \email[Author to whom correspondence should be addressed.
    Electronic mail: ]{lei.ren@duke.edu}

  \affiliation{Department of Radiation Oncoology, Duke University School of
    Medicine, Durham, NC 27710, USA}
  
  \ifdefined\review             % ----- review

    \begin{minipage}[t][0.96\textheight][b]{1.0\linewidth}
      \footnotesize \textsuperscript{a)}%
      A.K.\ Dubey and A.S.\ Iliopoulos have contributed equally to this
      study, and are considered as co-first authors.
      \\
      \footnotesize \textsuperscript{b)}%
      Author to whom correspondence should be addressed.
      Electronic mail: \href{mailto:lei.ren@duke.edu}{lei.ren@duke.edu}
    \end{minipage}
    \par
    \vspace*{-\textheight}
    \resetlinenumber[1]

  \fi
  
\fi\ifdefined\plain             % ---------- plain

  \author[1]{Abhishek~Kumar~Dubey%
             \footnote{A.K.\ Dubey and A.S.\ Iliopoulos have contributed
               equally to this study, and are considered as co-first
               authors.}}
  % [https://tex.stackexchange.com/a/173331/11788]
  \newcommand\CoAuthorMark{\footnotemark[\arabic{footnote}]}
  \author[1]{Alexandros-Stavros~Iliopoulos%
             \protect\CoAuthorMark}
  \author[1]{Xiaobai~Sun}
  \author[2]{Fang-Fang~Yin}
  \author[2]{Lei~Ren}

  \affil[1]{%
    Department~of~Computer~Science,
    Duke~University,
    Durham,~NC~27705,~USA}
  \affil[2]{%
    Department~of~Radiation~Oncology,
    Duke~University~School~of~Medicine,
    Durham,~NC~27710,~USA}

\fi

% ----- authors
  
  \ifdefined\review             % ----- (draft)
  
    % ----- date
    \date{February 25, 2017}

    % ----- title
    \maketitle

    % ----- table of contents
    \clearpage
    \tableofcontents
    \clearpage
  
    % ----- abstract
    \phantomsection
    \addcontentsline{toc}{section}{Abstract}
    \begin{abstract}
        \label{sec:abstract}

\noindent
\textbf{Purpose:}
Often, the inverse deformation vector field~(DVF) is needed together with
the corresponding forward DVF in 4D reconstruction and dose calculation,
adaptive radiation therapy, and simultaneous deformable registration.
This study aims at improving both accuracy and efficiency of iterative
algorithms for DVF inversion, and advancing our understanding of
divergence and latency conditions.

% --------------------------------------------------

\noindent
\textbf{Method:}
We introduce a framework of fixed-point iteration algorithms with active
feedback control for DVF inversion.
Based on rigorous convergence analysis, we design control mechanisms for
modulating the inverse consistency (IC) residual of the current iterate, to
be used as feedback into the next iterate.  
The control is designed adaptively to the input DVF with the objective to 
enlarge the convergence area and expedite convergence.
Three particular settings of feedback control are introduced: constant
value over the domain throughout the iteration; alternating values between
iteration steps; and spatially variant values.
We also introduce three spectral measures of the displacement Jacobian for
characterizing a DVF.
These measures reveal the critical role of what we term the
non-translational displacement component (NTDC) of the DVF.
We carry out inversion experiments with an analytical DVF pair, and with
DVFs associated with thoracic CT images of $6$ patients at end of
expiration and end of inspiration.
%
% --------------------------------------------------

\noindent
\textbf{Results:}
NTDC-adaptive iterations are shown to attain a larger convergence region at
a faster pace compared to previous non-adaptive DVF inversion iteration
algorithms.
By our numerical experiments, alternating control yields smaller IC
residuals and inversion errors than constant control.  Spatially variant
control renders smaller residuals and errors by at least an order of
magnitude, compared to other schemes, in no more than $10$ steps.
Inversion results also show remarkable quantitative agreement with
analysis-based predictions.

% --------------------------------------------------

\noindent
\textbf{Conclusion:}
Our analysis captures properties of DVF data associated with clinical CT
images, and provides new understanding of iterative DVF inversion
algorithms with a simple residual feedback control.  Adaptive control is
necessary and highly effective in the presence of non-small NTDCs.  The
adaptive iterations or the spectral measures, or both, may potentially be
incorporated into deformable image registration methods.

% ----- abstract
    \end{abstract}

  \else                         % ----- (plain)
  
    % ----- no date
    \date{}

    % ----- single-column title & abstract
    \maketitle
    \begin{abstract}

\noindent
\textbf{Purpose:}
Often, the inverse deformation vector field~(DVF) is needed together with
the corresponding forward DVF in 4D reconstruction and dose calculation,
adaptive radiation therapy, and simultaneous deformable registration.
This study aims at improving both accuracy and efficiency of iterative
algorithms for DVF inversion, and advancing our understanding of
divergence and latency conditions.

% --------------------------------------------------

\noindent
\textbf{Method:}
We introduce a framework of fixed-point iteration algorithms with active
feedback control for DVF inversion.
Based on rigorous convergence analysis, we design control mechanisms for
modulating the inverse consistency (IC) residual of the current iterate, to
be used as feedback into the next iterate.  
The control is designed adaptively to the input DVF with the objective to 
enlarge the convergence area and expedite convergence.
Three particular settings of feedback control are introduced: constant
value over the domain throughout the iteration; alternating values between
iteration steps; and spatially variant values.
We also introduce three spectral measures of the displacement Jacobian for
characterizing a DVF.
These measures reveal the critical role of what we term the
non-translational displacement component (NTDC) of the DVF.
% in our analysis and adaptive iteration design.
%
We carry out inversion experiments with an analytical DVF pair, and with
DVFs associated with thoracic CT images of $6$ patients at end of
expiration and end of inspiration.
%
% --------------------------------------------------

\noindent
\textbf{Results:}
NTDC-adaptive iterations are shown to attain a larger convergence region at
a faster pace compared to previous non-adaptive DVF inversion iteration
algorithms.
By our numerical experiments, alternating control yields smaller IC
residuals and inversion errors than constant control.  Spatially variant
control renders smaller residuals and errors by at least an order of
magnitude, compared to other schemes, in no more than $10$ steps.
Inversion results also show remarkable quantitative agreement with
analysis-based predictions.

% --------------------------------------------------

\noindent
\textbf{Conclusion:}
Our analysis captures properties of DVF data associated with clinical CT
images, and provides new understanding of iterative DVF inversion
algorithms with a simple residual feedback control.  Adaptive control is
necessary and highly effective in the presence of non-small NTDCs.  The
adaptive iterations or the spectral measures, or both, may potentially be
incorporated into deformable image registration methods.

% ----- single-column title & abstract
    \end{abstract}
    \twocolumn

  \fi
  
\fi

%%%%%%%%%%%%%%%%%%%%%%%%%%%%%%%%%%%%%%%%%%%%%%%%%%
%%% DOCUMENT

% ==================== INTRODUCTION
%
\section{Introduction}
\label{sec:introduction}

We consider numerical inversion of a deformation vector field (DVF).
Inverse DVFs are needed, together with their respective forward DVFs, to
map images, structure contours, or doses back and forth in applications
such as 4D image reconstruction~\cite{ren2014limited}, dose accumulation
calculations and multi-modality treatment planning in adaptive
radiotherapy~\cite{yan_pseudoinverse_2010, vercauteren2013deformation,
  rivest-henault_robust_2015, gustafsson2017assessment}, and cardiac
functional analysis~\cite{fung2017motion}.
DVF inversion is also a fundamental operation in simultaneous and symmetric
registration methods~\cite{christensen2001consistent, leow2005inverse,
  avants_symmetric_2008, sotiras2013deformable, heinrich_deformable_2016}.
An important consideration is ensuring that the forward and reverse
mappings are inverse-consistent~\cite{christensen1999consistent,
  sotiras2013deformable}.
Theoretical guarantees of convergence and computational efficiency have
been a long-standing open problem with DVF inversion.

A consistent pair of forward and reverse mappings can be obtained via
one-way deformable registration followed by a DVF inversion
process\cite{chen2008simple, yan_pseudoinverse_2010}, or via simultaneous,
symmetric two-way registration methods~\cite{christensen2001consistent,
  leow2005inverse, avants_symmetric_2008, heinrich_deformable_2016}.
The former, asymmetric approach is often preferred in certain clinical
applications with limited time window, in part because it is shown to be
faster empirically, and in part because of the non-negligible asymmetry in
clinical image quality.  One of the images may be more adversely affected
by noise or artifacts than the other.  Such asymmetry makes one-way
registration better or worse depending on the mapping direction, due to the
high sensitivity of registration methods to noise or variation in imaging
conditions~\cite{sabuncu_asymmetric_2009, yan_pseudoinverse_2010,
  fung2017motion}.
We consider the asymmetric approach to be an effective means to
counterbalancing the asymmetry in image quality.

DVF inversion is often involved also in simultaneous registration, which
results in both forward and reverse
mappings~\cite{christensen2001consistent, leow2005inverse,
  avants_symmetric_2008, heinrich_deformable_2016}.
Inverse consistency~(IC) between forward and reverse mappings for
deformable registration was addressed in the early work of
Thirion~\cite{thirion1998image} and
Christensen~\cite{christensen1999consistent}.
The IC condition has since
been incorporated in various deformable registration models.
It is either used as an explicit constraint attached to an optimization
model~\cite{christensen2001consistent}, or employed implicitly and
approximately in numerical iterations~\cite{leow2005inverse}.
The registration process may involve multiple intermediate transformations
and their composition.  DVF inversion is used to ensure that the
transformations, intermediate as well as final, meet the IC condition.
Many studies on simultaneous estimation of forward-inverse DVF pairs can be
found in the survey by Sotiras \etal~\cite{sotiras2013deformable} and
references therein.

With a provided DVF as input, numerical inversion of the DVF can be
governed by the IC condition and carried out in displacement space, rather
than in image space.  The inversion relationship is inherently nonlinear.
One therefore resorts to iterative solution methods, except in certain
cases.
There is a close relationship between IC residuals (i.e., deviations from
the IC condition) and inversion errors in the iterative estimates.  We will
leverage this relationship to improve upon the inverse DVF iterates by
using the IC residuals as feedback into the iteration.

Two particular and influential iteration algorithms for DVF inversion were
developed by Christensen and Johnson~\cite{christensen2001consistent} and
Chen \etal~\cite{chen2008simple}.
The iteration by Christensen and Johnson~\cite{christensen2001consistent}
is often effective, and is notable in making use of residual feedback.  The
iteration is closely related to the residual method by
Thirion~\cite{thirion1998image}, which is based on a heuristic to enforce
that DVF estimates are bijective (invertible).
The condition under which the iteration converges or fails to converge was
hitherto unknown.  Chen \etal~\cite{chen2008simple} departed
from heuristic design.  They introduced a particular fixed-point
iteration for DVF inversion, and identified a convergence condition.  
The iteration was not compared to the earlier algorithm by Christensen and
Johnson~\cite{christensen2001consistent}, and there is no mention of
residual feedback.

We make the following key contributions to the understanding and
convergence control of iterative DVF inversion algorithms.
\begin{inparaenum}[(i)] 
\item We present a framework of iteration algorithms with simple and
  adaptive residual feedback control.  This includes the two precursor
  algorithms, in both format and analysis.
\item The framework is underpinned by a unified analysis of error
  propagation and convergence.  The analysis enables connections and
  comparisons among iteration algorithms for DVF inversion, and leads to
  the design of more effective ones.
\item We characterize the critical role of what we introduce as the
  non-translational displacement component (NTDC) in error propagation, and
  provide quantitative NTDC measures.
  When the NTDC is non-small, adaptive residual feedback control is
  necessary to guarantee convergence.
  This insight is new.
\end{inparaenum} 
We assess our findings experimentally with synthetic DVF data, and patient
DVF data obtained from thoracic CT images.

The rest of the document is organized as follows.  In \cref{sec:method}, we
introduce our algorithm framework, provide formal analysis, present three
practical feedback control schemes, and discuss three spectral measures for
NTDC characterization.  In \cref{sec:experiments}, we describe and assess
the patient DVF data used in our experiments.  Pre-inversion assessment of
control schemes and post-inversion evaluation of results with the patient
DVFs are provided in \cref{sec:results}.  A direct evaluation using
analytical DVF data is included in
\cref{sec:appendix-casestudy-analyticaldvf}.  We conclude the paper in
\cref{sec:discussion} with additional discussion on the clinical utility of
our algorithms and analysis.

% ==================== METHODS
%
\section{Methods}
\label{sec:method}

% ---------- Preliminaries
%
\subsection{DVF inversion preliminaries}
\label{subsec:preliminaries}

DVF inversion can be phrased as follows.  A reference and a target image,
denoted by $\idRef$ and $\idTgt$, respectively, can be related to one
another by two non-linear transformations.  Denote by $\Omega$ the image
domain, $\Omega \subset \mathbb{R}^3$.  The forward transformation,
$\f \colon \Omega \to \Omega$, maps the voxels of the reference image
$\idRef$ onto those of the target image $\idTgt$ via
$\f(\x') = \x' + \F(\x')$,
where $\F(\x')$ is the forward 3D displacement at $\x' \in \Omega$.
Conversely, the inverse transformation, $\g \colon \Omega \to \Omega$, maps
the voxels of $\idTgt$ back to $\idRef$ via
$\g(\x) = \x + \G(\x)$,
where $\G(\x)$ is the inverse 3D displacement at $\x \in \Omega$.
DVF inversion means obtaining $\G$ given $\F$.
The two transformations are inverse to each other:
$(\f \circ \g) (\x) = \x$,
and
$(\g \circ \f) (\x') = \x'$
for $\x, \x' \in \Omega$.
Consequently, the forward and inverse DVFs satisfy the simultaneous
\emph{inverse consistency}~(IC) condition:
\begin{subequations}
  \label{eq:consistency-condition}
  \begin{align}
  \label{eq:consistency-condition-G}
  \G(\x) + \F(\x + \G(\x)) &= \mbs{0},
  \\
   \label{eq:consistency-condition-F}
  \F(\x') + \G(\x' + \F(\x'))  &= \mbs{0}.
  \end{align}
\end{subequations}
The IC condition governs iterative DVF inversion.

% ----- error-residual relationship

\subsubsection{Inversion error \& inverse consistency\\residual}
\label{subsub:error-residual-relation}

Denote by $\Ghat$ an estimate of the inverse DVF $\G_*$. The unknown error
in the estimate,
\begin{equation}
\label{eq:iteratate-error} 
  \E(\x) = \Ghat(\x) - \G_*(\x) , 
\end{equation} 
is manifested in the inverse consistency~(IC) residual,
\begin{align}
\label{eq:residual-R-G}
  \R_{\G}(\x)  = 
  \Ghat(\x) + \F (\x + \Ghat(\x) ),
\end{align}
which is computationally available.  Qualitatively, the residual is zero if
and only if the inversion error is zero.  In order to use the IC residual
as feedback for improving the inverse DVF estimate, we investigate the
quantitative relationship between inversion error and IC residual.

Assume in analysis that the deformation transformation $\f$ is
differentiable.  By the mean value theorem, the IC residual and the
inversion error can be related by
\begin{equation}
  \label{eq:err-residual-relation} 
  \R_{\G}(\x) = \J_{\f}(\xxi) \, \E(\x) ,
\end{equation}
where $\J_\f(\xxi)$ is the forward transformation Jacobian evaluated at
$\xxi$, which lies between $\x + \G_*(\x)$ and $\x + \Ghat(\x)$.  When
$\E(\x)$ is small, $\J_{\f}(\xxi)$ can be numerically approximated by
$\J_{\f}(\x + \Ghat(\x))$.  The Jacobian $\J_{\f}$ is spatially variant
over $\Omega$, except in some special cases.
Provided with a forward DVF $\F$, we will rely on the IC residual $\R_{\G}$
and relationship \cref{eq:err-residual-relation} to improve upon the
inverse estimate $\Ghat$ via an iterative process.

By \cref{eq:consistency-condition-F}, one shall also consider the other IC
residual,
\begin{equation}
  \label{eq:residual_f}
  \RF(\x') = \F(\x') + \Ghat( \x' + \F(\x') ) .
\end{equation}
We omit a rigorous analysis of the differential relationship between $\RF$
and the estimation error $\E$.  In a nutshell, the residual is spatially
related to the error through the following mapping:
\begin{equation}
  \label{eq:ru-of-target-domain}
  \R_{\F}( \x + \Ghat(\x) ) = \E( \x + \R_{\G}(\x) ) .
\end{equation}
The residual $\R_{\G}$ in \cref{eq:ru-of-target-domain} can be made
sufficiently small if the iteration converges.
We will use $\R_{\F}(\x + \Ghat(\x))$ in addition to $\R_{\G}(\x)$ to
quantitatively assess inverse DVF estimates, $\Ghat$.

\subsubsection{Non-translational displacement component (NTDC)}
\label{subsub:NTDC}

We introduce the decomposition of a DVF into translational and
non-translational components, to elucidate the relationship between
inversion errors and  IC residuals.
By \cref{eq:err-residual-relation}, the estimate error is related to the IC
residual via $\J_{\f}$. 
The transformation Jacobian $\J_{\f}$ is the displacement
Jacobian $\J_{\F}$ shifted by the identity:
\begin{equation}
  \label{eq:jacobian-transformation-displacement}
  \J_{\f} = \I + \J_{\F}. 
\end{equation}
When $\J_{\F} = \mbs{0}$, then $\J_{\f}=\I$ and the residual $\R_{\G}$ is
equal to the estimate error $\E$.  The inverse displacement $\G$ can then
be obtained immediately by adding the residual to the current estimate,
regardless of the direction and magnitude of displacement $\F$.  In this
case, we consider the corresponding displacement $\F$ as translational.
When $\J_{\F} \neq \mbs{0}$, there is a non-translational component in the
displacement.  The non-translational displacement component (NTDC) is
responsible for the nontrivial, non-transparent relationship between the
estimate error and the IC residual.

The NTDC Jacobian is used in DVF characterization, pre-inversion
convergence analysis and prediction, and adaptive feedback control design.
Conceptually, we decompose the displacement into translational $\F_{\text{t}}$ 
and non-translational $\F_{\text{nt}}$ components: $\F(\x) = \F_{\text{t}} + \F_{\text{nt}}(\x)$.
These components are identified by their respective contributions to the
Jacobian $\J_{\F}$. Only non-translational components contribute to the
Jacobian, i.e., $\J_{\F_{\text{t}}} = \mbs{0}$ and
\mbox{$\J_{\F_{\text{nt}}} = \J_{\F}$}. 
We may therefore refer to the NTDC Jacobian as the displacement Jacobian $\J_{\F}$,
with the understanding that the translational component plays no part in
it.
We will introduce in \cref{subsec:spectral-NTDC-characterization} spectral
measures for characterizing the NTDC, and provide an explicit criterion for 
considering the NTDC as non-small.

% ---------- NTDC-adaptive iteration
%
\subsection{NTDC-adaptive iteration}
\label{sec:iteration}

% ----- residual feedback framework

\subsubsection{Active feedback control framework}
\label{sec:feedback-control}

We introduce a family of fixed-point iterations for DVF inversion, using
the IC residual $\R_{\G}$ as feedback.  
In the rest of this section, we denote the IC residual simply by $\R$.
Feedback control is exercised to suppress the estimate error, based on
error propagation analysis and~\cref{eq:err-residual-relation}.
Assume a forward DVF $\F$ is provided over $\Omega$.  At step
$k = 0,1,2,\ldots$, we compute the residual $\R_k$ associated with the
current estimate $\G_k$ and get the next estimate by
\begin{equation}
  \label{eq:fpim-general}
  \G_{k+1} (\x)  =
  \G_{k} (\x) - \B_k(\x) \, \R_{k}(\x) , 
\end{equation}
where the term $\B_{k}(\x)\, \R_{k}(\x)$ is the modulated residual, and
$\B_k(\x)$ is a $3 \times 3$ feedback control matrix associated with
$\x \in \Omega$. 
In this paper, we consider the control mechanism in its simplest form:
$\B_k(\x)$ is an isotropic scaling matrix, $\B_k(\x) = (1-\mu_k(\x))
\I$. Iteration~\cref{eq:fpim-general} then takes the form
\begin{equation}
  \label{eq:fpim-scaling-control}
  \G_{k+1} (\x)  =
  \G_{k} (\x) - (1 - \mu_{k}(\x)) \, \R_{k}(\x). 
\end{equation}
We refer to $\mu_{k}(\x)$ as the feedback control parameter. 
In what follows, we will introduce three particular adaptive control
schemes: constant parameter value over the domain and throughout the
iteration, alternating values between iteration steps, and spatially
variant values.

When the control parameter is spatially uniform, i.e., it does not vary
with $\x$, iteration~\cref{eq:fpim-scaling-control} becomes
\begin{equation}
  \label{eq:fpim-single-parameter}
  \G_{k+1} (\x)  = 
  \G_{k} (\x)
  - (1 - \mu_k) \, \R_{k}(\x). 
\end{equation} 
Control is stationary if $\mu_k = \mu$ for some constant $\mu$, and
non-stationary otherwise.  The iterations with constant-value control can
be further divided into non-adaptive (pre-fixed constants) and adaptive
ones.
The two precursor algorithms for DVF
inversion~\cite{christensen2001consistent,chen2008simple} mentioned in
\cref{sec:introduction} both adhere to the form of
iteration~\cref{eq:fpim-single-parameter} with constant non-adaptive
(pre-fixed) control values.
Specifically, $\mu_{k}=0.5$ yields the algorithm of Christensen and Johnson~\cite{christensen2001consistent},
and $\mu_k=0$ yields the algorithm of Chen \etal.~\cite{chen2008simple}.

% ----- spectral analysis

\subsubsection{Spectral analysis}
\label{sec:spectral-analysis}

We provide analytical apparatus for designing residual feedback control
in the simple form of \cref{eq:fpim-scaling-control} in order to
guarantee convergence and improve convergence speed.
By \cref{eq:err-residual-relation} and \cref{eq:fpim-scaling-control},
inverse estimate errors \cref{eq:iteratate-error} propagate throughout
the iteration by the equation
\begin{subequations} 
  \label{eq:Ek-propagation} 
  \begin{align}
    \label{eq:Ek-propagation-E}
    \E_{k+1}(\x) 
    &= \EPM_k(\x; \mu) \, \E_{k}(\x) , 
    \\
    \label{eq:Ek-propagation-P}
    \EPM_k(\x; \mu) 
    &= \I - (1 - \mu) \J_{\f} (\xxi_k) ,
 %      = \mu \I - (1 - \mu) \J_{\F}(\xxi_k) ,
  \end{align}
\end{subequations} 
where $\mu$ is short for $\mu_{k}(\x)$, and $\EPM_{k}(\x; \mu)$ is the
one-step error propagation matrix at $\xxi_k$, which lies between
$\x+\G_{\ast}(\x)$ and $\x+\G_{k}(\x)$.  The propagation matrix depends
on the value of $\mu$ and varies during the iteration process.  If
\begin{align} 
  \label{eq:sufficient-convergence-condition} 
  \rho\left( \EPM_k(\x; \mu) \right) \leq \rho_{\sup} , 
  && k = 0, 1, 2, \ldots, 
\end{align} 
for some $\rho_{\sup} \in [0, 1)$, then the iteration converges.  Here,
$\rho(\EPM)$ denotes the spectral radius of the propagation matrix
$\mbs{P}$, i.e., the ratio between successive errors in magnitude.

Consider the special case where the displacement is translational, i.e.,
$\J_{\F}(\x) = \mbs{0}$, $\J_{\f}(\x) = \I$, and
$\EPM_k(\x; \mu) = \mu \I$.  The iteration converges with any
$\mu \in (-1,1)$, and converges faster when $\mu$ is closer to $0$. At
the mid-range value, $\mu=0$, the iteration renders the inverse DVF in
one step.  We focus our study on deformations with non-translational
components, without excluding the case of translational displacement.

%%  --- to surmount a challenge described below 

Feedback control design based on convergence analysis is challenging.
The point-wise error sequences \cref{eq:Ek-propagation-E}, each associated
with a voxel location $\x$, cover the whole domain $\Omega$.  Although the
sequences depend on the initial guess, a sequence
$\E_k(\x), k = 0,1,2, \ldots$, converges to zero if it can be guaranteed
that the associated scalar sequence $\rho(\EPM_k(\x; \mu))$ at
$\xxi_k(\x)$, $k=0,1,2,\ldots $, is bounded from above below $1$; see
\cref{eq:Ek-propagation-P,eq:sufficient-convergence-condition}.
Even with a fixed value of $\mu$, tracking every scalar sequence of
$\rho(\EPM_k(\x;\mu))$ at unknown and spatially varying mean-value
locations $\xxi_k(\x)$ is implausible. We surmount this challenge by the
following novel approach.

We take a covering-and-partitioning approach.
Spec\-ifically, with any value of the control parameter $\mu$, we consider
the deformation Jacobian everywhere over $\Omega$.  We define the
infinitesimal contraction matrix at all $\x \in \Omega$:
\begin{equation}
  \label{eq:infinitesimal-Q}
  \ICM(\x; \mu)
  = \I - (1 - \mu) \J_{\f} (\x)
  = \mu \I - (1-\mu) \J_{\F}(\x) .
\end{equation}
Any particular value of the control parameter $\mu$ partitions $\Omega$
into a contraction region,
\begin{equation} 
  \label{eq:contraction-region} 
  \Omega_{\text{c}}(\mu) = 
  \setcond{ \x }{ 
    \rho\left( \ICM(\x; \mu) \right) < 1 , \; \x \in \Omega
  } ,
\end{equation} 
and its complement $\Omega-\Omega_{\text{c}}(\mu)$, the non-contraction
region.
The error \cref{eq:Ek-propagation-E} converges to zero if all the
mean-value locations $\xxi_k$ fall within the contraction region.  If
$\Omega_{\text{c}}(\mu) = \Omega$, i.e., the contraction region covers the
entire domain (or, more strictly speaking,
$\sup_{\x \in \Omega} \rho\left( \ICM(\x; \mu) \right) <1$), then the
iteration converges.
By the covering-and-partitioning approach, we reduce feedback control
design to finding the control parameter values that yield the largest
contraction region over $\Omega$.

Consider the special value $\mu = 0$, with which
the spectral radius
$\rho( \ICM(\x; 0 ) ) = \rho( \J_{\F}(\x) )$.  If
$\rho\left( \J_{\F}(\x) \right ) < 1$, then $\x$ lies in the contraction
region. Otherwise, $\x$ is in the non-contraction region,
\begin{equation} 
  \label{eq:nonsmall-NTDC} 
  \rho\left( \J_{\F}(\x) \right) \geq 1, 
\end{equation} 
and a positive feedback control value ($\mu>0$) is warranted.  Formally, we
define by~\cref{eq:nonsmall-NTDC} the concept of a non-small NTDC at $\x$:
the NTDCs are considered non-small where the spectral radius of the
displacement Jacobian is equal to or greater than $1$.  Whenever the
mean-value location $\xxi_k$, at which $\J_{\F}(\xxi_k)=\EPM_{k}(\x;0)$,
falls in a non-small-NTDC region, the error in the next iterate is
magnified if active feedback control is not applied.
%
%%  ... describe contracting conditions in terms of eigenvalues 
%
We describe in the rest of this section how to materialize the sufficient
condition $\Omega_{\text{c}}(\mu) = \Omega$ for guaranteed convergence of
all iterative sequences.
We translate the condition into expressions that relate the parameter $\mu$
to the eigenvalues of the transformation Jacobian,
$\lambda_j(\x) = \lambda_j( \J_{\f}(\x) )$, $j=1,2,3$.

% real-eigenvalue case 
%
Consider first the case where the eigenvalues are all real and positive
over $\Omega$.
The condition $\Omega_{\text{c}}(\mu) = \Omega$ is achieved by any specific
value of $\mu$ in the range
\begin{equation}
  \label{eq:mu-range-real-case} 
  \max \left\{ -1, 1 - 2 \min_{\x \in \Omega, \, j=1,2,3} 
  \frac{ 1  }{ \lambda_j(\x)  }\right\}  < \mu < 1.
\end{equation} 
When $\mu$ is negative, $(1-\mu)>1$ and the impact of $\J_{\F}$ on error
contraction is over-relaxed; see~\cref{eq:infinitesimal-Q}.  
When $\mu < -1$, a small spectral radius of $\ICM(\x; \mu) $ is the result of
algebraic cancellation.  To avoid severe cancellation and subsequent
instability in numerical computations, we bound the control parameter value
from below by $-1$.

% complex-eigenvalue case
%
Consider next the presence of complex eigenvalues, which are prevalent in
DVFs associated with patient CT images (see
\cref{tab:patient-dataset-characterization}
in~\cref{sec:patient-data-characterization}).  This should not be a
surprise; the eigenvalues of a plane rotation matrix are complex, for
instance.  Complex eigenvalues of a real-valued matrix exist in conjugate
pairs.
For any fixed value $\mu < 1$, the error contraction condition
$ \rho(\ICM(\x; \mu)) < 1 $ becomes
\begin{align}
  \label{eq:contraction-condition-complex-forward}
  2 \real\left( \lambda_j(\x) \right) > (1-\mu) |\lambda_j(\x)|^2 , 
  &&
  j = 1, 2, 3, 
\end{align}
for $\x \in \Omega$. 
This condition immediately rejects singular Jacobians, and gives rise to
the local feasible parameter range
\begin{equation} 
\label{eq:local-range-of-mu} 
  \max \left\{ -1, 1 - 2 \reciprocalGap(\x) \right\} < \mu < 1, 
\end{equation} 
where
\begin{equation} 
  \label{eq:positive-gamma} 
  \reciprocalGap(\x)
  = \min_{j} \frac{ 
    \real\left( \lambda_j(\x) \right)
  }{
    |\lambda_j(\x)|^2
  }  
  = \min_{j} \real\left( \lambda_j^{-1}(\x) \right) 
  > 0. 
\end{equation} 
We refer to \cref{eq:positive-gamma} as the controllability condition.  It
has two equivalent expressions: one in terms of the eigenvalues, and one in
terms of the reciprocal eigenvalues.

A few remarks about condition \cref{eq:positive-gamma} are in order.  The
condition rejects any Jacobian with eigenvalues on the imaginary axis or in
the left half of the complex plane, in which case the determinant of the
Jacobian may be negative, zero, or even positive.
The condition is necessary and sufficient for the parameter range
\cref{eq:local-range-of-mu} to be non-empty, and hence for feasible
values of $\mu$ to exist.
% 
% add after finishing the review response letter 
% 
One can locate the few cases where the condition is violated.  Violations
of the controllability condition anywhere in the image domain most likely
manifest as artifacts introduced by the forward DVF generation process.  In
other words, this condition shall be recognized as a rule for local
regularization, with respect to DVF inversion, not a limitation imposed by
the control mechanism \cref{eq:fpim-scaling-control}.

% ----- optimal control parameter value

\subsubsection{Maximal error suppression}
\label{subsub:optimal-parameters}

% FIGURE: P-matrix spectrum - optimal scaling ========== [FINAL VERSION]
% 
\ifdefined\medphys\ifdefined\review\else

\begin{figure*}
  \centering
  \begin{minipage}{\widetwocol}
    \vspace*{-\valignImgSkip}
    \centering
    %
    % ---------- 2-POINT DISC (COMPLEX)
    % 
    \subfloat[\label{fig:propagation-spectrum-optimal-mu-complex} 
    case C1]{%
      \includegraphics[width=0.48\linewidth]{%
        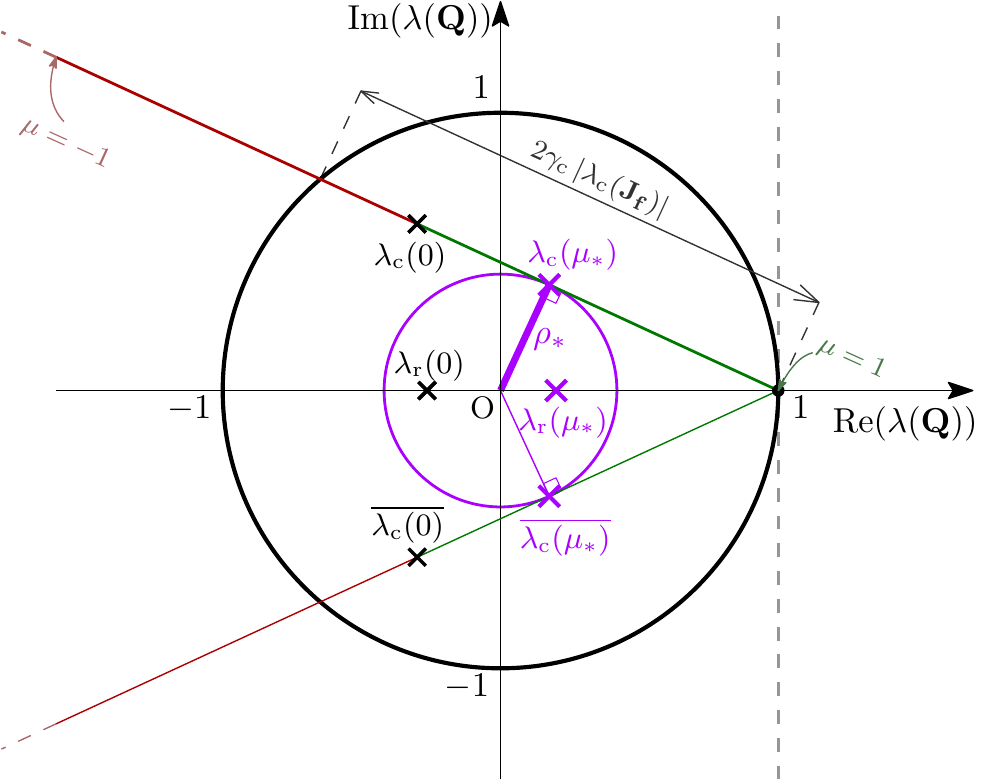}}%
    \hfill
    % 
    % ---------- 3-POINT DISC (COMPLEX + REAL)
    % 
    \subfloat[\label{fig:propagation-spectrum-optimal-mu-3point}
    case C2]{%
      \includegraphics[width=0.48\linewidth]{%
        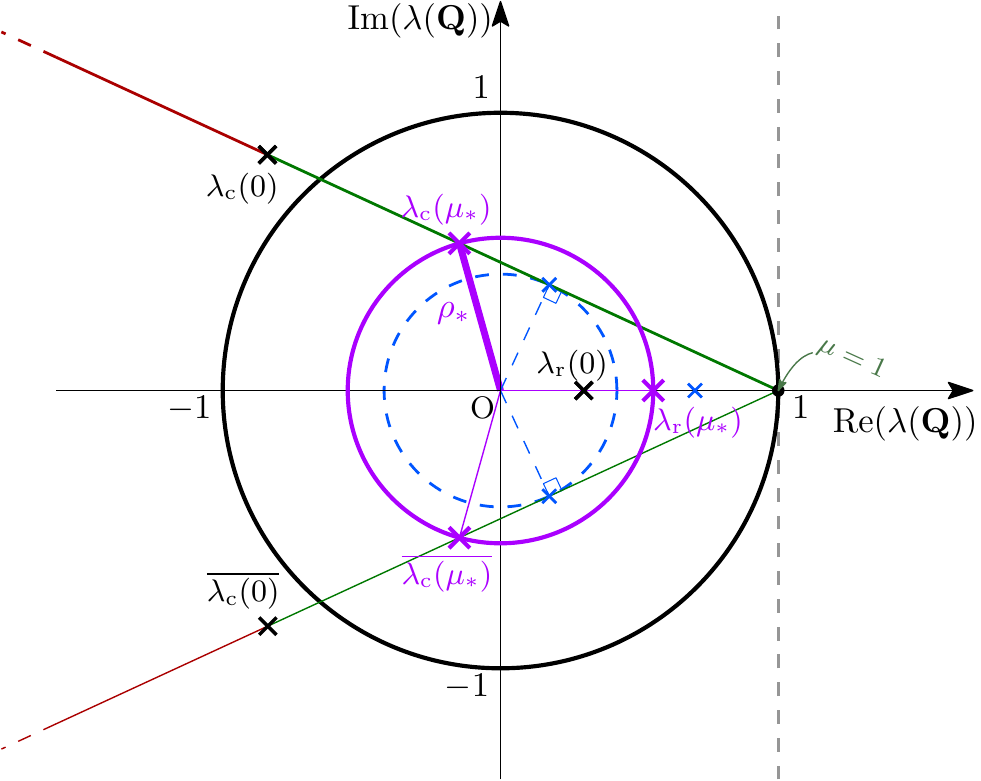}}%
  \end{minipage}
  \caption{%
    Geometric outline of \cref{eq:mu-star-complex-cases,%
      eq:rho-star-complex-case}.  See the explanation in
    \cref{subsub:optimal-parameters}.  The notation $\lambda(\mu)$ is short
    for $\lambda(\ICM(\mu))$.%
  }
  \label{fig:propagation-spectrum-optimal-mu}
\end{figure*}

\fi\fi

We consider now how to determine, over an infinitesimal neighborhood of
$\x$, the control parameter value $\mu_{\ast}(\x)$ such that estimate
errors are suppressed as much as possible.  Specifically, we minimize the
spectral radius of the local infinitesimal contraction matrix, i.e.,
$\mu_{\ast}(\x) = \argmin_{\mu} \rho(\ICM(\x;\mu))$.  By
\cref{eq:infinitesimal-Q}, the spectrum of $\ICM(\x;\mu)$ is that of the
displacement Jacobian $\J_{\F}(\x)$ scaled by $(\mu-1)$ and shifted by $1$.
Since the analysis concerns the variation of $\rho(\ICM(\x;\mu))$ with
parameter $\mu$ at any fixed point $\x$, we simplify the notation
$\ICM(\x;\mu)$ to $\ICM(\mu)$, and similarly with notation for other
related quantities, omitting $\x$ for the rest of this sub-section.  To
further simplify the expressions below, we use the eigenvalues of
$\J_{\f}$, which are the NTDC eigenvalues shifted by $1$; see
\cref{eq:jacobian-transformation-displacement}.

Assume the controllability condition~\cref{eq:positive-gamma} holds, i.e.,
$\reciprocalGap > 0$; otherwise, there exists no feasible parameter value.
When the three eigenvalues of $\J_{\f}$ are all real and positive, it is
straightforward to verify that
\begin{subequations}
  \label{eq:opt-control-real-case}
  \begin{gather} 
    \label{eq:mu-star-real-case}
    \mu_{\ast} = 1 - \frac{2}{ \lambda_{\max} + \lambda_{\min} } ,
    \qquad\qquad\qquad \text{\emph{(case R)}}
    \intertext{and} 
    \label{eq:rho-star-real-case}
    \rho_{\ast} = \min_{\mu} \rho( \ICM(\mu) ) =
    \frac{ \lambda_{\max} - \lambda_{\min} }
    { \lambda_{\max} + \lambda_{\min} } < 1.
  \end{gather}
\end{subequations}
In contrast, in the presence of complex eigenvalues, which appear in a
conjugate pair, $\lambda_{\text{c}}$ and $\overline{\lambda_{\text{c}}}$,
the geometric positions of the conjugate pair relative to the real
eigenvalue $\lambda_{\text{r}}$ are not maintained after scaling and
shifting.
Define
\begin{equation} 
  \label{eq:gamma-c} 
  \reciprocalGap_{\text{c}} = 
  \frac{ \real( \lambda_{\text{c}} ) }{ \abs{\lambda_{\text{c}}}^2 } 
  = \real( \lambda_{\text{c}}^{-1}) ;
\end{equation} 
then,
$\reciprocalGap = \min\{ \reciprocalGap_{\text{c}}, \lambda_{\text{r}}^{-1}
\}$.
By our analysis, the geometric relationship between the complex and real
eigenvalues can be put into two mutually exclusive cases: case \emph{C1} if
$\abs{ 1 - \reciprocalGap_{\text{c}} \lambda_{\text{r}} } \leq \abs{ 1 -
\reciprocalGap_{\text{c}} \lambda_{\text{c}} }$, and case \emph{C2} otherwise.
The optimal control parameter value for each case is
\begin{subequations}
  \label{eq:opt-control-complex-cases}
  \begin{gather}
    \label{eq:mu-star-complex-cases}
    \mu_{\ast} = 
    \left\{
      \begin{aligned}[c]
        & 1 - \reciprocalGap_{\text{c}} ,
        & \text{\emph{(case C1)}}
        \\
        & 1 - 
        \frac{ \real( \lambda_{\text{c}} - \lambda_{\text{r}} ) }
        { |\lambda_{\text{c}}| - \lambda_{\text{r}} }
        \frac{ 2 }
        { |\lambda_{\text{c}}| + \lambda_{\text{r}} } .
        & \text{\emph{(case C2)}}
      \end{aligned}
    \right.
    \intertext{The optimal spectral radius is, for both cases,}
    \label{eq:rho-star-complex-case}
    \rho_{\ast} = \min_{\mu} \rho( \ICM(\mu) ) 
    = \abs{ 1 - (1 - \mu_{\ast}) \lambda_{\text{c}} } < 1 . 
  \end{gather}
\end{subequations}

% FIGURE: P-matrix spectrum - optimal scaling ========== [DRAFT/REVIEW VERSION]
% 
\ifdefined\medphys\ifdefined\review

\begin{figure*}
  \centering
  \begin{minipage}{\widetwocol}
    \vspace*{-\valignImgSkip}
    \centering
    %
    % ---------- 2-POINT DISC (COMPLEX)
    % 
    \subfloat[\label{fig:propagation-spectrum-optimal-mu-complex} 
    case C1]{%
      \includegraphics[width=0.48\linewidth]{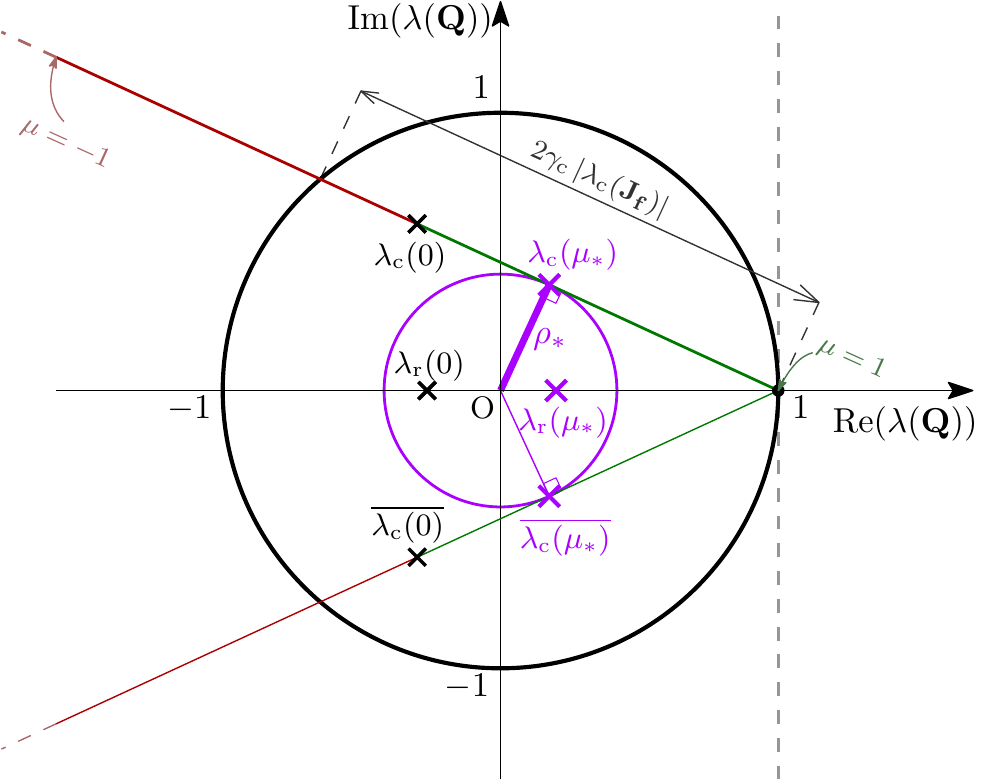}}%
    \hfill
    % 
    % ---------- 3-POINT DISC (COMPLEX + REAL)
    % 
    \subfloat[\label{fig:propagation-spectrum-optimal-mu-3point}
    case C2]{%
      \includegraphics[width=0.48\linewidth]{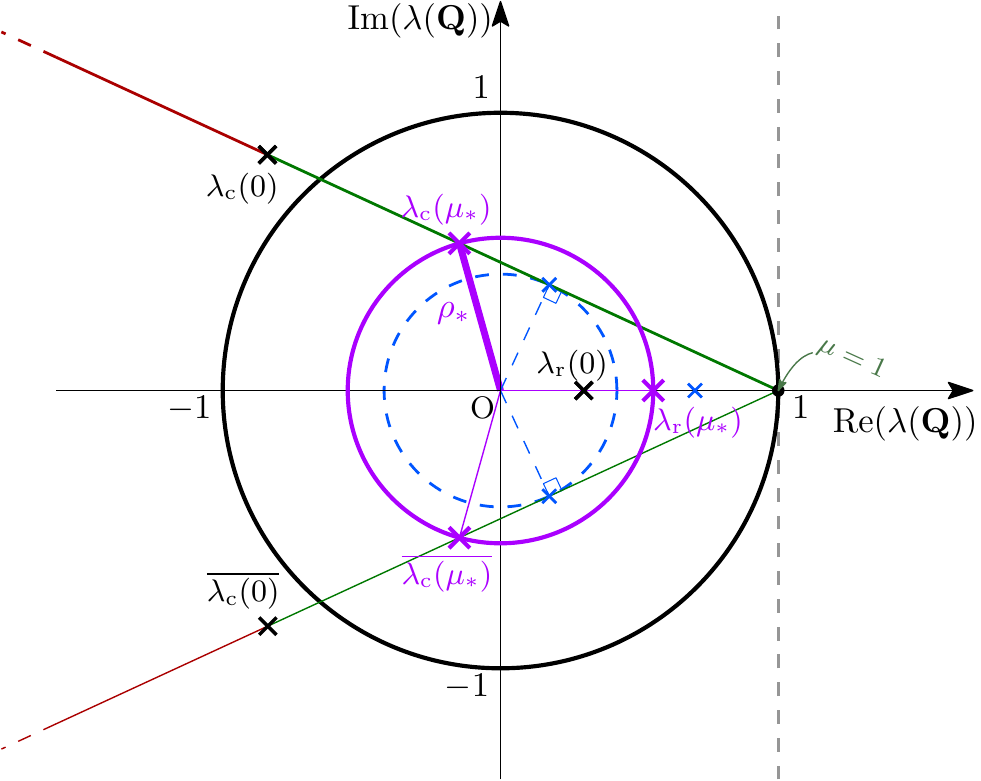}}%
  \end{minipage}
  \caption{%
    Geometric outline of \cref{eq:mu-star-complex-cases,%
      eq:rho-star-complex-case}.  See the explanation in
    \cref{subsub:optimal-parameters}.  The notation $\lambda(\mu)$ is short
    for $\lambda(\ICM(\mu))$.%
  }
  \label{fig:propagation-spectrum-optimal-mu}
\end{figure*}

\fi\else

\begin{figure*}
  \centering
  \begin{minipage}{\widetwocol}
    \vspace*{-\valignImgSkip}
    \centering
    %
    % ---------- 2-POINT DISC (COMPLEX)
    % 
    \subfloat[\label{fig:propagation-spectrum-optimal-mu-complex} 
    case C1]{%
      \includegraphics[width=0.48\linewidth]{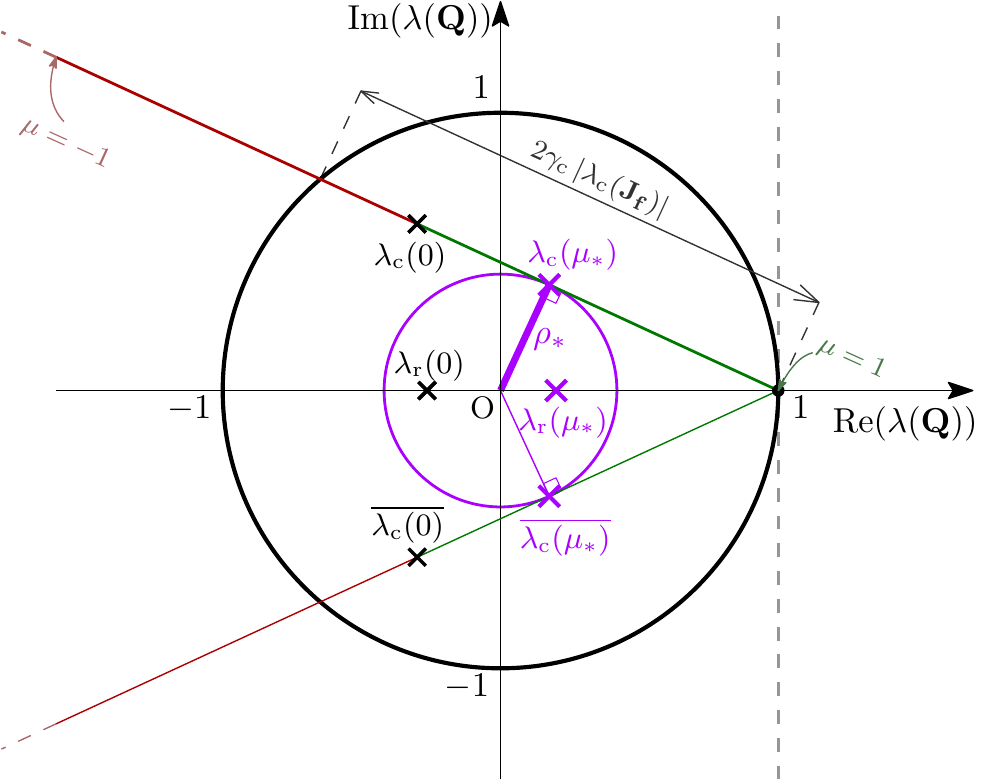}}%
    \hfill
    % 
    % ---------- 3-POINT DISC (COMPLEX + REAL)
    % 
    \subfloat[\label{fig:propagation-spectrum-optimal-mu-3point}
    case C2]{%
      \includegraphics[width=0.48\linewidth]{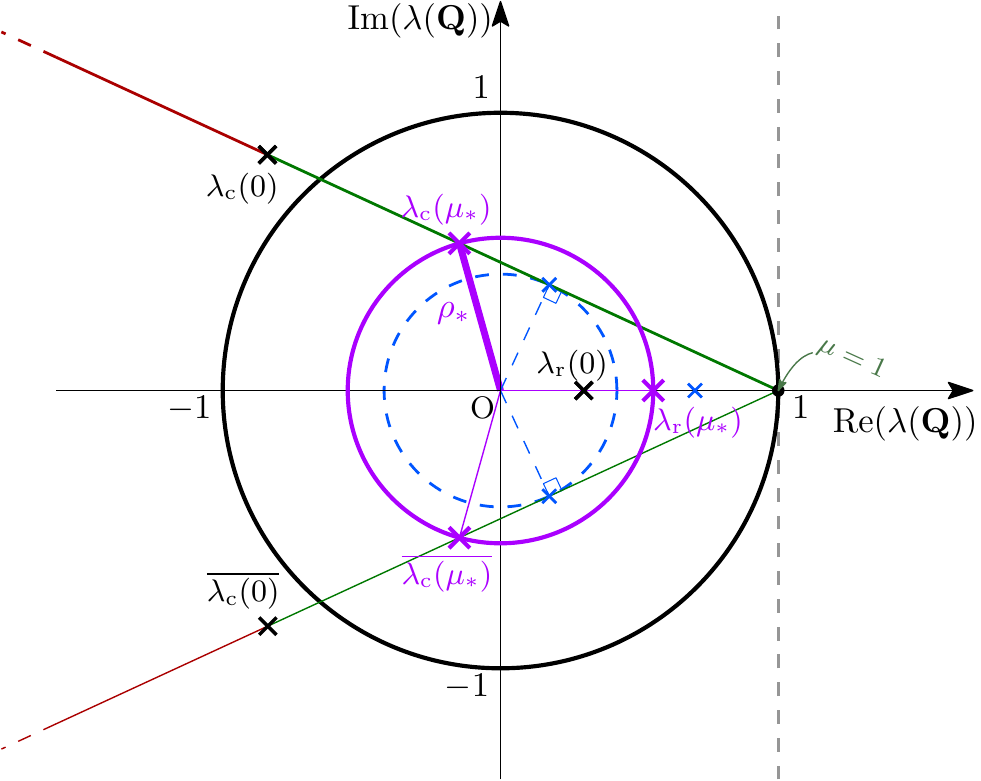}}%
  \end{minipage}
  \caption{%
    Geometric outline of \cref{eq:mu-star-complex-cases,%
      eq:rho-star-complex-case}.  See the explanation in
    \cref{subsub:optimal-parameters}.  The notation $\lambda(\mu)$ is short
    for $\lambda(\ICM(\mu))$.%
  }
  \label{fig:propagation-spectrum-optimal-mu}
\end{figure*}

\fi

The proof of \cref{eq:opt-control-complex-cases} could be lengthy and
tedious in words.  We provide instead a geometric explanation with two
drawings in \cref{fig:propagation-spectrum-optimal-mu}; one for each
case. The drawings illustrate how the eigenvalues of $\J_{\f}$ are scaled
and shifted onto those of $\ICM(\mu)$ in the complex plane, and how to
locate the optimal scaling value.
In each drawing, we locate first the complex eigenvalues of
$\ICM(0) = -\J_{\F}$.  We connect them by straight lines to $(1,0)$, where
the multiple eigenvalues of $\ICM(1)$ reside.  For any $\mu$, the conjugate
pair $\lambda_{\text{c}}(\ICM(\mu))$ lie on the two conjugate lines.  In
case \emph{C1}, the optimal value $\mu_{\ast}$ maps $\lambda_\text{c}$ onto
the very point at which these lines meet the tangential circle centered at
the origin; this gives the minimal spectral radius $\rho_{\ast}$, since the
real eigenvalue by the same scaling and shifting falls inside the circle.
In case \emph{C2}, the real eigenvalue is outside the circle (dashed blue
in \cref{fig:propagation-spectrum-optimal-mu-3point}).  We rescale the
eigenvalues until all three eigenvalues are on the same circle, which is of
the minimal spectral radius.

% ----- parameter value settings

\subsubsection{Control parameter schemes}
\label{sub:parameter-value-selection}

% mid-range control parameter value

\paragraph{Mid-range parameter value.}
\label{subsub:mid-range-parameter}

We present an adaptive scheme for determining a constant value, including
its existence condition, for the control parameter such that uniform
convergence is guaranteed over a sub-region or neighborhood
$\nhood \subseteq \Omega$.
First, we extend point-wise quantities to region-wise quantities.
Specifically, let
$ \inlineMathDisplay \reciprocalGap(\nhood) = \min_{\x \in \nhood}
\reciprocalGap(\x)$.
Computationally, $\reciprocalGap(\nhood) $ can be obtained easily with a
minimum filter.  If $\reciprocalGap(\nhood) > 0$, then the control parameter range,
$\left( \max\{-1, 1-2\reciprocalGap(\nhood)\}, 1 \right)$, for uniform
convergence over $\nhood$ is non-empty.  Any value in this range can serve
as a constant value for the control parameter, guaranteeing convergence.
We may use in particular the mid-range value.  When
$1 - 2\reciprocalGap > -1$, the mid-range value over $\nhood$ is simply
\begin{equation}
  \label{eq:mid-range-value} 
  \mu_{\text{m}}(\nhood) = 1 - \reciprocalGap(\nhood) ,
\end{equation}
and can be easily determined. If $\gamma(\nhood)=1$, then
$ \mu_{\text{m}}(\nhood) = 0 $; if $\gamma(\nhood)=0.5$, then
$ \mu_{\text{m}}(\nhood) = 0.5 $.  The mid-range scheme
\cref{eq:mid-range-value} is adaptive to any DVF with $\gamma > 0$ over
$\nhood$. It also leads to the next control scheme. 

% alternating control parameter values

\paragraph{Alternating parameter values.}
\label{subsec:alternating}

Convergence can be made faster, within the parameter range for uniform
convergence, by a simple modification: allowing the control parameter to
take two alternating values.
The idea is to exploit the non\--uniform spectral structure of the DVF while keeping
the control spatially uniform.
As we shall show in \cref{sec:experiments} with DVFs from patient images,
the local mid-range values, or locally optimal values, over the entire
image domain $\Omega$ may be grouped into two sub-ranges in the convergence
parameter range; one at the lower end, and one at the higher end. 
A simple alternating scheme is to use a value $\mu_{\rm e}$ at even
steps and another value $\mu_{\rm o}$ at odd steps.  Convergence 
can be analyzed via two-step error propagation, 
\begin{equation}
  \label{eq:propagation-alternating}
  \E_{2(k+1)}(\x) = 
  \underbrace{
    \EPM_{2k+1}(\x; \mu_{\text{o}}) \, 
    \EPM_{2k}(\x; \mu_{\text{e}})
  }_{\EPM_{\text{oe}}(\x)}% \, 
  \E_{2k}(\x) ,
\end{equation}
where $\EPM_{k'}(\x;\mu) = \I - (1 - \mu_{k'}) \J_{\f}(\xxi_{k'})$ and
$\xxi_{k'}$ lies between $\x + \G_{*}(\x)$ and $\x + \G_{k'}(\x)$, at
iteration step $k'$.
The spectral radius $\rho_{\text{oe}}$ of the two-step propagation
matrix %$\EPM_{\text{oe}}$ as in~\cref{eq:propagation-alternating} 
can be bounded from above by
$ \rho_{\text{o}} \rho_{\text{e}}, $
where $\rho_{\text{o}}$ and $\rho_{\text{e}}$ are the spectral radii of the
odd- and even-step propagation matrices ($\EPM_{2k+1}(\x;\mu_{\text{o}})$
and $\EPM_{2k}(\x;\mu_{\text{e}})$), respectively.  Thus, the contraction
condition is maintained.  Improvement in convergence speed is due to the
suppression of local errors, which are aggressively suppressed at odd
(even) steps without being enlarged at even (odd) steps.
% 
% adaptive control parameter values

\paragraph{Spatially variant parameter values.}
\label{subsub:adaptive-control}

The non-uni\-form spectral structure of the transformation Jacobian can be
better exploited by letting the control parameter vary spatially with $\x$
over $\Omega$.  This can be achieved by determining the parameter value at
$\x \in \Omega$ by a local neighborhood $\nhood(\x)$, such that the entire
image domain is covered, $\bigcup_{\x} \nhood(\x) = \Omega$.
The neighborhood size need not be greater than the maximal displacement
length, which is known in advance.  The parameter value at $\x$ can be the
mid-range value over $\nhood(\x)$, or the locally optimal value when the
neighborhood is small enough.
The iteration with spatially variant control is essentially non-stationary,
because the value of $\mu_k(\x)$ depends on the location of
$\x + \G_k(\x)$.

This scheme incurs extra but modest cost in two parts.  First, we create a
parameter map once for all iteration steps, by calculating the Jacobians,
eigenvalues, and control parameter values over $\Omega$.  This
preprocessing step takes about the same time in execution as a single
iteration step.  Then, the look-up of parameter values over the domain at
each iteration step takes no more than $5\%$ of the cost for 3D
vector-field interpolation.  The overall cost is outweighed by the gain in
practice; see \cref{sec:results}.

% adaptive control parameter values

\paragraph{Remarks.}
\label{subsub:remarks}
% flatex input: [tex/parameter-value-selection-note.tex]
% parameter-value-selection-note.tex

The mid-range or alternating control value settings may be preferable to
the locally optimal setting in the initial steps of the inversion
iteration.
Both uniform parameter settings are easy to implement and guarantee
convergence over the entire domain with arbitrary initialization.
On the other hand, spatially variant control yields faster convergence, but
is relatively more sensitive to the initial guess, depending on the size of
the local neighborhoods used in determining the control parameter values.
To circumvent this issue of sensitivity, we may use uniform control for a
few initial steps and switch to spatially variant control afterwards.
%
% ---------- Spectral NTDC characterization
%
\subsection{Spectral NTDC characterization}
\label{subsec:spectral-NTDC-characterization}

We discuss in this section how we characterize and evaluate
non-translational displacement components in a given DVF, with respect to
DVF inversion.  The NTDCs over $\Omega$ can be fully described by the
eigenvalues of the displacement Jacobians $\J_{\F}(\x) $ in the complex
plane.  Rather, we employ the following three real-valued scalar functions
for their informative properties.
\begin{inparaenum}[(i)]
\item The determinant of $\J_{\f}(\x)$, which is commonly used for
  deformation
  characterization~\cite{chen2008simple,christensen2001consistent}. The
  transformation $\f$ is invertible if and only if
  $\det{\J_{\f}(\x)} \neq 0$.
\item The spectral radius of $\J_{\F}(\x)$.  Where $\rho(\J_{\F}(\x)) = 0$,
  the DVF is locally translational; where $\rho(\J_{\F}(\x)) \geq 1$, the
  NTDC is non-small.
\item The algebraic control index
  \begin{equation}
    \label{eq:algebraic-index}
    1 - 2 \reciprocalGap(\x) ,
  \end{equation}
  where $\reciprocalGap(\x)$ is defined in \cref{eq:positive-gamma}.
\end{inparaenum}

The algebraic control index is informative in several ways.  First, it
offers an equivalent criterion to \cref{eq:nonsmall-NTDC} 
on whether the NTDC at $\x$ is non-small:
\begin{equation} 
  \label{eq:equivalent-NTDC-tests}
   1- 2\reciprocalGap(\x) > 0 \iff \rho(\J_{\F}(\x)) \geq 1 .
\end{equation}
Second, the necessary and sufficient condition for the existence of
feasible control parameter values is
\begin{equation} 
  \label{eq:existence-condition}
   1- 2\reciprocalGap(\x) < 1 
   \iff 
   \gamma(\x) > 0. 
\end{equation}
That is, the algebraic control index distinguishes non-small NTDCs from
small ones and furthermore tells whether or not a non-small NTDC 
can be put under control by a single-parameter control mechanism \cref{eq:fpim-scaling-control}.
Third, the index is a lower bound to all feasible control values; see
\cref{eq:local-range-of-mu}. 
It can be employed directly for locating the mid-range parameter value
(\cref{subsub:mid-range-parameter}), as well as for selecting alternating
values (\cref{subsec:alternating}).

The algebraic control index \cref{eq:algebraic-index} falls a little
short of replacing entirely the roles of the other two measures, in two
particular circumstances.
When $\det{\J_{\f}(\x)} = 0$, the transformation is locally singular;
the controllability condition is violated in this case and the index 
is not well-defined.
When $\rho(\J_{\F}(\x)) = 0$, the displacement at $\x$ is locally
translational; the algebraic control index is equal to $-1$ in this case,
but the converse is not necessarily true.
We use all three measures for data assessment in
\cref{sec:patient-data-characterization}.

% ==================== EXPERIMENTS
%
\section{Experiments}
\label{sec:experiments}

We report experimental results on numerical DVF inversion, using clinical
and analytical data.
In this section, we describe the experimental set-up and present 
characterization measures of the clinical DVF data.
Results with different inversion iteration schemes are presented in
\cref{sec:results} with the clinical DVF data, and in
\cref{sec:appendix-casestudy-analyticaldvf} with the analytical DVF data.

% ---------- Datasets
%
\subsection{Dataset description}
\label{sec:datasets}

\begin{figure}
  \centering
  \begin{minipage}{\wideonecol}
    \centering
    % 
    % ==================== SLICE LABELS
    % 
    \begin{minipage}[t]{\dvfThirdAxial}
      \centering
      axial
    \end{minipage}%
    \hfill
    \begin{minipage}[t]{\dvfThirdCoronal}
      \centering
      coronal
    \end{minipage}%
    \hfill
    \begin{minipage}[t]{\dvfThirdSagittal}
      \centering
      sagittal
    \end{minipage}%
    \\
    %  
    % ========== DISPLACEMENT QUIVER-PLOTS
    % 
    \begin{minipage}{\dvfThirdAxial}
      \centering
      \includegraphics[width=\linewidth]{%
        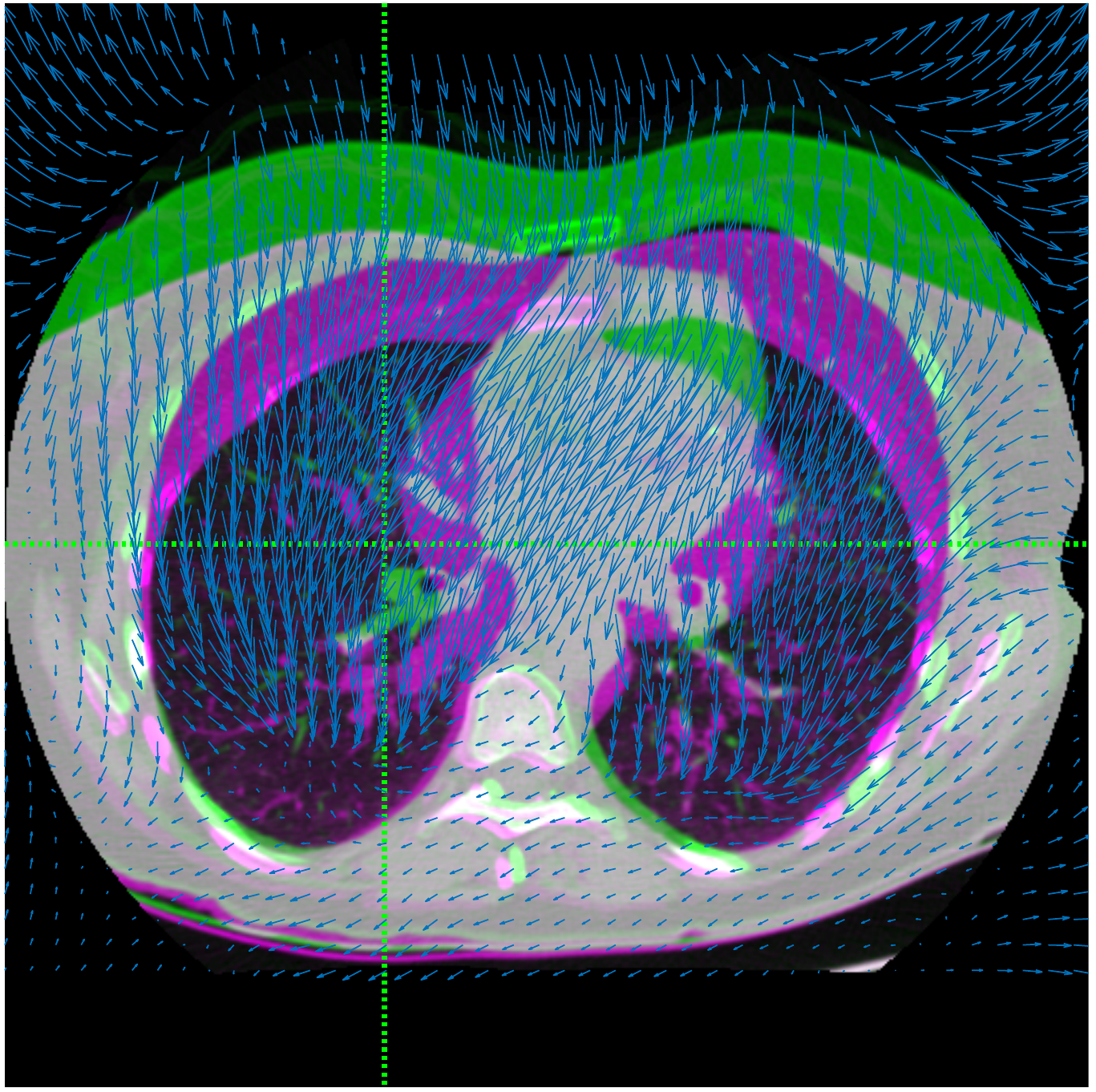}%
    \end{minipage}%
    \hfill
    \begin{minipage}{\dvfThirdCoronal}
      \centering
      \includegraphics[width=\linewidth]{%
        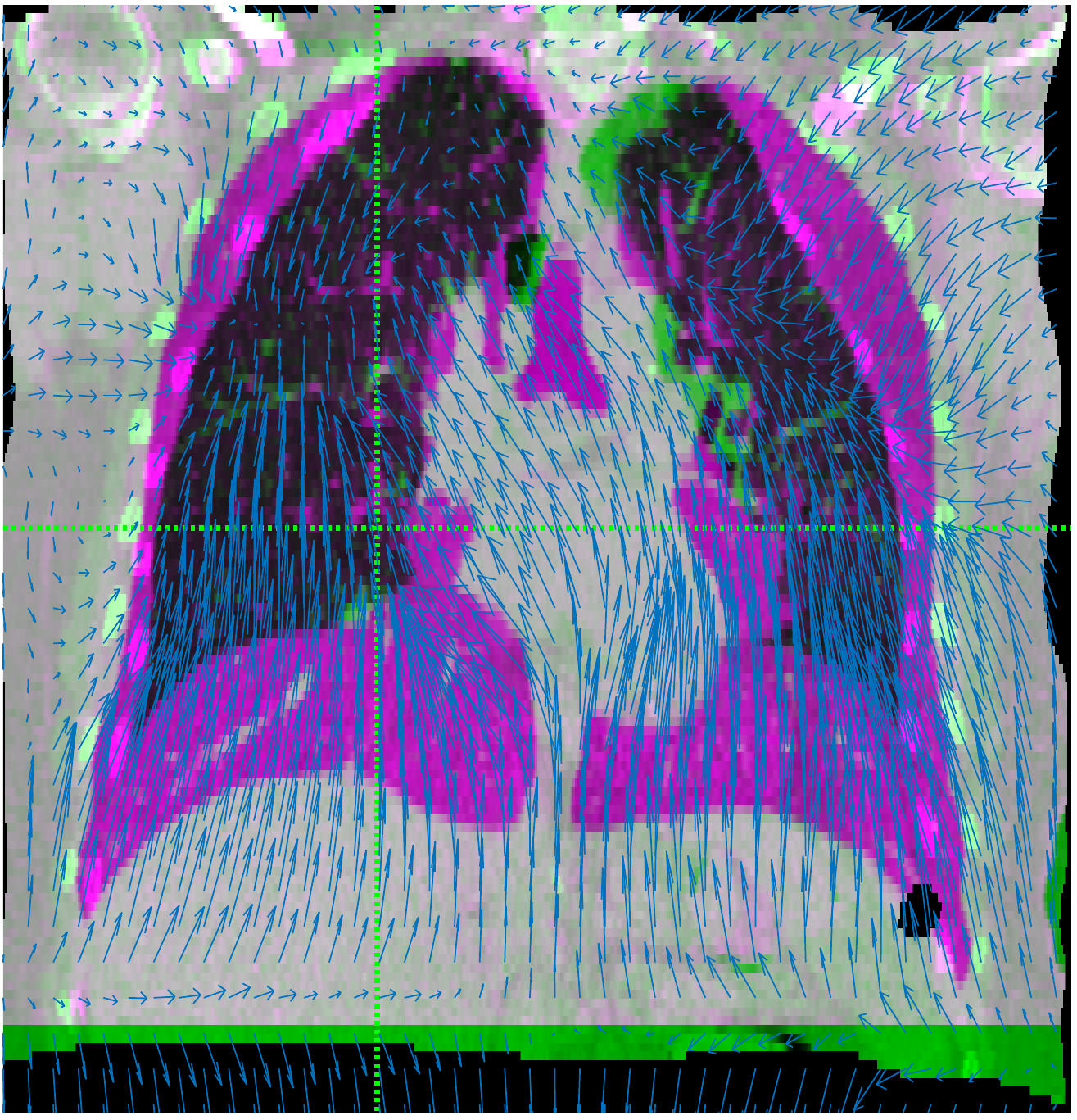}%
    \end{minipage}%
    \hfill
    \begin{minipage}{\dvfThirdSagittal}
      \centering
      \includegraphics[width=\linewidth]{%
        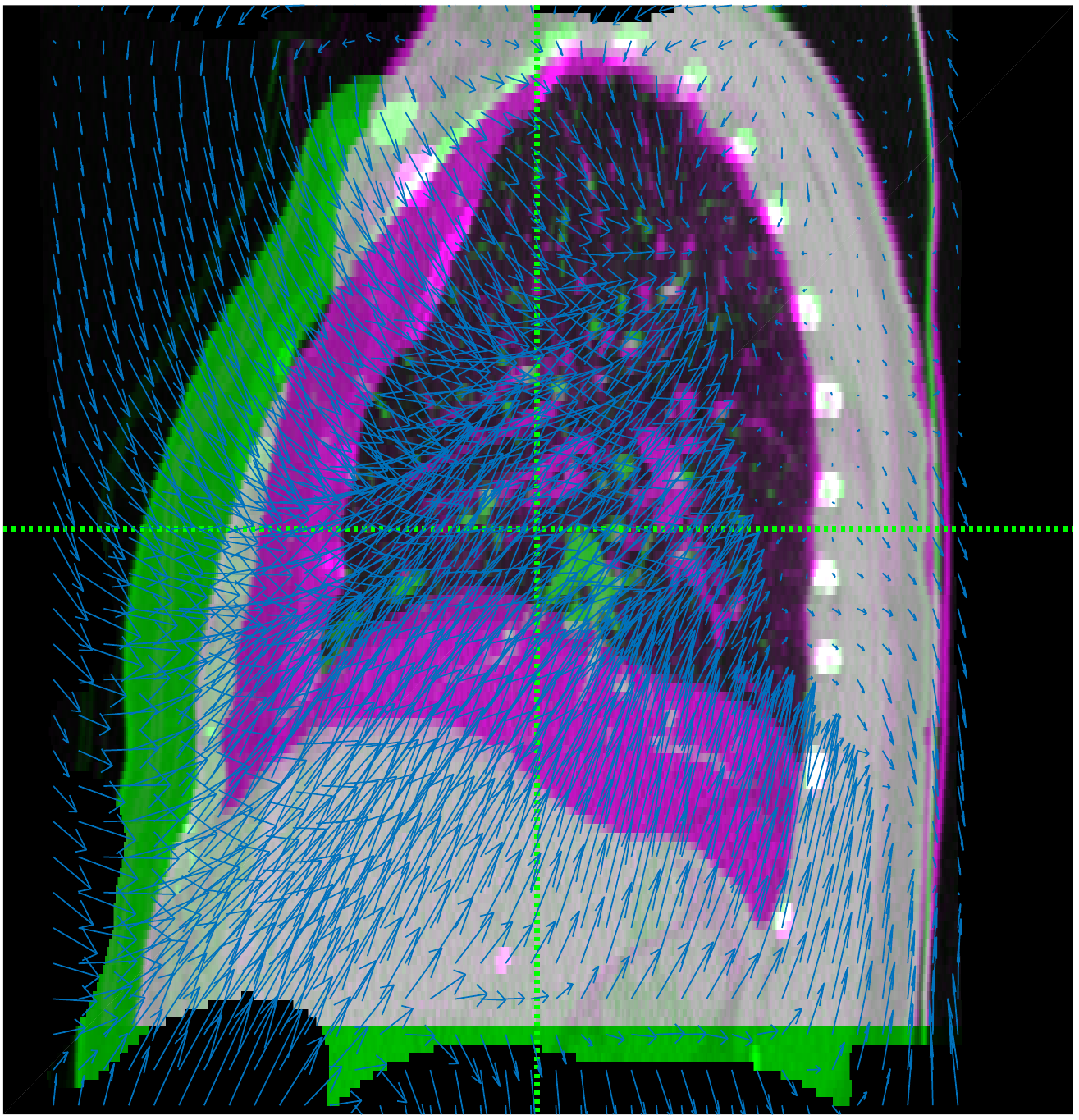}%
    \end{minipage}%
  \end{minipage}%
  % 
  % 
  % ==================== CAPTION
  % 
  \caption{%
    The forward DVF with patient COPD4, visualized by quiver-plots in
    axial, coronal, and sagittal slices, against an overlay of the
    reference CT image (EE, magenta) and target image (EI, green).
    The displayed quiver is spatially down-sampled by a factor
    of 12 along the LR and AP axes and a factor of 6 along the SI axis.}
  \label{fig:patient-ref-tgt-overlay}
\end{figure}

The experiments presented here are carried out with 6 pairs of thoracic CT
images at end of expiration (EE) and end of inspiration (EI).  The image
data are from the CT image collection available through the website of the
Deformable Image Registration Laboratory (DIR-Lab) at the University of
Texas Medical Branch~\cite{0031-9155-58-9-2861, 0031-9155-55-1-018}.
The data collection includes two sets of patient images, with 10 patients
in each set.  One dataset contains 4DCT images acquired at the University
of Texas MD Anderson Cancer Center as part of radiotherapy planning for
thoracic malignancy treatment~\cite{0031-9155-58-9-2861}. In-slice spatial
resolution is $(\unit{0.96}{\milli\meter})^2$ and slice thickness is
$\unit{2.5}{\milli\meter}$.
The other set contains EE and EI breath-hold CT images, taken from the
COPDGene study archive of the National Heart, Lung, and Blood
Institute~\cite{0031-9155-55-1-018}. In-slice spatial resolution ranges
from $(\unit{0.586}{\milli\meter})^2$ to $(\unit{0.647}{\milli\meter})^2$
and slice thickness is $\unit{2.5}{\milli\meter}$.
We refer to the data associated with each patient by the assigned label in
the DIR-Lab website collections.  We have selected the images of 1 patient
from the 4DCT set and 5 patients from the COPD set.

Our selection was based on variations in displacement and spectral measures
(\cref{subsec:spectral-NTDC-characterization}) of associated DVFs, such
that we present the cases that pose a bigger challenge to DVF inversion.
The forward DVF for each patient was obtained by one-way deformable image
registration with the Velocity software (Varian Medical Systems, Palo Alto,
CA, USA).  The EE image is used as reference (primary) and the EI image as
target (secondary), since the EE image is less susceptible to respiratory
phase binning or motion artifacts.
The 4DCT DVFs exhibit only small NTDCs, posing little challenge to
effective feedback control design, whereas the COPD DVFs exhibit more
diverse spatial and spectral measures.  We select the five COPD DVFs with
the largest amplitudes and variation in displacements to test the inversion
iterations with the more challenging cases.  We also select one DVF as
representative of the 4DCT DVFs, using the same criterion.
In what follows, we will provide measures and results in summaries for each
of the $6$ DVFs.  We will also provide in detail results for COPD4 DVF,
which is the most challenging case among the six DVFs with respect to
inversion, as indicated by the highest control index values in
\cref{tab:patient-dataset-characterization}.
The COPD4 DVF is displayed in \cref{fig:patient-ref-tgt-overlay} via
displacement-vector quivers in axial, coronal, and sagittal slices, over a
magenta-green overlay of the reference and target images.

% ---------- Evaluation measures
%
\subsection{Evaluation measures}
\label{sec:domain-aggregation}

We assess feedback control parameter settings for the DVF inversion
iteration with pre- and post-inversion evaluation measures.  For
pre-inversion evaluation, we use the spectral radii of the infinitesimal
contraction matrix~\cref{eq:infinitesimal-Q} over $\Omega$ to assess the
error contraction area and ratio by each control setting.  For
post-inversion evaluation, we use the two IC residuals, $\R_{\G}$ and
$\R_{\F}$ (\cref{subsub:error-residual-relation}), measured in point-wise
magnitudes.
In addition to inversion evaluation measures, we report characterization
measures of the forward DVFs.
All measures are scalar fields over the displacement domain.

We define the valid displacement domain $\Omega$ as follows.  Given the
image domain $\Omega_0$ for an input DVF, we get
\begin{equation}
  \label{eq:omega-def}
  \Omega = \Omega_{0} - (\Omega_{1} \cup \Omega_{2}) ,
\end{equation}
where
$\Omega_{1} = \setcond{ \x }{ \x \in \Omega_{0} , \; \f(\x) \not\in \Omega_{0}}$
and $\Omega_{2} = \Omega_{0} - \setcond{ \f(\x) }{ \x \in \Omega_{0} }$.
That is, we exclude regions that are either mapped outside the original
domain or not overlapping with the transformed domain.

Regarding summaries of scalar-field measures, we address two issues at
once.  First, a summary shall take into consideration the uncertainty in
numerically provided DVF data due to regional delineation, noise,
artifacts, and outliers.  Second, it shall reflect spatial variation in the
measure field and not obscure non-negligible changes in relatively small
regions.
Taking into account these two concerns, we summarize voxel-wise scalar
measures over $\Omega$ via multiple percentiles (upper-bound
values).
Specifically, let $\scalarfield$ be a scalar field over $\Omega$, bounded
from below by $\scalarfield_{\min}$.  Let
$\Omega(\scalarfield = \tau) = \setcond{\x }{ \scalarfield(\x) = \tau , \;
  \x \in \Omega }$
be the level set or iso-contour set of $\scalarfield$ at value $\tau$, and
$p(\scalarfield=\tau) = \abs{ \Omega(\scalarfield=\tau)} /\abs{\Omega}$ be
the density of the level set.  The $\prctile$-th percentile value of
$\scalarfield$ is defined as
\begin{equation}
\label{eq:prctile}
  \scalarfield[\prctile\%] =
  \inf_{\tau} \setcond{ \tau }{ 
    \int_{\scalarfield_{\min}}^{\tau} 
     p( \scalarfield = \tau') \ud \tau' > \beta\% } .
\end{equation}
In practical computation, we approximate $\scalarfield[\prctile\%]$ via a
discrete histogram of $\scalarfield$.
For all six patient DVFs, we report evaluation summaries with
box-and-whisker plots showing 2nd, 10th, 50th, 90th, and 98th percentiles;
and DVF characterization summaries in a table with 50th, 90th, and 98th
percentiles.  For the COPD4 DVF, we also display image slices of each
volumetric measure field.

% ---------- DVF characterization
%
\subsection{DVF characterization}
\label{sec:patient-data-characterization}

\begin{table}
  \centering
  \caption{% 
    Characterization summary of 6 patient DVFs by displacement lengths
    (along LR, AP, and SI axes) and spectral measures (determinant of the
    deformation Jacobian, spectral radius of the displacement Jacobian, and 
    algebraic control index \cref{eq:algebraic-index}).  The rightmost
    column shows the fraction of voxels associated with complex
    eigenvalues.
    Percentiles in the other columns are defined as per~\cref{eq:prctile}.%
  } 
  \label{tab:patient-dataset-characterization}
      \footnotesize
  %
  % \medskip
  %
  \begin{tabularwithnotes}{
    @{}
    l @{\hspace{1em}}
    d{2.0}
    @{\hspace{1em}}
    d{2.2} d{2.2} d{2.2}
    @{\hspace{1em}}
    d{2.2} d{2.2} d{2.2}
    @{\hspace{1em}}
    c
    @{}
    }%
    {%
      \tnote[\S]{$(100-\beta)$-th percentiles (50th, 10th,
        2nd) shown in the case of $\det{\J_{\f}}$}%
    }
    \toprule
    \multicolumn{1}{@{}c@{\hspace{1em}}}{}
    & \multicolumn{1}{@{}c@{\hspace{1em}}}{$\beta$}
    & \multicolumn{3}{@{}c@{\hspace{1em}}}{$\F(\x)$ \ (\milli\meter)}
    & \multicolumn{3}{@{}c@{\hspace{1em}}}{\textbf{spectral measures}}
    & \multicolumn{1}{@{}c@{}}{%
      \multirow{2}{*}{\vspace*{-4pt}\parbox{2.5em}{$\imag(\lambda)$ $\ne 0$}}}
    \\
    \cmidrule(r{3\trimCrule}){3-5}
    \cmidrule(r{3\trimCrule}){6-8}
    & \multicolumn{1}{@{}c@{\hspace{1em}}}{}
    & \multicolumn{1}{@{}c}{LR}
    & \multicolumn{1}{c}{AP}
    & \multicolumn{1}{c@{\hspace{1em}}}{SI}
    & \multicolumn{1}{@{}c}{$\det{\J_{\f}}$\tmark[\S]}
    & \multicolumn{1}{c}{$\rho(\J_{\F})$}
    & \multicolumn{1}{c@{\hspace{1em}}}{$1 \!-\! 2\reciprocalGap$}
    & 
    \\
    \midrule
    \multirow{3}{*}{\rotatebox{90}{4DCT7}}
    & 50   &    0.7 &  0.9 &  1.3   &   1.0  &  0.1 & -0.9 
    & \multirow{3}{*}{74\%}
    \\ 
    & 90   &    2.3 &  3.1 &  7.8   &   0.8  &  0.3 & -0.7
    \\
    & 98   &    4.3 &  5.3 & 13.3   &   0.6  &  0.4 & -0.5
    %
    % \\[\tabvskip]
    % % 
    % \multirow{3}{*}{4DCT8}
    % % 
    % & 50   &    0.7 &  0.8 &  0.9   &   1.0  &  0.1 & -0.9 
    % & \multirow{3}{*}{55\%}
    % \\
    % & 90   &    2.3 &  3.2 &  4.5   &   0.9  &  0.2 & -0.7
    % \\
    % & 98   &    4.0 &  6.5 &  11.3   &   0.7  &  0.4 & -0.5
    % % 
    \\[\tabvskip]
    \multirow{3}{*}{\rotatebox{90}{COPD1}}
    & 50   &    3.0 & 12.1 &  7.7   &   0.9  &  0.3 & -0.8 
    & \multirow{3}{*}{51\%}
    \\
    & 90   &   8.3  & 32.1 & 25.9   &   0.4  &  0.7 & -0.3
    \\
    & 98   &   14.0 & 37.4 & 33.6   &   \bf0.\bf1  &  \bf1.\bf0 & \bf0.\bf0
    \\[\tabvskip]
    \multirow{3}{*}{\rotatebox{90}{COPD4}}
    & 50   &    3.6 & 8.0 & 11.9   &   0.9  &  0.4 & -0.7 
    & \multirow{3}{*}{44\%}
    \\
    & 90   &   10.2 & 23.9 & 34.6   &   0.3  &  0.9 & -0.1
    \\
    & 98   &   14.9 & 29.1 & 49.5   &   \bf0.\bf0  &  \bf1.\bf4 &  \bf0.\bf3
    \\[\tabvskip]
    \multirow{3}{*}{\rotatebox{90}{COPD5}}
    & 50   &    2.3 & 8.7 &  10.1   &   1.0  &  0.3 & -0.7 
    & \multirow{3}{*}{58\%}
    \\
    & 90   &   7.7 & 30.9 & 31.2   &   0.4  &  0.8 & -0.2
    \\
    & 98   &   12.5 & 39.2 & 43.0   &   \bf0.\bf1  &  \bf1.\bf2 & \bf0.\bf1
    \\[\tabvskip]
    \multirow{3}{*}{\rotatebox{90}{COPD6}}
    & 50   &    2.7 & 21.6 &  8.9   &   0.9  &  0.3 & -0.7 
    & \multirow{3}{*}{69\%}
    \\
    & 90   &   7.9 & 26.0 & 21.6   &   0.4  &  0.7 & -0.3
    \\
    & 98   &   11.9 & 32.7 & 33.1   &   \bf0.\bf1  &  \bf1.\bf0 & \bf0.\bf0
    \\[\tabvskip]
    \multirow{3}{*}{\rotatebox{90}{COPD8}}
    & 50   &    1.8 &  6.2 &  6.7   &   0.9  &  0.3 & -0.7 
    & \multirow{3}{*}{57\%}
    \\
    & 90   &    6.3 & 15.2 & 23.5   &   0.4  &  0.8 & -0.3
    \\
    & 98   &   10.0 & 20.3 & 38.2   &   \bf0.\bf0  &  \bf1.\bf3 &  \bf0.\bf3
    \\
    \bottomrule
  \end{tabularwithnotes}
\end{table}

\Cref{tab:patient-dataset-characterization} lists spatial and spectral
measures of the six forward DVFs.  The spatial measures are displacement
lengths along the LR, AP, and SI axes.  The spectral measures were
described in \cref{subsec:spectral-NTDC-characterization}.

The COPD DVFs have non-small NTDCs over $2\%$ to $10\%$ of the domain, by
the criterion $\rho(\J_{\F}) \geq 1$ or the equivalent criterion on the
control index, $1 - 2\reciprocalGap \geq 0$.  They also have regions with
zero or negative determinants.
Regions with non-small NTDCs seem to concentrate primarily, but not
exclusively, around the diaphragm and chest wall where there is
substantial motion due to inspiration.
In order to guarantee convergence of the inversion iteration over no less
than $98\%$ of $\Omega$, it is necessary to exercise adequate residual
feedback control.

\begin{figure}
  \centering
  \begin{minipage}{\wideonecol}
    \centering
    % 
    % ==================== SLICE LABELS
    % 
    \begin{minipage}{\dctextmargin}
        \centering
        \text{}
    \end{minipage}
    \begin{minipage}{\dcwidthImg}
      \centering
      % 
      % ---------- images
      % 
      \begin{minipage}[t]{\dcThirdAxial}
        \centering
        axial
      \end{minipage}%
      \hfill
      \begin{minipage}[t]{\dcThirdCoronal}
        \centering
        coronal
      \end{minipage}%
      \hfill
      \begin{minipage}[t]{\dcThirdSagittal}
        \centering
        sagittal
      \end{minipage}%
    \end{minipage}%
    \hfill
    % 
    % ---------- colorbar
    % 
    \begin{minipage}{\dcwidthCbar}
      \phantom{.}
    \end{minipage}%
    \\
    % 
    % ==================== DET(J) HEAT-MAPS
    % 
    \begin{minipage}{\dctextmargin}
        \centering
        \rotatebox{90}{\text{\normalsize $\det{\J_{\f}}$}}
    \end{minipage}
    \begin{minipage}{\dcwidthImg}
      \centering
      % 
      % ---------- images
      % 
      \begin{minipage}{\dcThirdAxial}
        \centering
        \includegraphics[width=\linewidth]{%
          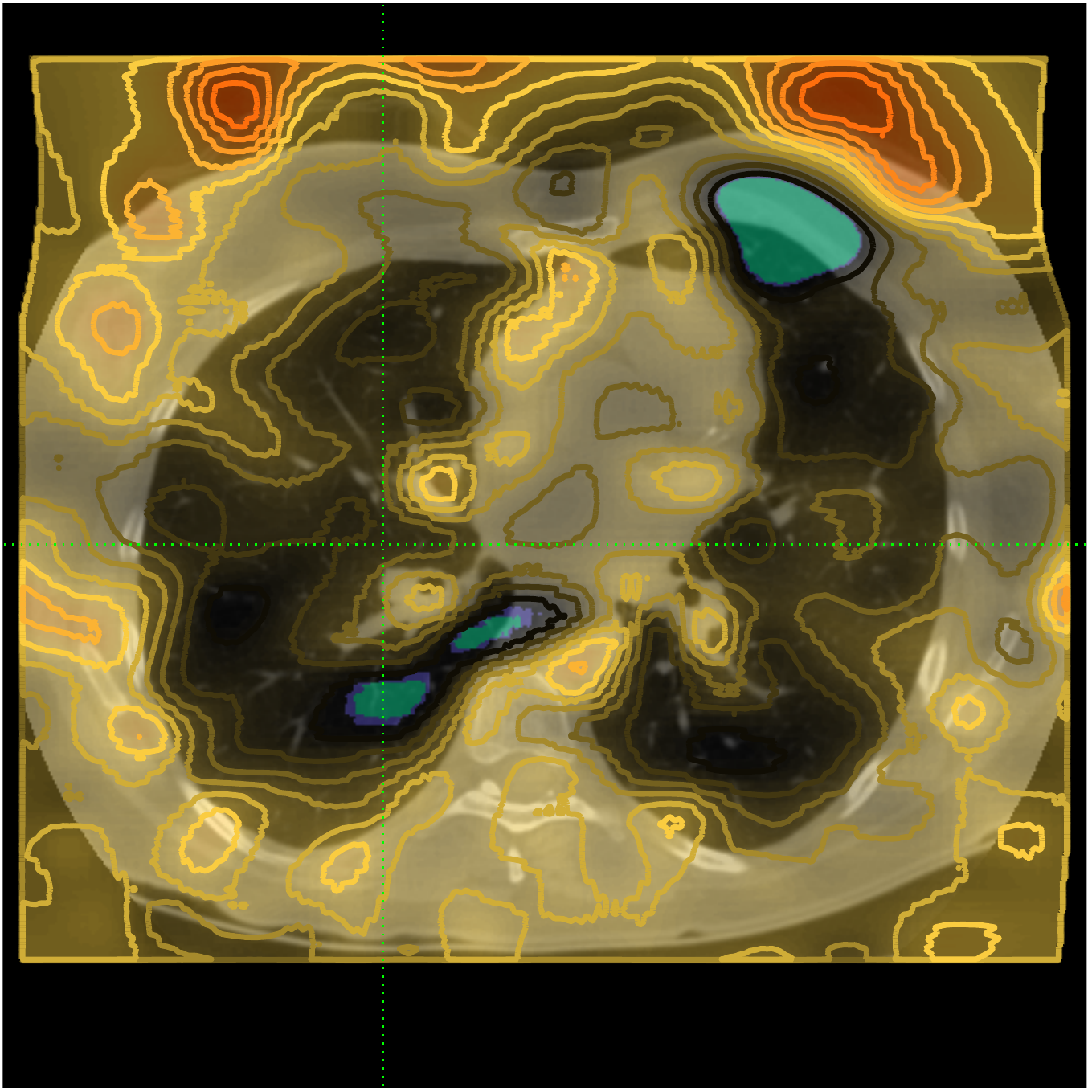}%
      \end{minipage}%
      \hfill
      \begin{minipage}{\dcThirdCoronal}
        \centering
        \includegraphics[width=\linewidth]{%
          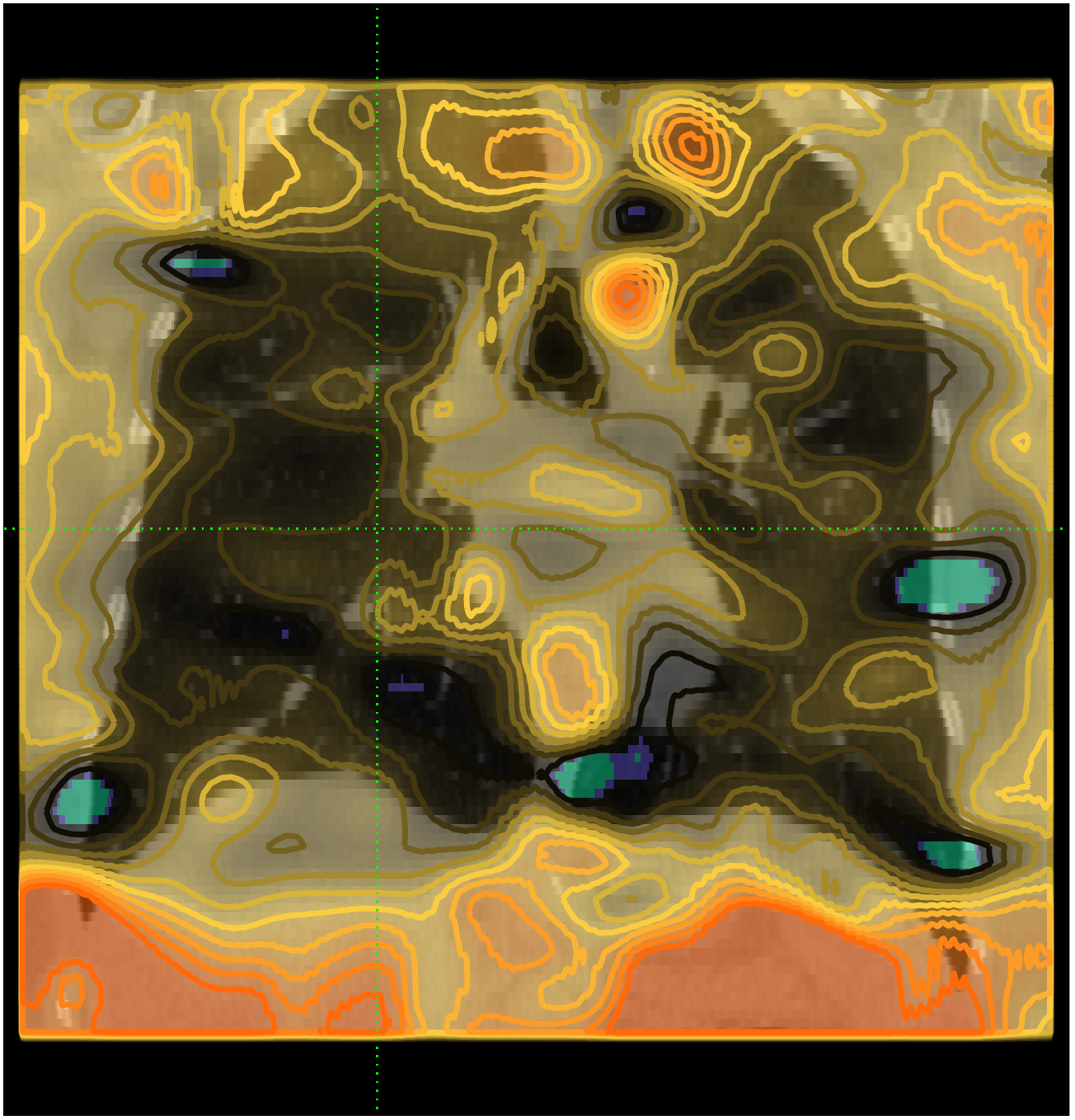}%
      \end{minipage}%
      \hfill
      \begin{minipage}{\dcThirdSagittal}
        \centering
        \includegraphics[width=\linewidth]{%
          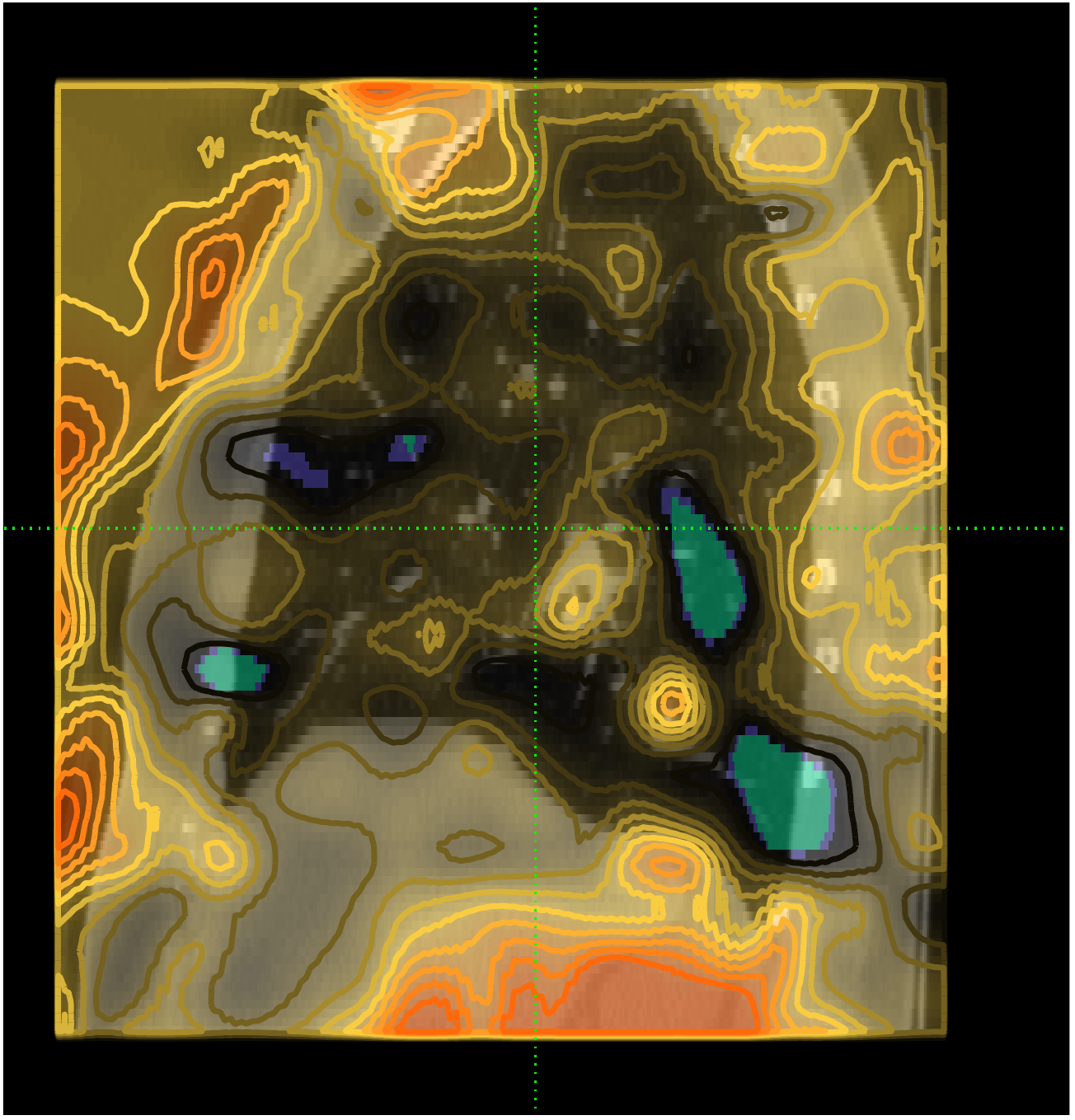}%
      \end{minipage}%
    \end{minipage}
    \hfill
    % 
    % ---------- colorbar
    % 
    \begin{minipage}{\dcwidthCbar}
      \centering
      \includegraphics[width=0.95\linewidth]{%
        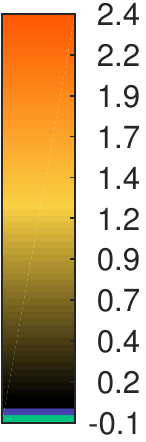}%
    \end{minipage}%
    \\
    % 
    % ==================== LAMBDA_MAX HEAT-MAPS
    % 
    \begin{minipage}{\dctextmargin}
        \centering
        \rotatebox{90}{\text{\normalsize $\rho( \J_{\F} )$}}
    \end{minipage}
    \begin{minipage}{\dcwidthImg}
      \centering
      % 
      % ---------- images
      % 
      \begin{minipage}{\dcThirdAxial}
        \centering
        \includegraphics[width=\linewidth]{%
          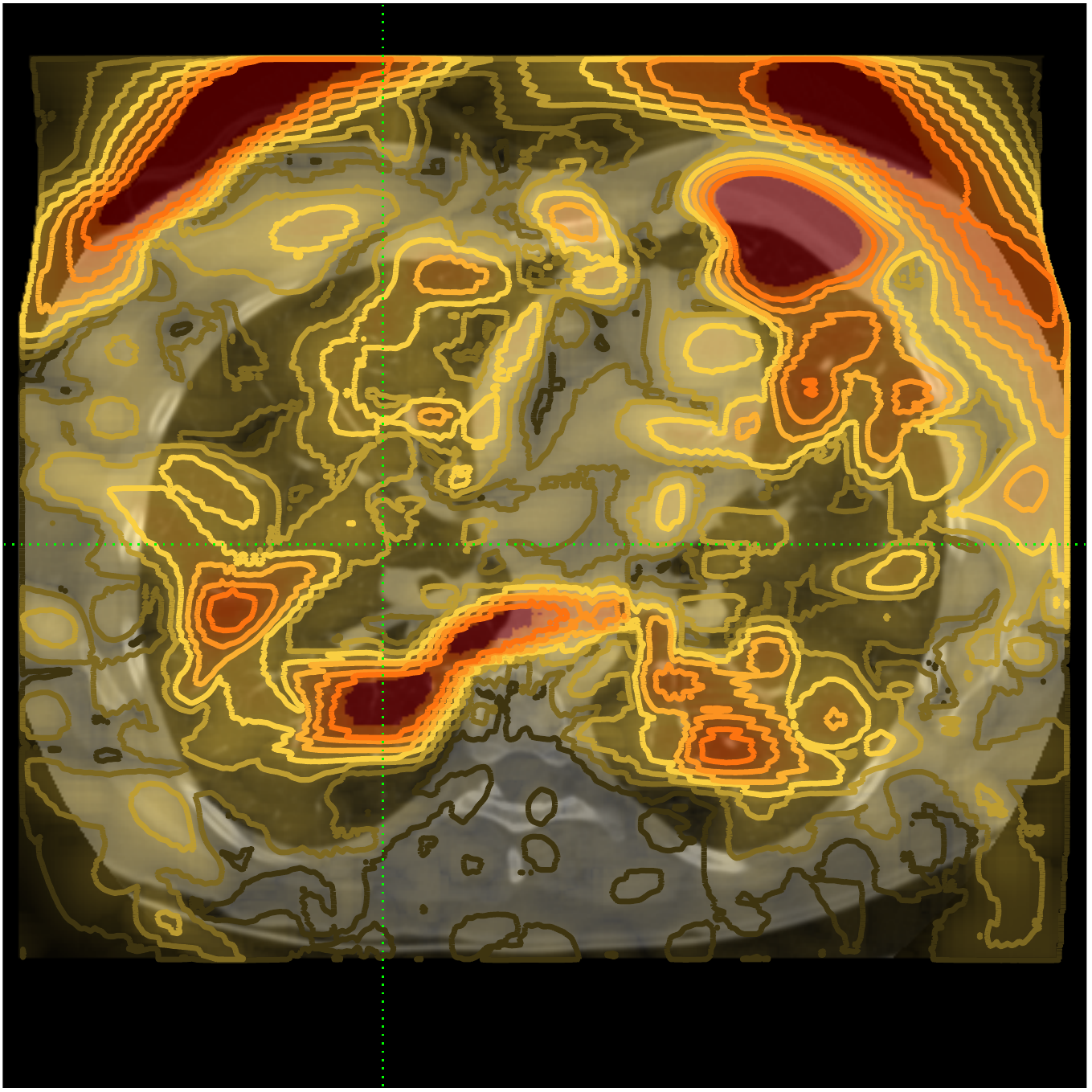}%
      \end{minipage}%
      \hfill
      \begin{minipage}{\dcThirdCoronal}
        \centering
        \includegraphics[width=\linewidth]{%
          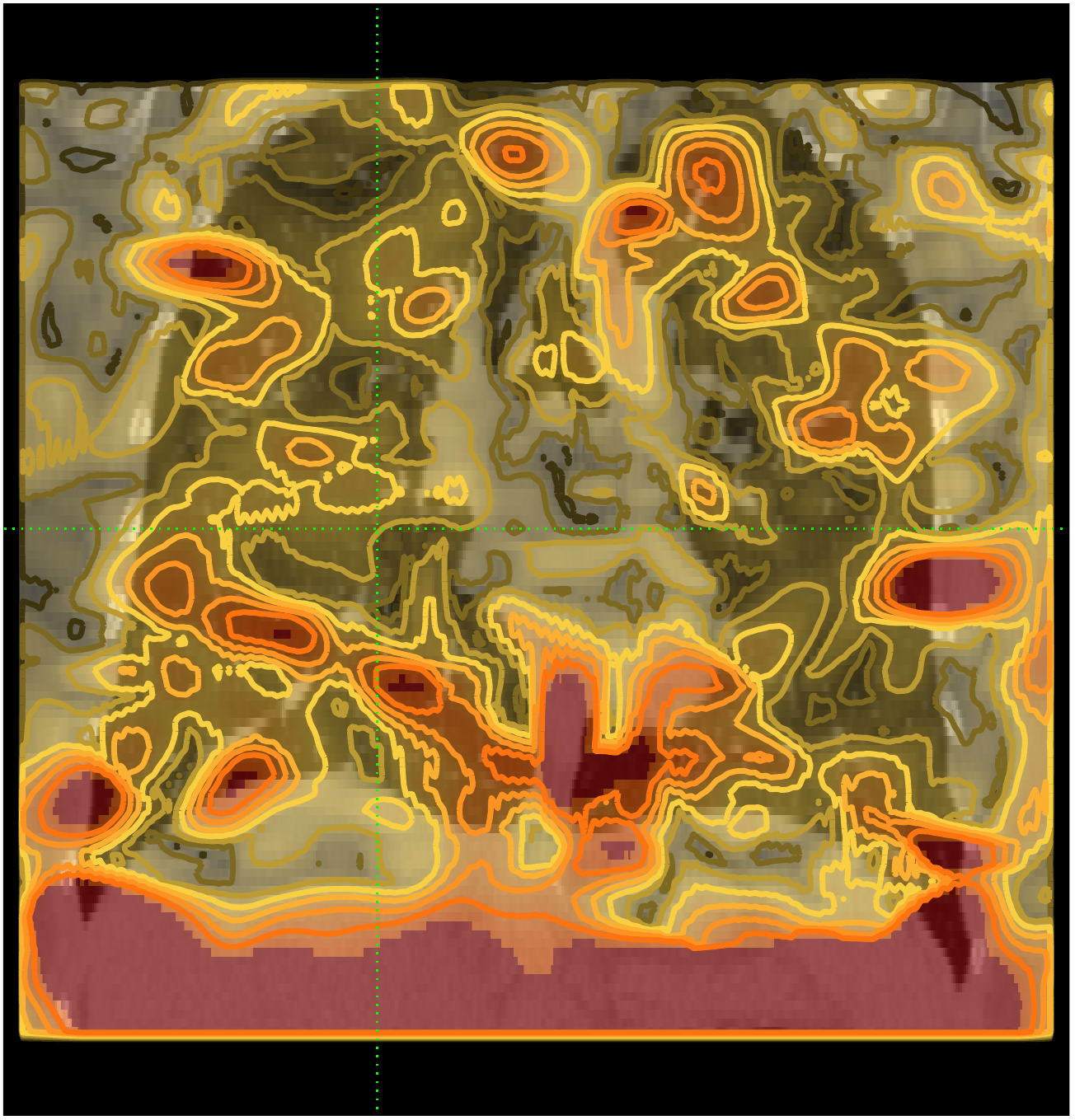}%
      \end{minipage}%
      \hfill
      \begin{minipage}{\dcThirdSagittal}
        \centering
        \includegraphics[width=\linewidth]{%
          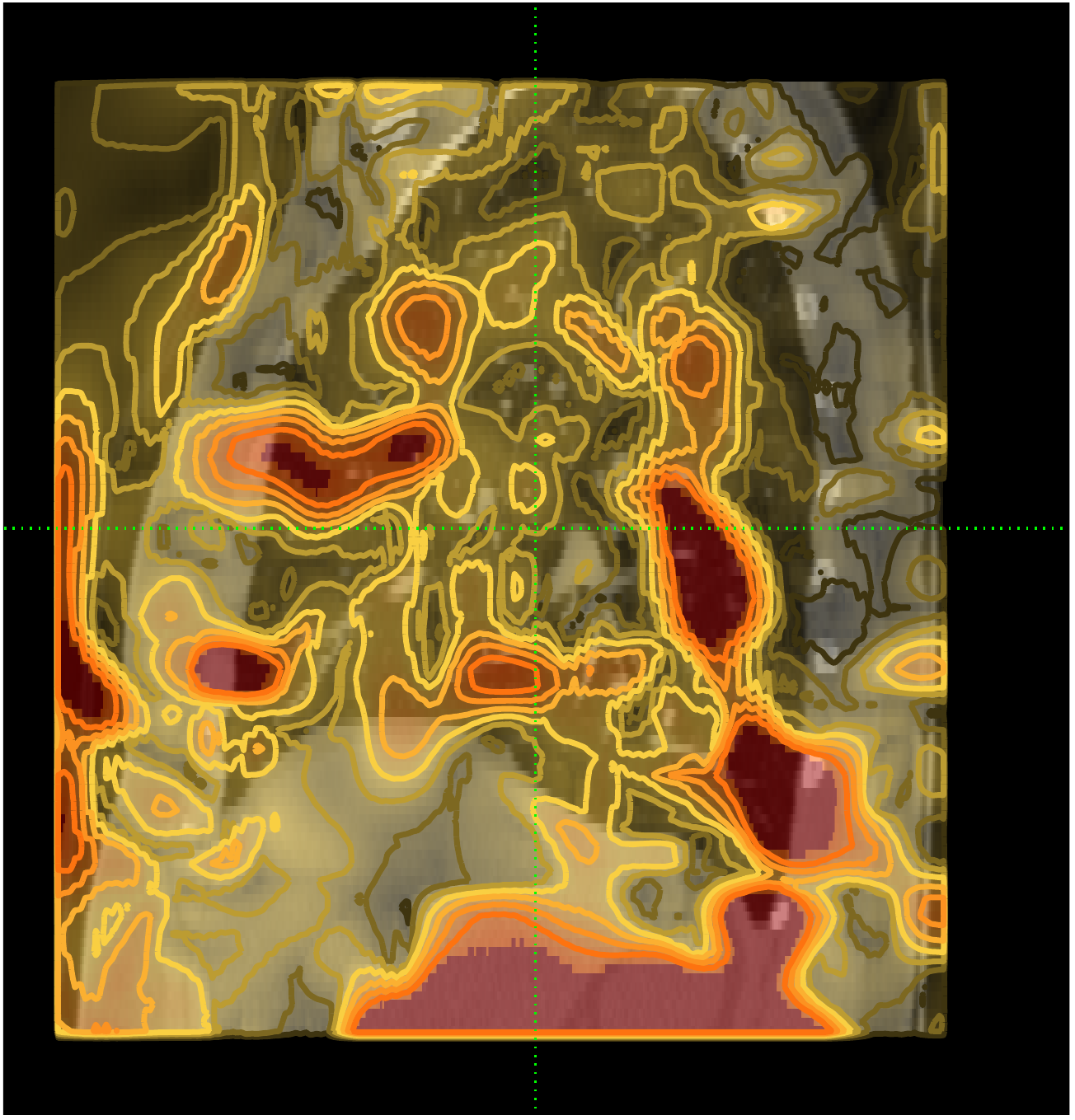}%
      \end{minipage}%
    \end{minipage}%
    \hfill
    % 
    % ---------- colorbar
    % 
    \begin{minipage}{\dcwidthCbar}
      \centering
      \includegraphics[width=0.9\linewidth]{%
        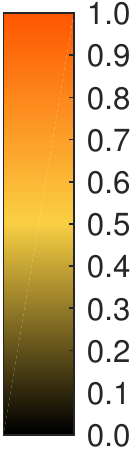}%
    \end{minipage}%
    \\
    % 
    % ==================== KAPPA HEAT-MAPS
    % 
    \begin{minipage}{\dctextmargin}
        \centering
        \rotatebox{90}{\normalsize $1 - 2\reciprocalGap$}
    \end{minipage}
    \begin{minipage}{\dcwidthImg}
      \centering
      % 
      % ---------- images
      % 
      \begin{minipage}{\dcThirdAxial}
        \centering
        \includegraphics[width=\linewidth]{%
          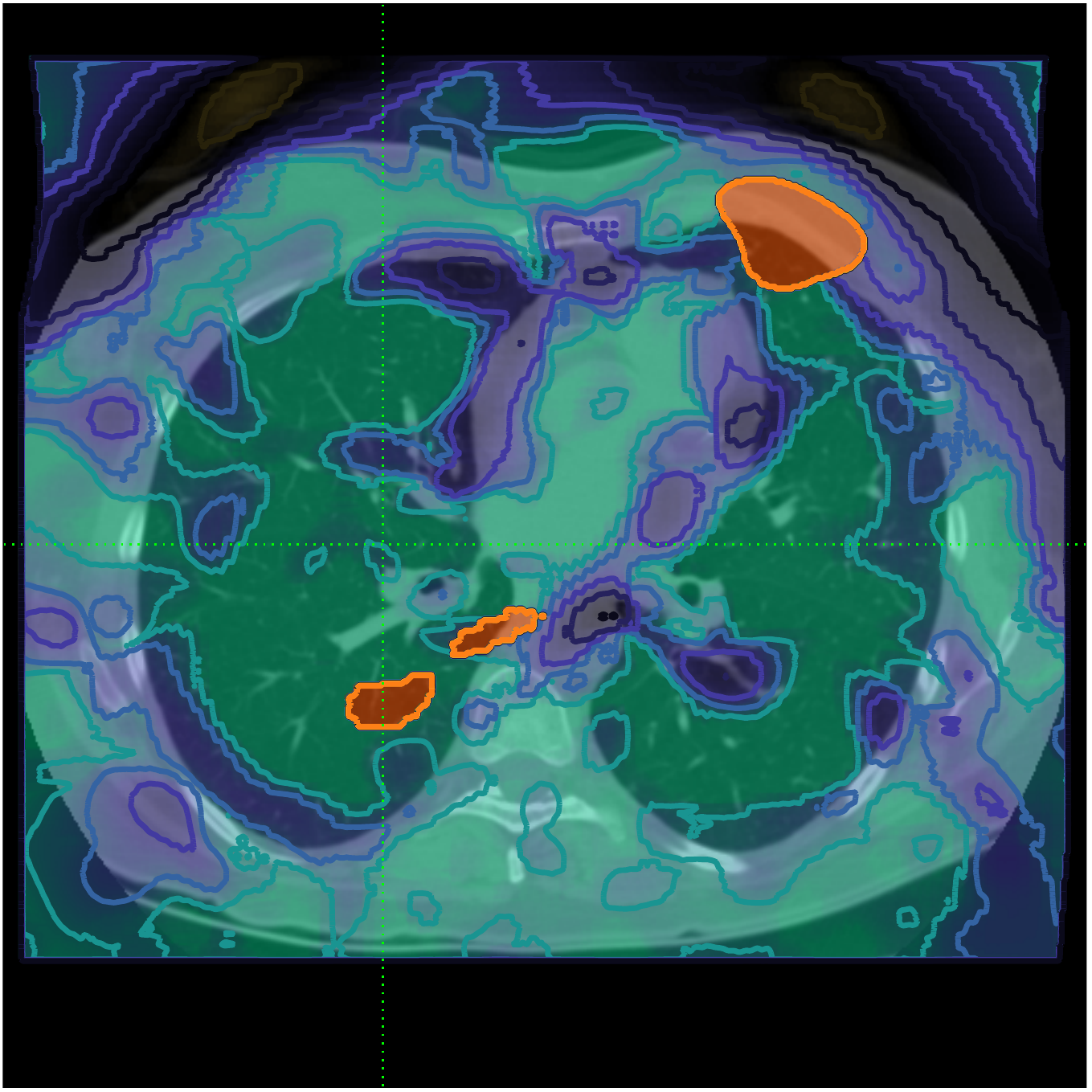}%
      \end{minipage}%
      \hfill
      \begin{minipage}{\dcThirdCoronal}
        \centering
        \includegraphics[width=\linewidth]{%
          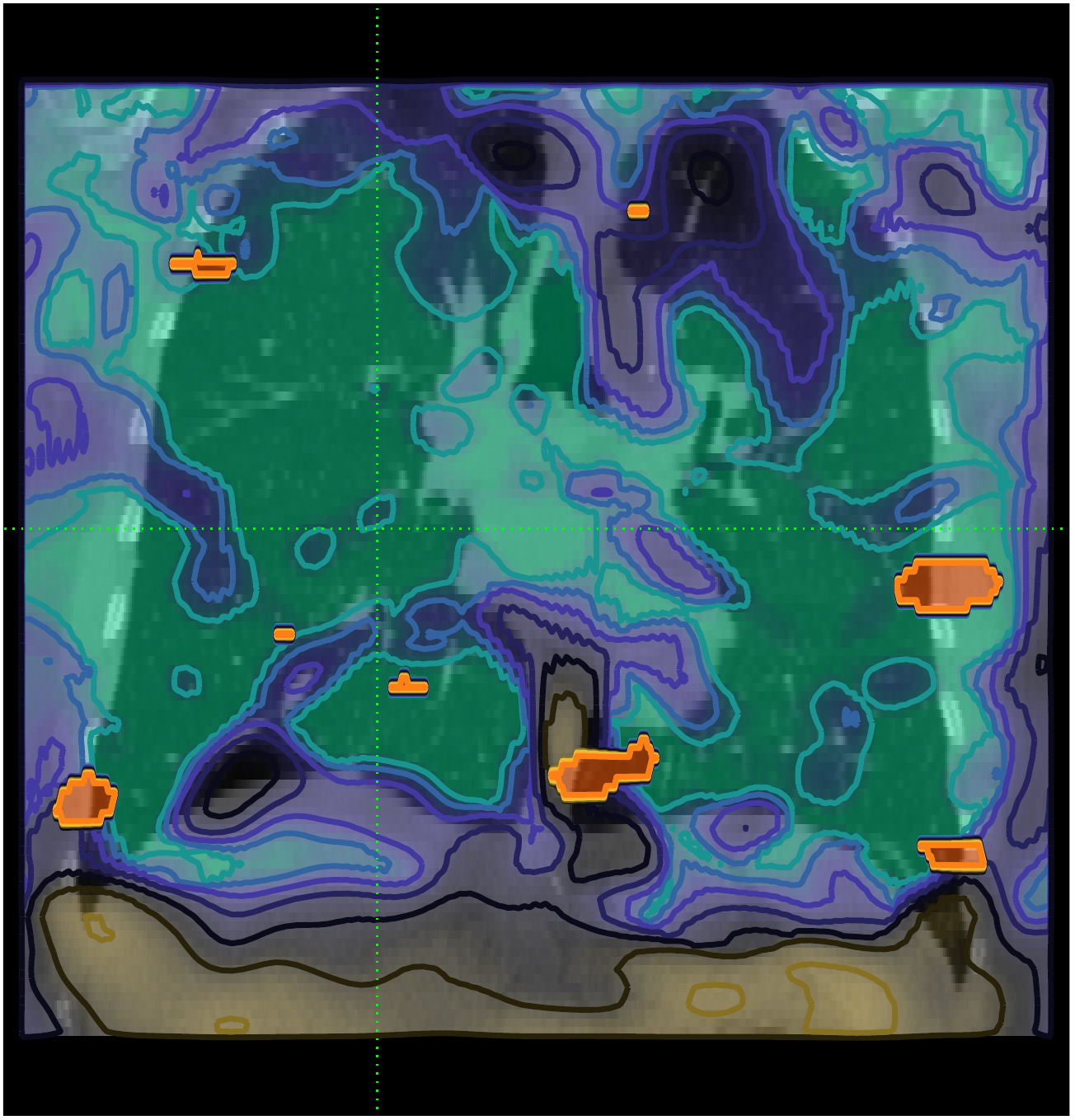}%
      \end{minipage}%
      \hfill
      \begin{minipage}{\dcThirdSagittal}
        \centering
        \includegraphics[width=\linewidth]{%
          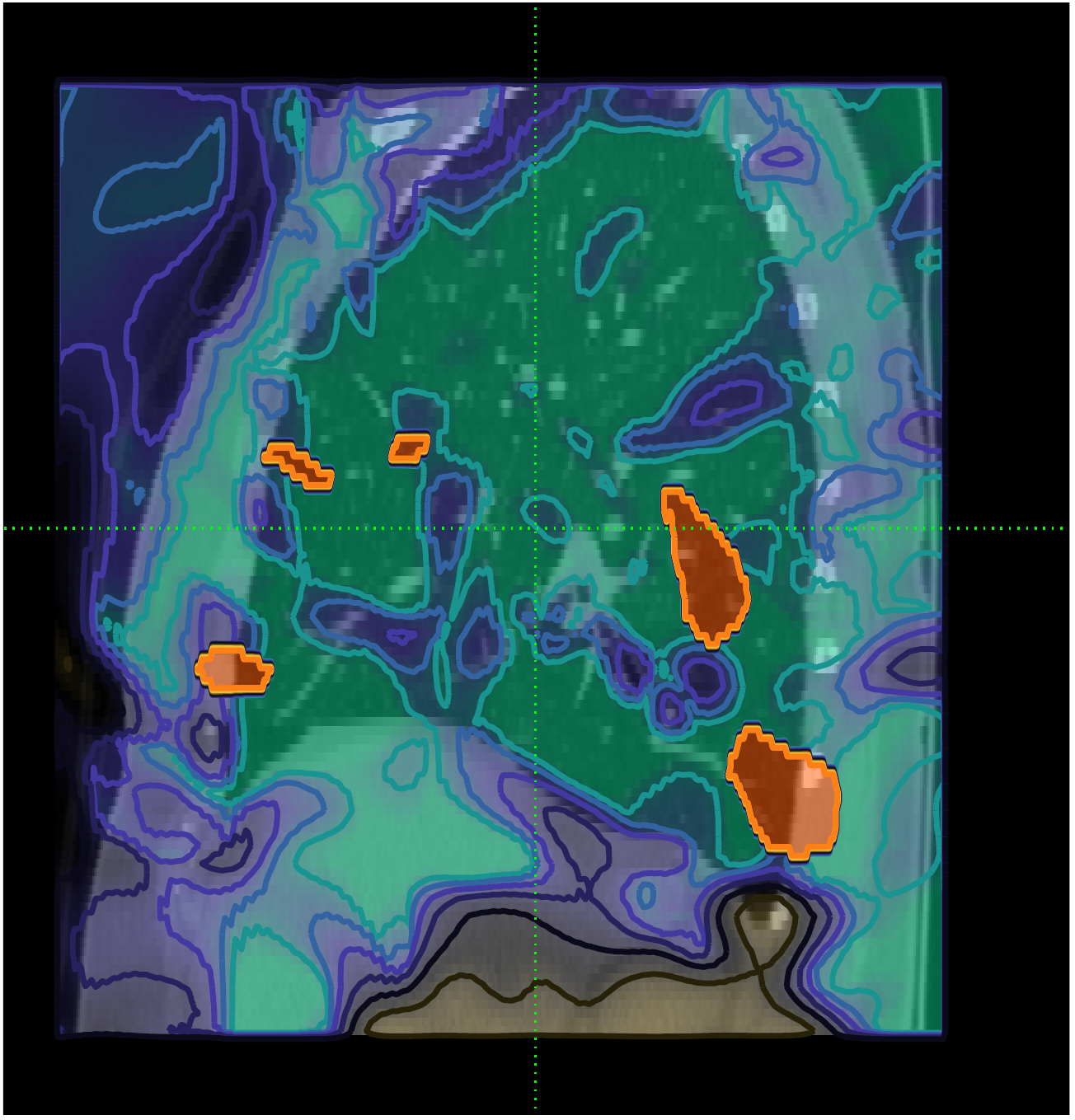}%
      \end{minipage}%
    \end{minipage}
    \hfill
    % 
    % ---------- colormap
    % 
    \begin{minipage}{\dcwidthCbar}
      \centering
      \includegraphics[width=0.95\linewidth]{%
        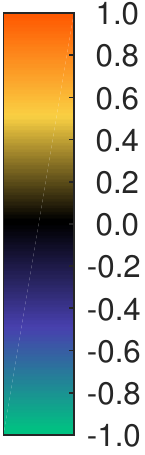}%
    \end{minipage}
  \end{minipage}%
  % 
  % 
  % ==================== CAPTION
  % 
  \caption{%
    Spatial variation of three spectral measures
    (\cref{subsec:spectral-NTDC-characterization}) with the COPD4 DVF, in
    volumetric, contoured heat-maps over the reference domain:
    $\abs{\J_{\f}(\x)}$ (top); $\rho(\J_{\F}(\x))$ (middle); and $1 -
    2\reciprocalGap(\x)$ (bottom). 
    The display range for the determinant map is determined by its 90th
    percentile value as per \cref{eq:prctile}. 
    Zero-valued (black) regions in the determinant show where the
    deformation transformation is non-invertible.  The maps of 
    $\rho(\J_{\F})$ show that NTDCs are observed almost everywhere. 
    The orange spots in the bottom maps show where the control index is
    greater than one; they coincide with the black and blue spots in the
    top maps, which indicate zero or negative determinant values.%
  }
  \label{fig:patient-jacobian-heatmaps}
\end{figure}

We present in \cref{fig:patient-jacobian-heatmaps} a detailed view of the
spectral measures of the COPD4 DVF, via contoured heat-maps over image
slices.
As discussed in \cref{subsec:spectral-NTDC-characterization}, the control
index maps (bottom row) are indeed most informative.  Regions with
non-small NTDCs ($1 - 2\reciprocalGap(\x) \geq 0$) are in the
gray-to-orange color range, and the salient orange spots indicate where the
controllability condition \cref{eq:positive-gamma} is violated
($1 - 2\reciprocalGap(\x) \geq 1$).
The other two maps provide complementary or mutually confirming
information.
The determinant maps (top row) show negative and zero values in blue and
black, respectively; these are within the orange regions in the bottom
maps.
The maps in the middle row show the spectral radii of NTDCs.  Regions with
non-small NTDCs ($\rho(\J_{\F}(\x)) \geq 1$) are highlighted in red.  
% , which are within the reddish regions with larger determinant values in
% the top maps and the gray-to-orange regions in the bottom maps.
%
The controllability condition holds over more than $98\%$ of $\Omega$.  We
suspect that the problematic spots where it is violated were artifacts of
the registration process.

The 4DCT7 DVF has small NTDCs: $\rho(\J_{\F})$ is small, and
$\det{\J_{\f}}$ is not far from $1$.  The DVF satisfies the controllability
condition virtually everywhere.

% ==================== RESULTS
%
\section{Results}
\label{sec:results}

We present results for iterative DVF inversion by
\cref{eq:fpim-scaling-control} with three types of feedback control
schemes:
\begin{inparaenum}[(i)]
\item uniform, constant parameter $\mu$, with each of $4$ values in
  $\{ 0,\smsp 0.3,\smsp 0.5,\smsp 0.7 \}$, which includes the two precursor
  algorithms~\cite{chen2008simple,christensen2001consistent};
\item alternating parameter values
  $ \mu_{\text{oe}}=(\mu_{\text{o}}, \mu_{\text{e}}) $, where
  $\mu_{\text{o}}$ and $\mu_{\text{e}}$ are adaptively set as the mid-range
  values at the 50th and 98th percentiles, respectively (see
  \cref{subsub:mid-range-parameter,subsec:alternating}, and $1 -
  2\reciprocalGap$ in \cref{tab:patient-dataset-characterization}); and
\item spatially variant control $\mu_{\ast}(\x)$ with locally optimal
  values by \cref{eq:mu-star-real-case} or 
  \cref{eq:mu-star-complex-cases}.
\end{inparaenum} 
All six control parameter settings are applied to each of the six DVFs.

Adaptive feedback control with the mid-range parameter value
(\cref{subsub:mid-range-parameter}), although not reported as another
control scheme along the $6$ schemes listed above, is actually used in
more than one way in the experiments.
First, the 50th and 98th percentile mid-range values are employed in
alternating-values control scheme.
Second, the initial guess for each DVF inversion iteration is set to
$\G_{0}(\x) = (\mu_{\text{m}}[98\%] - 1) \F(\x)$ with all $6$ control
schemes, where $\mu_{\text{m}}[98\%]$ is the $98$th percentile mid-range
value.  This is equivalent to taking a single step with mid-range value
control and zero initial guess.

This data-adaptive initialization yields a better initial estimate, and
mitigates an out-of-boundary issue with small-magnitude control parameter
values.
Consider, for instance, the iteration with $\mu = 0$ and zero initial
guess.  Even within the contraction region, it is likely that $\x - \F(\x)$
falls outside the image boundary, and the iteration fails at the second
step, $\G_2(\x) = - \F(\x - \F(\x))$, over a large number of voxels, such
as those on or below the diaphragm, close to the inferior boundary.

We present in \cref{sec:contracting-region-rate} pre-inversion evaluation
of contraction area and ratio with each control scheme, and in
\cref{sec:results-IC-residuals} post-inversion evaluation of IC residuals
with the inverse DVF estimates.

% ---------- Contraction area and rate
%
\subsection{Contraction area \& ratio}
\label{sec:contracting-region-rate}

\begin{figure}
  \centering
  \includegraphics[width=0.35\textwidth]{%
    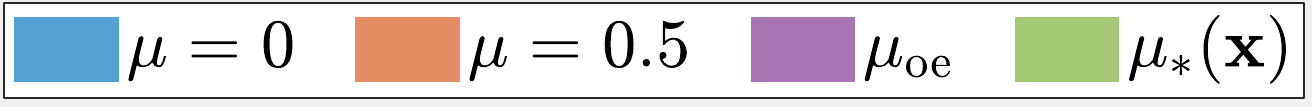}
  \\
  \includegraphics[width=\widthConvergenceFig]{%
    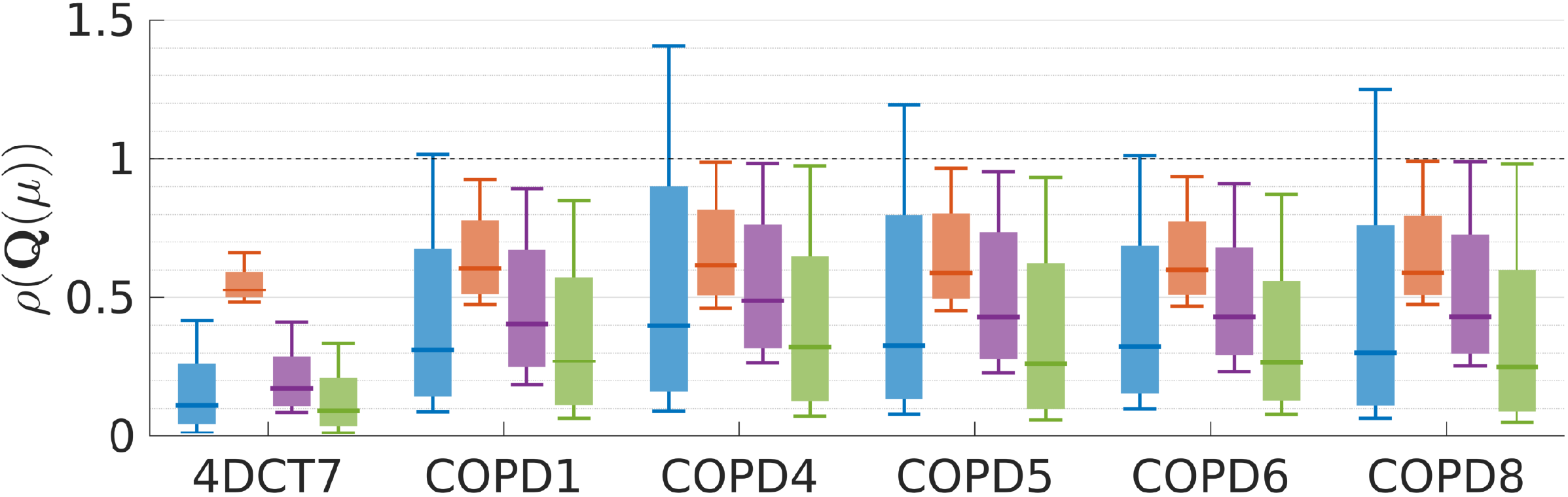}
  \caption{%
    Error contraction ratio percentiles over $\Omega$ for each of the 6
    DVFs, with 4 feedback control parameter settings: constant
    $\mu = 0$ and $\mu = 0.5$, alternating $\mu_{\text{oe}}$, and spatially
    variant $\mu_{\ast}(\x)$.
    Box whiskers indicate the 2nd and 98th percentiles; the 10th and 90th
    percentiles are at the low and high ends of each box; and the median
    (50th percentile) is marked by a horizontal bar through each box.%
  }
  \label{fig:rho-stats-all-patients}
\end{figure}

\begin{figure}
  \centering
  \begin{minipage}{\wideonecol}
    \centering
    %
    % ==================== colorbar
    %
    \hspace*{0.8em}
    \includegraphics[width=0.97\linewidth]{%
        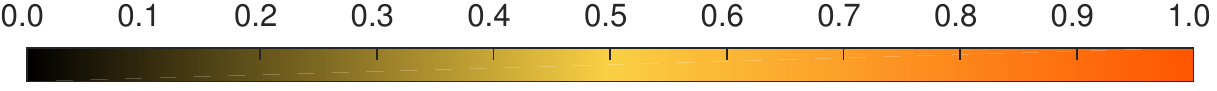}%
    \\[-0.7em]
    %
    % ==================== SLICE LABELS
    %
    % ----- images
    % 
      % 
      \begin{minipage}{\rhotextmargin}
        \centering
        \text{}
      \end{minipage}%
      \begin{minipage}{\rhoThirdAxial}
        \centering
        axial
      \end{minipage}%
      \hfill
      \begin{minipage}{\rhoThirdCoronal}
        \centering
        coronal
      \end{minipage}%
      \hfill
      \begin{minipage}{\rhoThirdSagittal}
        \centering
        sagittal
      \end{minipage}%
    \\
    % 
    % ==================== RHO MAPS (MU = 0)
    %
    % ----- images
    % 
      \begin{minipage}{\rhotextmargin}
        \centering
        \rotatebox[origin=c]{90}{\normalsize$\mu = 0$}
      \end{minipage}%
      % 
      % \hfill
      %
      \begin{minipage}{\rhoThirdAxial}
        \centering
        \includegraphics[width=\linewidth]{%
          patient4_spectralradius_mu_pt_0-axial}
      \end{minipage}%
      \hfill
      \begin{minipage}{\rhoThirdCoronal}
        \centering
        \includegraphics[width=\linewidth]{%
          patient4_spectralradius_mu_pt_0-coronal}
      \end{minipage}%
      \hfill
      \begin{minipage}{\rhoThirdSagittal}
        \centering
        \includegraphics[width=\linewidth]{%
          patient4_spectralradius_mu_pt_0-sagittal}
      \end{minipage}%
    \\
    % 
    % ==================== RHO MAPS (MU = 0.5)
    %
    % ----- images
    % 
      % 
      \begin{minipage}{\rhotextmargin}
        \centering
        \begin{turn}{90}
            {\normalsize $\mu=0.5$}
        \end{turn}
      \end{minipage}%
      \begin{minipage}{\rhoThirdAxial}
        \centering
        \includegraphics[width=\linewidth]{%
          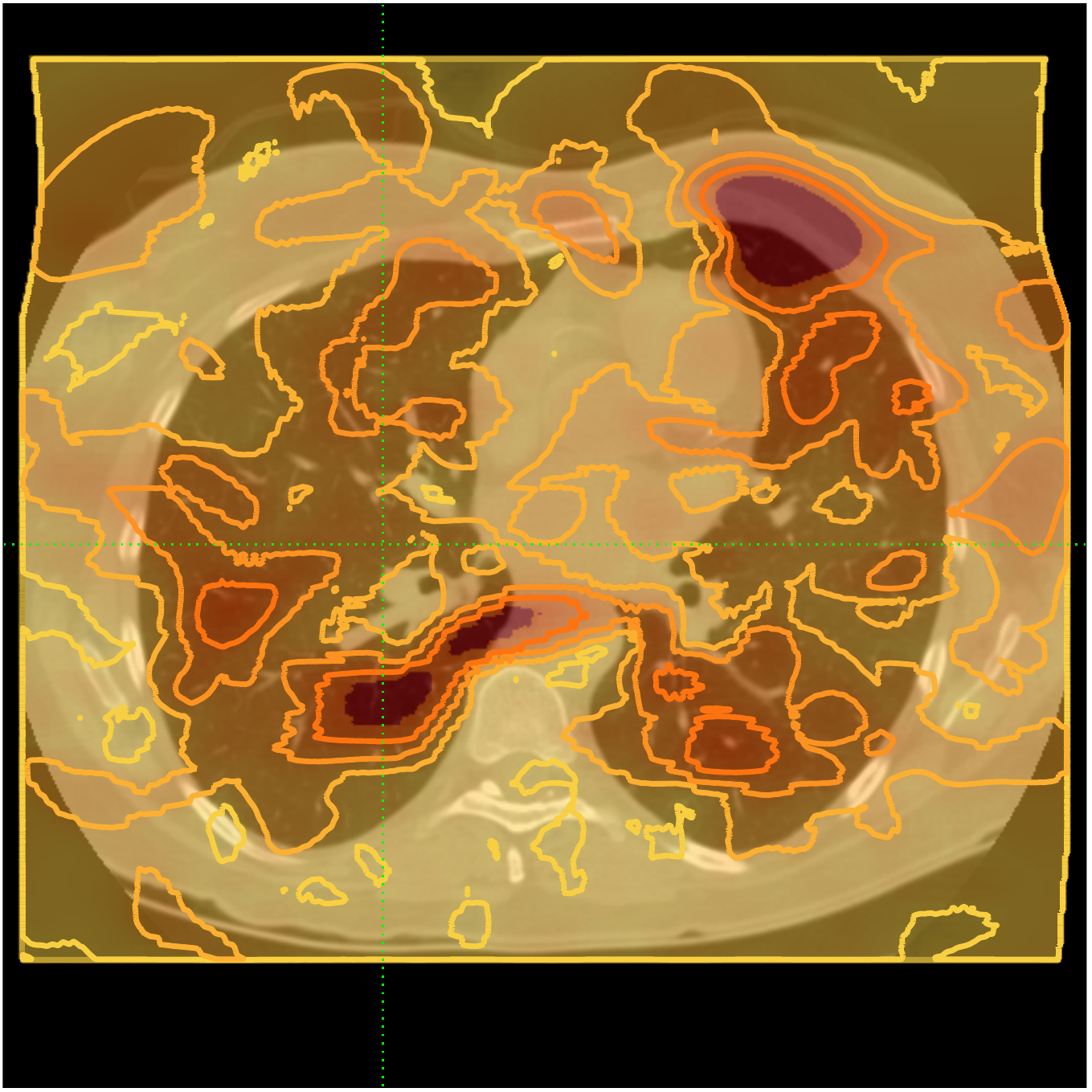}
      \end{minipage}%
      \hfill
      \begin{minipage}{\rhoThirdCoronal}
        \centering
        \includegraphics[width=\linewidth]{%
          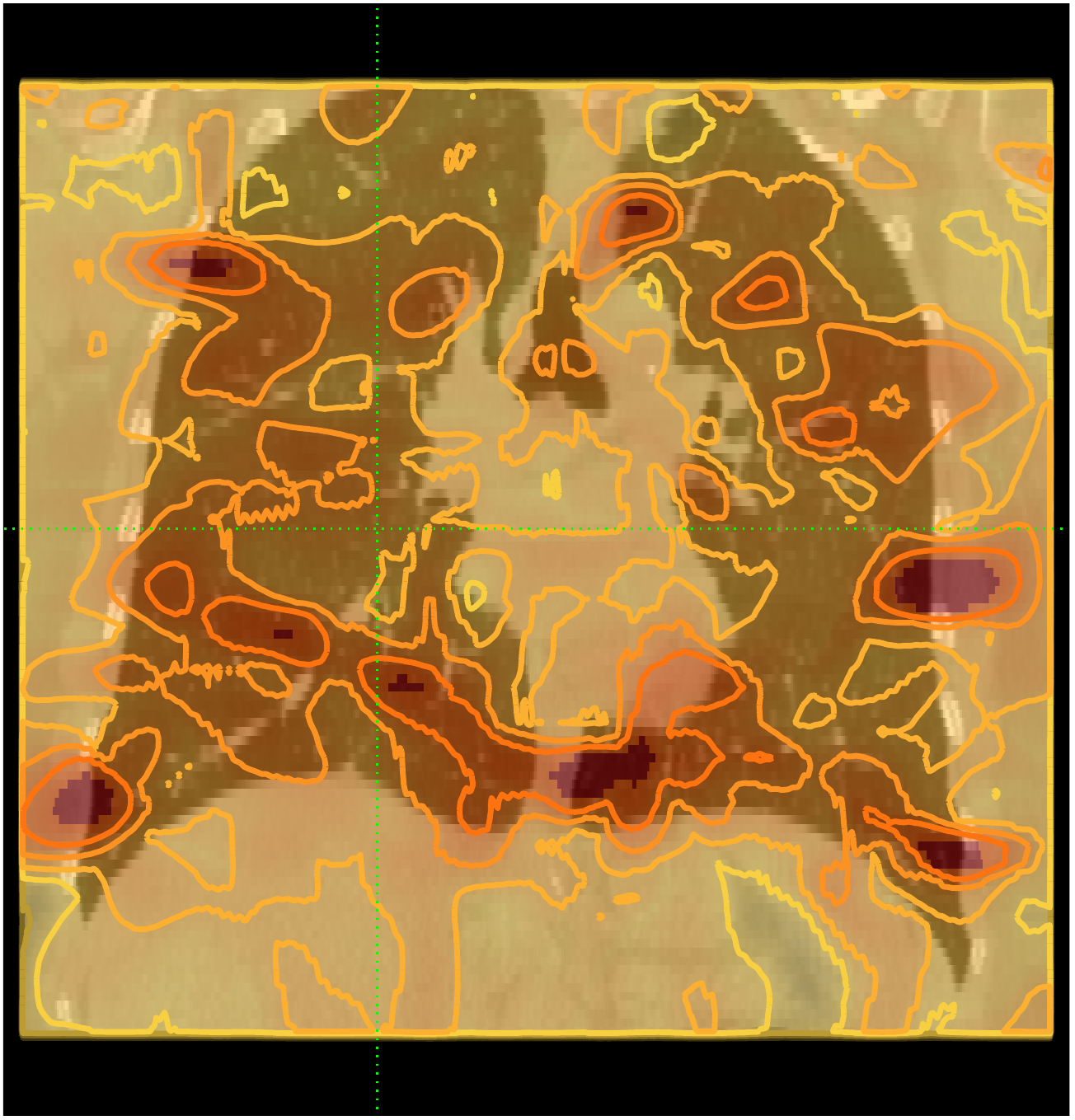}
      \end{minipage}%
      \hfill
      \begin{minipage}{\rhoThirdSagittal}
        \centering
        \includegraphics[width=\linewidth]{%
          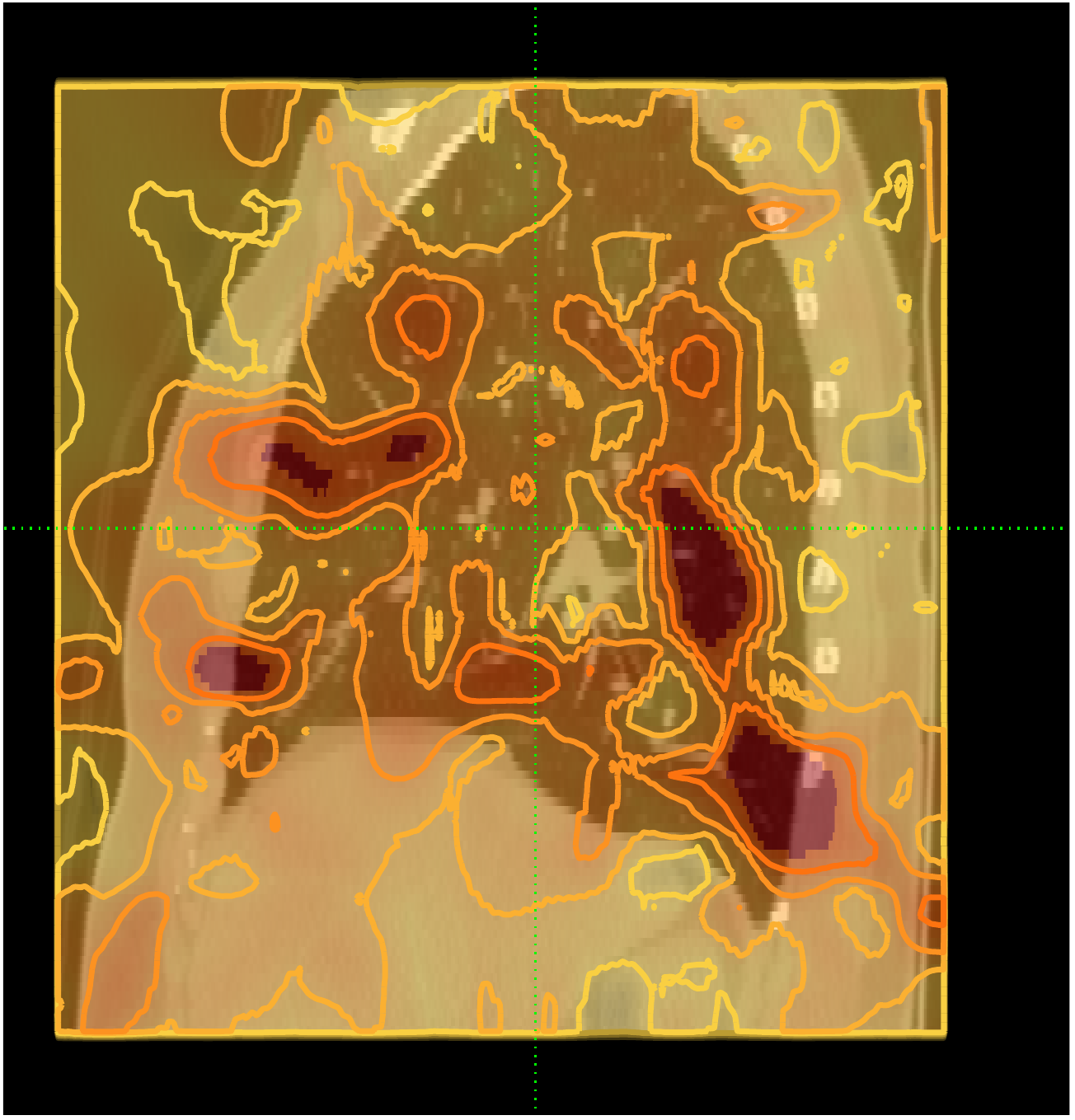}
      \end{minipage}%
    \\
    % 
    % ==================== RHO MAPS (MU map)
    %
      \begin{minipage}{\rhotextmargin}
        \centering
        \begin{turn}{90}
            {\normalsize $\mu_{\text{oe}}$}
        \end{turn}
      \end{minipage}%
      \begin{minipage}{\rhoThirdAxial}
        \centering
        \includegraphics[width=\linewidth]{%
          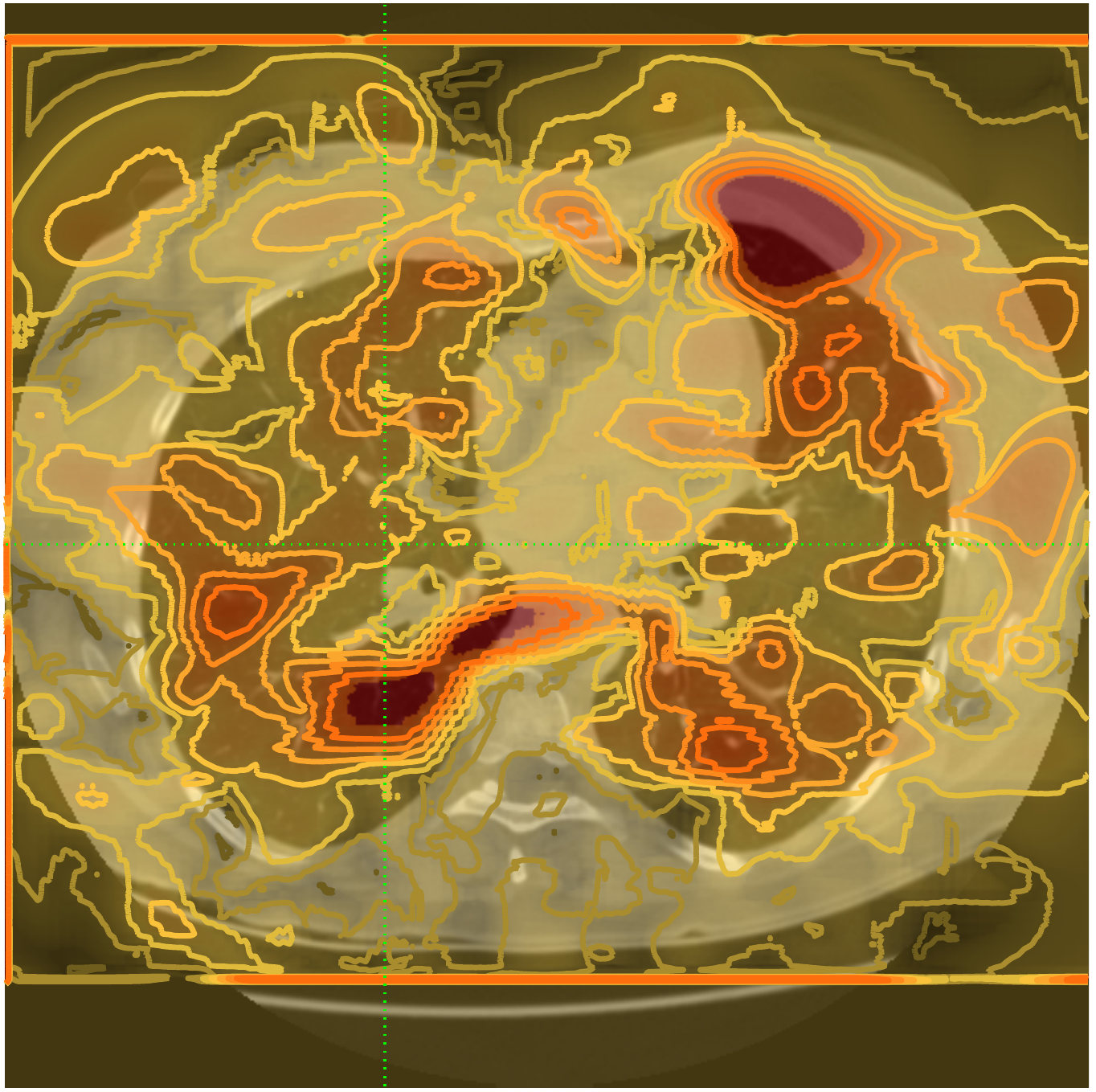}
      \end{minipage}%
      \hfill
      \begin{minipage}{\rhoThirdCoronal}
        \centering
        \includegraphics[width=\linewidth]{%
          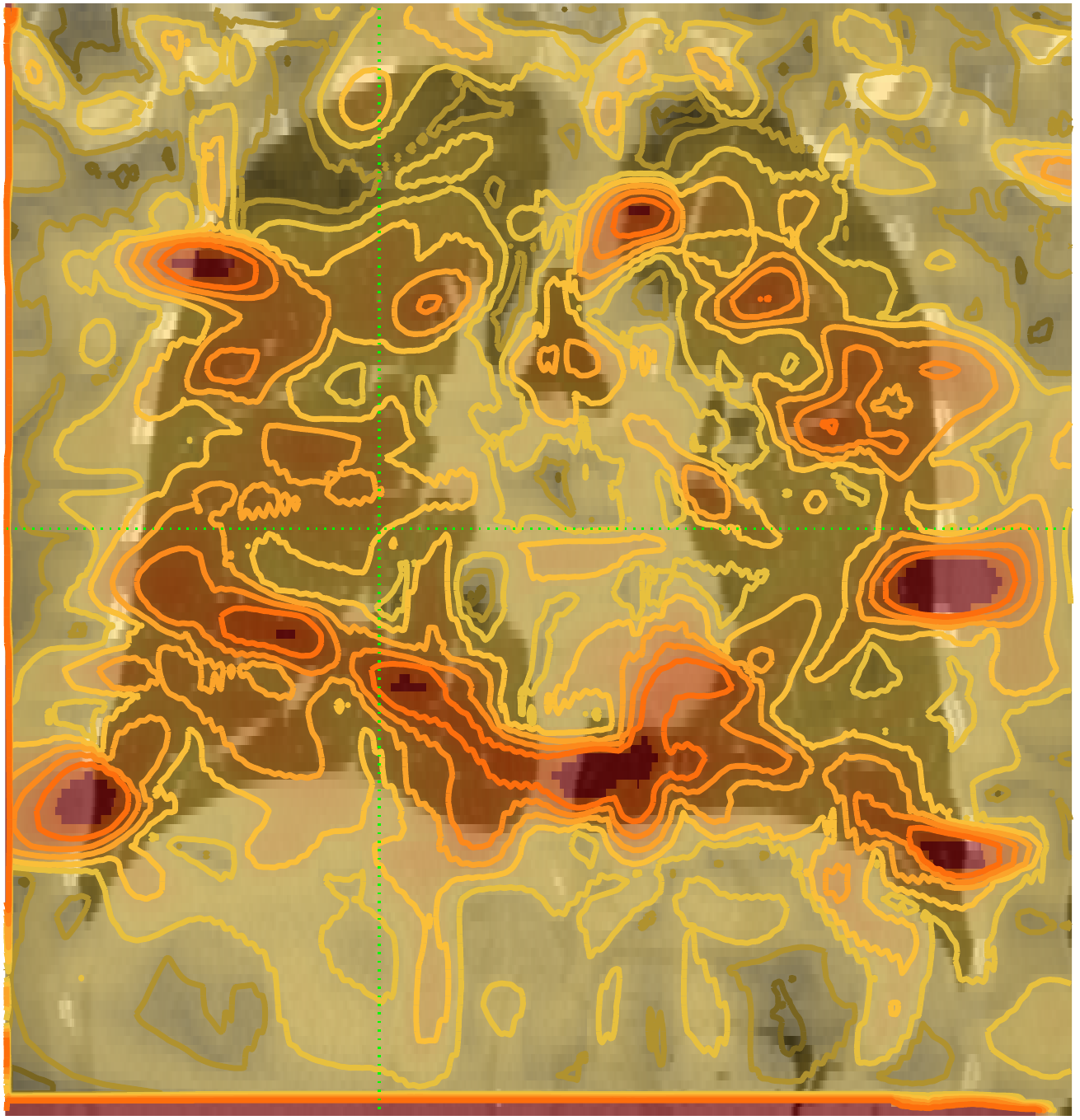}
      \end{minipage}%
      \hfill
      \begin{minipage}{\rhoThirdSagittal}
        \centering
        \includegraphics[width=\linewidth]{%
          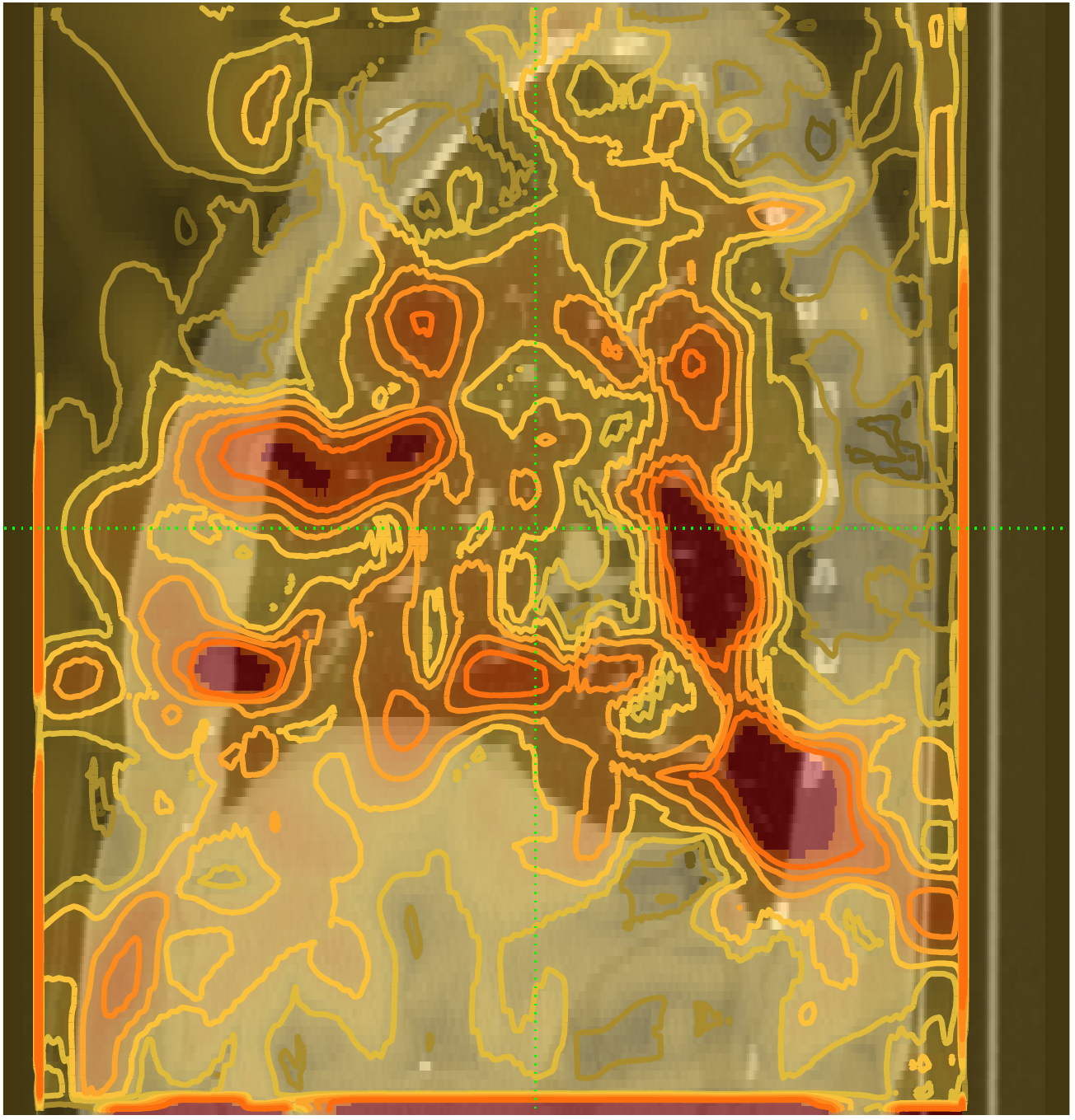}
      \end{minipage}%
    \\
    % 
    % 
    % ==================== RHO MAPS (MU map)
    %
    % ----- images
    % 
      \begin{minipage}{\rhotextmargin}
        \centering
        \begin{turn}{90}
            {\normalsize $\mu_{*}(\x)$}
        \end{turn}
      \end{minipage}%
      \begin{minipage}{\rhoThirdAxial}
        \centering
        \includegraphics[width=\linewidth]{%
          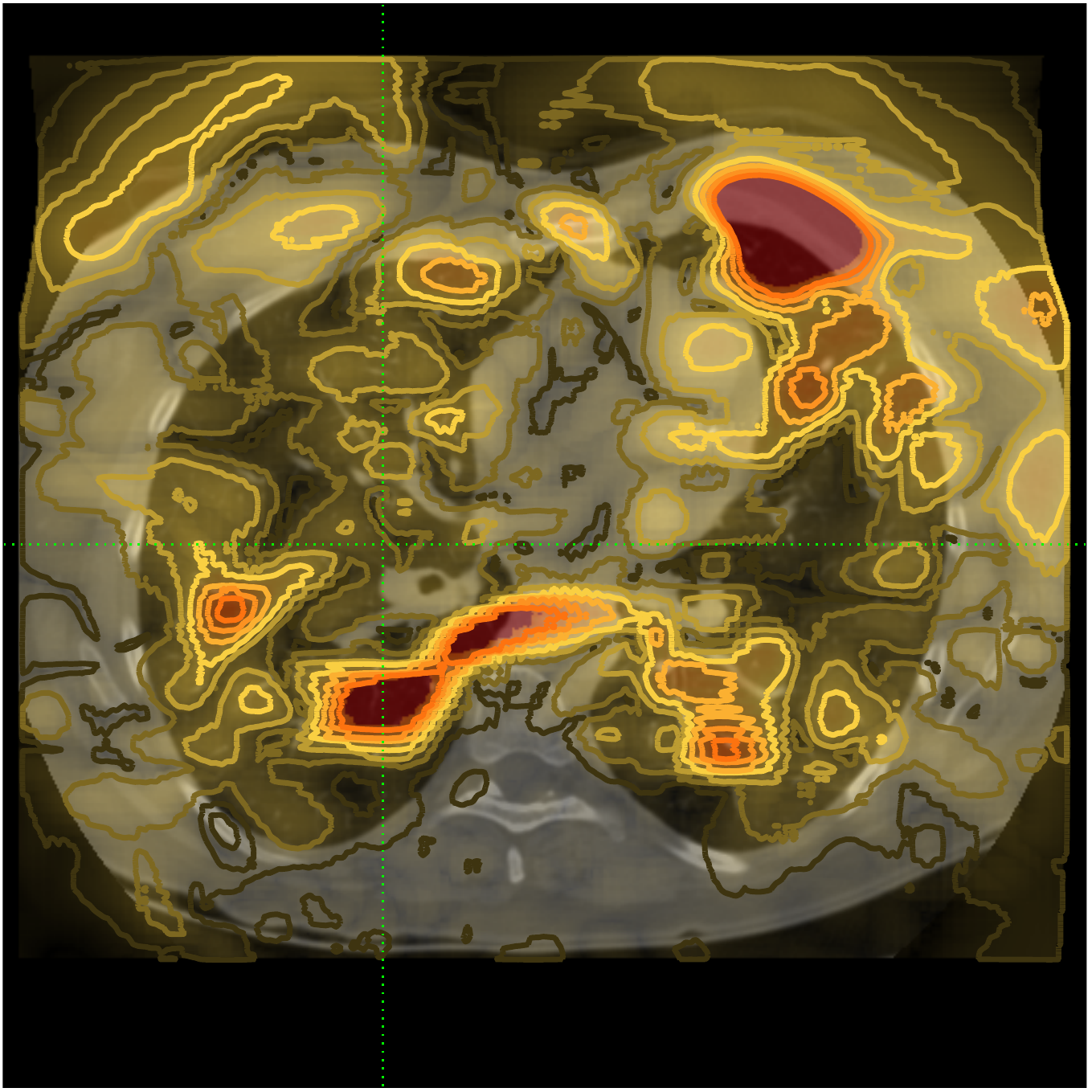}
      \end{minipage}%
      \hfill
      \begin{minipage}{\rhoThirdCoronal}
        \centering
        \includegraphics[width=\linewidth]{%
          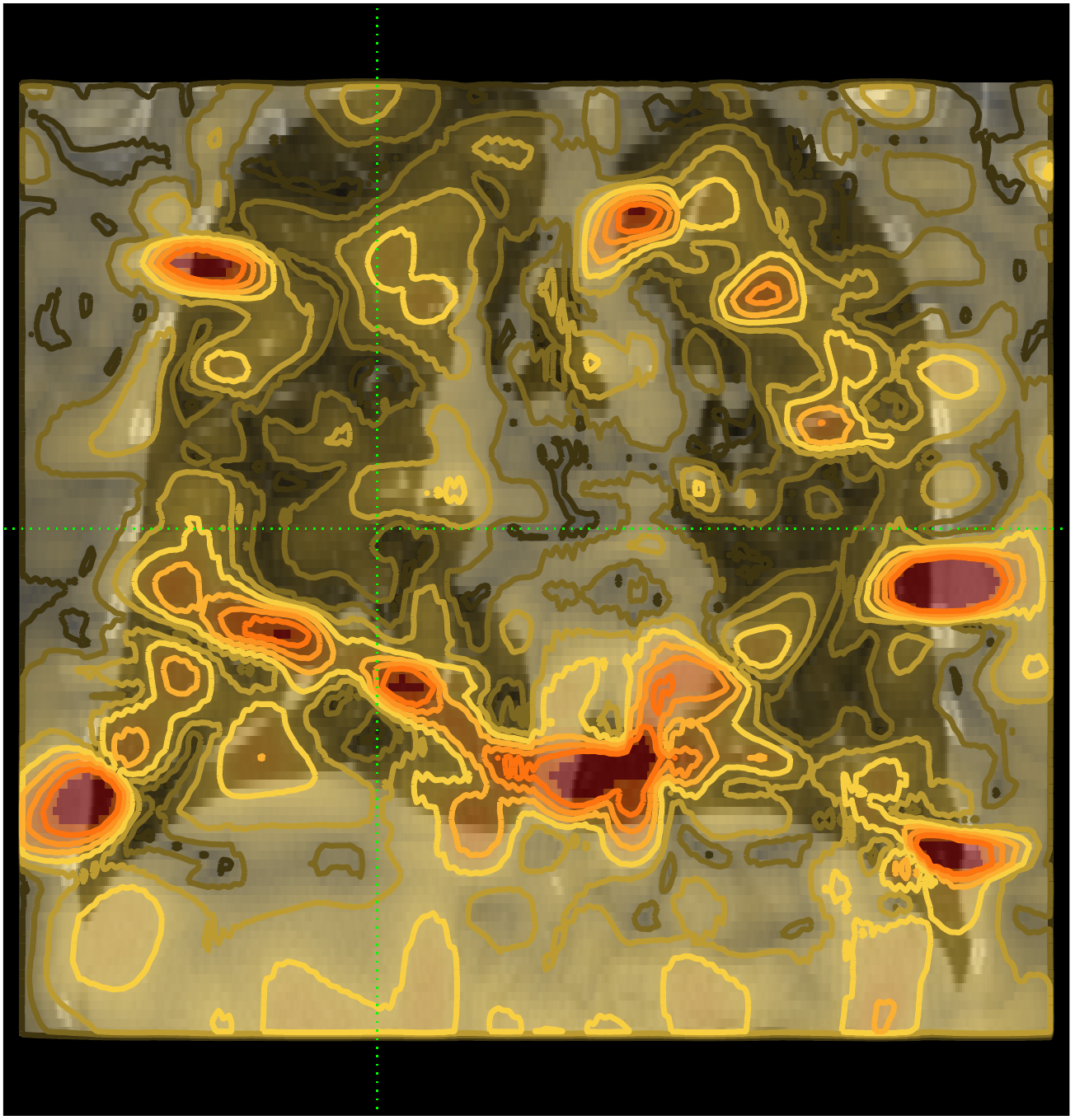}
      \end{minipage}%
      \hfill
      \begin{minipage}{\rhoThirdSagittal}
        \centering
        \includegraphics[width=\linewidth]{%
          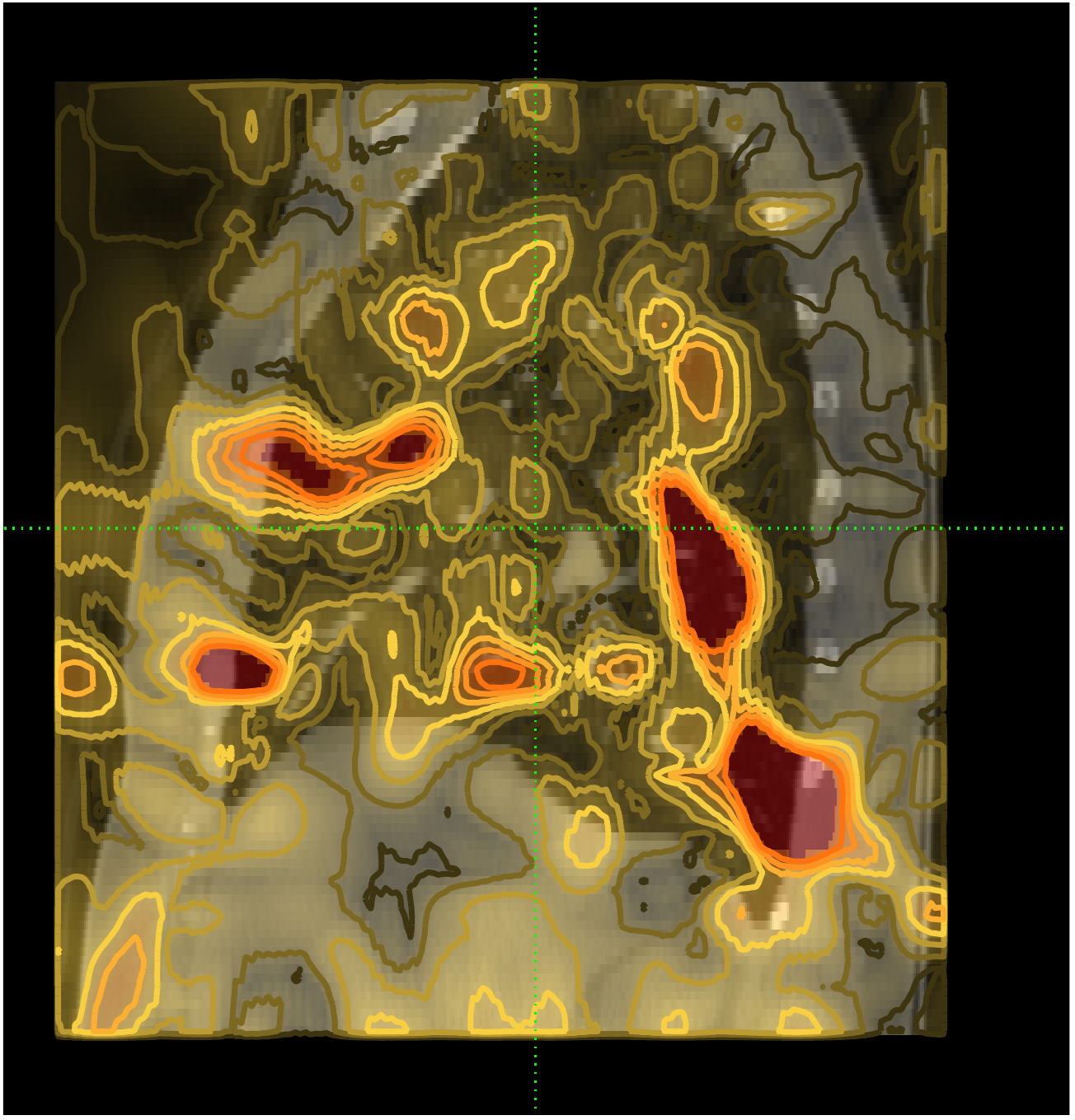}
      \end{minipage}%
  \end{minipage}
  %
  % 
  % ==================== CAPTION
  % 
  \caption{%
    Volumetric heap-maps of the error contraction ratio $\rho(\ICM(\x;\mu))$
    over the reference domain, with the COPD4 DVF.  Comparison between $4$
    different feedback control parameter settings:
    $\mu = 0$;
    $\mu = 0.5$;
    alternating values, $\mu_{\text{oe}}$, with $\mu_{\text{o}} = 0.15$ at odd
    steps and $\mu_{\text{e}} = 0.65$ at even steps; and 
    spatially variant values, $\mu_{\ast}(\x)$.
    Errors are suppressed more aggressively over the darker regions where
    $\rho(\ICM(\x;\mu))$ is small, but are enlarged over the red regions,
    where $\rho(\ICM(\x;\mu)) \ge 1$.%
  }
  \label{fig:contraction-regions-copd4}
\end{figure}

We provide in \cref{fig:rho-stats-all-patients} a summary of error
contraction ratios, i.e., the spectral radii of the infinitesimal
contraction matrices, for each of the $6$ DVFs and with each of the
following $4$ control settings: $\mu = 0$, $\mu = 0.5$, alternating values
$\mu_{\text{oe}}$, and spatially variant values $\mu_{\ast}(\x)$.  The
latter two are adaptive.
Lower contraction ratios indicate faster error suppression.
With the 4DCT7 DVF, the contraction ratios indicate convergence by all
schemes, with the iteration with $\mu = 0.5$ at a much slower pace.
With the COPD DVFs, except COPD4, the two constant, non-adaptive schemes
are comparable to each other at the 90th percentile.  At the 95th (not
shown) and 98th percentiles, the scheme with $\mu = 0$ fails to contract
over non-small NTDC regions, whereas the scheme with $\mu = 0.5$ maintains
contraction ratios under $1$.  The value $0.5$ happens to be in the control
parameter range of each COPD DVF.
The adaptive control scheme with alternating values for each DVF is better
than the non-adaptive schemes up to 90th percentile and in fact up to 95th
percentile (not shown).  If it is desirable that the contraction region
cover 98\% or more of the domain, one shall use the high-percentile
mid-range value, which happens to be between $0.5$ and $0.65$ for each COPD
DVF.
The spatially variant scheme yields the lowest contraction ratios at all
percentiles and with each DVF.

We display in \cref{fig:contraction-regions-copd4} a comparison between the
$4$ control parameter settings in the spatial variation of error
contraction ratios with the COPD4 DVF.
The regions in red are non-contraction regions.  Failure to contract is
either due to violation of the controllability condition or due to
inadequate feedback control (cf.\ \cref{fig:patient-jacobian-heatmaps,%
  fig:contraction-regions-copd4}).  The violation regions are common to all
heat-maps, and correspond to the orange spots in the control index maps in
\cref{fig:patient-jacobian-heatmaps}.  The scheme with $\mu=0$ has the
smallest contraction area, indicating the failure of this non-adaptive
control setting over the controllable region.
Spatially variant control $\mu_{\ast}(\x)$ yields the largest contraction
area and lowest contraction ratios.

\ifdefined\medphys\ifdefined\review\else

\begin{figure*}
  \centering
  % 
  % ==================== LEGEND
  % 
  \subfloat{%
    \includegraphics[width=0.55\linewidth]{%
      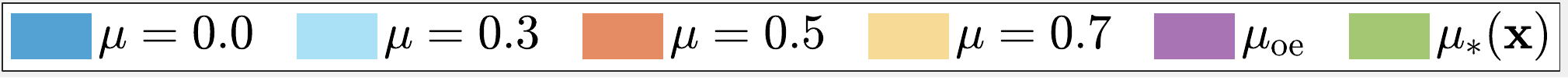}}
  \\
  %
  % ==================== 4DCT7
  % 
  \subfloat{%[\label{fig:residual-percentiles-boxplots-4dct7}][4DCT7]{%
    \includegraphics[width=\halftwocol]{%
      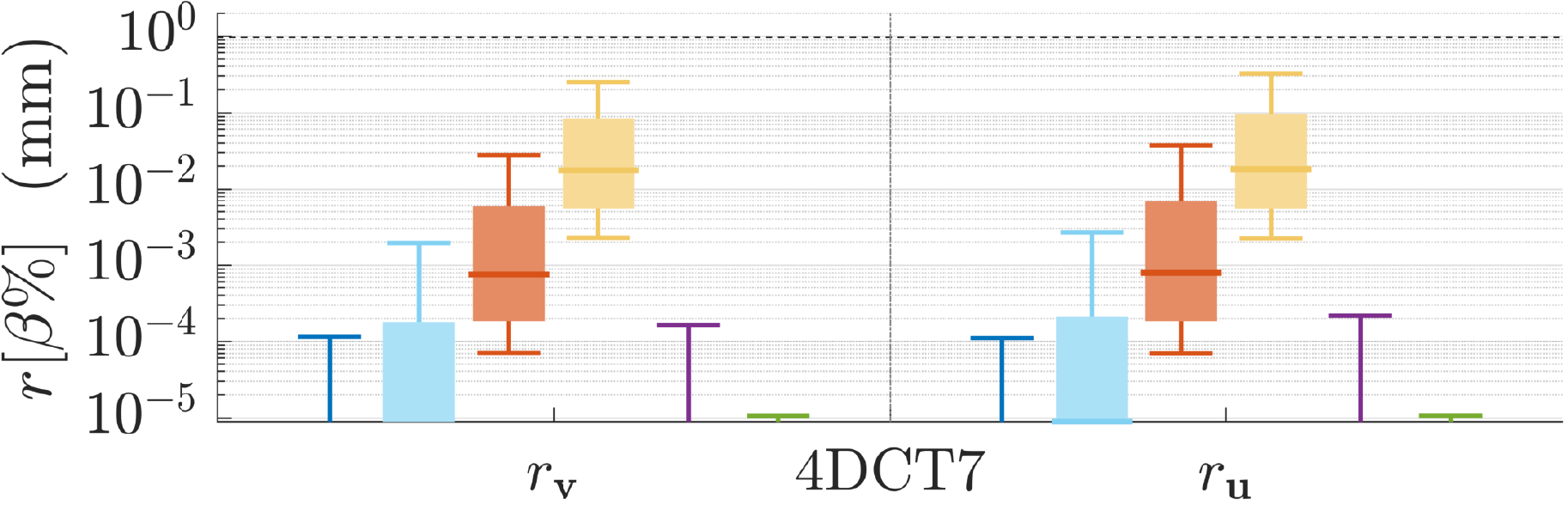}}%
  \hfill
  % 
  % ==================== COPD1
  % 
  \subfloat{%[\label{fig:residual-percentiles-boxplots-copd1}][COPD1]{%
    \includegraphics[width=\halftwocol]{%
      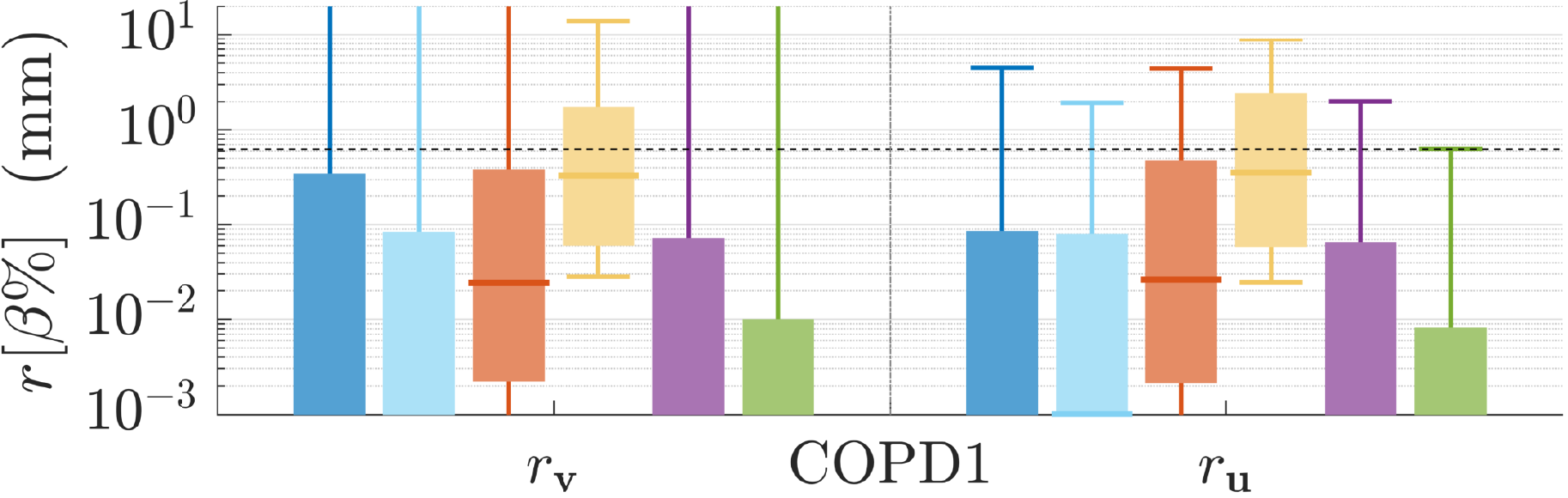}}%
  \\
  %
  % ==================== COPD4
  % 
  \subfloat{%[\label{fig:residual-percentiles-boxplots-copd4}][COPD4]{%
    \includegraphics[width=\halftwocol]{%
      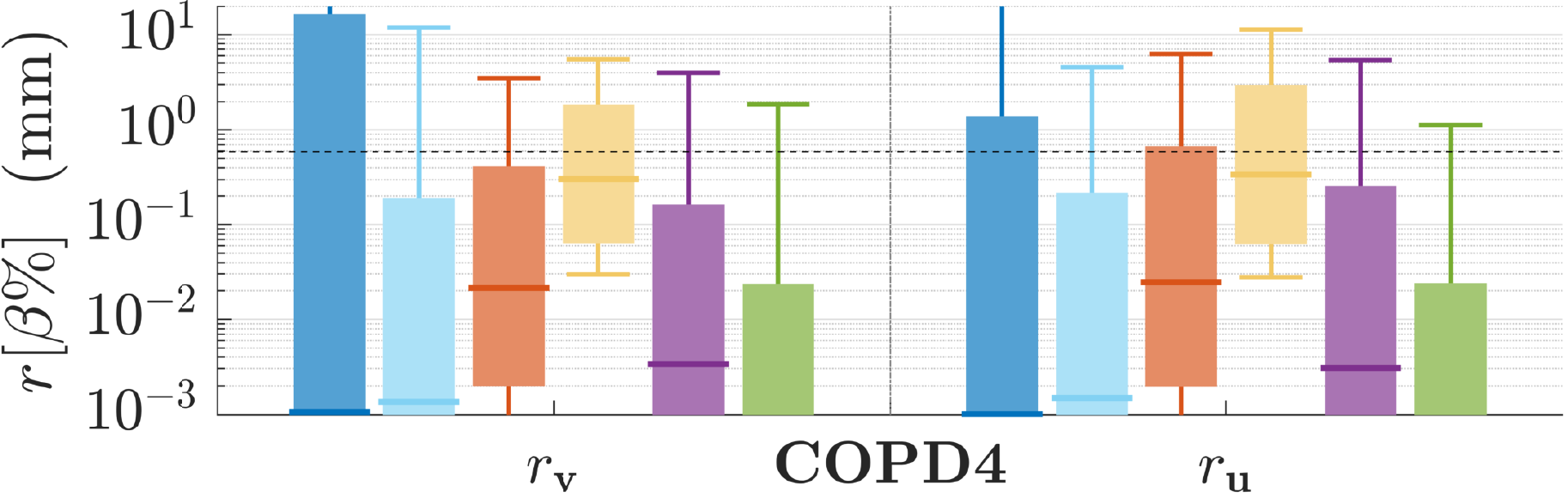}}%
  \hfill
  % 
  % ==================== COPD5
  % 
  \subfloat{%[\label{fig:residual-percentiles-boxplots-copd5}][COPD5]{%
    \includegraphics[width=\halftwocol]{%
      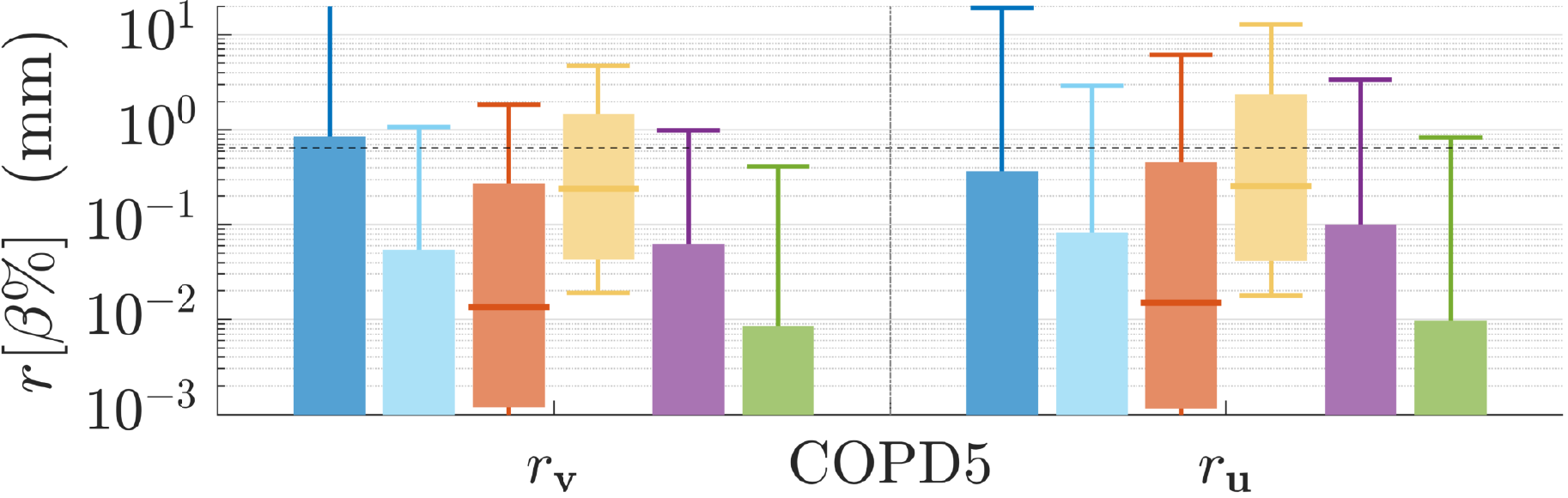}}%
  \\
  %
  % ==================== COPD6
  % 
  \subfloat{%[\label{fig:residual-percentiles-boxplots-copd6}][COPD6]{%
    \includegraphics[width=\halftwocol]{%
      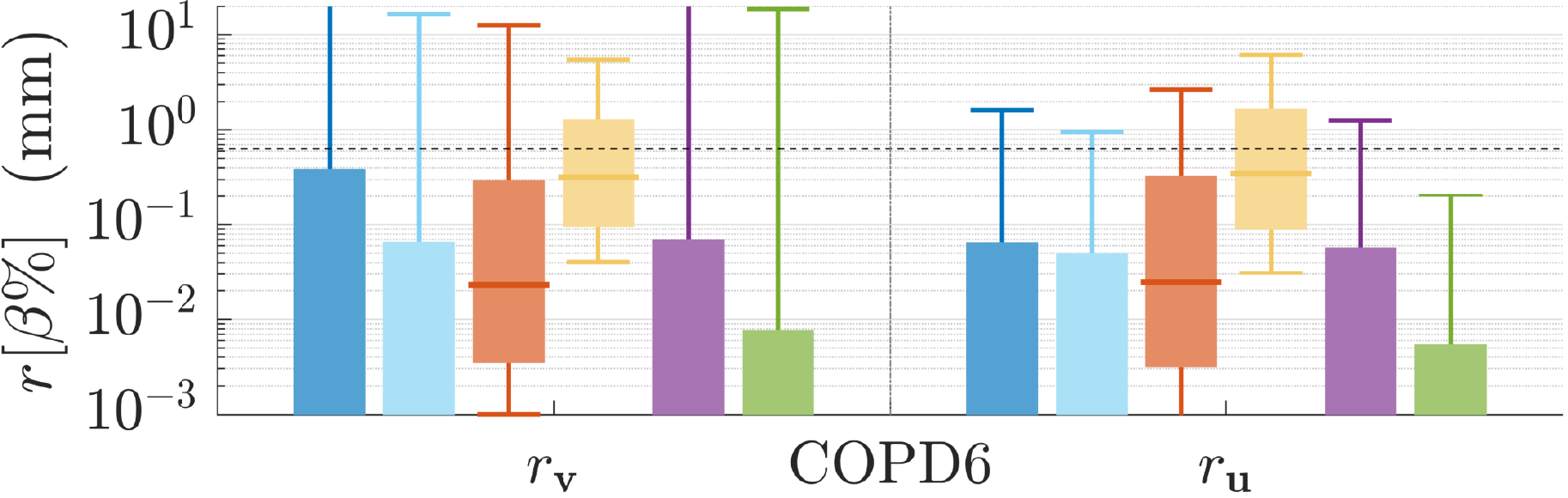}}%
  \hfill
  % 
  % ==================== COPD7
  % 
  \subfloat{%[\label{fig:residual-percentiles-boxplots-copd8}][COPD8]{%
    \includegraphics[width=\halftwocol]{%
      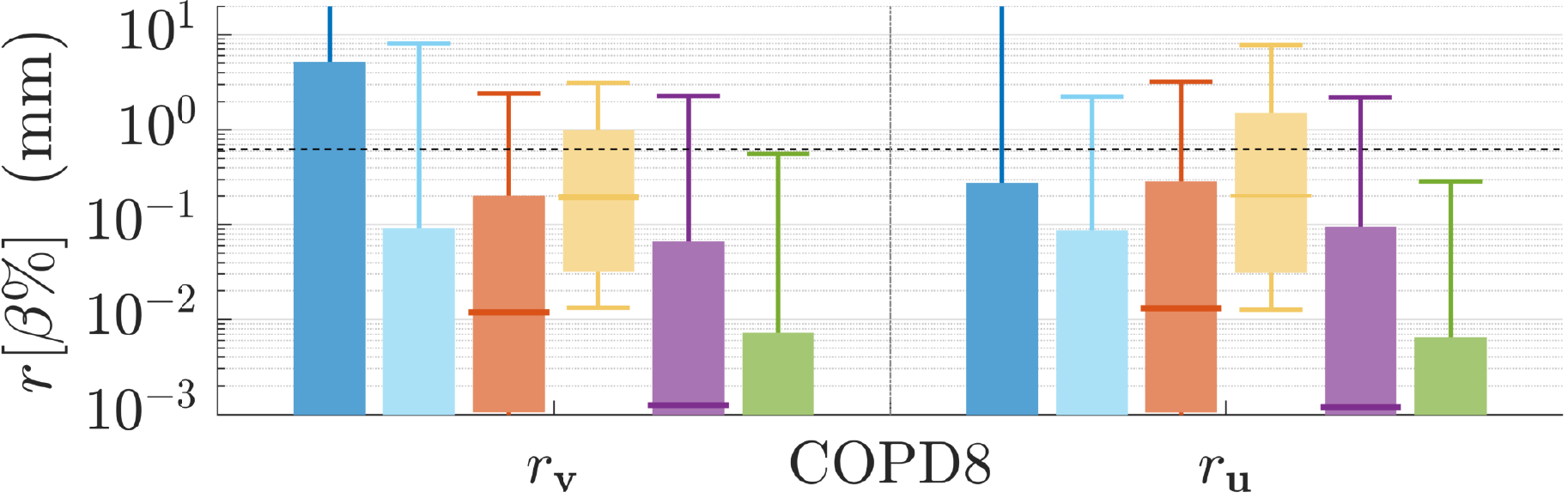}}%
  % 
  % 
  % 
  % ==================== CAPTION
  % 
  \caption{%
    Inverse consistency residuals $r_{\G}(\x)$ and $r_{\F}(\x + \G(\x))$,
    reported in magnitude percentiles \cref{eq:prctile} in $\log$-scale
    over the target domain, at the 10th iteration step with each patient
    and feedback control setting.
    Box whiskers indicate the 2nd and 98th percentiles; the 10th and 90th
    percentiles are at the low and high ends of each box; and the median
    (50th percentile) is marked by a horizontal bar through each box.
    The $y$-axis range is truncated to \unit{[10^{-5}, 2]}{\milli\meter}
    for the 4DCT7 plot, and to \unit{[0.001, 20]}{\milli\meter} for the
    COPD plots; percentile values outside these ranges are not shown.
    The horizontal dashed line in each plot indicates the in-slice
    resolution of the corresponding CT image.}
  \label{fig:residual-percentiles-boxplots}
\end{figure*}

\fi\fi

% ---------- IC residuals
%
\subsection{Inverse consistency residuals}
\label{sec:results-IC-residuals}
\ifdefined\medphys
  \ifdefined\review

\begin{figure*}
  \centering
  % 
  % ==================== LEGEND
  % 
  \subfloat{%
    \includegraphics[width=0.55\linewidth]{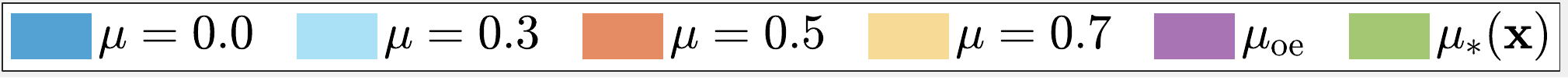}}
  \\
  %
  % ==================== 4DCT7
  % 
  \subfloat{%[\label{fig:residual-percentiles-boxplots-4dct7}][4DCT7]{%
    \includegraphics[width=\halftwocol]{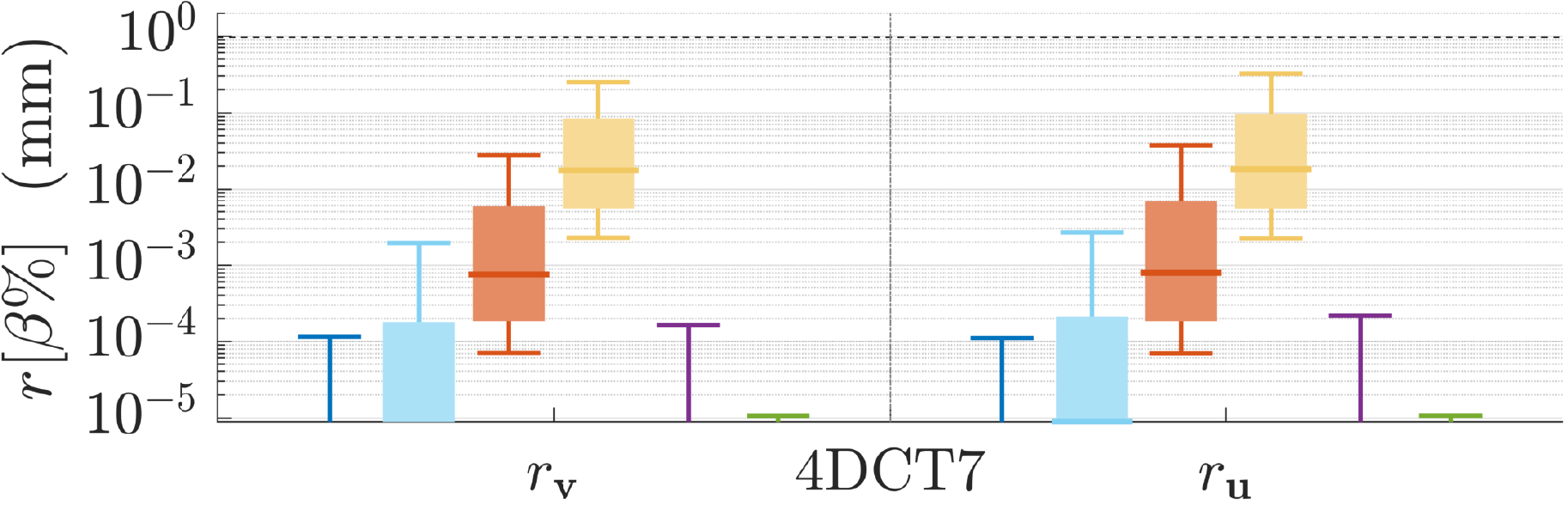}}%
  \hfill
  % 
  % ==================== COPD1
  % 
  \subfloat{%[\label{fig:residual-percentiles-boxplots-copd1}][COPD1]{%
    \includegraphics[width=\halftwocol]{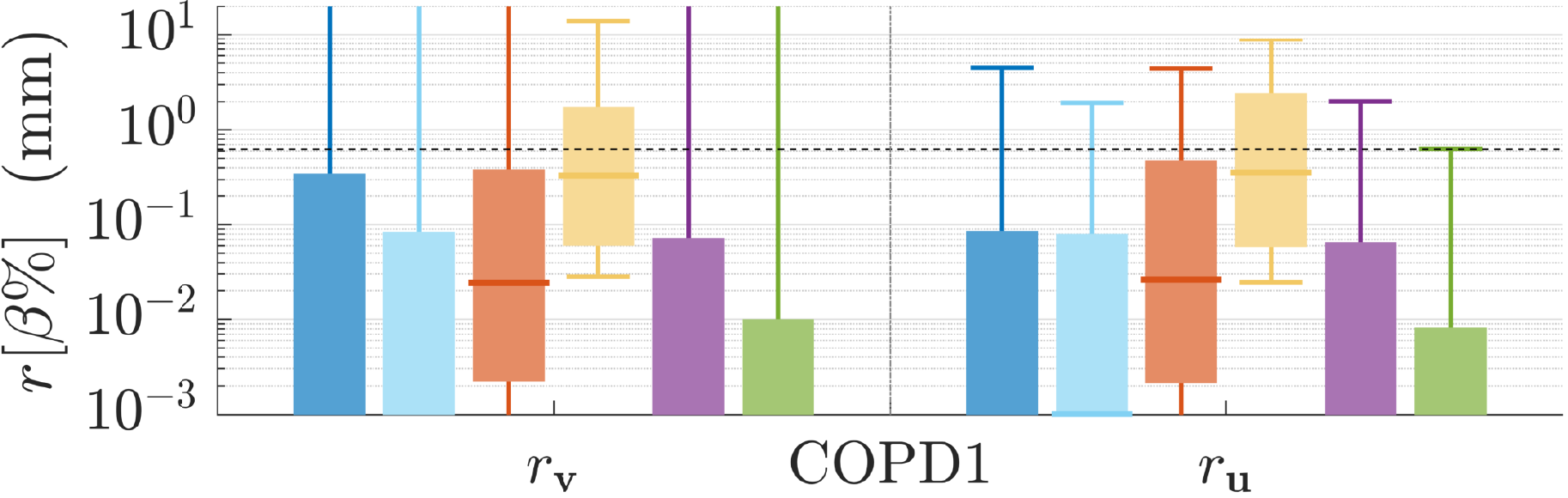}}%
  \\
  %
  % ==================== COPD4
  % 
  \subfloat{%[\label{fig:residual-percentiles-boxplots-copd4}][COPD4]{%
    \includegraphics[width=\halftwocol]{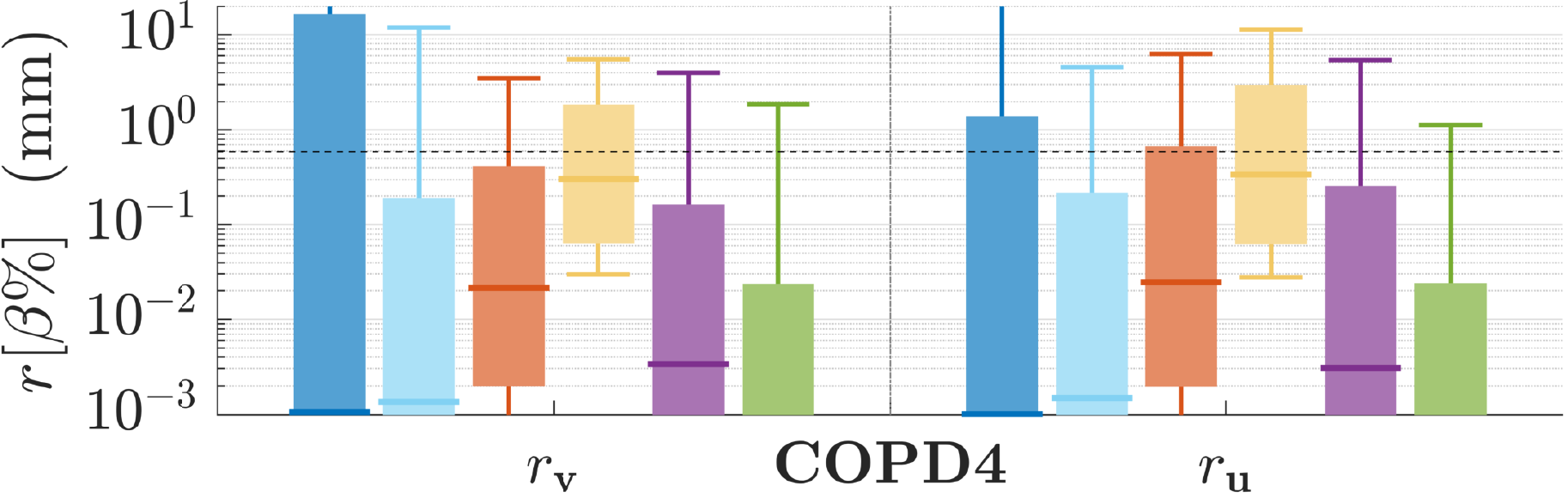}}%
  \hfill
  % 
  % ==================== COPD5
  % 
  \subfloat{%[\label{fig:residual-percentiles-boxplots-copd5}][COPD5]{%
    \includegraphics[width=\halftwocol]{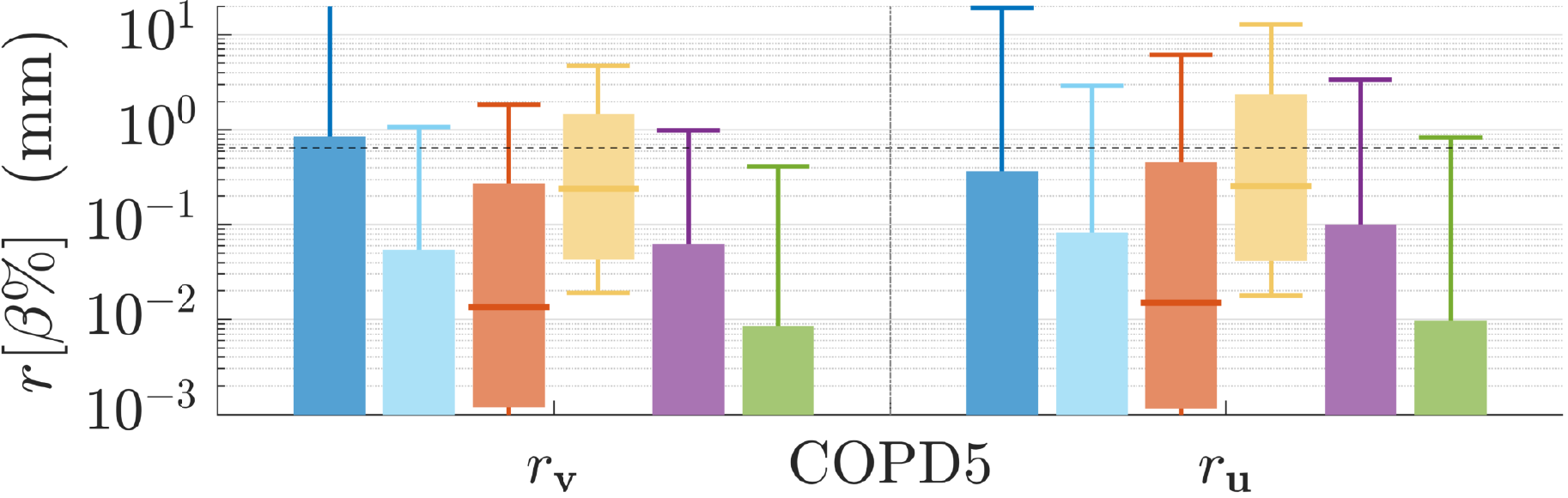}}%
  \\
  %
  % ==================== COPD6
  % 
  \subfloat{%[\label{fig:residual-percentiles-boxplots-copd6}][COPD6]{%
    \includegraphics[width=\halftwocol]{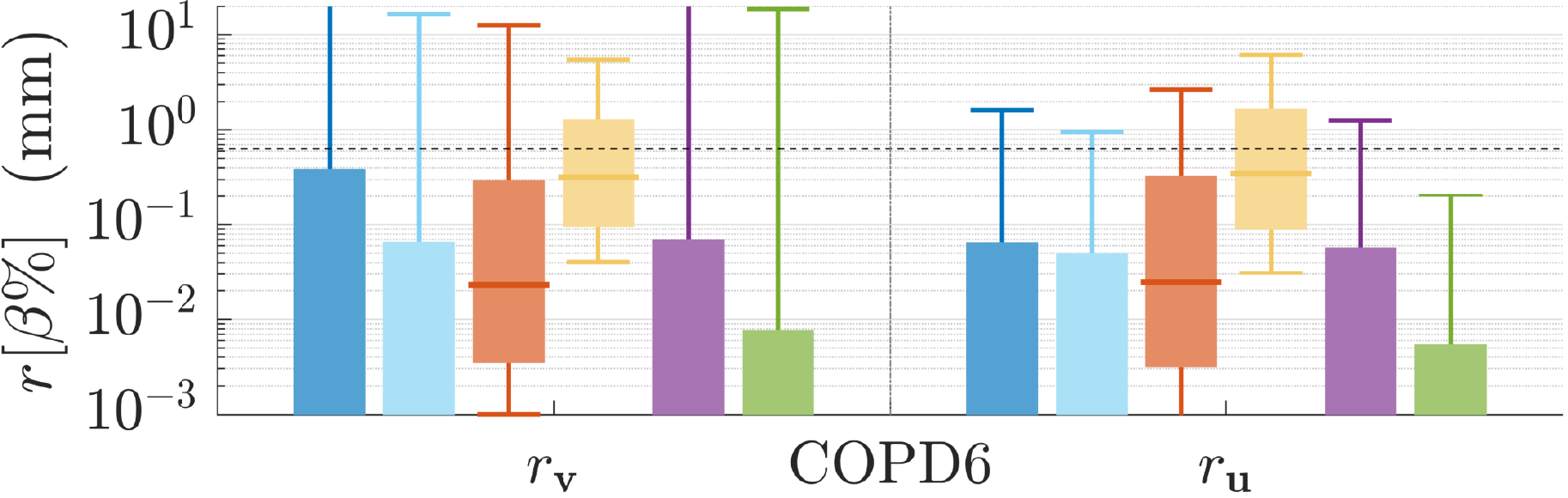}}%
  \hfill
  % 
  % ==================== COPD7
  % 
  \subfloat{%[\label{fig:residual-percentiles-boxplots-copd8}][COPD8]{%
    \includegraphics[width=\halftwocol]{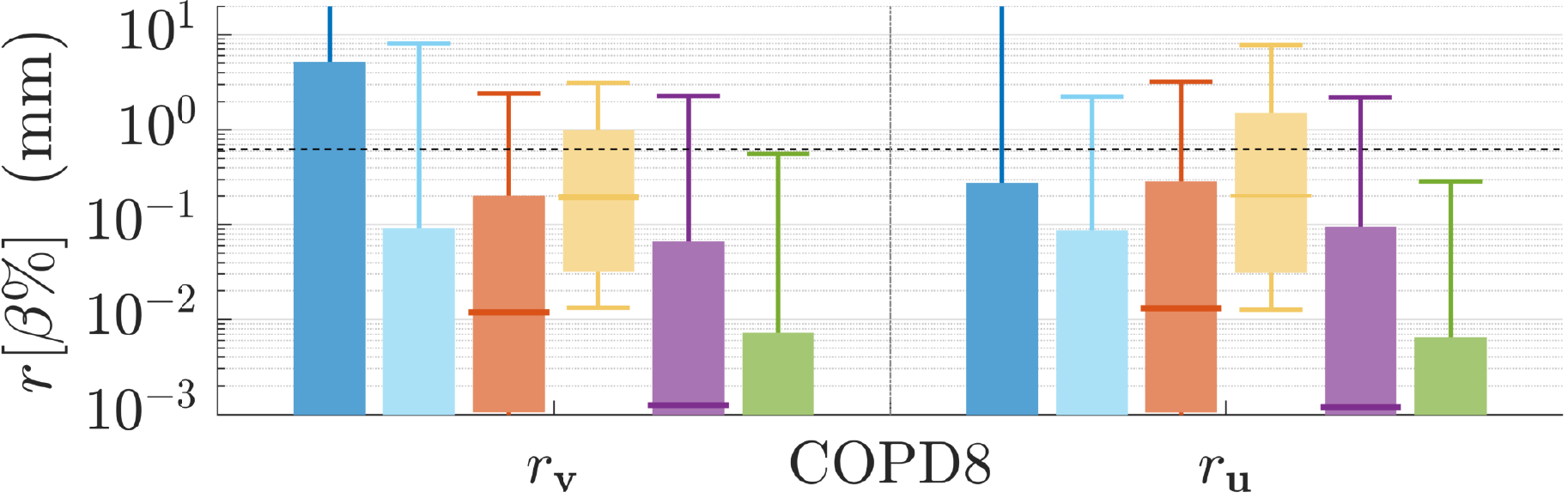}}%
  % 
  % 
  % 
  % ==================== CAPTION
  % 
  \caption{%
    Inverse consistency residuals $r_{\G}(\x)$ and $r_{\F}(\x + \G(\x))$,
    reported in magnitude percentiles \cref{eq:prctile} in $\log$-scale
    over the target domain, at the 10th iteration step with each patient
    and feedback control setting.
    Box whiskers indicate the 2nd and 98th percentiles; the 10th and 90th
    percentiles are at the low and high ends of each box; and the median
    (50th percentile) is marked by a horizontal bar through each box.
    The $y$-axis range is truncated to \unit{[10^{-5}, 2]}{\milli\meter}
    for the 4DCT7 plot, and to \unit{[0.001, 20]}{\milli\meter} for the
    COPD plots; percentile values outside these ranges are not shown.
    The horizontal dashed line in each plot indicates the in-slice
    resolution of the corresponding CT image.}
  \label{fig:residual-percentiles-boxplots}
\end{figure*}

  \fi
\else

\begin{figure*}
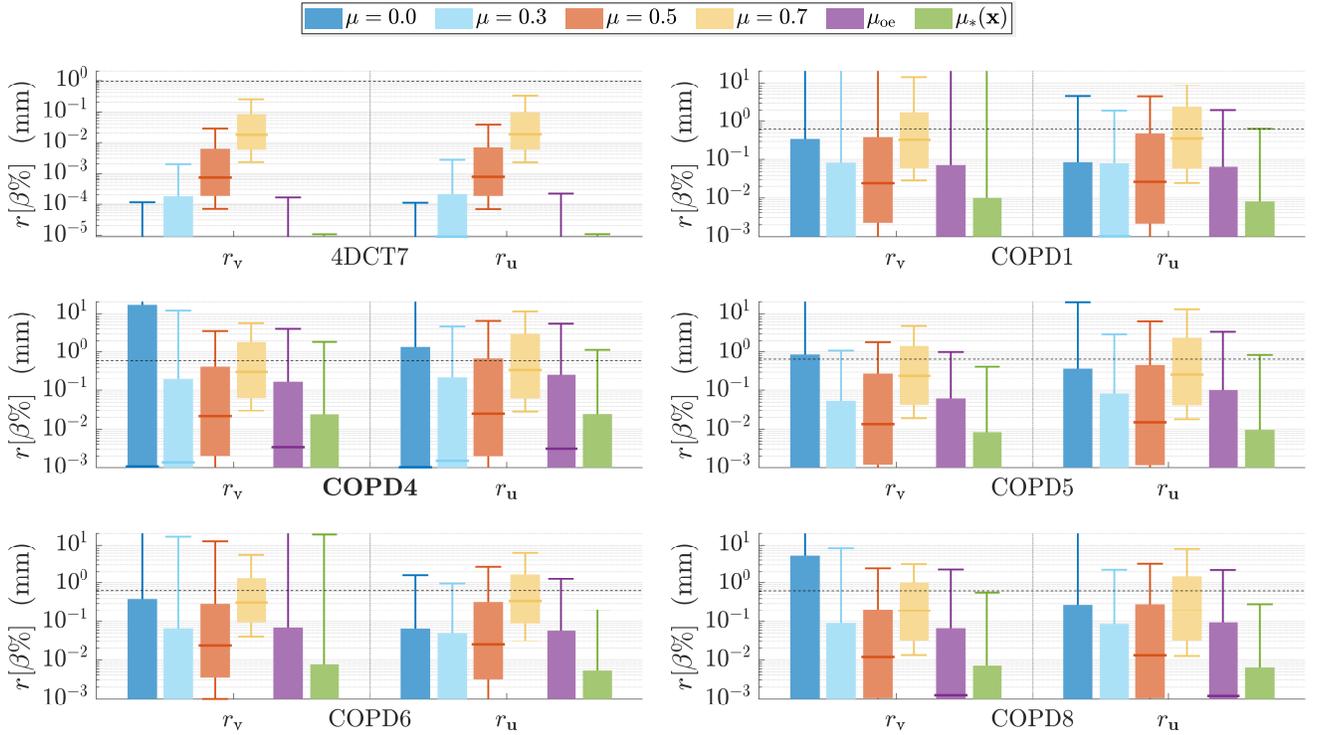

  \centering
  % 
  % ==================== LEGEND
  % 
  \subfloat{%
    \includegraphics[width=0.55\linewidth]{}}
  \\
  %
  % ==================== 4DCT7
  % 
  \subfloat{%[\label{fig:residual-percentiles-boxplots-4dct7}][4DCT7]{%
    \includegraphics[width=\halftwocol]{}}%
  \hfill
  % 
  % ==================== COPD1
  % 
  \subfloat{%[\label{fig:residual-percentiles-boxplots-copd1}][COPD1]{%
    \includegraphics[width=\halftwocol]{}}%
  \\
  %
  % ==================== COPD4
  % 
  \subfloat{%[\label{fig:residual-percentiles-boxplots-copd4}][COPD4]{%
    \includegraphics[width=\halftwocol]{}}%
  \hfill
  % 
  % ==================== COPD5
  % 
  \subfloat{%[\label{fig:residual-percentiles-boxplots-copd5}][COPD5]{%
    \includegraphics[width=\halftwocol]{}}%
  \\
  %
  % ==================== COPD6
  % 
  \subfloat{%[\label{fig:residual-percentiles-boxplots-copd6}][COPD6]{%
    \includegraphics[width=\halftwocol]{}}%
  \hfill
  % 
  % ==================== COPD7
  % 
  \subfloat{%[\label{fig:residual-percentiles-boxplots-copd8}][COPD8]{%
    \includegraphics[width=\halftwocol]{}}%
  % 
  % 
  % 
  % ==================== CAPTION
  % 
  \caption{%
    Inverse consistency residuals $r_{\G}(\x)$ and $r_{\F}(\x + \G(\x))$,
    reported in magnitude percentiles \cref{eq:prctile} in $\log$-scale
    over the target domain, at the 10th iteration step with each patient
    and feedback control setting.
    Box whiskers indicate the 2nd and 98th percentiles; the 10th and 90th
    percentiles are at the low and high ends of each box; and the median
    (50th percentile) is marked by a horizontal bar through each box.
    The $y$-axis range is truncated to \unit{[10^{-5}, 2]}{\milli\meter}
    for the 4DCT7 plot, and to \unit{[0.001, 20]}{\milli\meter} for the
    COPD plots; percentile values outside these ranges are not shown.
    The horizontal dashed line in each plot indicates the in-slice
    resolution of the corresponding CT image.}
  \label{fig:residual-percentiles-boxplots}
\end{figure*}

\fi

We summarize in \cref{fig:residual-percentiles-boxplots} post-inversion
evaluation of the iteration \cref{eq:fpim-scaling-control} by $6$ different
feedback control schemes, for each of the $6$ DVFs, in terms of the two IC
residuals in \cref{subsub:error-residual-relation}.  The residuals are
calculated at the 10th iteration step, measured by magnitude,
%$r(\x) = \sqrt{ r_{\LR}^2(\x) + r_{\AP}^2(\x) + r_{\SI}^2(\x) }$
and reported as percentile values in $\log$-scale.

With the 4DCT7 DVF, IC residuals are below the in-slice resolution with
every scheme.  They are larger with $\mu = 0.5$ due to slower convergence,
as expected by the pre-inversion evaluation, and even larger with
$\mu = 0.7$.

With the COPD DVFs, we note first that substantial spatial variation of the
residuals is observed for each DVF.
The residuals at the 50th and lower percentiles are well below
\unit{1}{\milli\meter} with all control settings.  How far below
\unit{1}{\milli\meter} is related to the convergence pace, since the
contraction region for each control setting covers much more than half of
the domain.  The iteration with constant $\mu=0.7$ is the slowest.
We observe the following among the $4$ constant, non-adaptive control
schemes.  With the same initialization, residual reduction does not
correlate linearly with the values of $\mu$: specifically for the 5 COPD
DVFs, the $\mu=0$ and $\mu=0.7$ settings are inferior to $\mu=0.3$ and
$\mu=0.5$ at high percentiles.  The latter two settings are comparable at
the 98th percentile, and the scheme with $\mu=0.3$ yields smaller residuals
up to the 90th.
No single constant-value scheme is as good as the adaptive scheme with
alternating values.
The residuals by the scheme with spatially variant values are much smaller
than by other schemes, by roughly an order of magnitude or more at the 90th
percentile.  More remarkably, the 98th percentile residuals are reduced to
below \unit{2}{\milli\meter} by only $10$ iteration steps with spatially
variant control.
%

% fig-xcat-residual_overlay.tex

\begin{figure*}[p]
  \centering
  %
  % ============================== COLORBAR
  % 
  \hspace*{1.6em}
  \includegraphics[width=0.97\linewidth]{%
    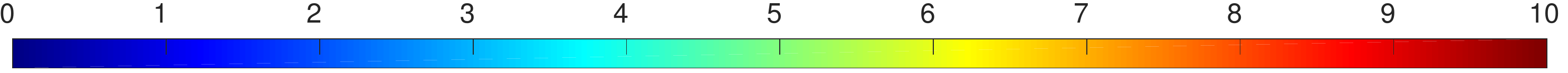}%
  \\[-0.5em]
  %
  % ============================== IC RESIDUAL R_V
  % 
  \begin{minipage}[t]{\ichalftwocolleft}
  %
  % ---------- SLICE PLANES
  %
  \begin{minipage}[t]{\ictextmarginleft}
    \centering
    \text{}
  \end{minipage}%
  \hfill
  \begin{minipage}{\icwidthImgleft}
    \centering
    \begin{minipage}[t]{\thirdAxialPR}
      \centering
      axial
    \end{minipage}%
    \hfill
    \begin{minipage}[t]{\thirdCoronalPR}
      \centering
      coronal
    \end{minipage}%
    \hfill
    \begin{minipage}[t]{\thirdSagittalPR}
      \centering
      sagittal
    \end{minipage}%
  \end{minipage}%
  \\
  % 
  % ---------- IMAGES (mu = 0)
  % 
  % 
  \begin{minipage}{\ictextmarginleft}
    \centering
    \rotatebox{90}{\normalsize$\mu=0$}
  \end{minipage}
  \hfill
  \begin{minipage}{\icwidthImgleft}
    \vspace*{-\valignImgSkip}
    \centering
    \subfloat{%
      \includegraphics[width=\thirdAxialPR]{%
    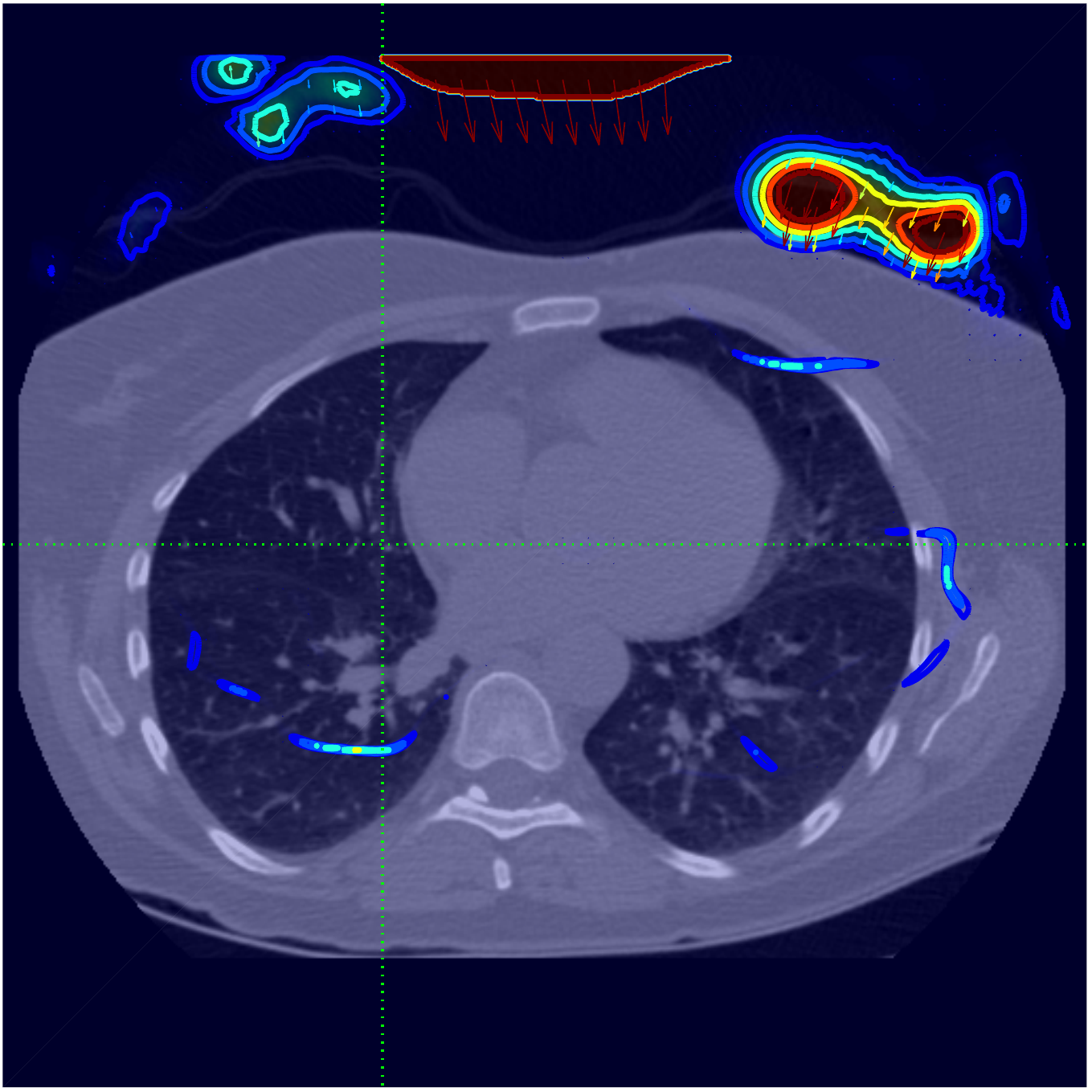}}%
    \hfill
    \subfloat{%
      \includegraphics[width=\thirdCoronalPR]{%
    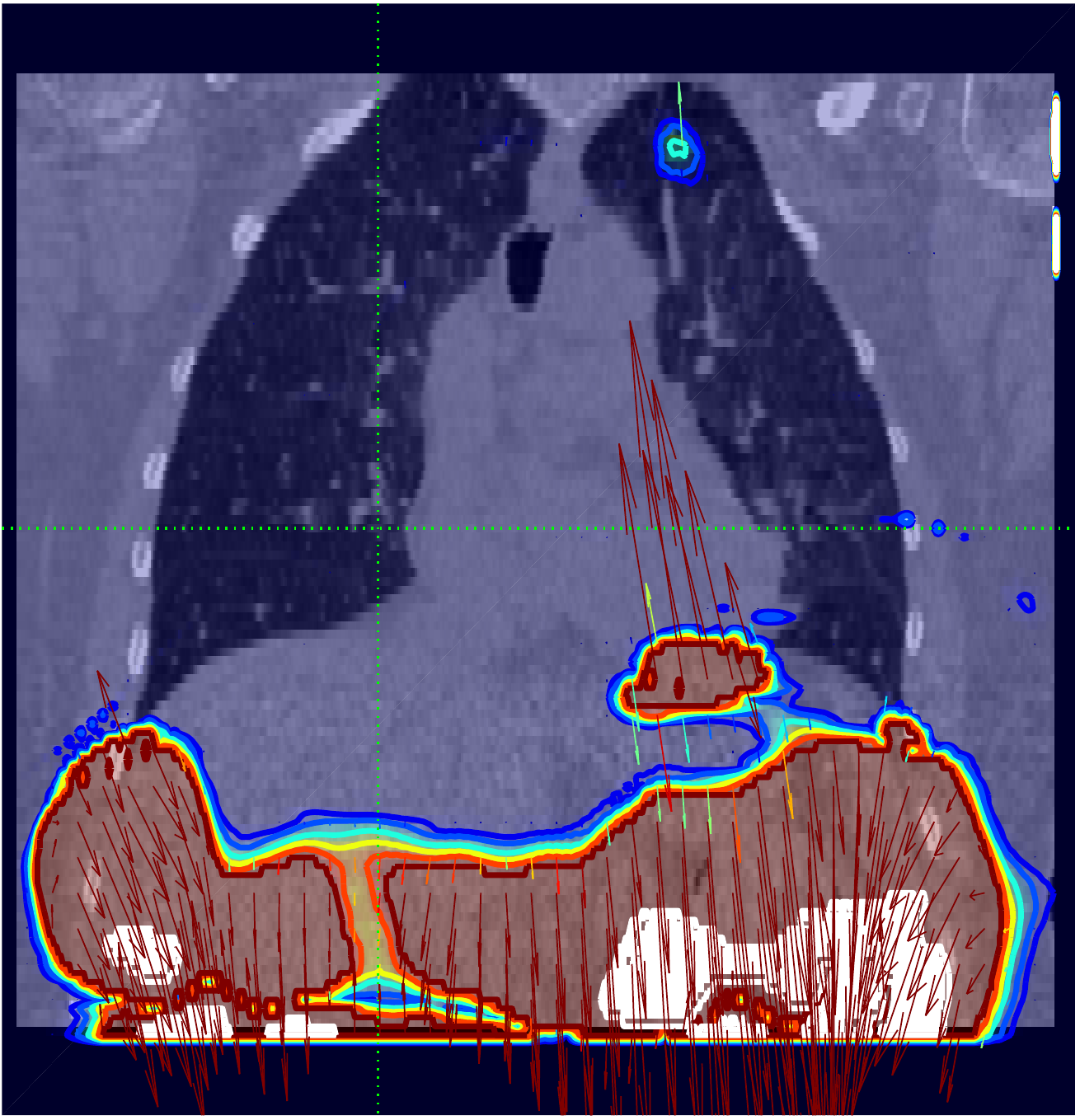}}% 
    \hfill
    \subfloat{%
      \includegraphics[width=\thirdSagittalPR]{%
    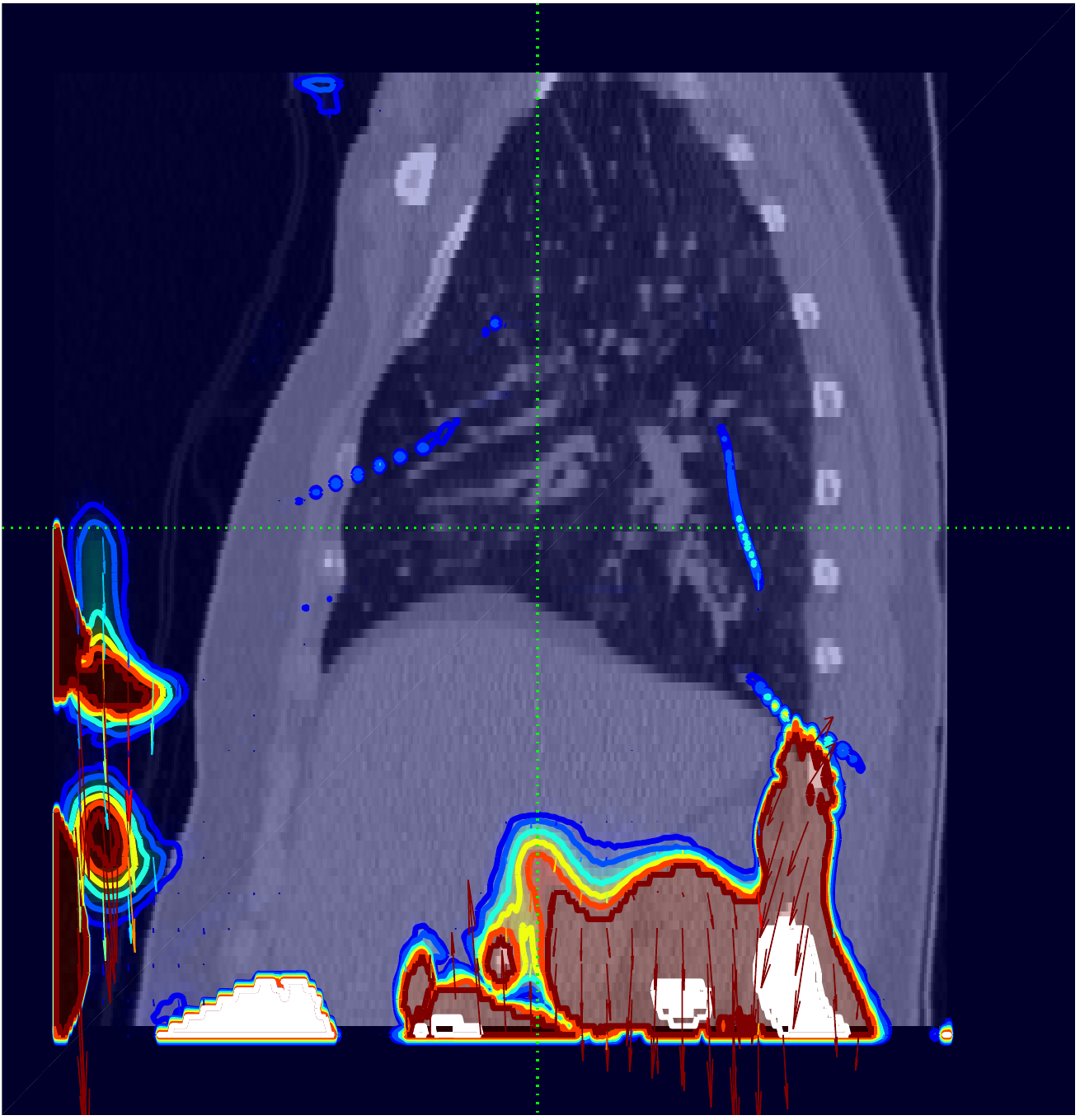}}% 
  \end{minipage}%
  \\
  %
  % ---------- IMAGES (mu_mu = 0.5)
  % 
  \begin{minipage}{\ictextmarginleft}
    \centering
    \rotatebox{90}{\normalsize$\mu=0.5$}
  \end{minipage}
  \hfill
  \begin{minipage}{\icwidthImgleft}
    \vspace*{-\valignImgSkip}
    \centering
    \subfloat{%
      \includegraphics[width=\thirdAxialPR]{%
    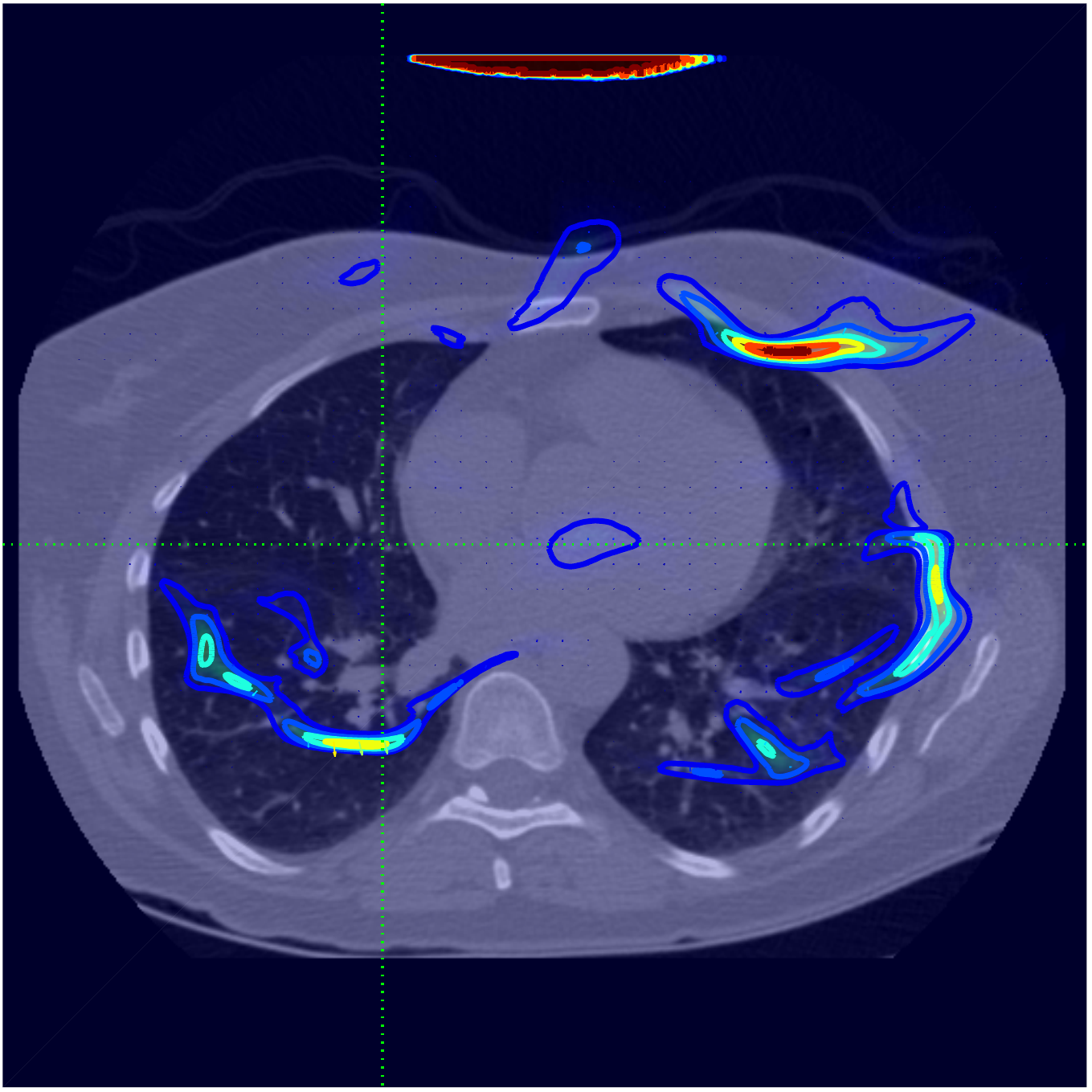}}%
    \hfill
    \subfloat{%
      \includegraphics[width=\thirdCoronalPR]{%
    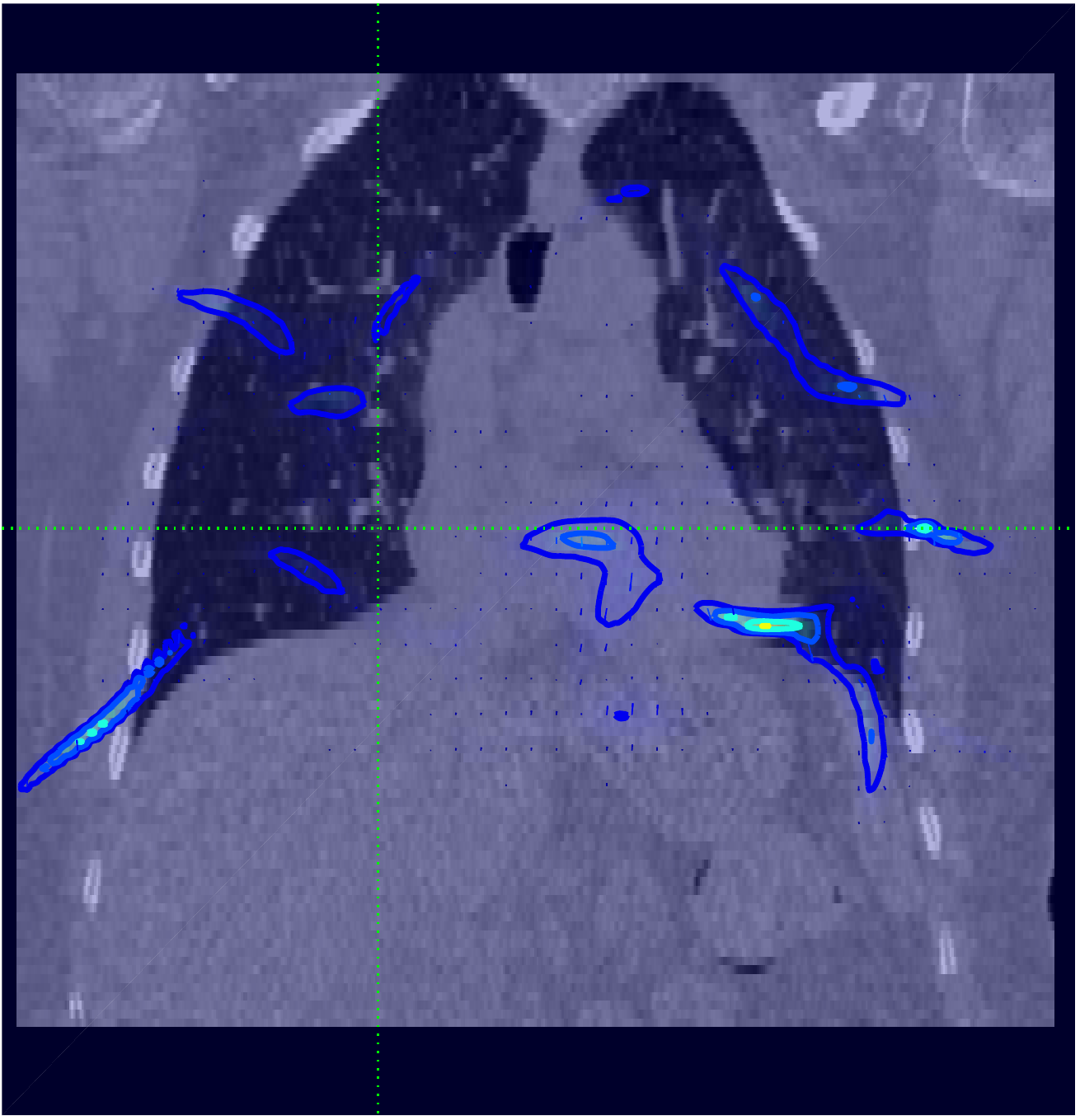}}% 
    \hfill
    \subfloat{%
      \includegraphics[width=\thirdSagittalPR]{%
    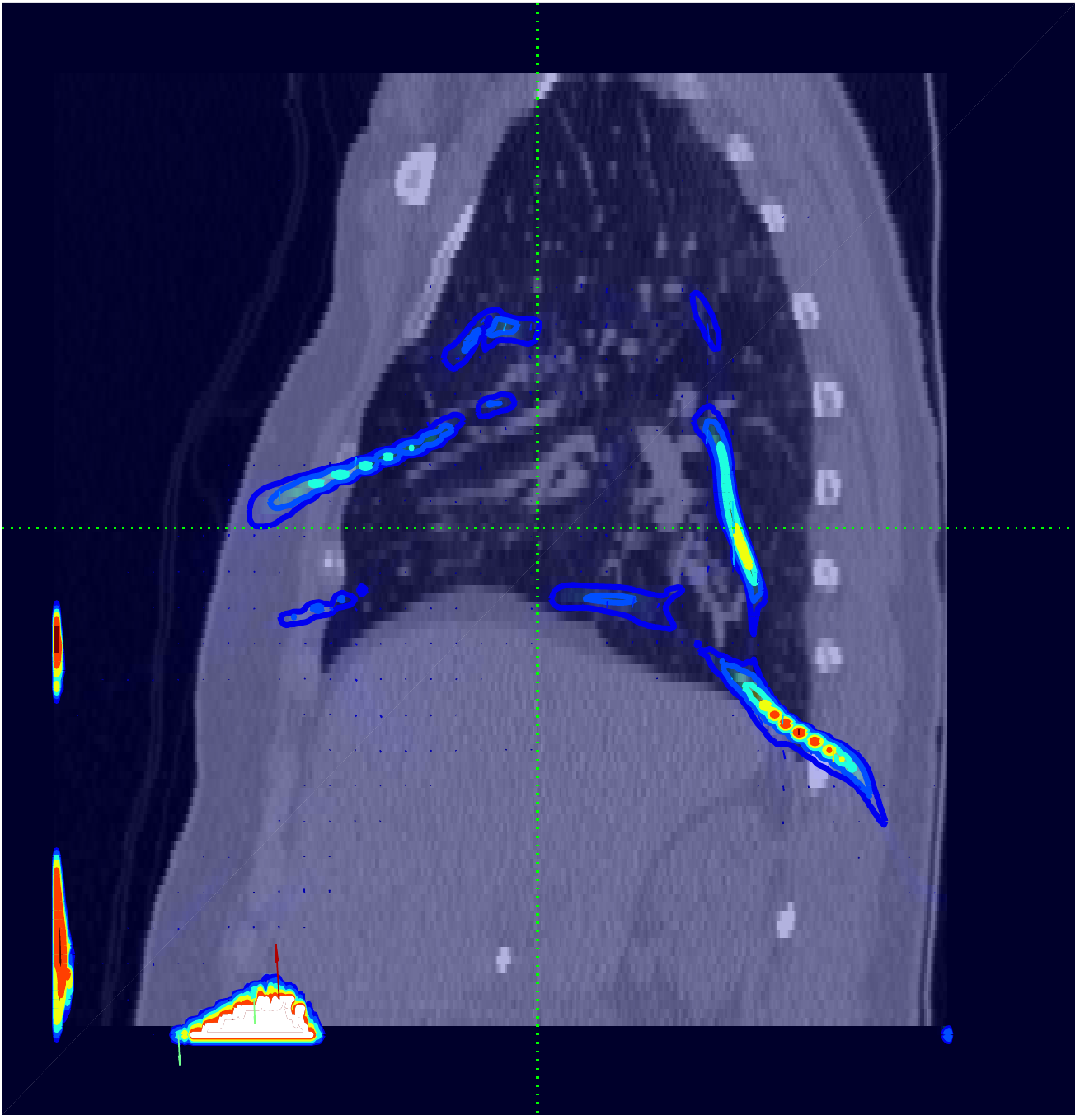}}% 
  \end{minipage}%
  \\
  %
  % ---------- IMAGES (mu_mu = oe)
  % 
  \begin{minipage}{\ictextmarginleft}
    \centering
    \rotatebox{90}{\normalsize$\mu_{\text{oe}}$}
  \end{minipage}
  \hfill
  \begin{minipage}{\icwidthImgleft}
    \vspace*{-\valignImgSkip}
    \centering
    \subfloat{%
      \includegraphics[width=\thirdAxialPR]{%
    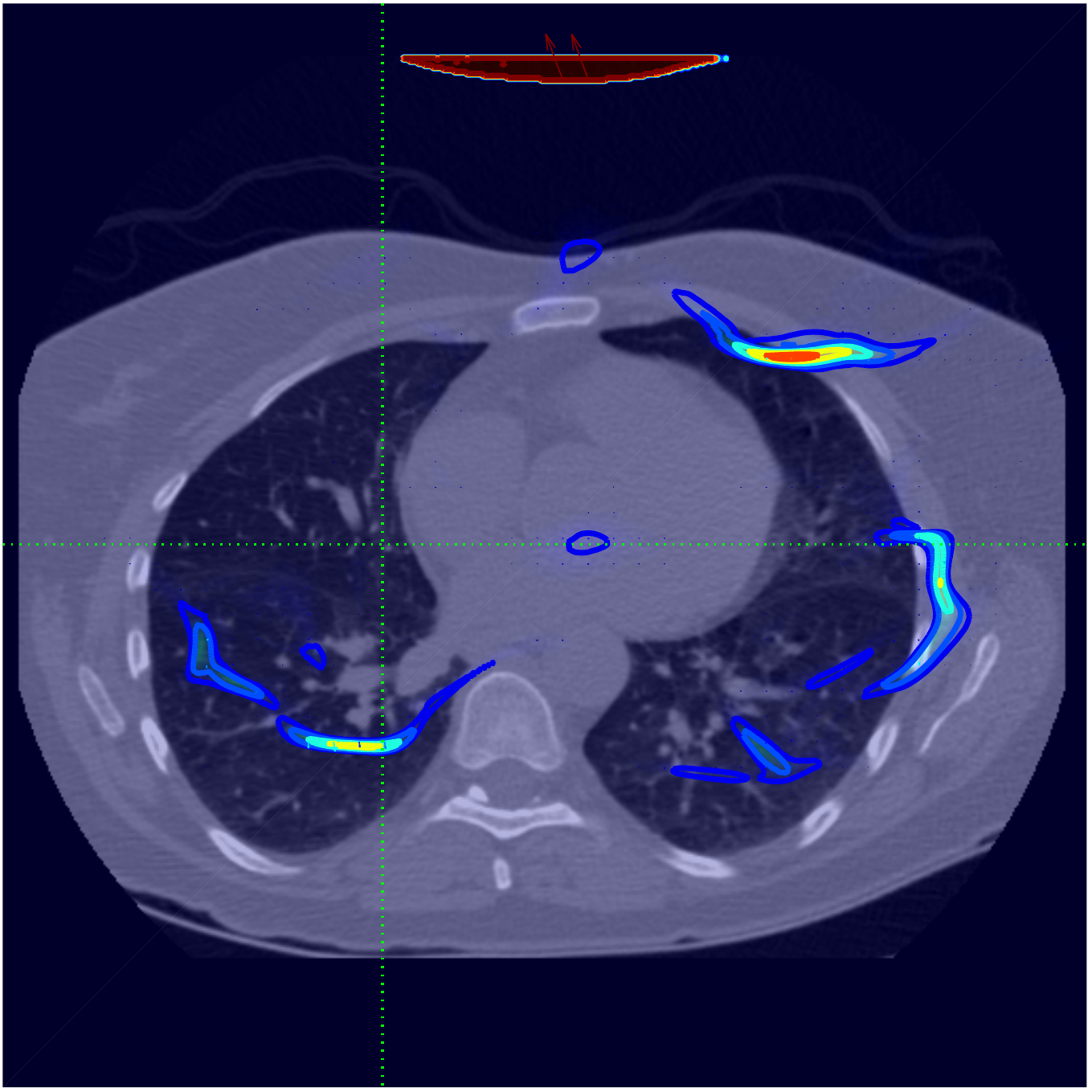}}%
    \hfill
    \subfloat{%
      \includegraphics[width=\thirdCoronalPR]{%
    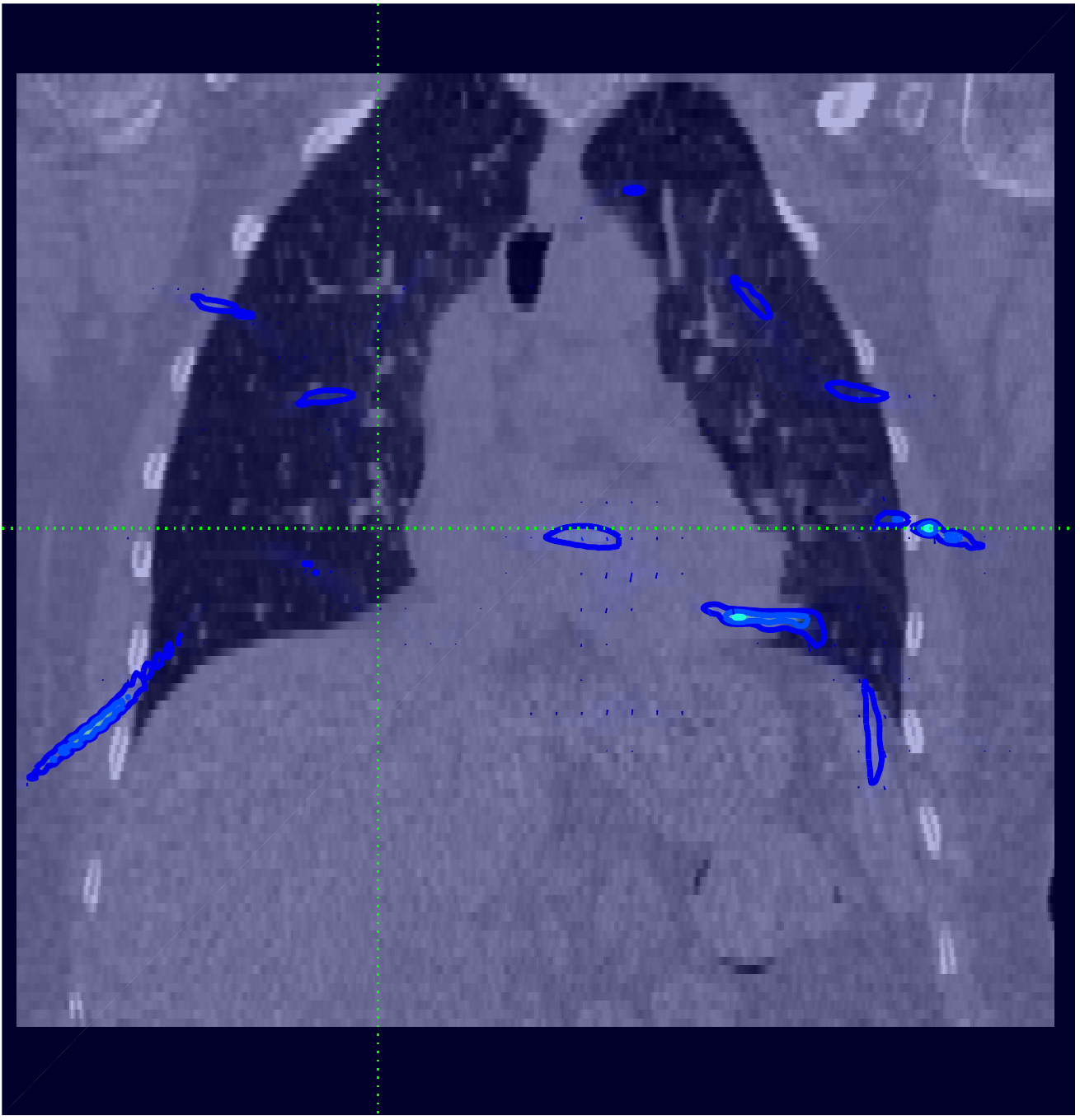}}% 
    \hfill
    \subfloat{%
      \includegraphics[width=\thirdSagittalPR]{%
    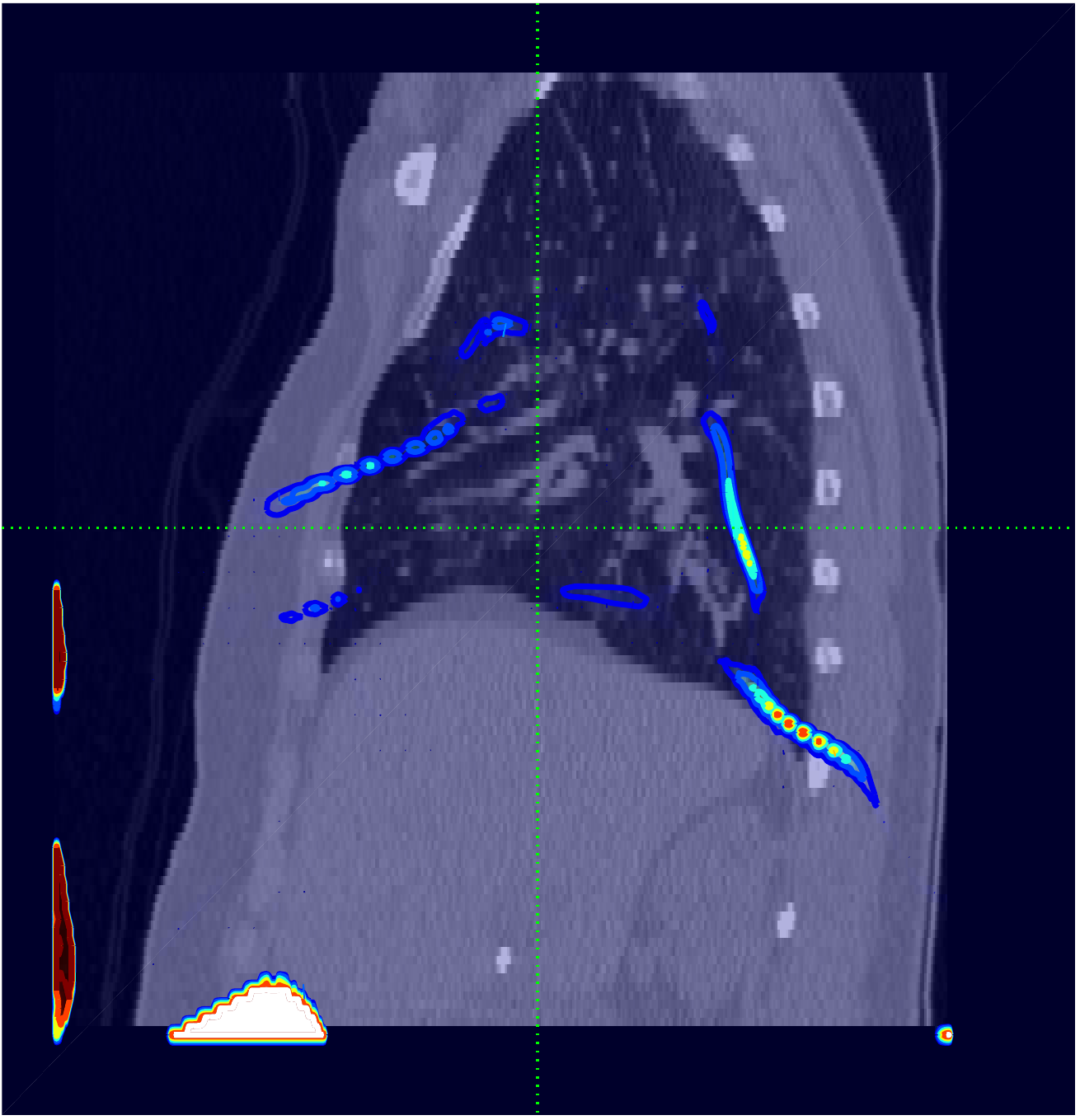}}% 
  \end{minipage}%
  \\
  %
  % ---------- IMAGES (mu_map)
  % 
  \begin{minipage}{\ictextmarginleft}
    \centering
    \rotatebox{90}{\normalsize$\mu_{*}(\x)$}
  \end{minipage}
  \hfill \hfill \hfill
  \begin{minipage}{\icwidthImgleft}
    \vspace*{-\valignImgSkip}
    \centering
    \subfloat{%
      \includegraphics[width=\thirdAxialPR]{%
    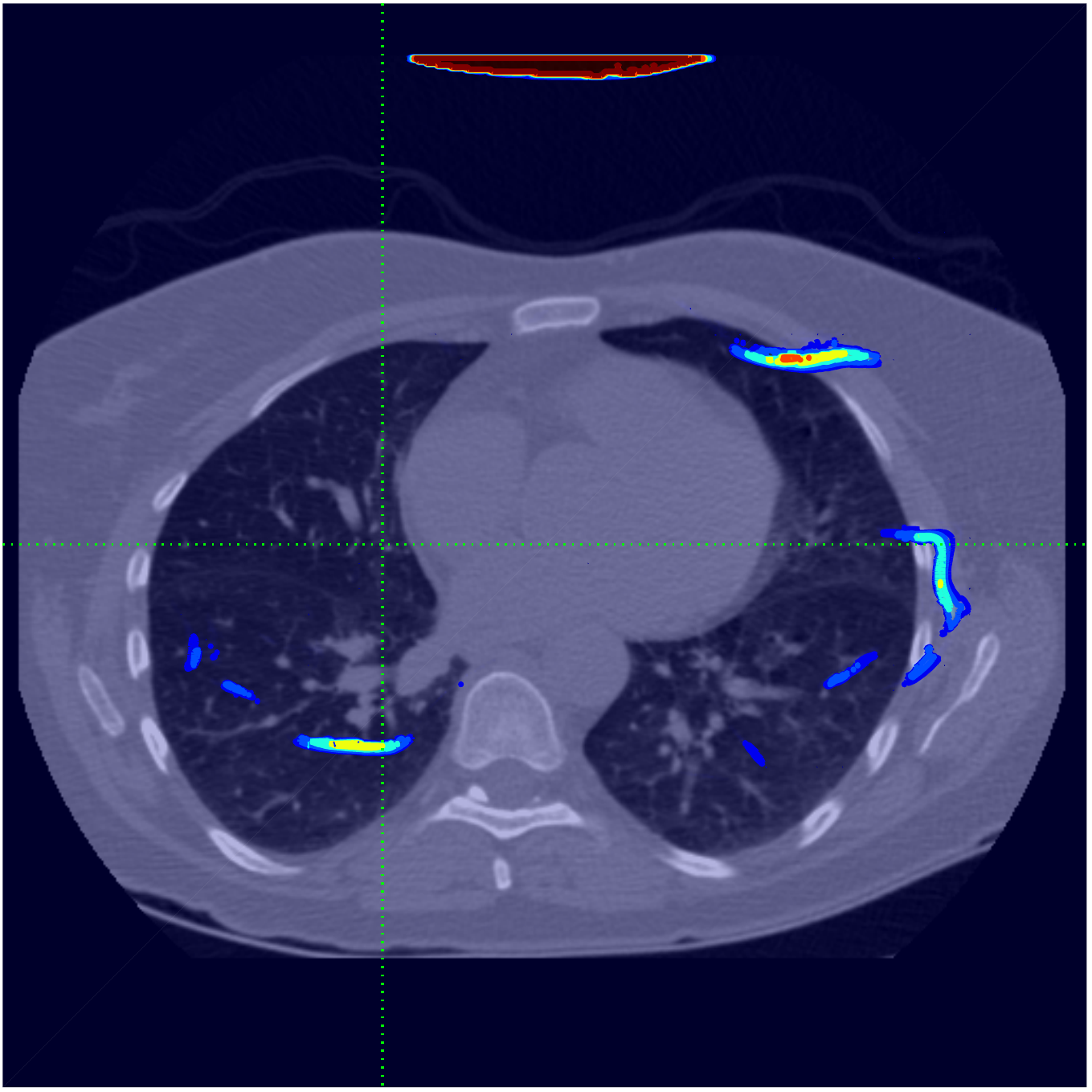}}%
    \hfill
    \subfloat{%
      \includegraphics[width=\thirdCoronalPR]{%
    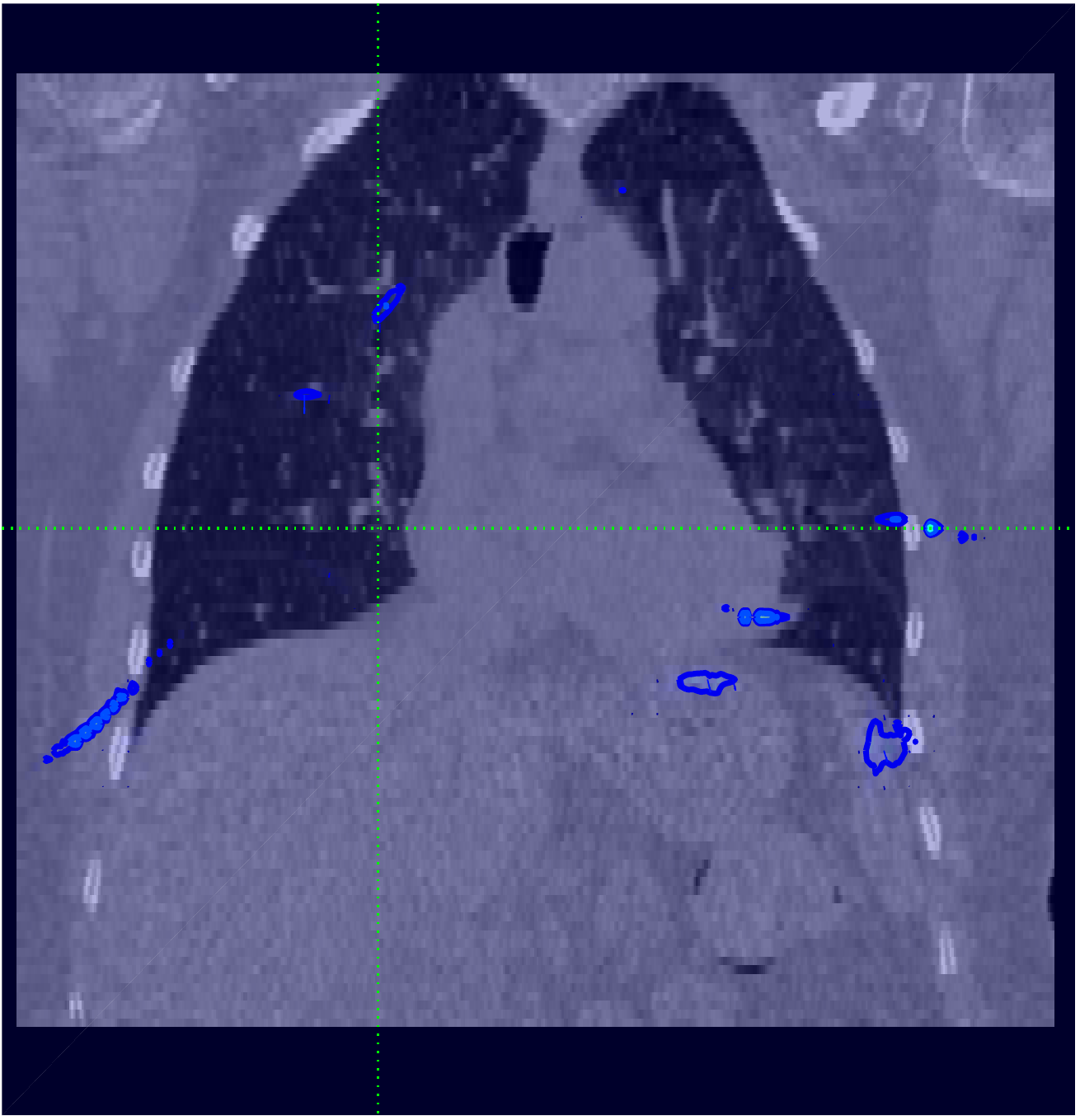}}% 
    \hfill
    \subfloat{%
      \includegraphics[width=\thirdSagittalPR]{%
    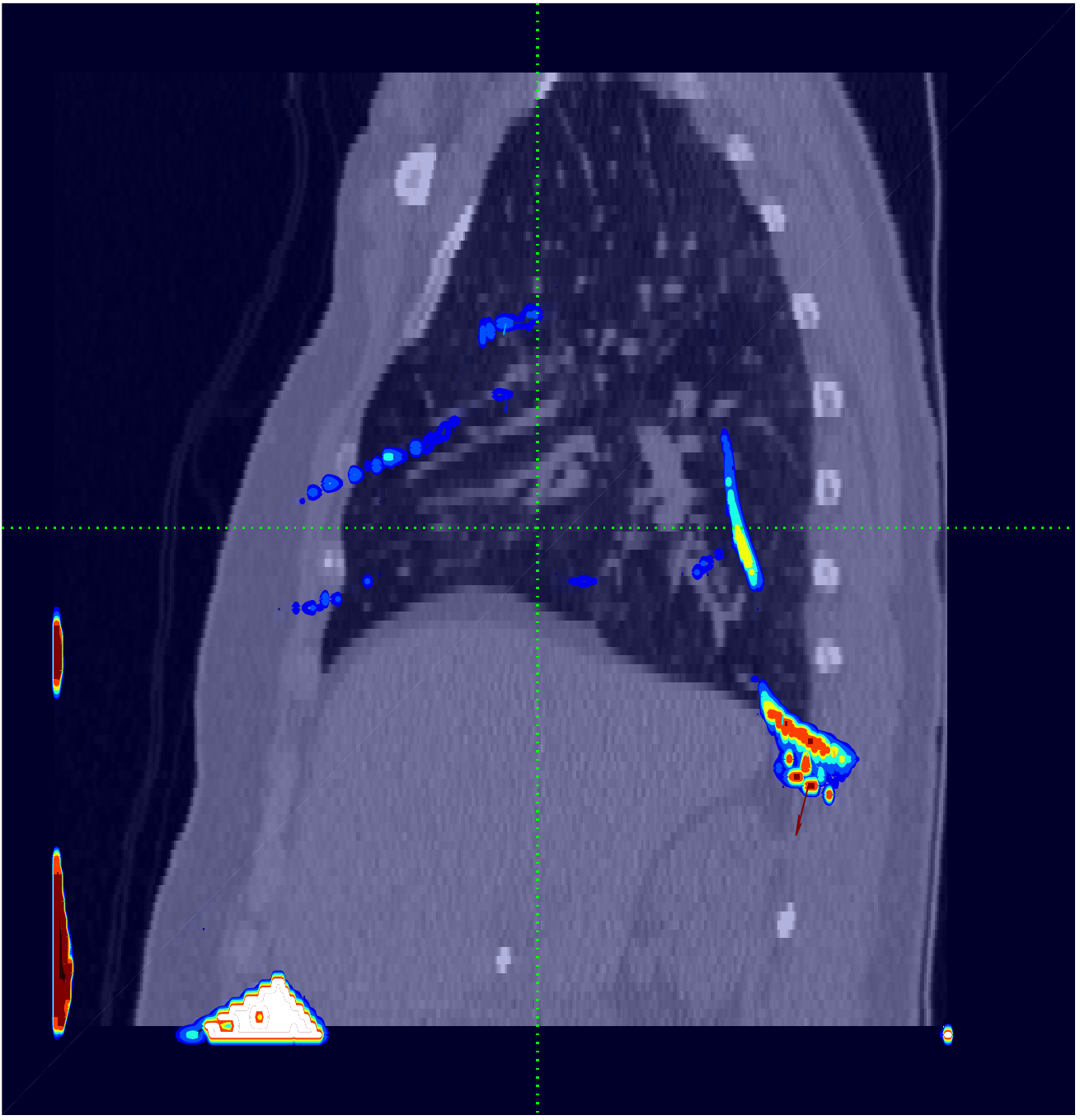}}% 
  \end{minipage}
  \\
  \begin{minipage}{\ictextmarginleft}
    \centering
    \text{}
  \end{minipage}
  \begin{minipage}{\icwidthImgleft}
    \centering\vspace*{2pt}
    \normalsize (a) $\R_{\G}$
  \end{minipage}
\end{minipage}%
\hfill
%
%
% ============================== IC RESIDUAL R_U
%
%
\begin{minipage}[t]{\ichalftwocolright}%
  \centering
  %
  % ---------- SLICE PLANES
  % 
  %
  \begin{minipage}{\icwidthImgright}%
    \centering
    \begin{minipage}[t]{\thirdAxialPR}
      \centering
      axial
    \end{minipage}%
    \hfill
    \begin{minipage}[t]{\thirdCoronalPR}
      \centering
      coronal
    \end{minipage}%
    \hfill
    \begin{minipage}[t]{\thirdSagittalPR}
      \centering
      sagittal
    \end{minipage}%
  \end{minipage}%
  \\
  %
  % ---------- IMAGES (mu_mu = 0)
  % 
  % 
  \begin{minipage}{\icwidthImgright}
    \vspace*{-\valignImgSkip}
    \centering
    \subfloat{%
      \includegraphics[width=\thirdAxialPR]{%
    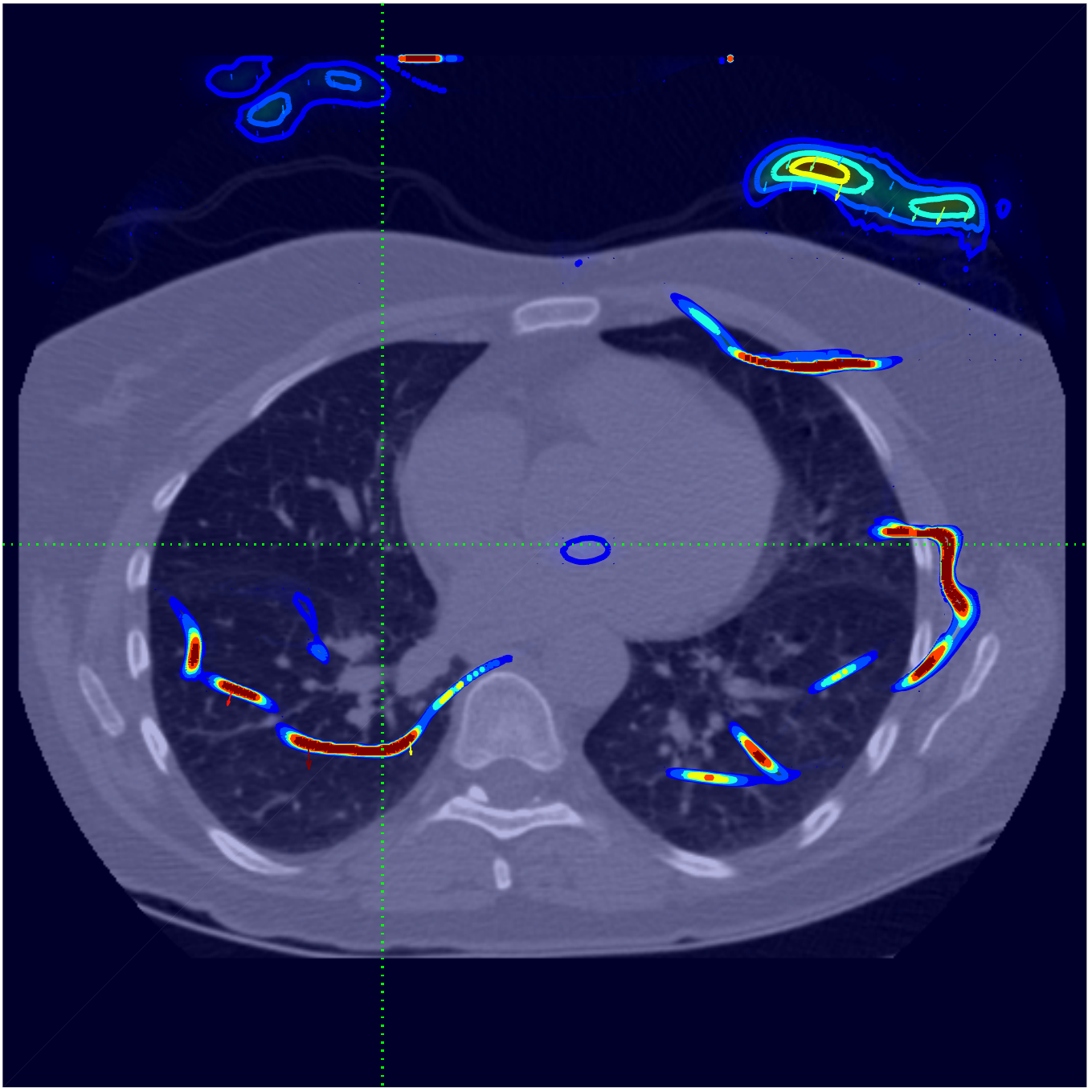}}%
    \hfill
    \subfloat{%
      \includegraphics[width=\thirdCoronalPR]{%
      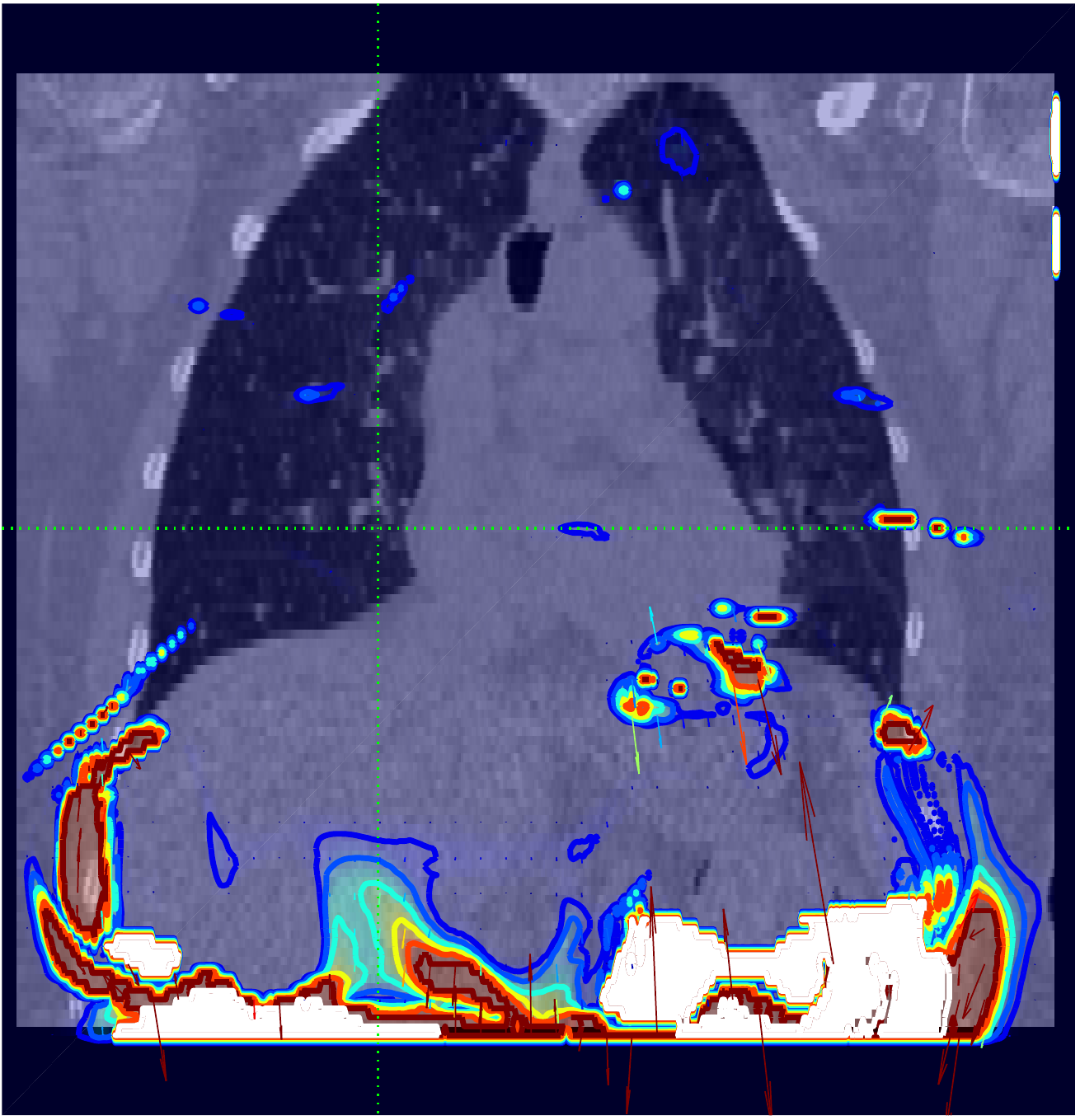}}% 
    \hfill
    \subfloat{%
      \includegraphics[width=\thirdSagittalPR]{%
     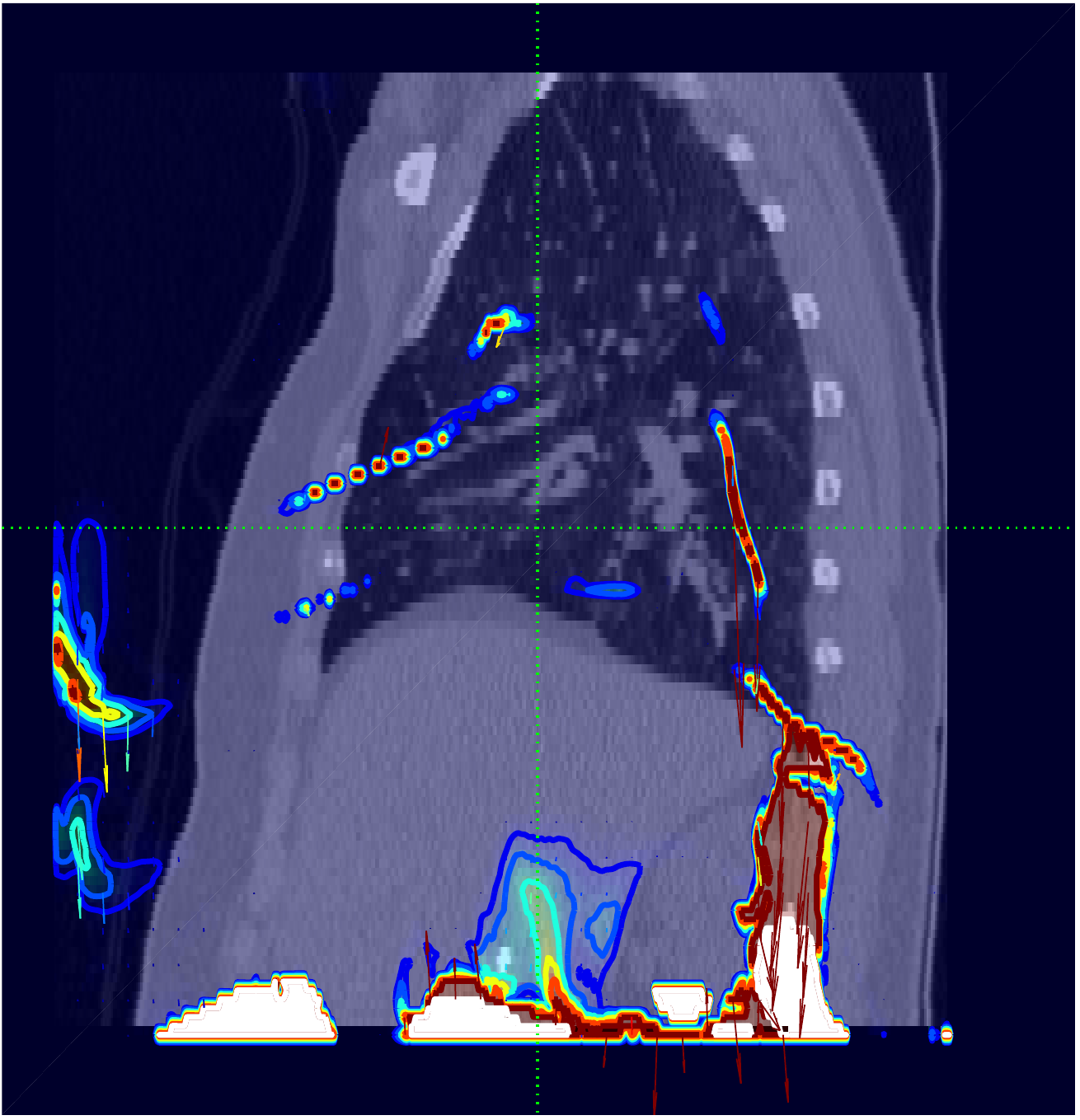}}% 
  \end{minipage}%
  \hfill\hfill\hfill
  \\
  % 
  % ---------- IMAGES (mu_mu = 0.5)
  % 
  \begin{minipage}{\icwidthImgright}
    \vspace*{-\valignImgSkip}
    \centering
    \subfloat{%
      \includegraphics[width=\thirdAxialPR]{%
    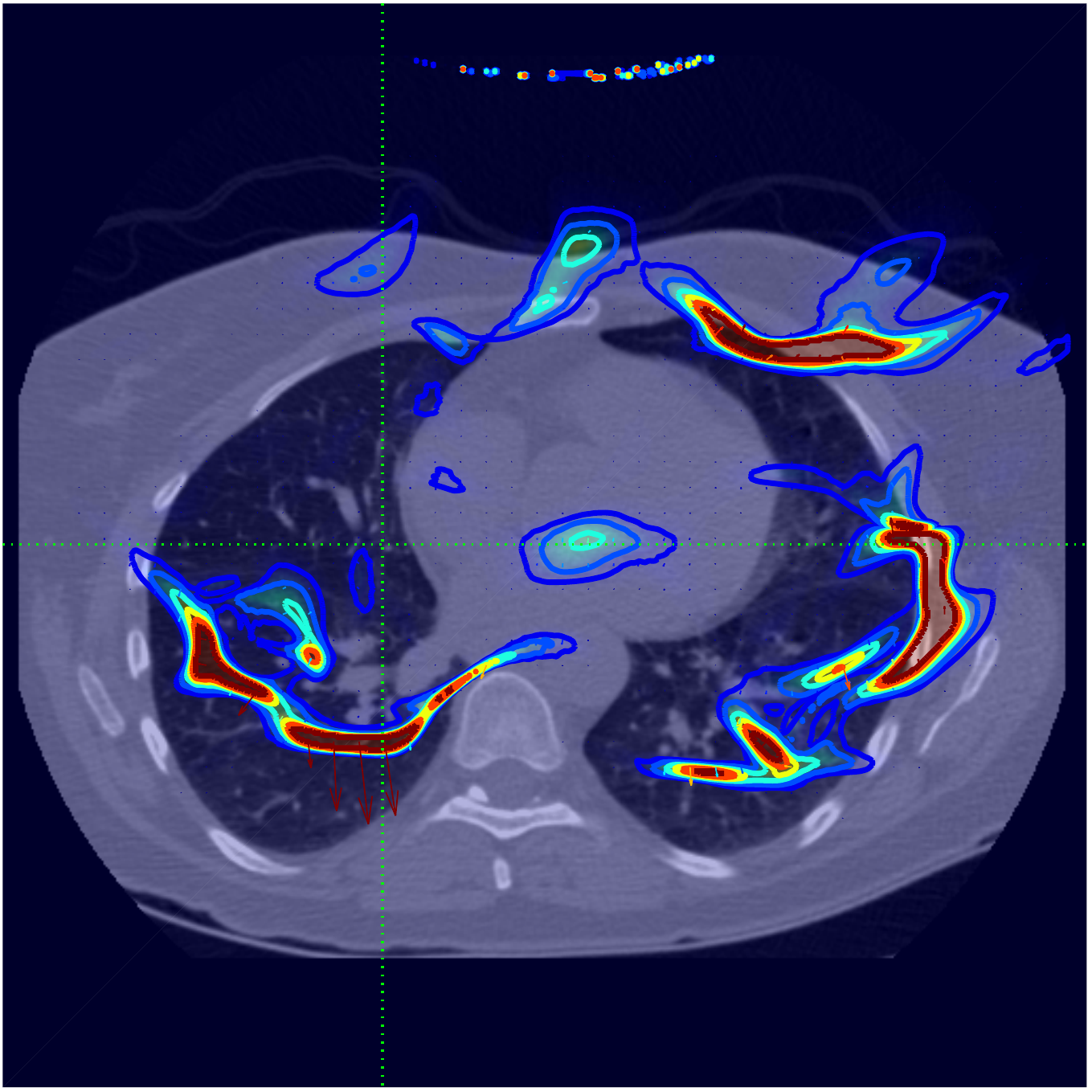}}%
    \hfill
    \subfloat{%
      \includegraphics[width=\thirdCoronalPR]{%
      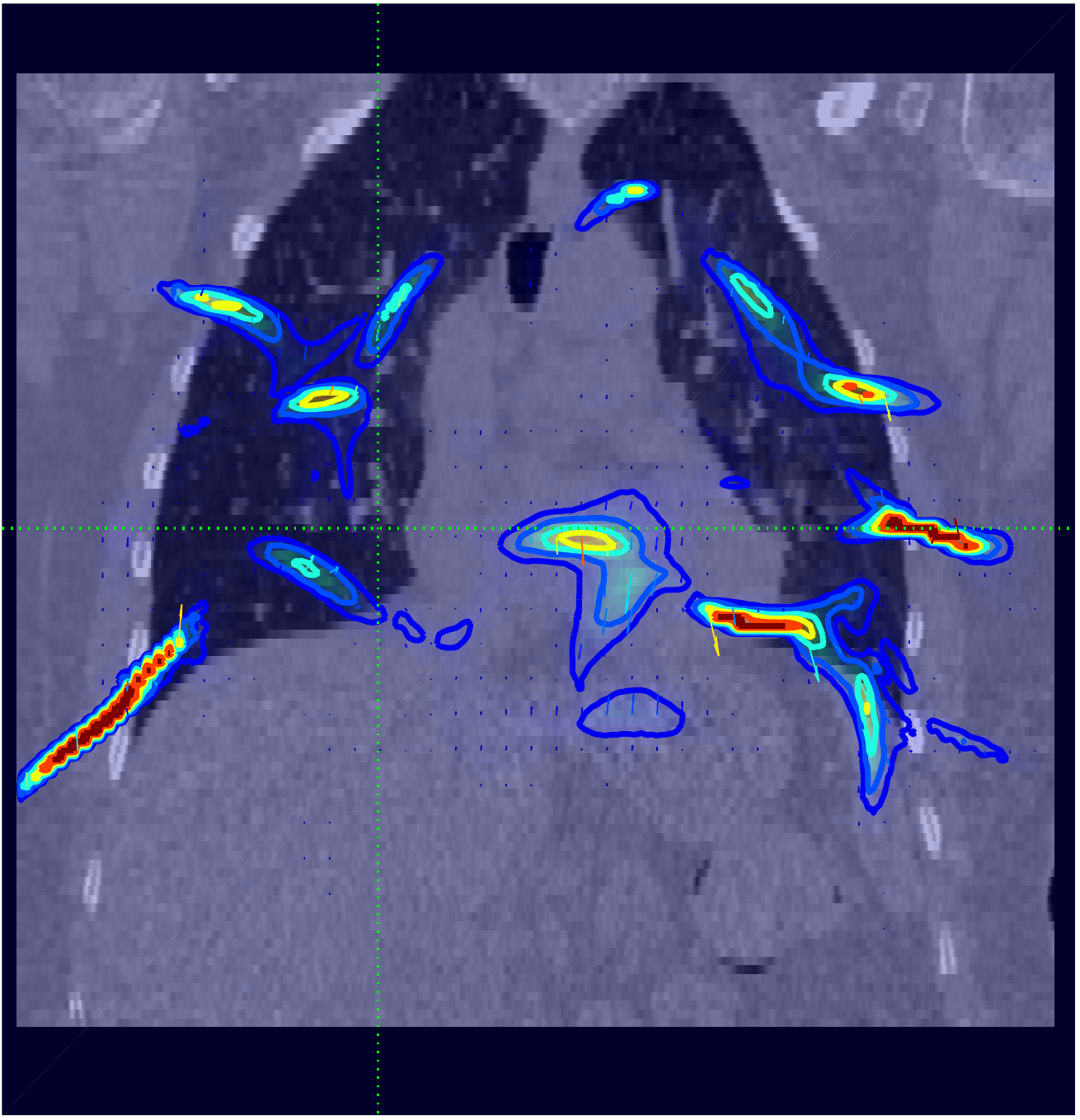}}% 
    \hfill
    \subfloat{%
      \includegraphics[width=\thirdSagittalPR]{%
     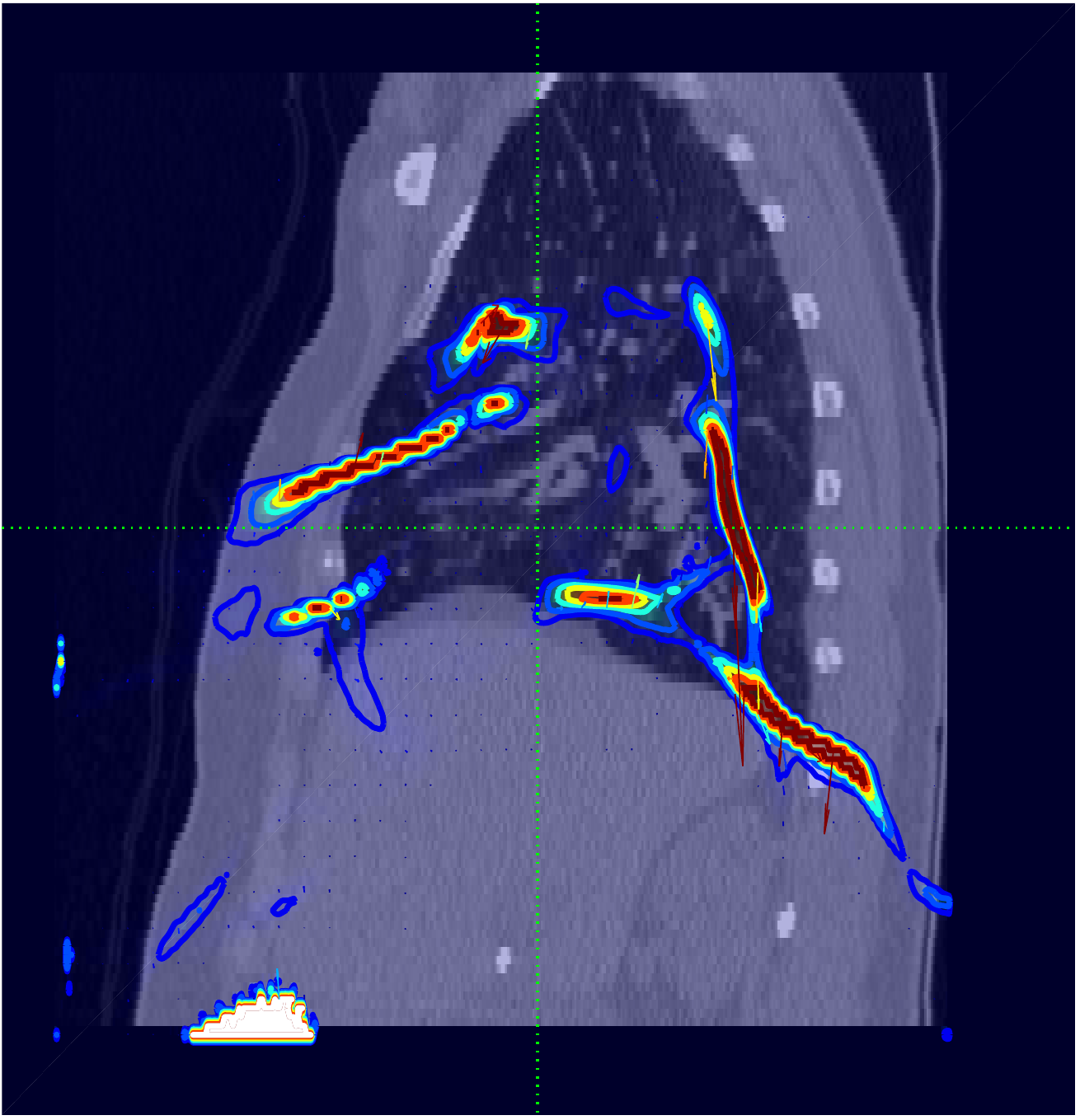}}% 
  \end{minipage}%
  \hfill\hfill\hfill
  \\
  % 
  % ----------- IMAGES (alternating mu)
  %
  %
  \begin{minipage}{\icwidthImgright}
    \vspace*{-\valignImgSkip}
    \centering
    \subfloat{%
      \includegraphics[width=\thirdAxialPR]{%
      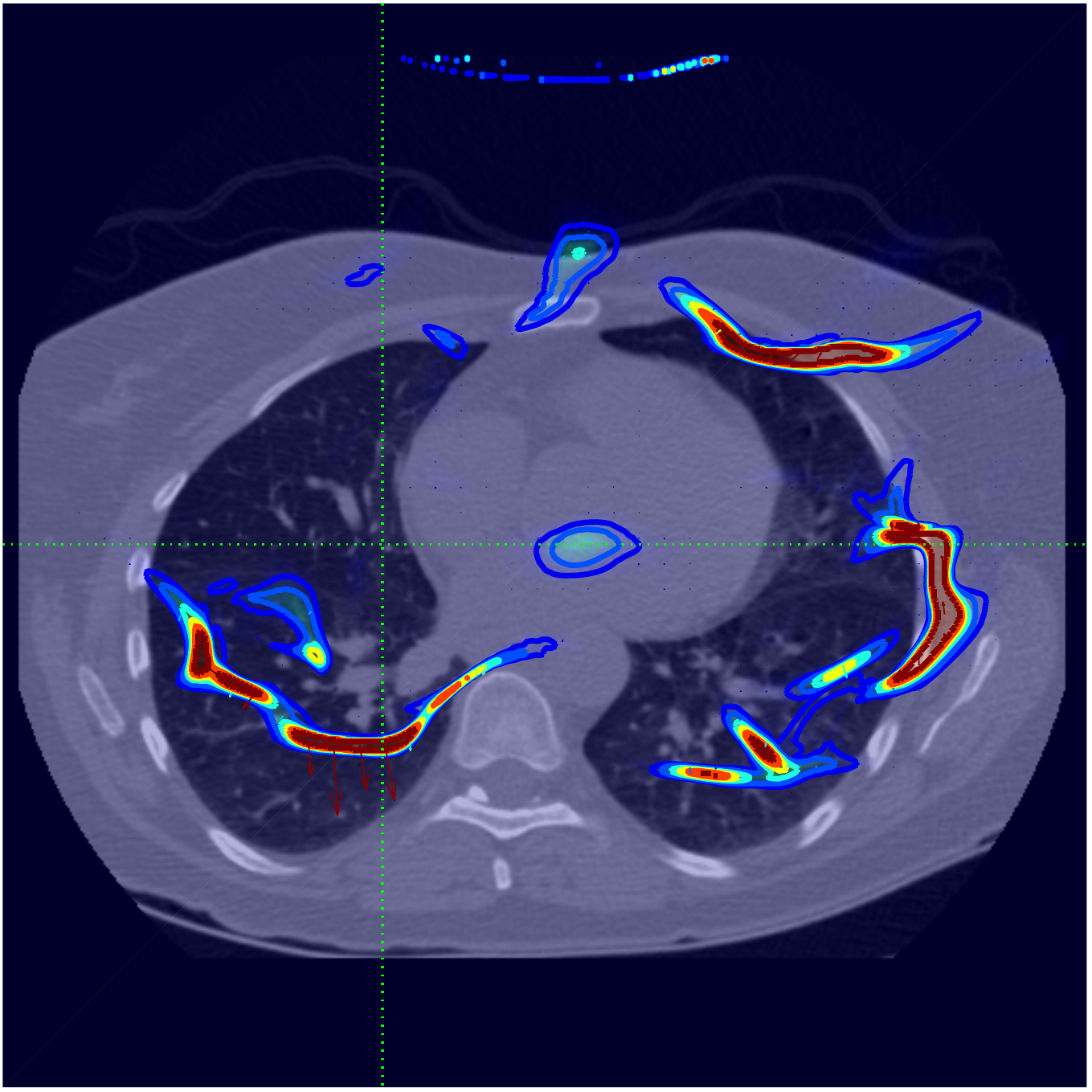}}%
    \hfill
    \subfloat{%
      \includegraphics[width=\thirdCoronalPR]{%
      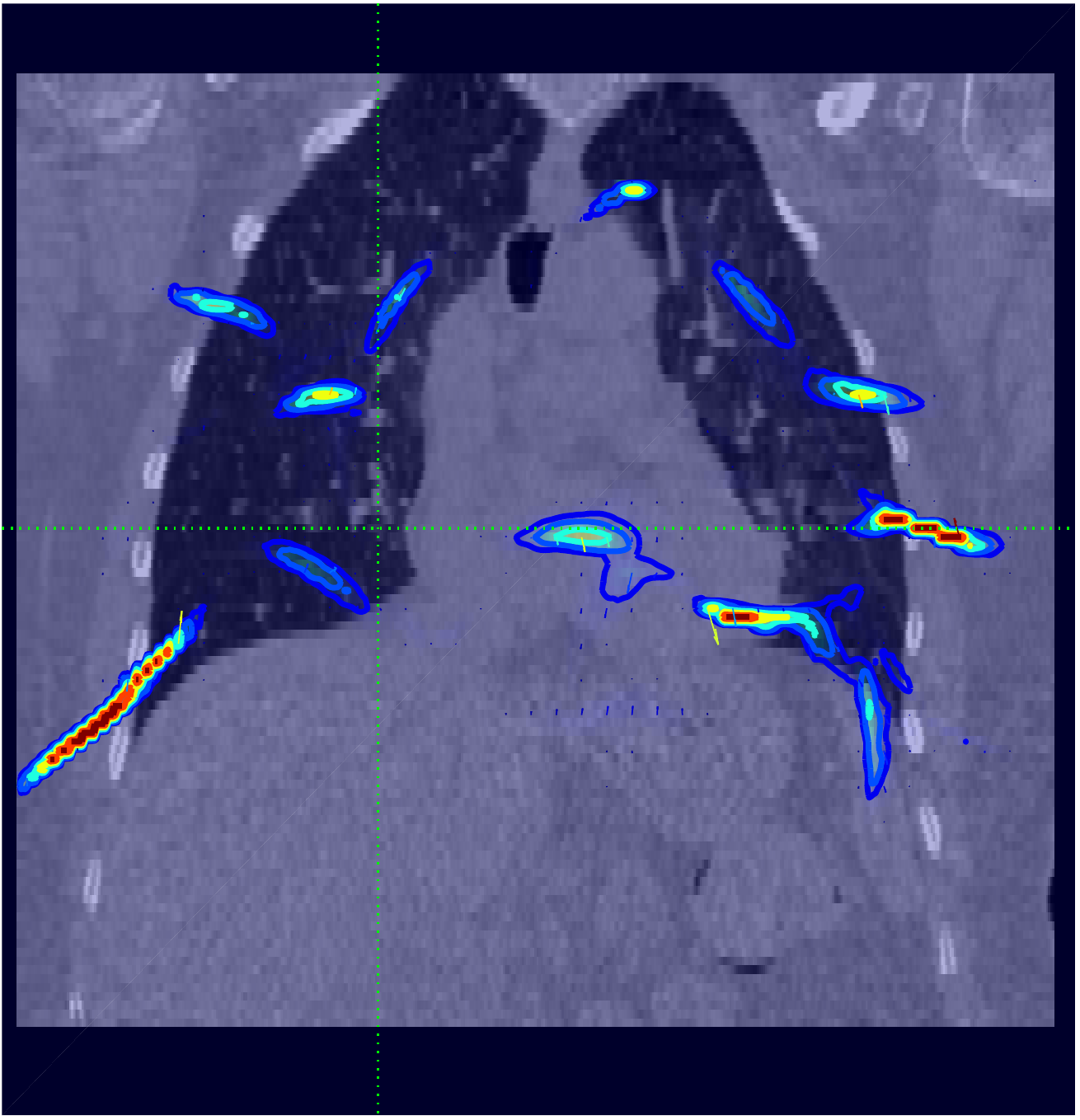}}%
    \hfill
    \subfloat{%
      \includegraphics[width=\thirdSagittalPR]{%
     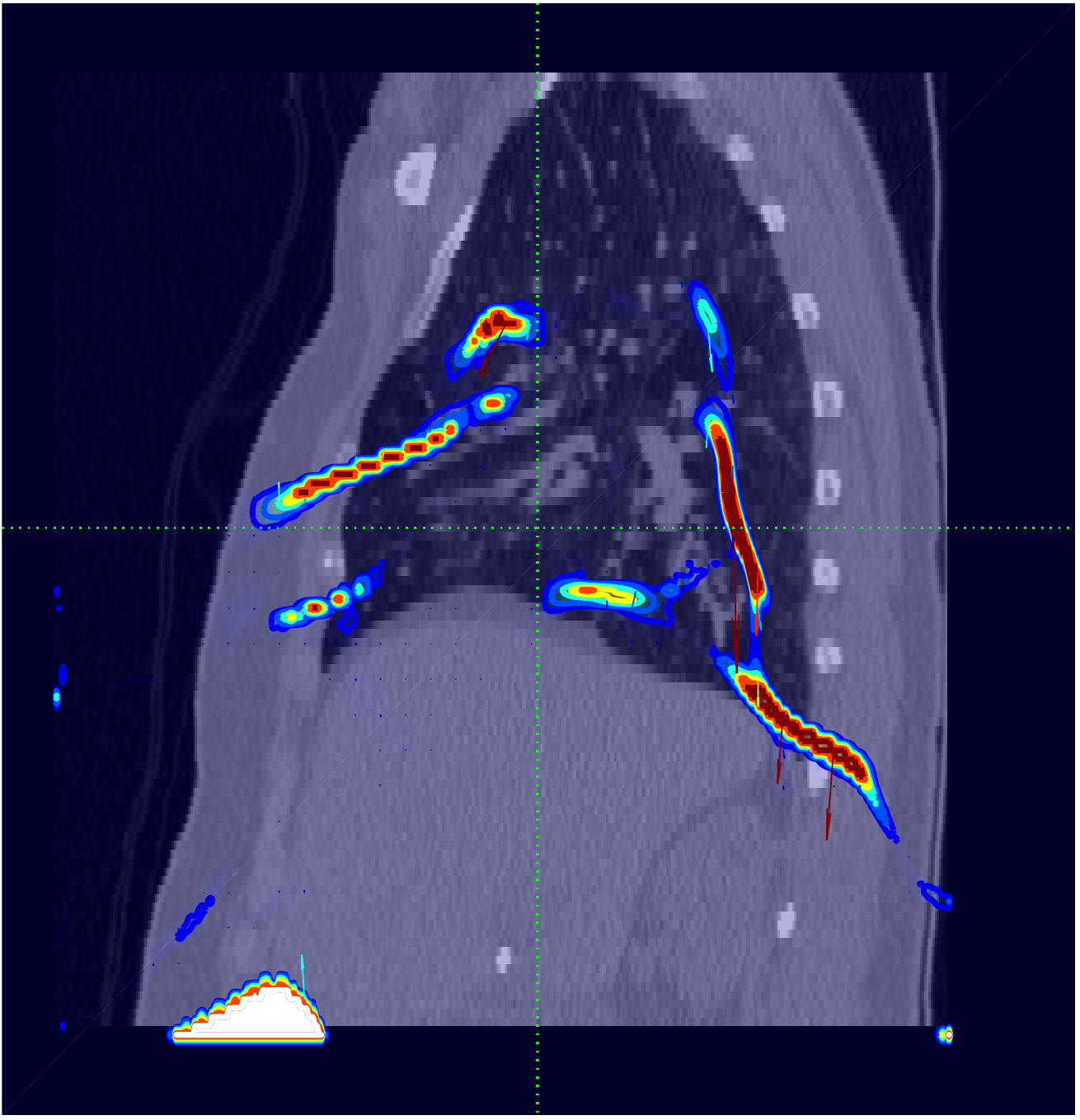}}%
  \end{minipage}%
   \\
  % % 
  % % ----------- IMAGES (mu*)
  % %
  % %
   \begin{minipage}{\icwidthImgright}
     \vspace*{-\valignImgSkip}
     \centering
     \subfloat{%
       \includegraphics[width=\thirdAxialPR]{%
       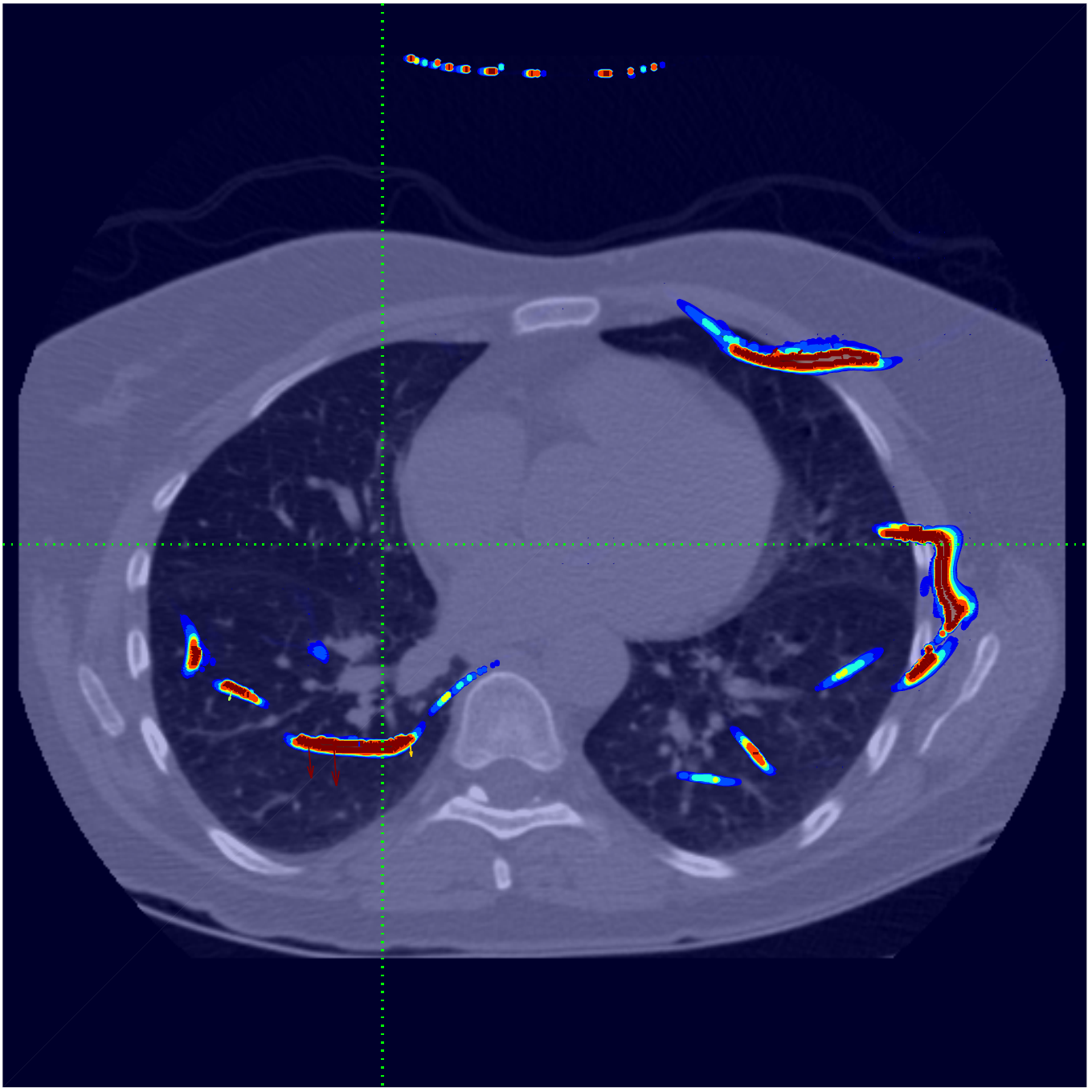}}%
     \hfill
     \subfloat{%
       \includegraphics[width=\thirdCoronalPR]{%
       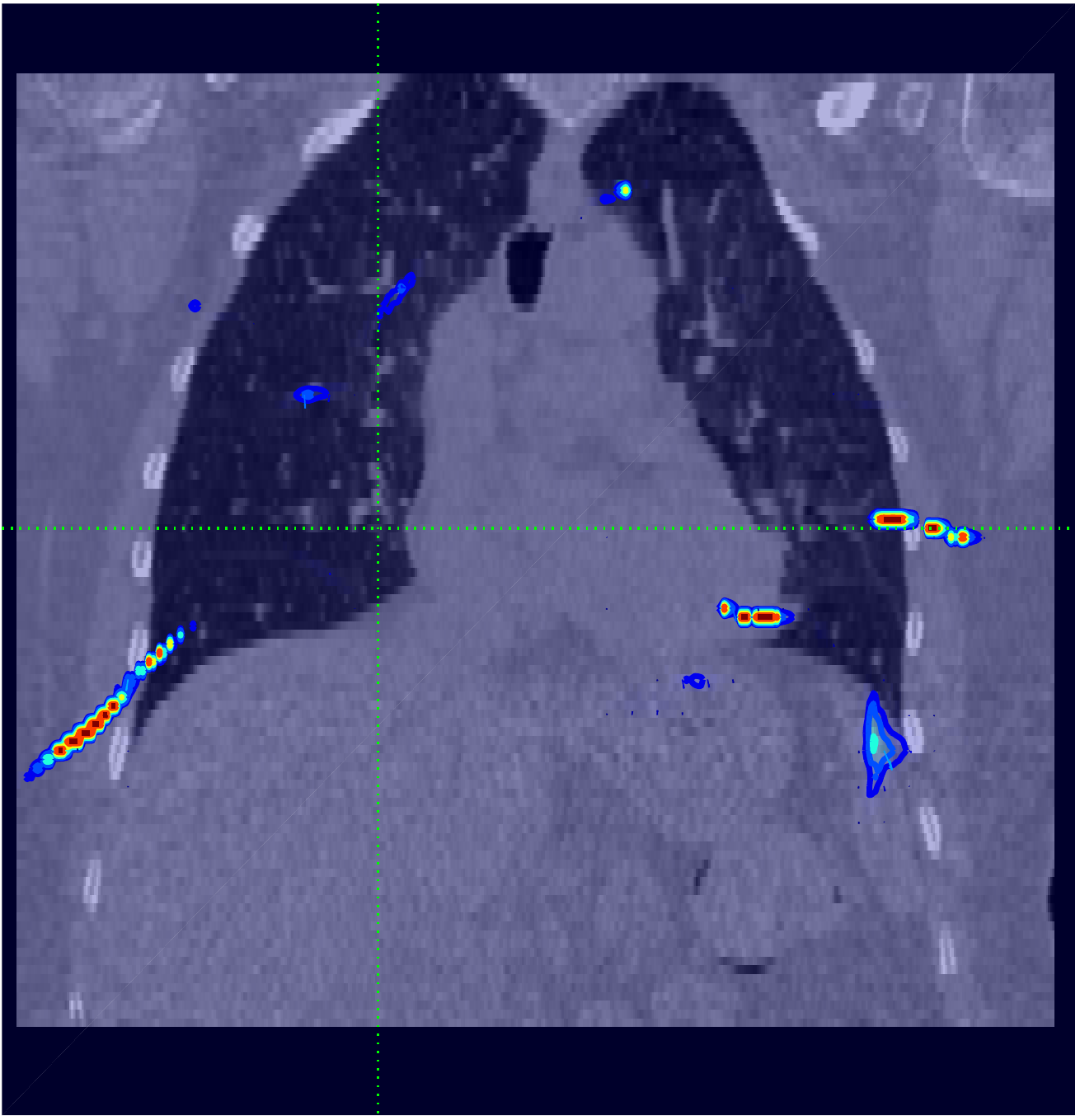}}%
     \hfill
     \subfloat{%
       \includegraphics[width=\thirdSagittalPR]{%
      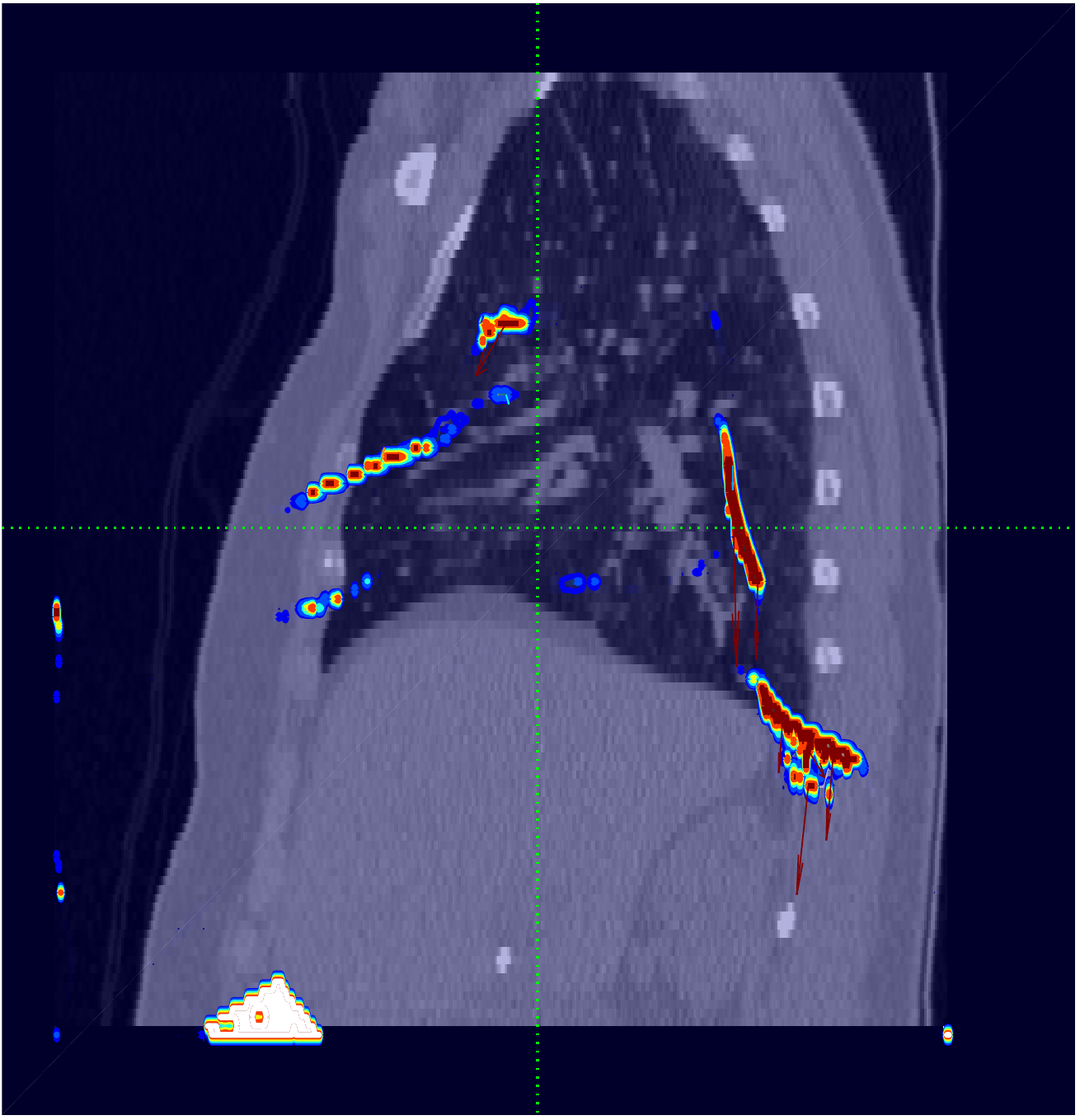}}%
   \end{minipage}%
   \hfill\hfill\hfill
  \\[\subcaptionSkip]
  % 
  % ---------- HARD-CODED SUB-CAPTION LABELS
  % 
  \begin{minipage}{\rwidthImgP}
    \centering
    \normalsize (b) $\R_{\F}$
  \end{minipage}
  \hfill\hfill\hfill
  % 
  % \hfill
  % 
  \begin{minipage}{\rwidthCbarP}
    \centering
    \hfill
  \end{minipage}
  % 
  % ---------- caption
  % 
  %
  \end{minipage}
  \caption{%
    Volumetric quiver-plots of $\R_{\G}(\x)$ and $\R_{\F}(\x + \G(\x))$, with
    contoured heat-maps of magnitude (in \milli\meter), at the 10th
    iteration step with the COPD4 DVF, overlaid on the target image.
    Comparison between four feedback control schemes: constant $\mu = 0$,
    constant $\mu = 0.5$, alternating $\mu = \mu_{\text{oe}}$ (with
    $\mu_{\text{o}} = 0.15$ and $\mu_{\text{e}} = 0.65$), and spatially
    variant $\mu_{\ast}(\x)$.
    The heat-map display range is truncated at \unit{10}{\milli\meter} for
    visual inspection.  White spots indicate regions where residual
    feedback entailed out-of-bounds values during the iteration.%
    \vspace*{-0.5em}
  }
  \label{fig:patient-copd4-residual-overlay-adjusted-k}
\end{figure*}

\begin{figure*}
  \centering
  \subfloat{%
    \includegraphics[width=0.40\textwidth]{%
      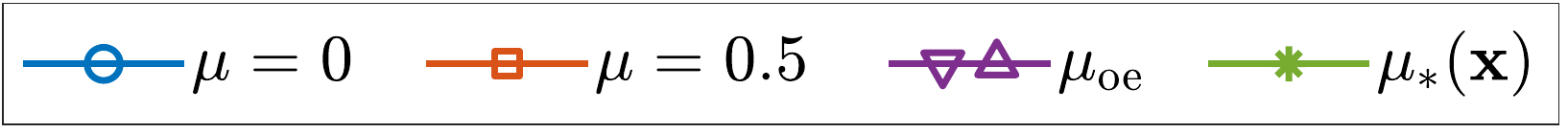}}%
  \\[0.5em]
  \begin{minipage}{0.90\linewidth}
    \vspace*{-\valignImgSkip}
    \centering
    \subfloat{%
      \includegraphics[width=\thirdonecol]{%
        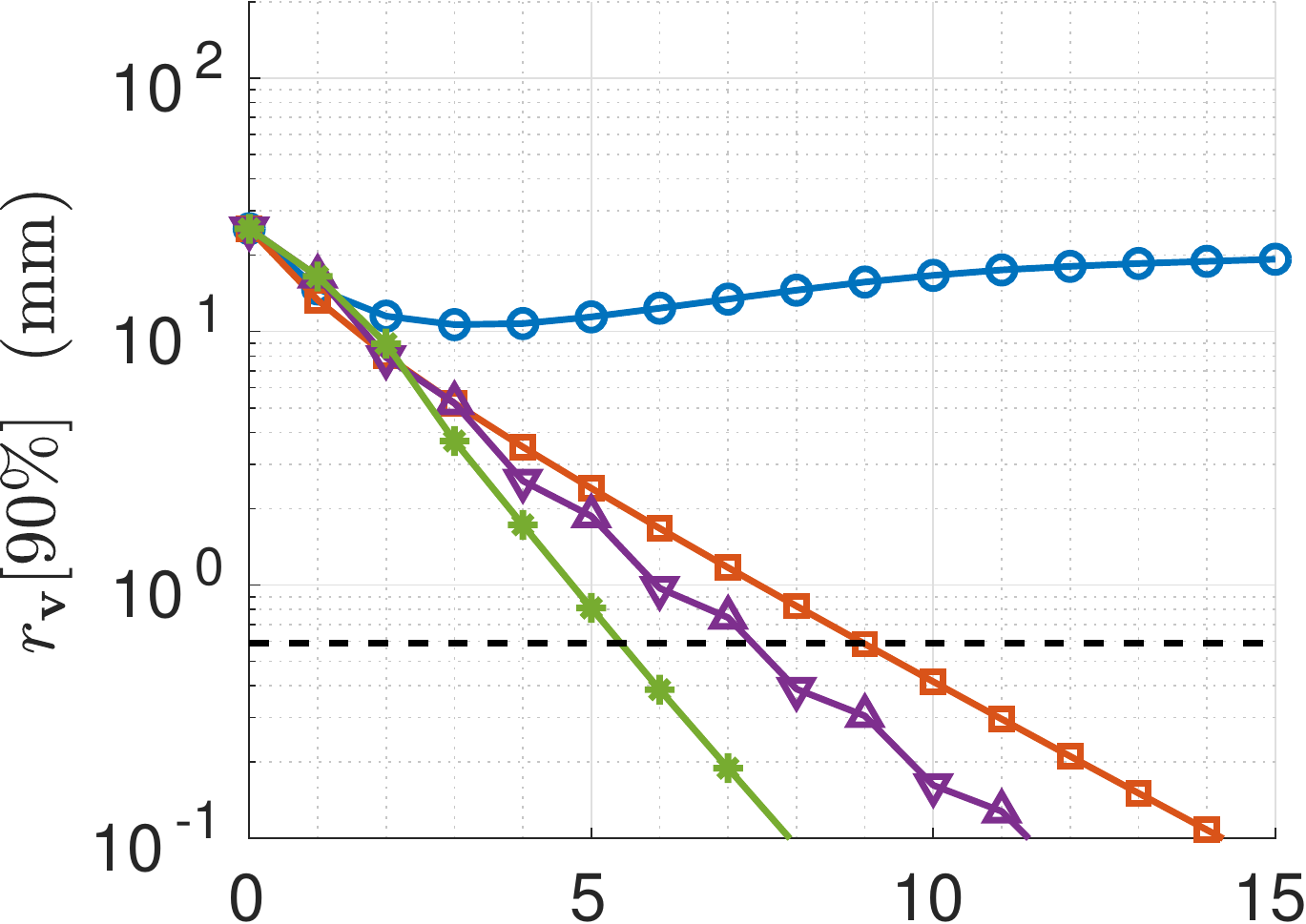}}%
    \hfill
    \subfloat{%
      \includegraphics[width=\thirdonecol]{%
        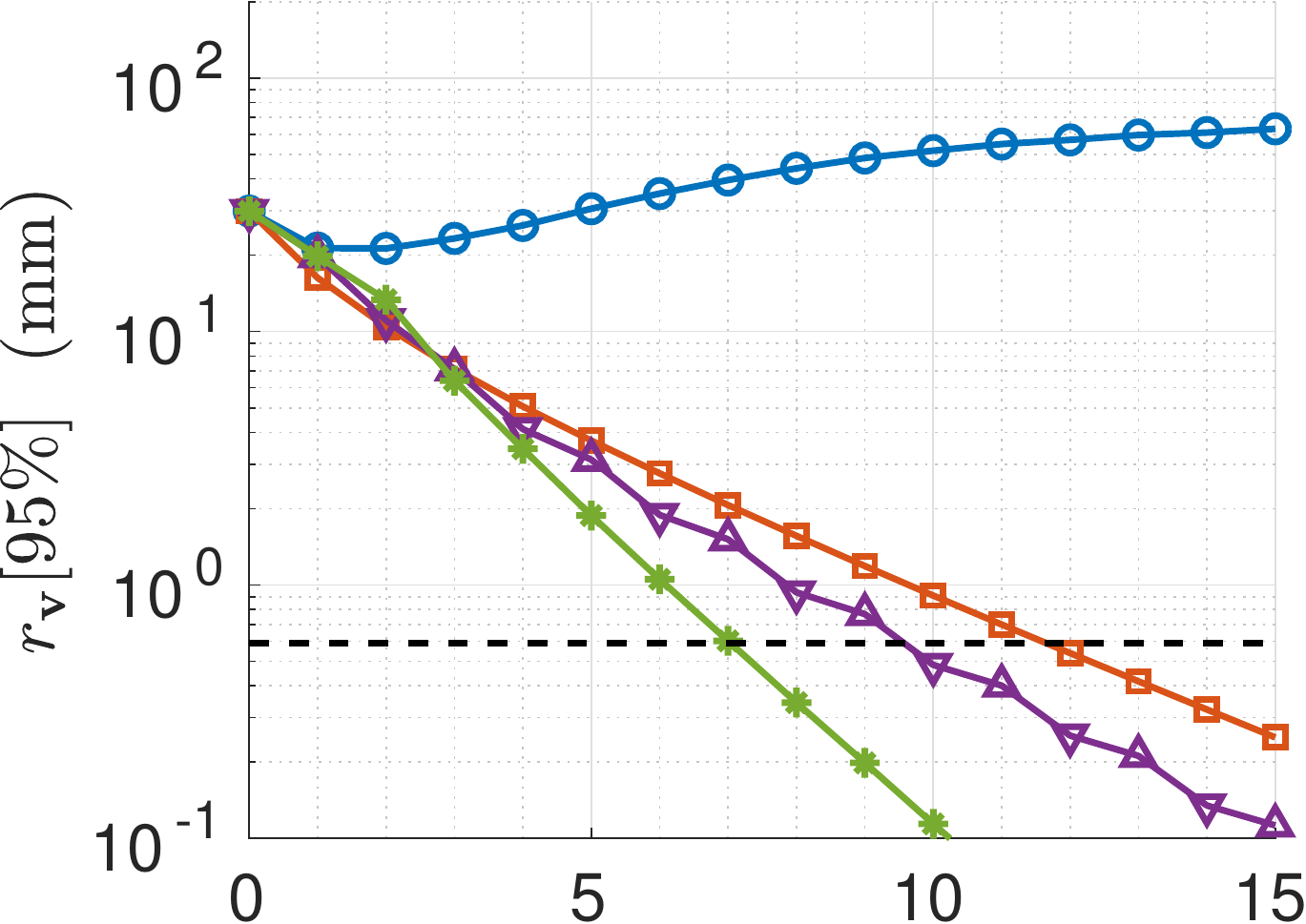}}%
    \hfill 
    \subfloat{%
      \includegraphics[width=\thirdonecol]{%
        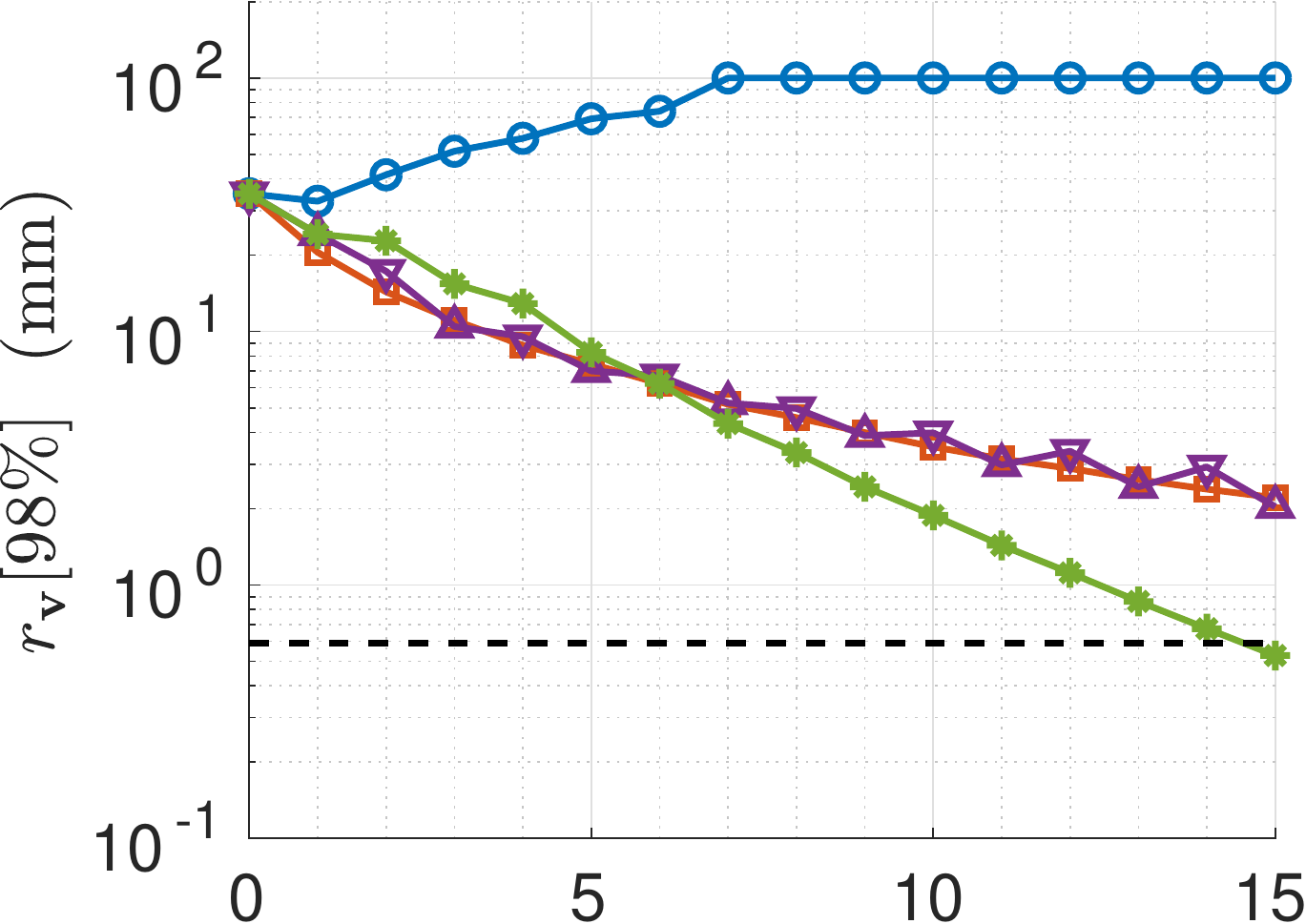}}%
    \\[-0.5em]
    \subfloat{%
      \includegraphics[width=\thirdonecol]{%
        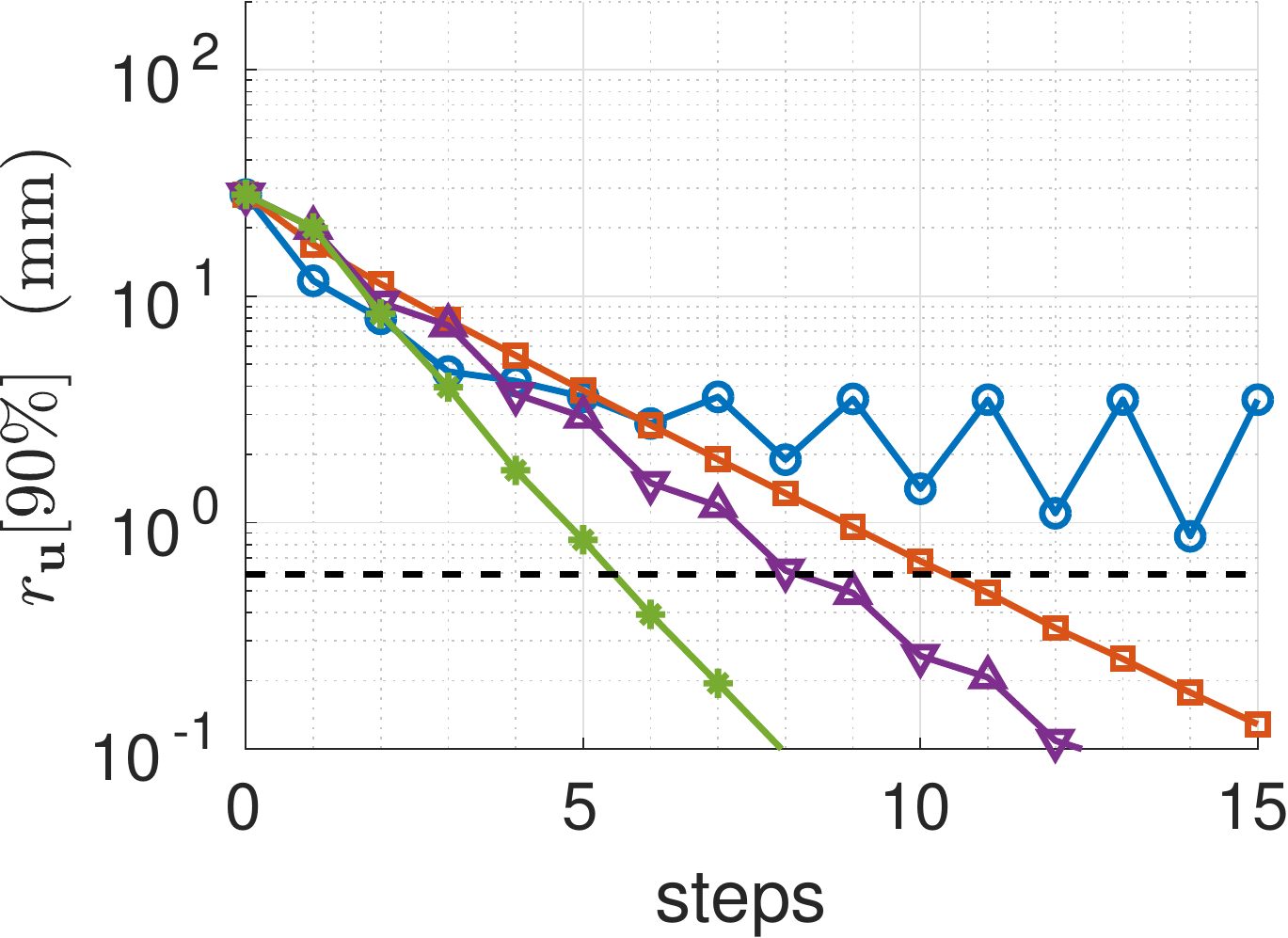}}%
    \hfill
    \subfloat{%
      \includegraphics[width=\thirdonecol]{%
        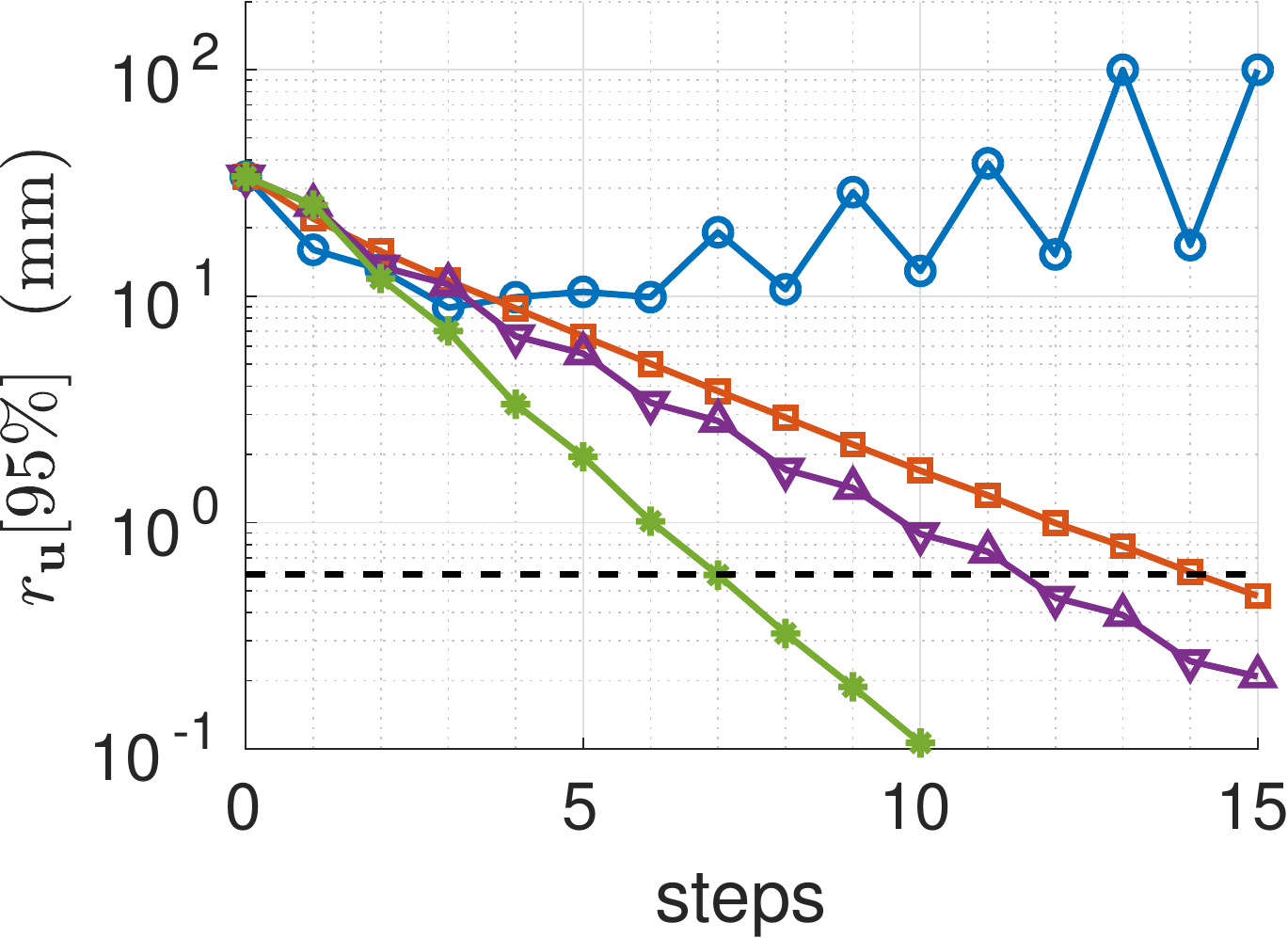}}%
    \hfill
    \subfloat{%
      \includegraphics[width=\thirdonecol]{%
        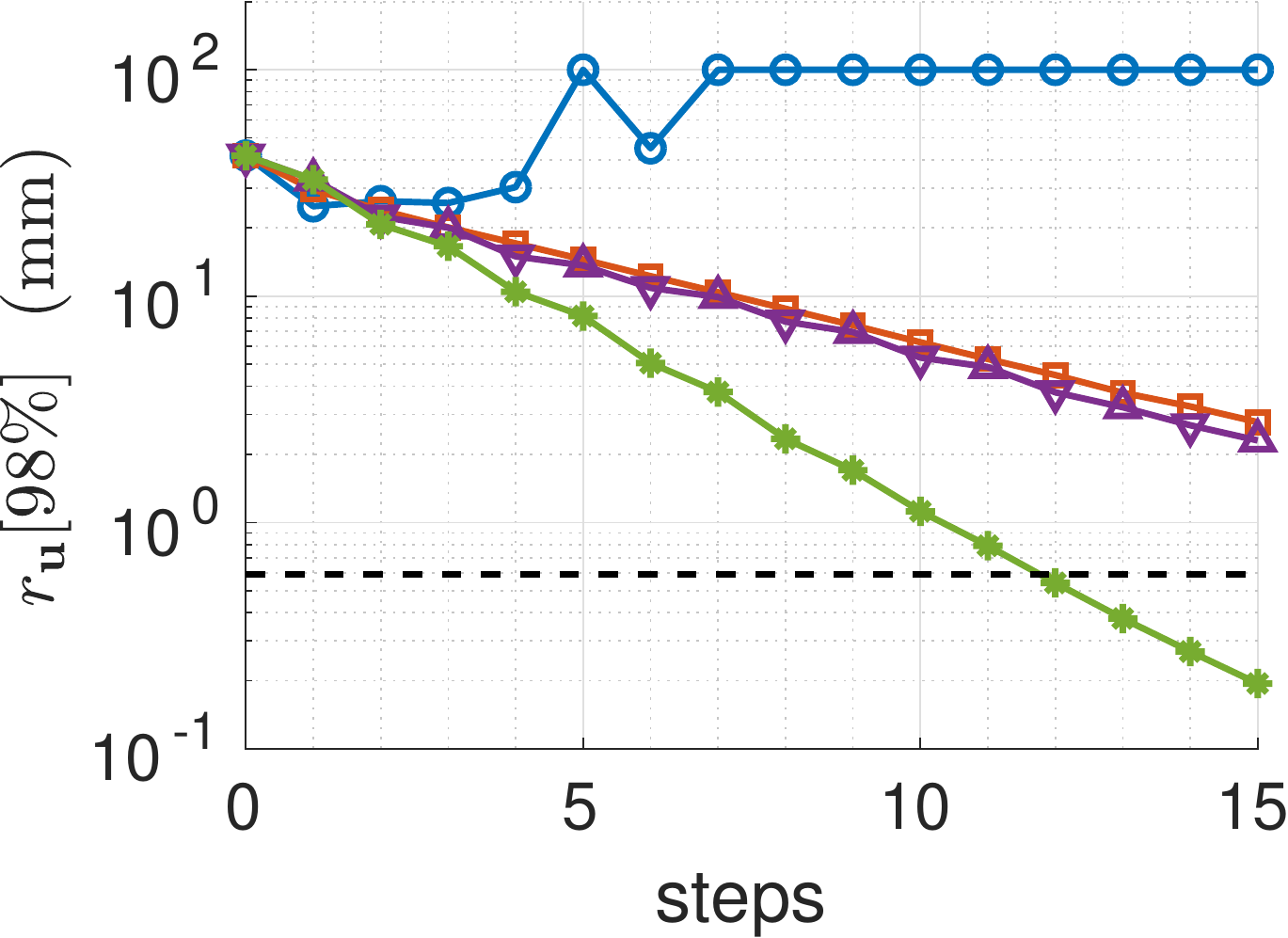}}%
  \end{minipage}
  \caption{%
    Pace of inverse consistency residual suppression during the first 15
    iteration steps with the COPD4 DVF, reported in percentiles of $r_{\G}(\x)$
    and $r_{\F}(\x + \G(\x))$ in $\log$-scale.
    Comparison between four feedback control schemes: constant $\mu = 0$,
    constant $\mu = 0.5$, alternating $\mu = \mu_{\text{oe}}$ (with
    $\mu_{\text{o}} = 0.15$ and $\mu_{\text{e}} = 0.65$), and spatially
    variant $\mu_{\ast}(\x)$.
    The black horizontal dashed lines mark the in-slice image resolution.
  }
  \label{fig:patient-copd4-error-vs-iter}
\end{figure*}

We provide in \cref{fig:patient-copd4-residual-overlay-adjusted-k,%
  fig:patient-copd4-error-vs-iter} %
a detailed comparison between $4$ feedback control settings with the COPD4
DVF.
\Cref{fig:patient-copd4-residual-overlay-adjusted-k} shows the spatial
variation of the IC residuals at the 10th iteration step, visualized as
volumetric quiver-plot and magnitude heat-map slices, overlaid on the
target CT image.
\Cref{fig:patient-copd4-error-vs-iter} shows the progression of residual
magnitudes in 90th, 95th, and 98th percentile values for the first $15$
iteration steps.
Both figures highlight the advantage of NTDC-adaptive iterations over
non-adaptive ones.
In particular, the spatially variant control scheme reduces the
residuals more rapidly and over the largest area. 
At the $10$th step, by
\cref{fig:patient-copd4-residual-overlay-adjusted-k}, it has removed or
substantially reduced the problematic spots visible in the other heat maps.
In fact, it renders residuals below the in-slice resolution
(\unit{0.59}{\milli\meter}) by $7$ steps at the 95th percentile, and by
$15$ steps at the 98th; the non-adaptive schemes take twice as many or more
steps to reach the same range, or fail to reach it.

% ==================== DISCUSSION
%
\section{Discussion and conclusion}
\label{sec:discussion}

We have elucidated and characterized the central role of non-translational
displacement components (NTDC) in iterative DVF inversion.  We have
developed a framework of NTDC-adaptive algorithms for DVF inversion with a
simple residual feedback control mechanism, and completed the framework
with rigorous convergence analysis.  Experimental results demonstrate the
superior performance of our adaptive control methodology, in both
convergence area and speed.
We have also found remarkable agreement between pre-inversion assessment of
control schemes, as arising from our analysis, and post-evaluation of
inversion results.

The clinical utility of our algorithms and analysis can be reflected in
multiple ways. 
\begin{inparaenum}[(i)]
\item NTDC-adaptive iteration algorithms enable quick and accurate
  estimation of an inverse DVF, which is valuable in a number of clinical
  applications, such as 4D image reconstruction and adaptive radiotherapy.
\item % Evaluation of a given DVF is of great interest and clinical
      % importance.
  The spectral measures in \cref{subsec:spectral-NTDC-characterization} can
  be used, independently of any inversion task, for evaluating DVFs
  generated with existing software under clinical application conditions.
\item The inversion algorithms, presented here as an asymmetric approach
  for generating the inverse of a forward mapping, can be incorporated with
  ease into existing software for simultaneous deformable image
  registration, which may consist of multiple forward and backward
  transformation stages~\cite{sotiras2013deformable}. It is also plausible
  to employ the spectral measures as local regularization terms at each
  transformation stage, complementary to global ones.  Additional,
  systematic studies are needed for such extensions.  Potential benefits
  include rapid refinement in inverse consistency, increased robustness to
  the asymmetry in image quality between registered images, and reduced
  registration artifacts.
\end{inparaenum}%

Lung deformations are of great clinical
concern~\cite{ren2014limited,yan_pseudoinverse_2010}.
DVFs obtained from thoracic CT images are used in turn to test our theory
and algorithms. 
Further testing, with similar as well as different types of deformations,
will better underscore the scope and impact of our theory and algorithms.

% ==================== ACKNOWLEDGEMENTS
%
\section*{Acknowledgements}
\label{sec:acknowledgements}
% flatex input: [tex/acknowledgements.tex]
% acknowledgements.tex 

This work was supported by the National Institutes of Health Grant
No.~R01-CA184173.
We thank the anonymous reviewers, Associate Editors, and Editor for their
valuable comments on our previous manuscripts.
We thank Dr.\ Richard Castillo for provision of the CT data used in this
study, and Dr.\ Animesh Srivastava for technical assistance.

% ==================== CONFLICTS OF INTEREST
%
\ifdefined\review
  \section*{Disclosure of conflicts of interest}
  \addcontentsline{toc}{section}{Conflicts of interest}
  \label{sec:conflict-interest}
  % flatex input: [tex/conflict-interest.tex]
% conflict-interest.tex

The authors have no relevant conflicts of interest to disclose.

\fi

%%%%%%%%%%%%%%%%%%%%%%%%%%%%%%%%%%%%%%%%%%%%%%%%%%
%%% REFERENCES

{\small
% citation formatting
\ifdefined\medphys              % ---------- AIP Medical Physics
  % (nothing required)
\fi\ifdefined\plain             % ---------- plain
%*flatex input: [ARTICLE-dvf-inversion-relaxation.bbl]
%

% flatex input end: [ARTICLE-dvf-inversion-relaxation.bbl]
%FLATEX-REM:  \bibliographystyle{abbrv}
\fi

\bibliographystyle{abbrv}
% additional sections
%\ifdefined\plain
%\input{tex/appendix-analysis-global-linear-DVF}
%\fi

% references
\ifdefined\medphys
  \ifdefined\review\else
    \section*{References}
    \vspace*{-1em}
  \fi
\fi
%FLATEX-REM:\bibliography{ref}

}

%%%%%%%%%%%%%%%%%%%%%%%%%%%%%%%%%%%%%%%%%%%%%%%%%%
%%% APPENDIX

\appendix

\section{Numerical inversion errors with an analytical DVF pair}
\label{sec:appendix-casestudy-analyticaldvf}

\begin{figure}
  \centering
  \begin{minipage}{\wideonecol}
    \vspace*{-\valignImgSkip}
    \centering
    \subfloat[\label{fig:rings-ref}]{%
      \includegraphics[width=\thirdonecol]{%
        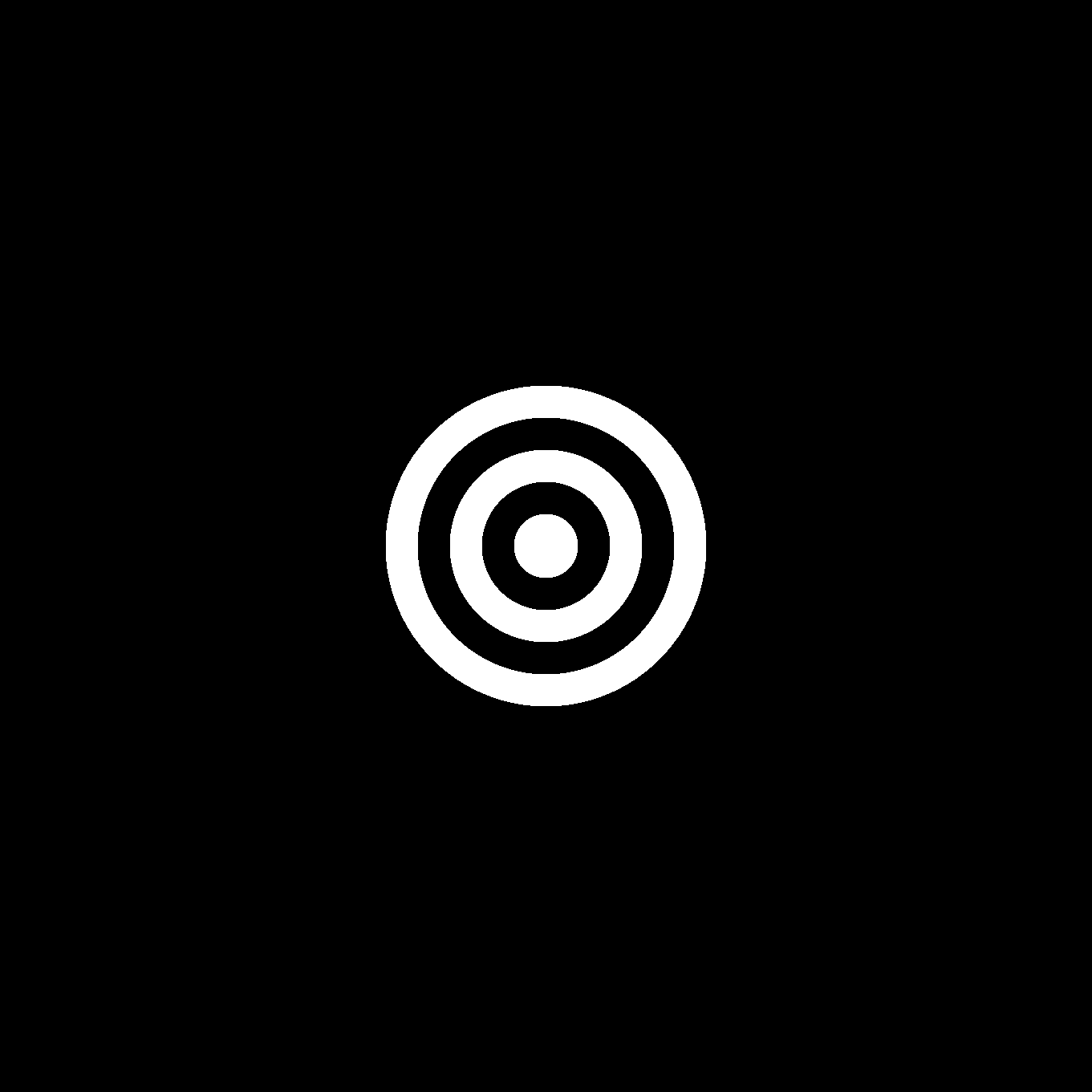}}%
    \hfill
    \subfloat[\label{fig:rings-deformed-small}]{%
      \includegraphics[width=\thirdonecol]{%
        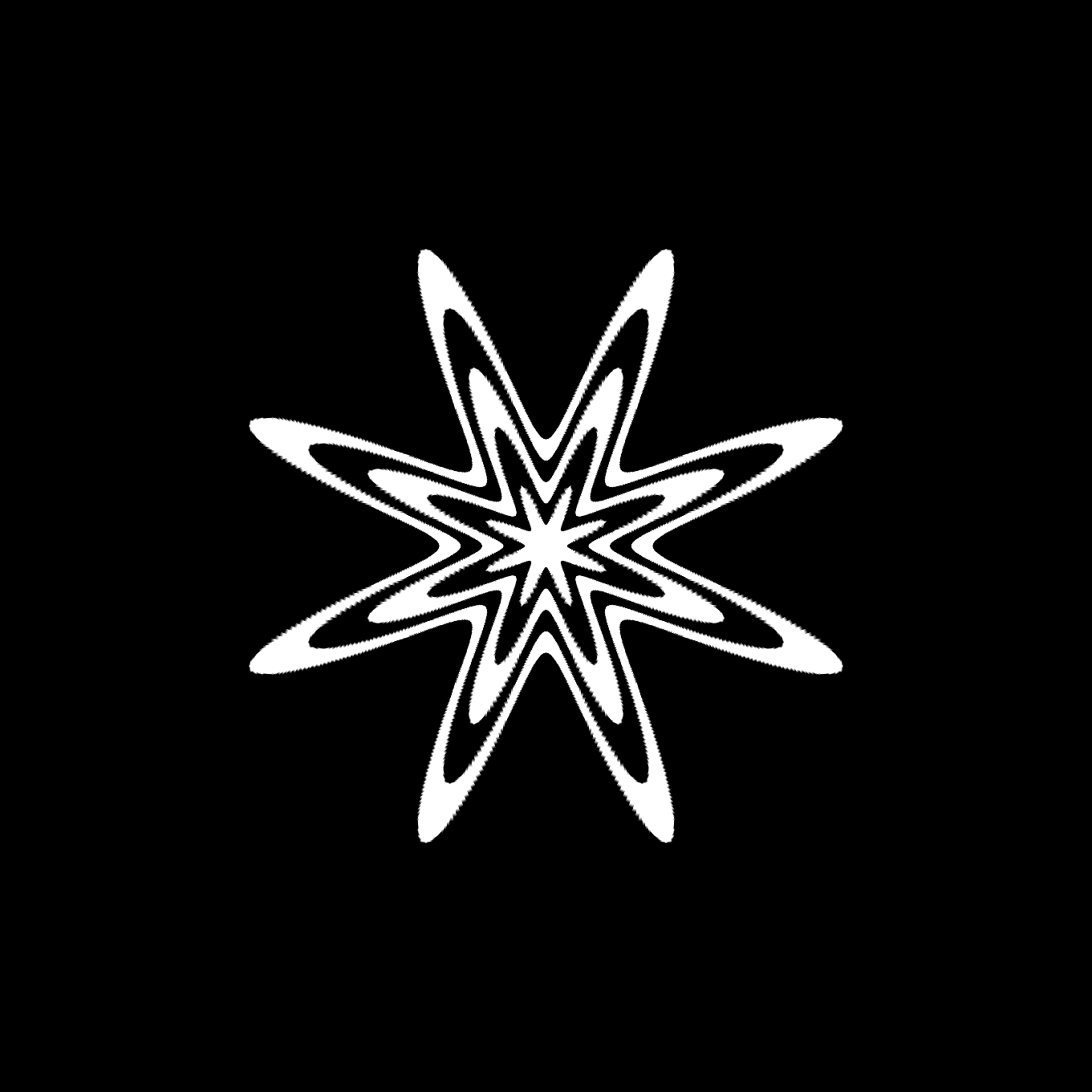}}%
    \hfill
    \subfloat[\label{fig:rings-deformed-large}]{%
      \includegraphics[width=\thirdonecol]{%
        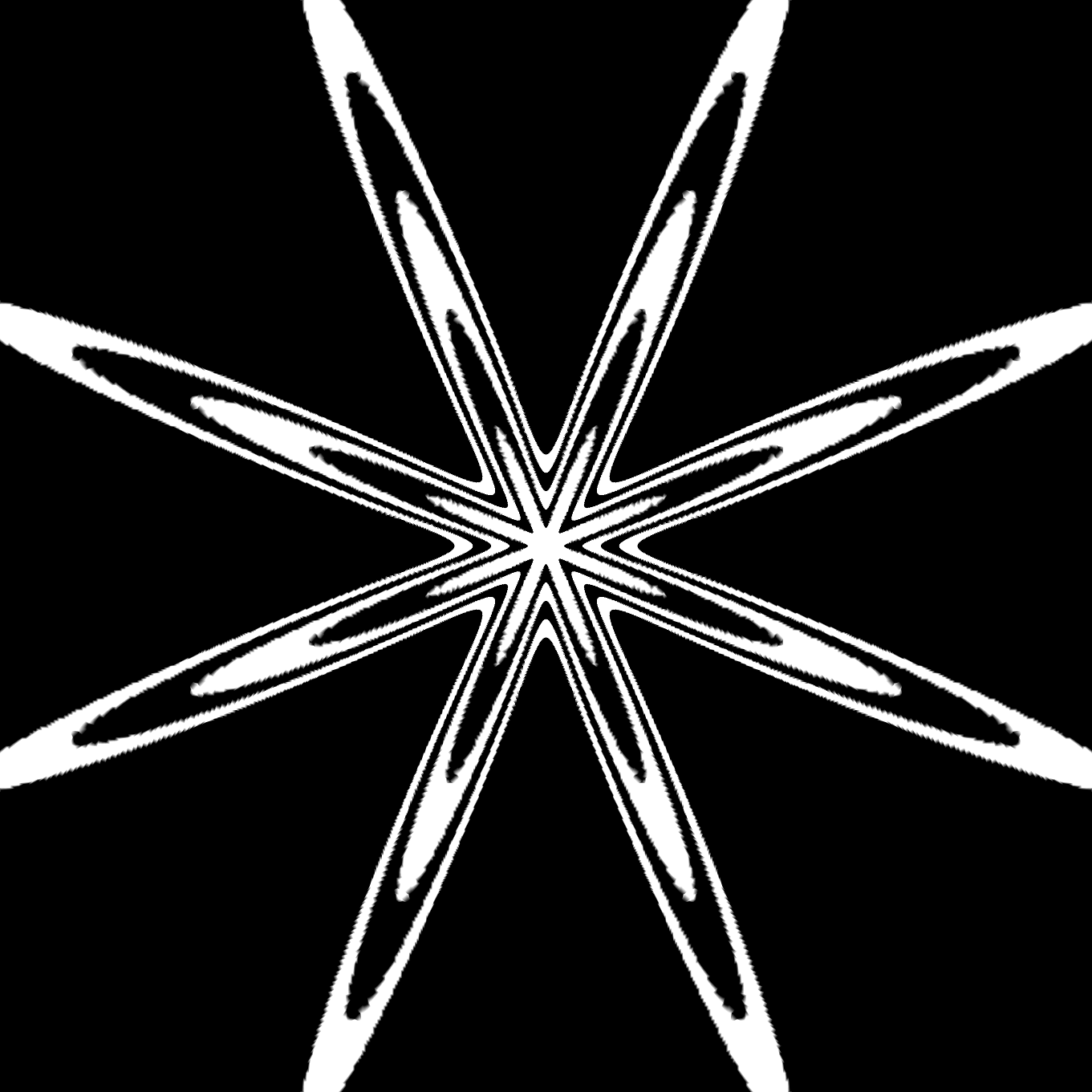}}%
  \end{minipage}
  \caption{%
    Image-space visualization of the analytical DVFs of
    \cref{eq:2D-analytic-DVFs}.
    \emph{(a)}~Reference image of concentric rings over $[-34,34]^2$.
    \emph{(b)}~Target image obtained by deforming the reference image with
    the forward DVF of \cref{eq:2D-analytic-DVFs-fwd} with parameter values
    $m = 8$ and $b = 0.5$.  %
    \emph{(c)}~Target image by the forward DVF with parameter values $m = 8$ and
    $b = 0.8$.
    % 
    %This visualization is re-created following the work of Chen
    %\protect\etal.~\cite{chen2008simple} in a square domain [-16?, 16?]%
  }
  \label{fig:rings}
\end{figure}

We provide comparisons in numerical inversion errors among several DVF
inversion algorithms in the framework of \cref{eq:fpim-scaling-control}.
The study of inversion errors, unlike IC residuals, requires the
ground-truth inverse DVF.
We use the 2D analytical DVFs introduced by Chen
\etal~\cite{chen2008simple}:
\ifdefined\medphys\ifdefined\review
  \hspace*{-1em}
\fi\fi
\begin{subequations}
  \label{eq:2D-analytic-DVFs} 
  \begin{align} 
    \label{eq:2D-analytic-DVFs-fwd} 
    \F(\x') &= \left( \frac{1}{1 + b \cos (m \theta(\x'))} -  1 \right) \x', 
    \\ 
    \label{eq:2D-analytic-DVFs-bwd} 
    \G(\x) &= b \cos (m  \theta(\x)) \, \x, 
  \end{align}
\end{subequations}
where $\x, \x'$ lie in a 2D domain $\Omega$, $b \in (0,1)$ is a radial
(stretch) parameter, $m \in \mathbb{N}$ is an angular (oscillation)
parameter, and $\theta(\x) \in [0,2\pi)$ is the angular coordinate of
$\x$ in the polar representation
$\tpose{\x} = \norm{\x} ( \cos \theta(\x), \sin \theta(\x) )$. The
angular coordinate is well-defined everywhere except at $\x=\mbs{0}$.
It is straightforward to verify that the analytical DVFs are inverse to
each other.
The DVFs were visualized in image space by Chen \etal~\cite{chen2008simple}
via deforming a specific reference image of concentric rings; we re-create
such images in \cref{fig:rings}.

%% ========== figure inclution ==================
% flatex input: [tex/fig-rings.tex]
% fig-rings.tex
% 

%% ========== figure inclution ==================
%% =============================================

\begin{figure}
  \centering
  \begin{minipage}{\wideonecol}
    \vspace*{-\valignImgSkip}
    \centering
    \begin{minipage}{0.325\linewidth}%
     \centering
      \hspace*{0.0em}
      \includegraphics[width=0.98\linewidth]{%
        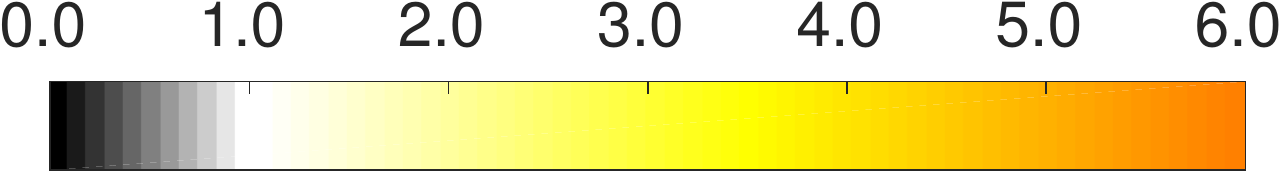}%
    \end{minipage}%
    \hfill
    \begin{minipage}{0.325\linewidth}%
    \centering
        \hspace*{0.0em}
        \includegraphics[width=0.98\linewidth]{%
        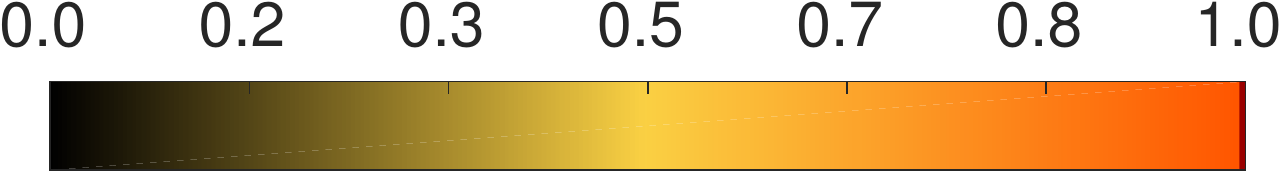}%
    \end{minipage}
    \hfill
    \begin{minipage}{0.325\linewidth}%
    \centering
      \hspace*{0.0em}
      \includegraphics[width=0.98\linewidth]{%
        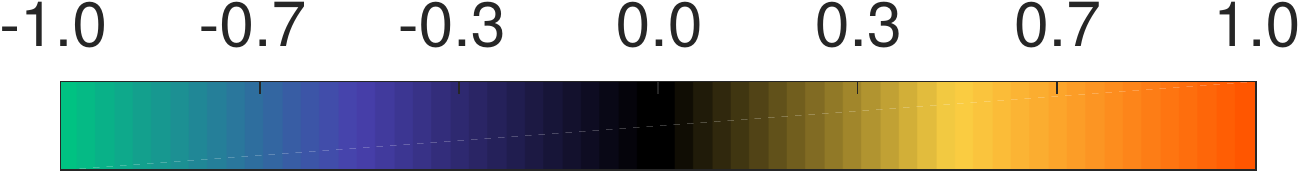}%
    \end{minipage}
    \hfill
    \\
    \vspace*{-\valignImgSkip}
    \subfloat[$|\J_{\mathbf{f}}(\x)|$]{%
      \includegraphics[width=\thirdonecol]{%
        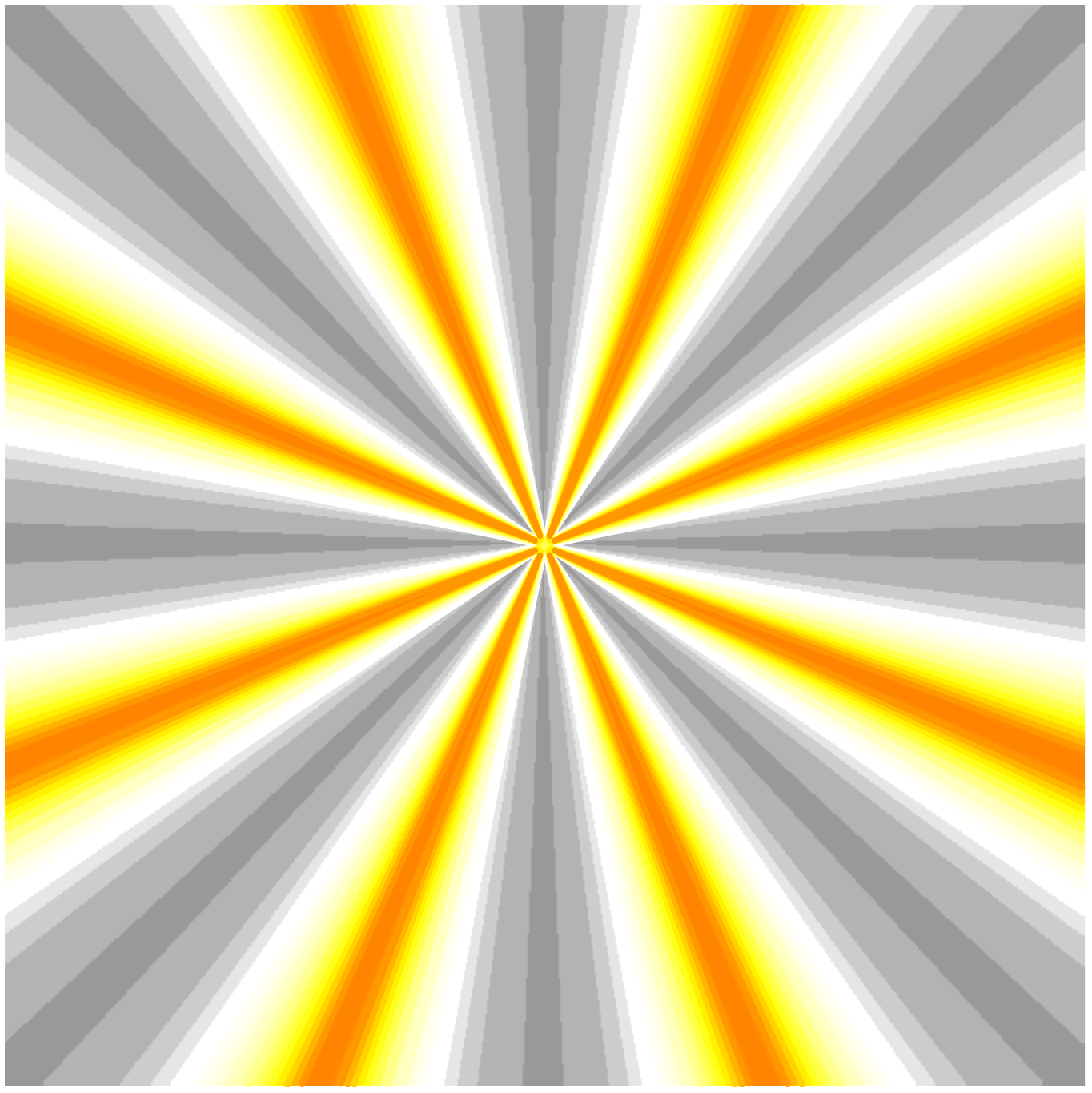}}%
    \hfill
    \subfloat[$\rho(\J_{\F}(\x))$]{%
      \includegraphics[width=\thirdonecol]{%
        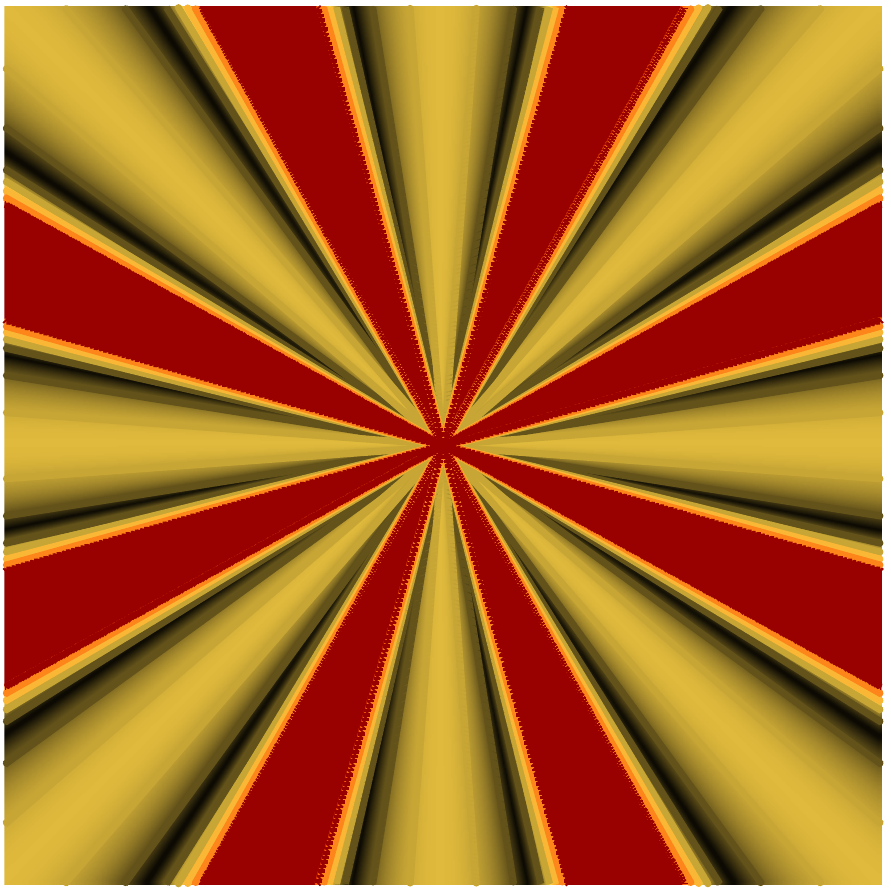}}%
    \hfill
    \subfloat[$1-2\gamma(\x)$]{%
      \includegraphics[width=\thirdonecol]{%
        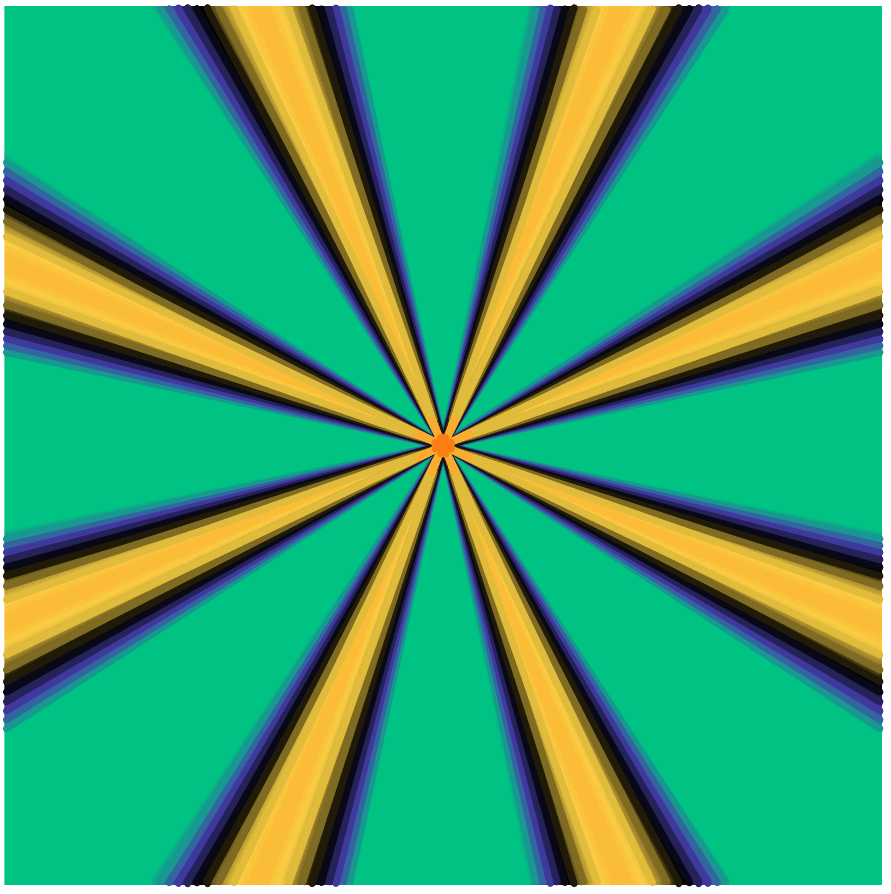}}%
  \end{minipage}
  \caption{%
    Spectral maps of three characterization measures
    (\cref{subsec:spectral-NTDC-characterization}) of the discretized DVF
    \cref{eq:2D-analytic-DVFs-fwd} over $\Omega = [-17, 17]^2$.
    \emph{(a)}~Determinant map.  Small (black) and large (orange)
    values show volume shrinkage and expansion, respectively. 
    \emph{(b)}~Spectral NTDC radius map.  Non-small NTDC regions, per
    \cref{eq:nonsmall-NTDC,eq:equivalent-NTDC-tests}, are highlighted in
    red.
    % 
    % The control $\mu=0$ is clearly inadequate in this region
    % suppressing the iterate errors.
    % 
    \emph{(c)}~Algebraic control index map.  The index range shows that
    global convergence is feasible with adaptive feedback control; see
    \cref{eq:local-range-of-mu,eq:existence-condition}.}
\label{fig:chendvf-datacharacterization}
\end{figure}

In numerical experiments, we discretize the analytical DVFs on a grid of
finite resolution, apply each inversion algorithm to the discretized
forward DVF, and compare the numerical inverse estimate to the inverse DVF.
Specifically, we set $m=8$ and $b=0.8$, and discretize $\F$ on a 2D grid
over $[-34,34]^2$ with spatial resolution $0.05$. We let $\Omega$ be the
sub-grid over $[-17,17]^2$, where both DVFs are valid in the sense of
\cref{eq:omega-def}.
We assess the spectral properties of the numerical DVF and present in
\cref{fig:chendvf-datacharacterization} its spectral maps by the three
characterization measures introduced
in~\cref{subsec:spectral-NTDC-characterization}.

% ============= spectral characterization maps

Four iterative inversion algorithms are examined.  The same algorithms are
compared in IC residuals in \cref{sec:results-IC-residuals}.  The first two
are existing work~\cite{chen2008simple, christensen2001consistent}; they
are non-adaptive and globally constant, with the control parameter values
$\mu=0$ and $\mu=0.5$. 
The next two are adaptive to the DVF.  One is non-stationary, using the
alternating parameter values $\mu_{o}$ and $\mu_{e}$
(\cref{subsec:alternating}), adaptively set to $50$th and $98$th
percentile mid-range values over $\Omega$.  The other is spatially
variant $\mu_{\ast}(\x)$ (\cref{subsub:adaptive-control}), using locally
optimal values (per
\cref{eq:mu-star-real-case,eq:mu-star-complex-cases}, customized to 2D
DVFs), except in a small neighborhood around $\x = \mbs{0}$ where the
mid-range value is used.

\begin{figure*}[p]
  \centering
  \subfloat{%
    \includegraphics[width=0.40\textwidth]{%
      legend_residual_progression_plots}}%
  \\[0.5em]
  \begin{minipage}{0.09\linewidth}
   \centering
   \vspace*{-0.8em}
   \rotatebox[origin=c]{90}{\footnotesize$\G^{[\text{a}]}_{0}$}\\
   \vspace*{4.2em}
   \rotatebox[origin=c]{90}{\footnotesize$\G^{[\text{b}]}_{0}$}
   \end{minipage}
   \hfill
  \begin{minipage}{0.90\linewidth}
    \vspace*{-\valignImgSkip}
    \centering
    \subfloat{%
      \includegraphics[width=\thirdonecol]{%
        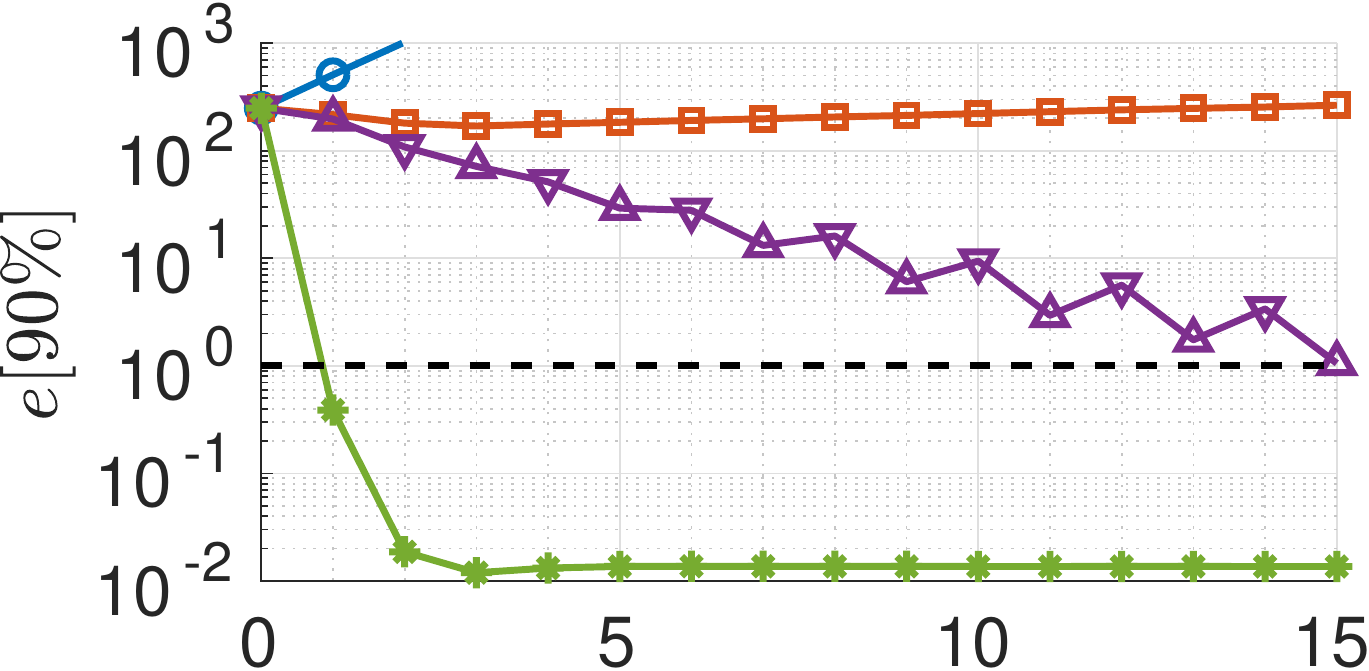}}%
    \hfill
    \subfloat{%
      \includegraphics[width=\thirdonecol]{%
        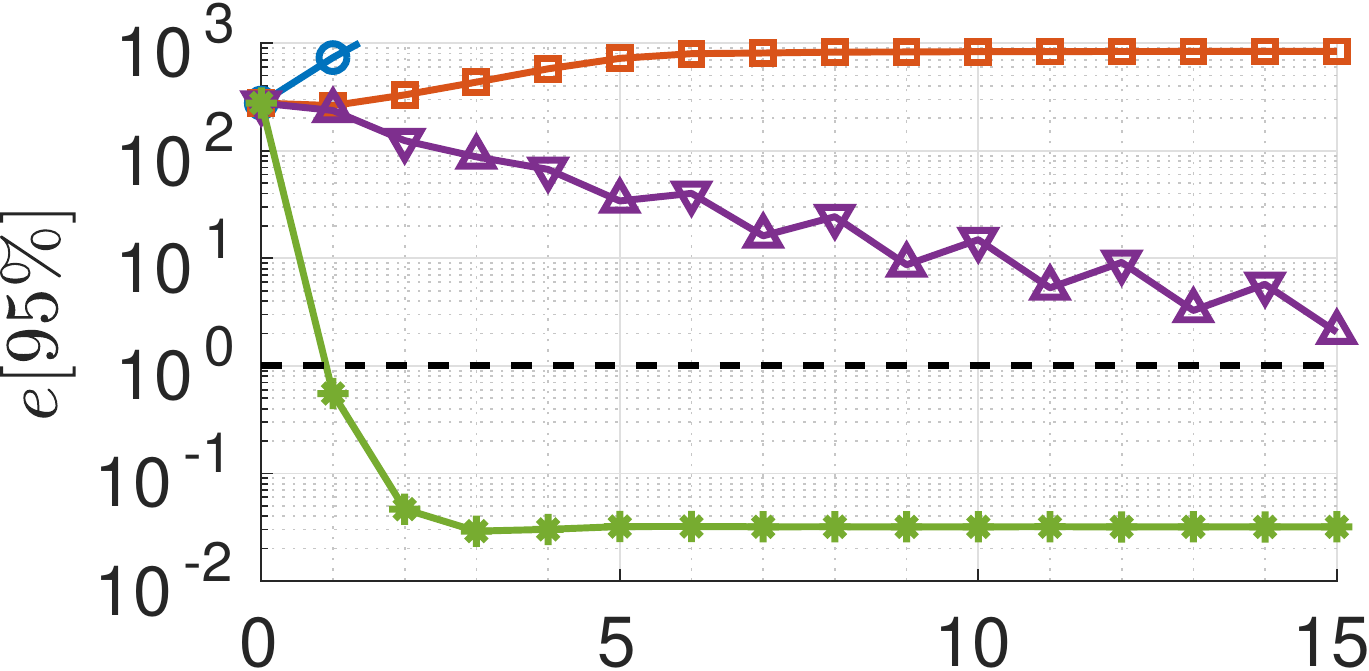}}%
    \hfill 
    \subfloat{%
      \includegraphics[width=\thirdonecol]{%
        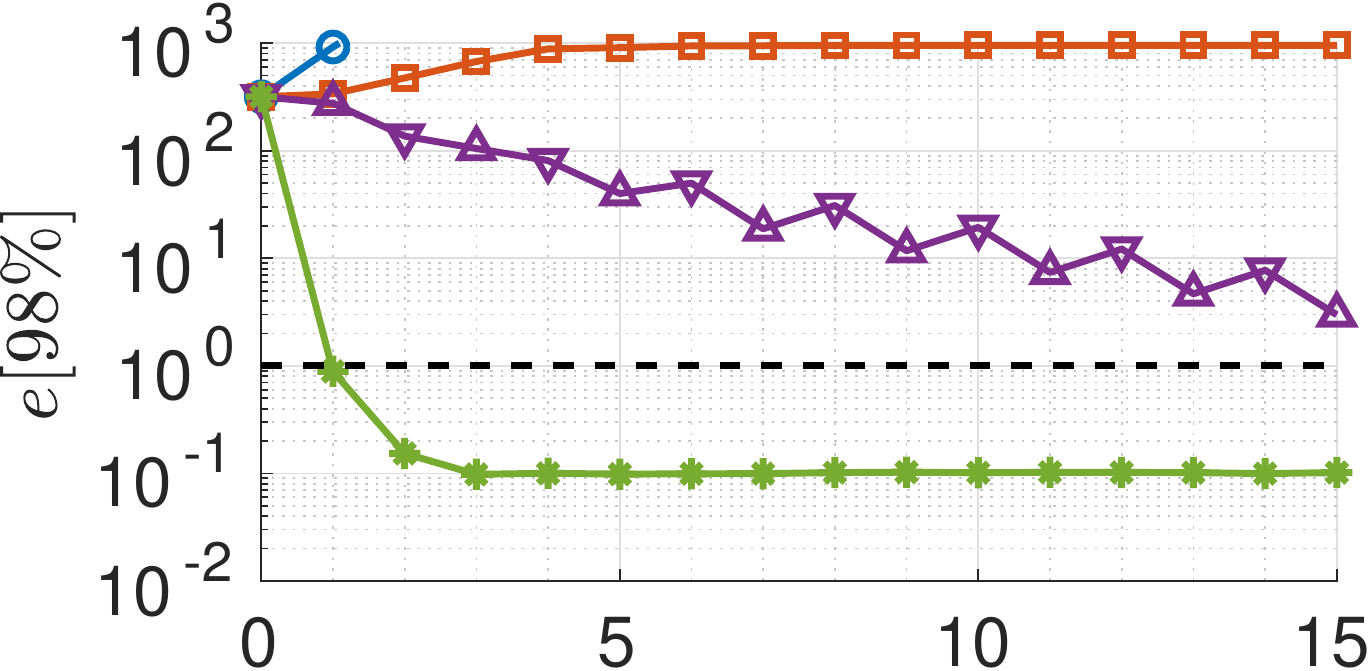}}%
    \\
    \subfloat{%
      \includegraphics[width=\thirdonecol]{%
        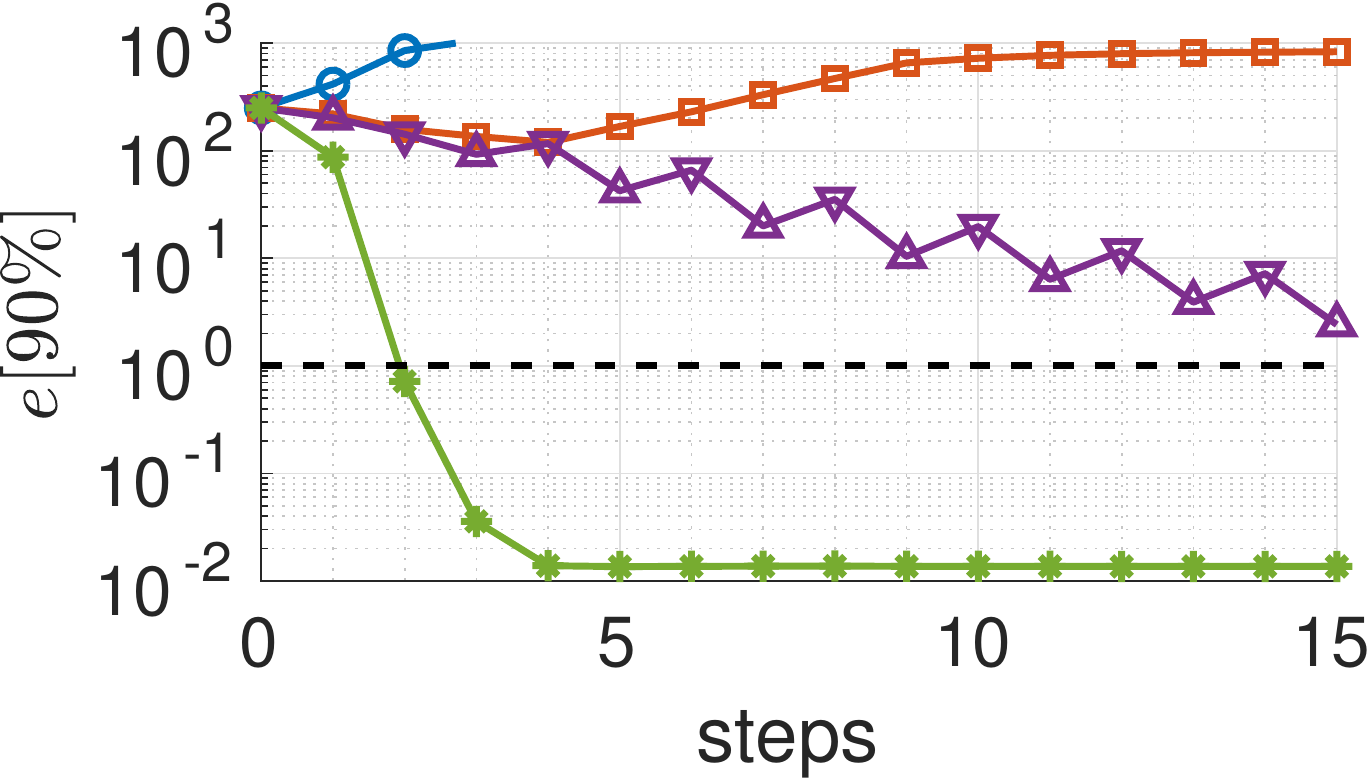}}%
    \hfill
    \subfloat{%
      \includegraphics[width=\thirdonecol]{%
        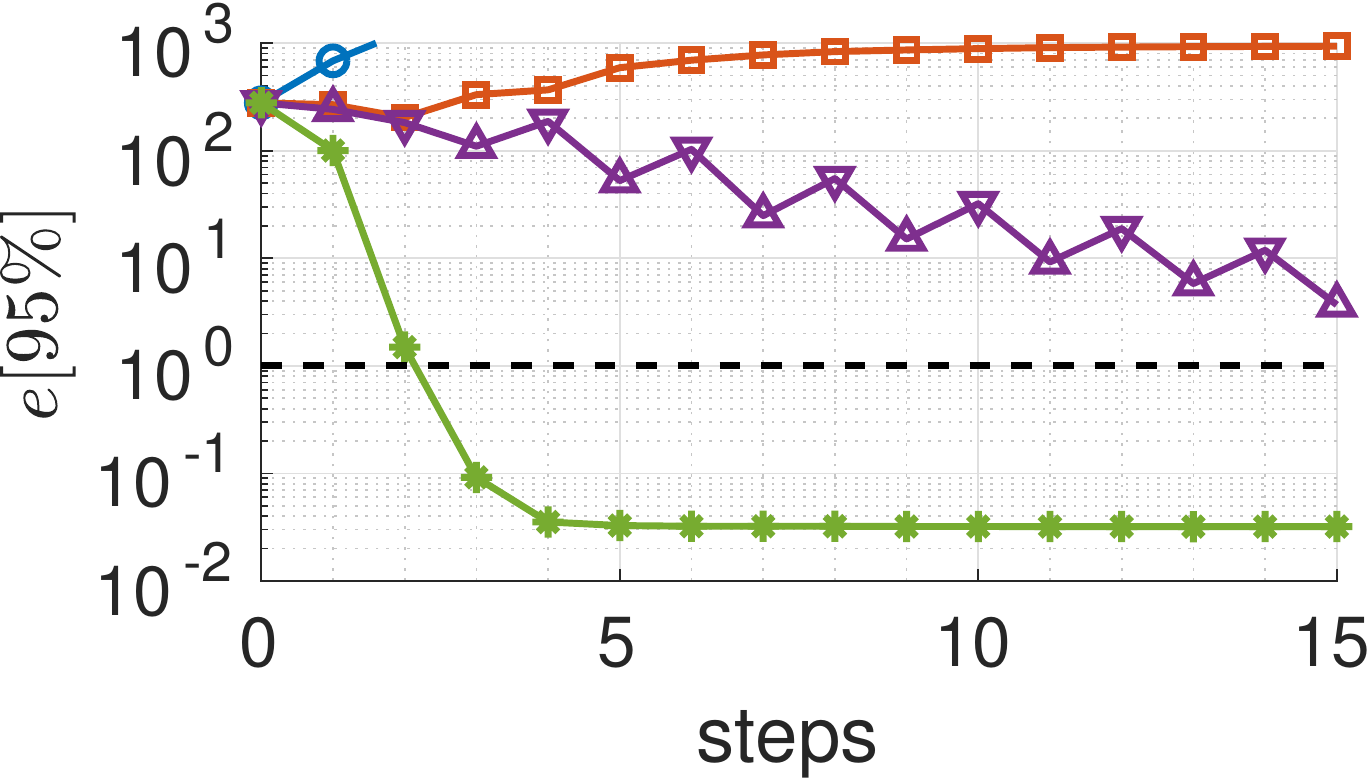}}%
    \hfill 
    \subfloat{%
      \includegraphics[width=\thirdonecol]{%
        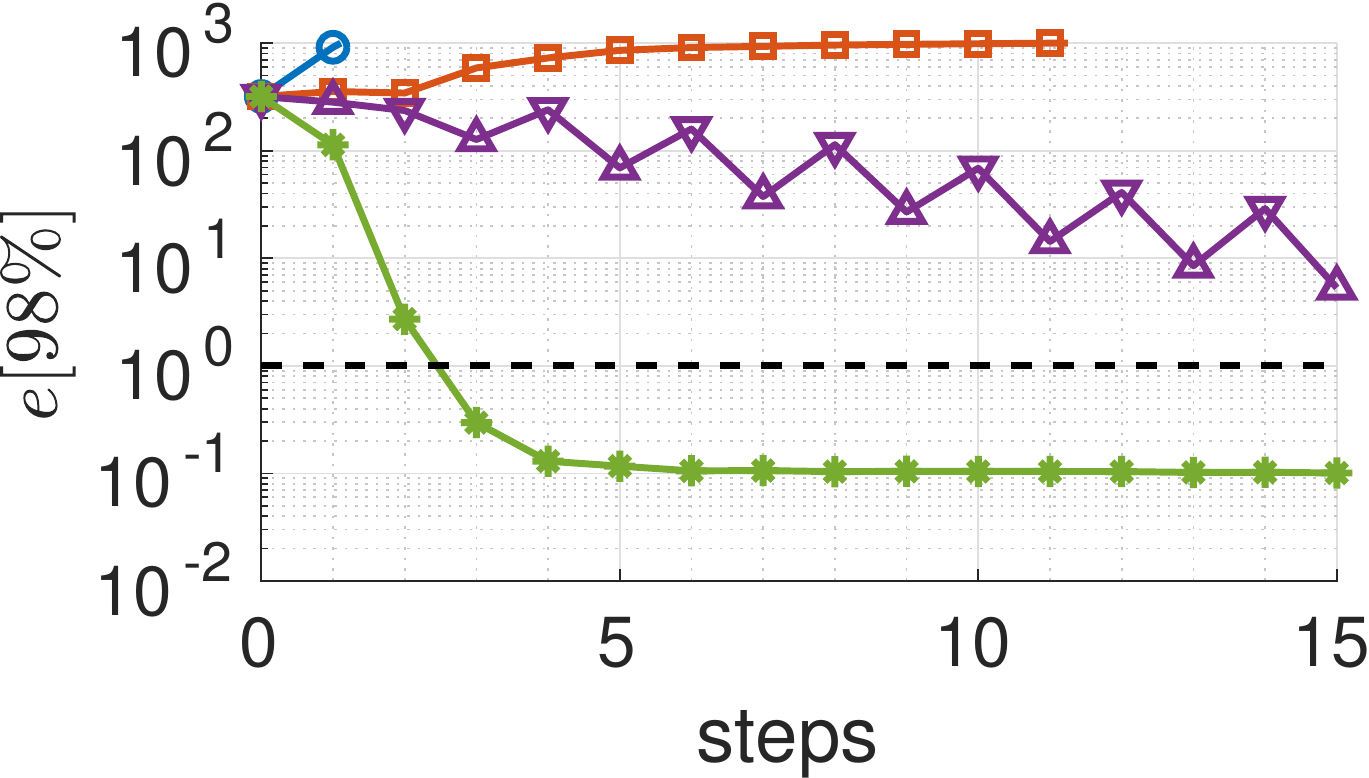}}%
  \end{minipage}
  \caption{%
    Inversion error magnitudes in three percentiles (90th left, 95th
    middle, and 98th right) during the first 15 steps of each iteration
    with the discretized DVF, with two different initial guesses (top and
    bottom). 
    Each plot shows the error sequence with each of the following four
    control schemes: constant $\mu = 0$, constant $\mu = 0.5$, alternating
    $\mu_{\text{oe}}$ with $\mu_{\text{o}} = 0$ and
    $\mu_{\text{e}} = 0.77$, and spatially variant $\mu_{\ast}(\x)$. 
    Error magnitudes,
    $e(\x) = \sqrt{ e^{2}_{\text{LR}} + e^{2}_{\text{AP}} +
      e^{2}_{\text{SI}} }$,
    are measured in pixel-length unit and plotted in $\log$-scale.}
  \label{fig:chendvf-error-vs-iter}
\end{figure*}

\begin{figure*}
  \centering
  \begin{minipage}{0.85\linewidth}
  % 
  %======================== COLORBAR
  %
  \hspace*{0.75em}
  \includegraphics[width=\linewidth]{%
  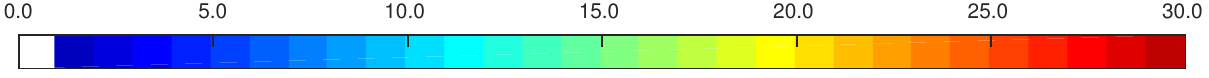}%
  \\[-0.5em]
 %
  % ============================== Error
  %
  \begin{minipage}[t]{\ichalftwocolleft}
    \centering
     \begin{minipage}[t]{\ictextmarginleft}
         \centering
         \text{}
    \end{minipage}%
    \hfill
    \begin{minipage}{\icwidthImgleft}
      \begin{minipage}{\thirdonecol}
        \centering
        1 step
      \end{minipage}%
      \hfill
      \begin{minipage}{\thirdonecol}
        \centering
        8 steps
      \end{minipage}%
      \hfill
      \begin{minipage}{\thirdonecol}
        \centering
        15 steps
      \end{minipage}%
    \end{minipage}
    \\
    % 
    % ==================== RHO MAPS (MU = 0)
    %
    % ----- images
    % 
    \begin{minipage}{\ictextmarginleft}
      \centering
      \rotatebox[origin=c]{90}{\normalsize$\mu = 0$}
    \end{minipage}%
      % 
    %\hfill
      %
    \begin{minipage}{\icwidthImgleft}
    \vspace*{-\valignImgSkip}
    \centering
      \subfloat{%
       \includegraphics[width=\thirdonecol]{%
          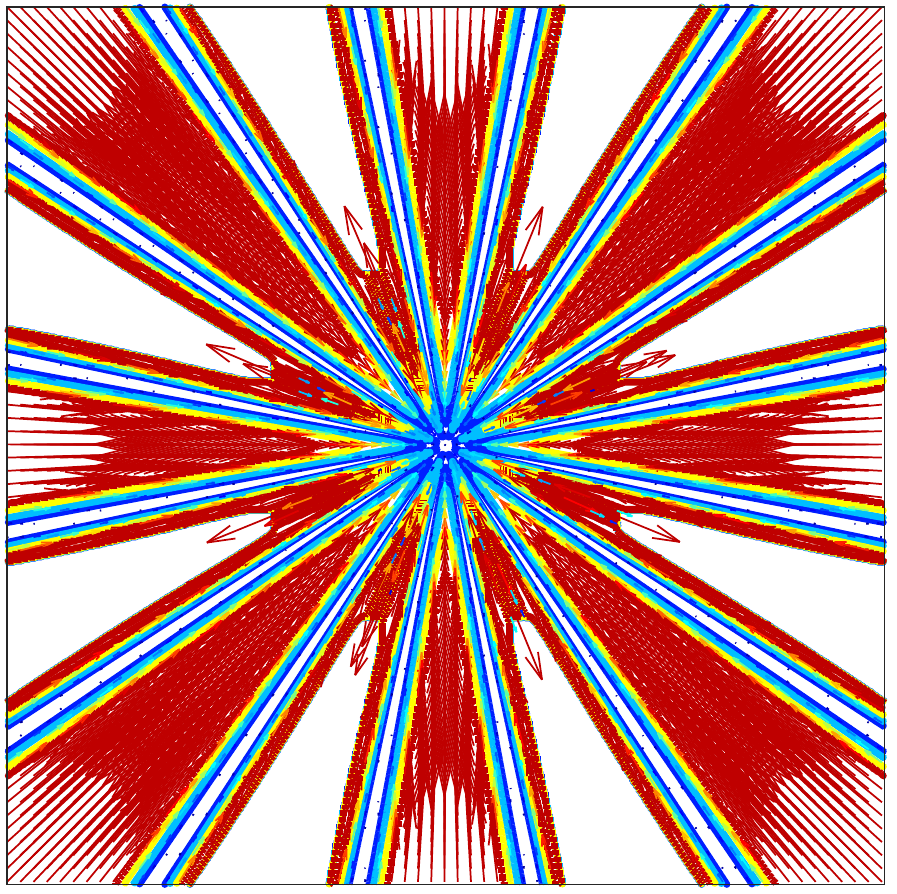}}%
      \hfill
      \subfloat{%
       \includegraphics[width=\thirdonecol]{%
          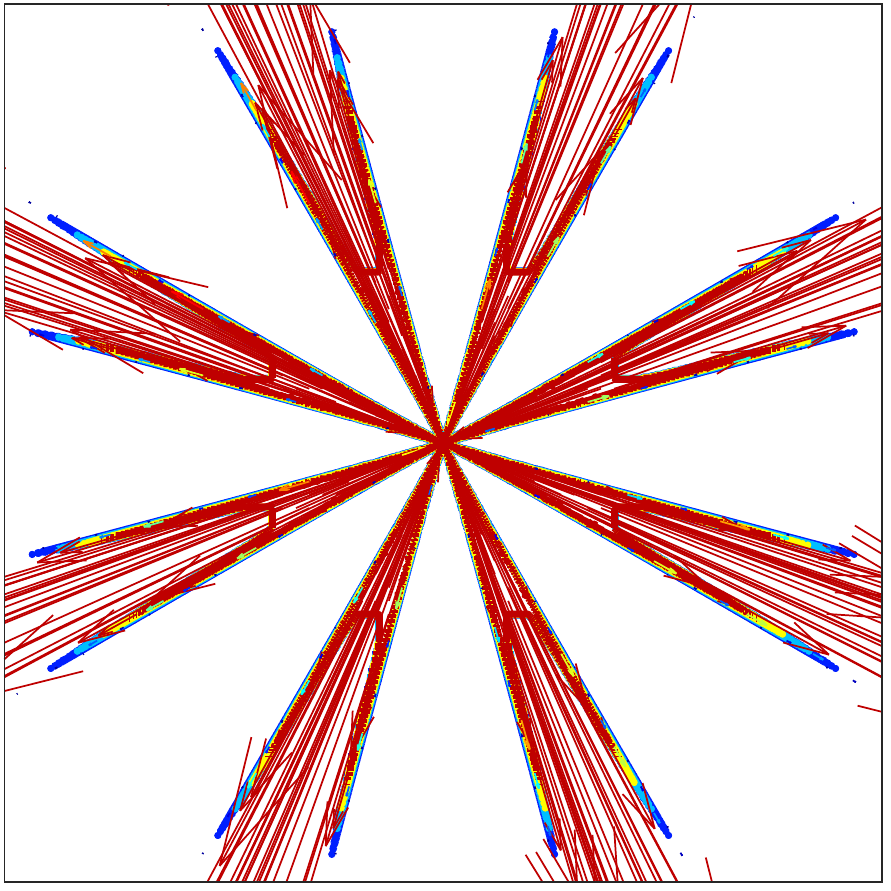}}%
      \hfill
      \subfloat{%
       \includegraphics[width=\thirdonecol]{%
          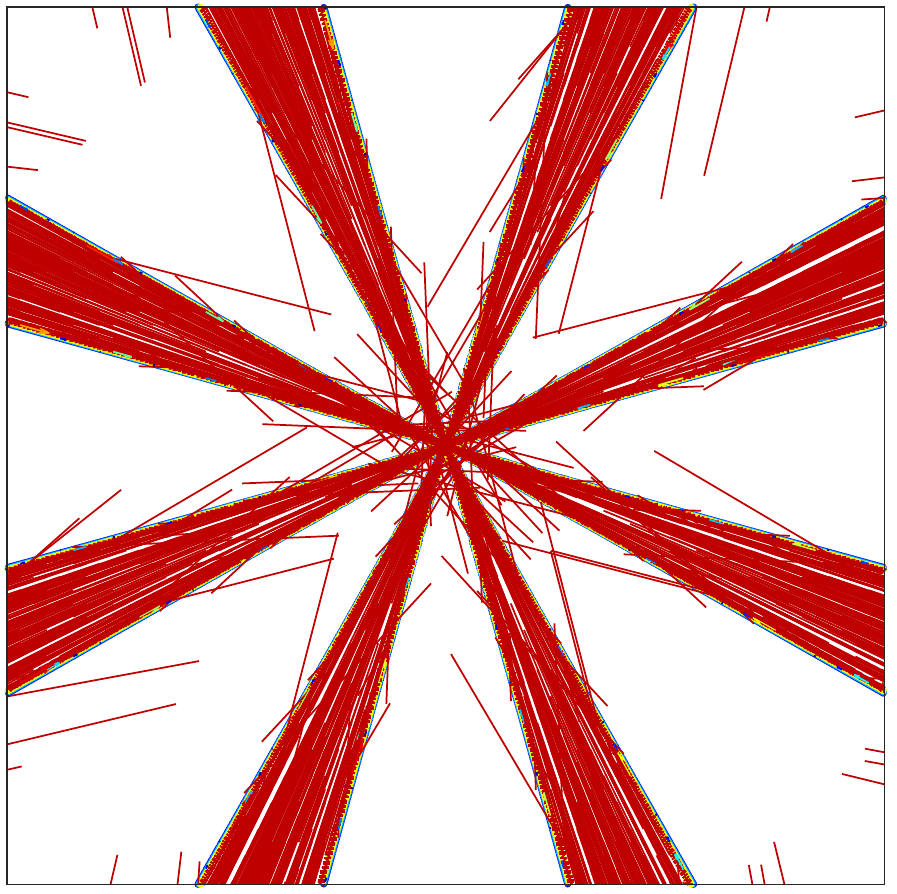}}%
  \end{minipage}%
\\
\vspace{-0.5em}%
    % 
    % ==================== RHO MAPS (MU = 0.5)
    %
    % ----- images
    % 
      % 
      \begin{minipage}{\ictextmarginleft}
        \centering
        \begin{turn}{90}
            {\normalsize $\mu=0.5$}
        \end{turn}
      \end{minipage}%
      \begin{minipage}{\icwidthImgleft} 
      \subfloat{%
        \includegraphics[width=\thirdonecol]{%
        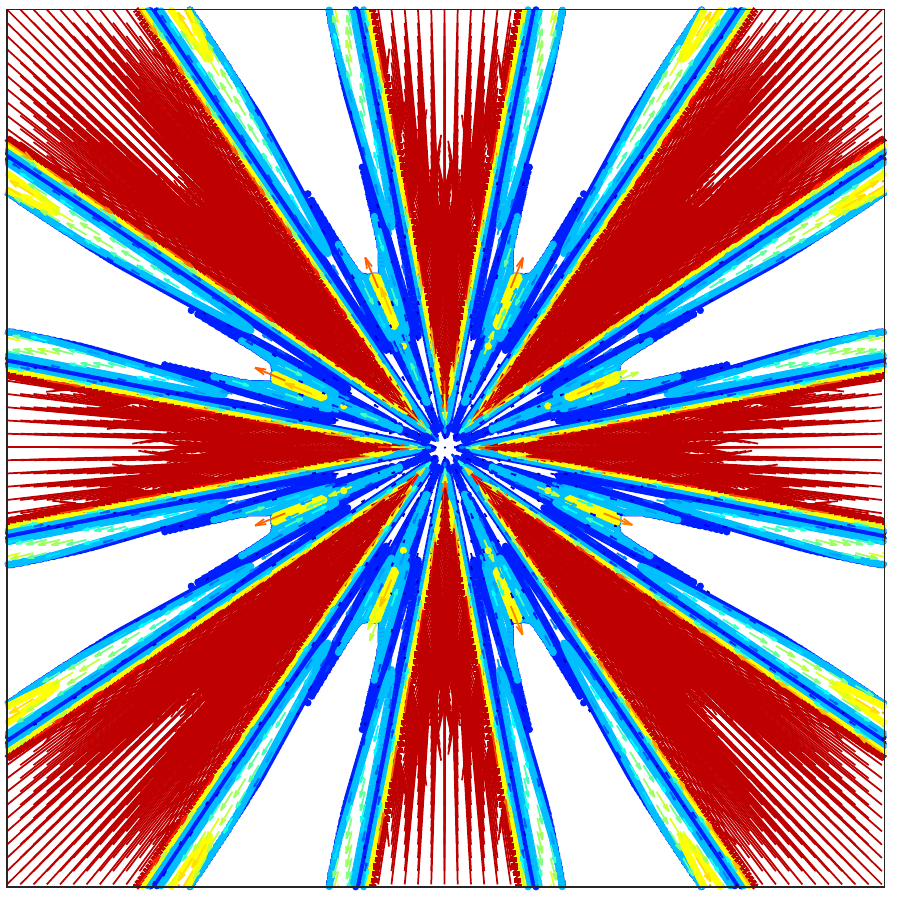}}%
      \hfill
      \subfloat{%
        \includegraphics[width=\thirdonecol]{%
        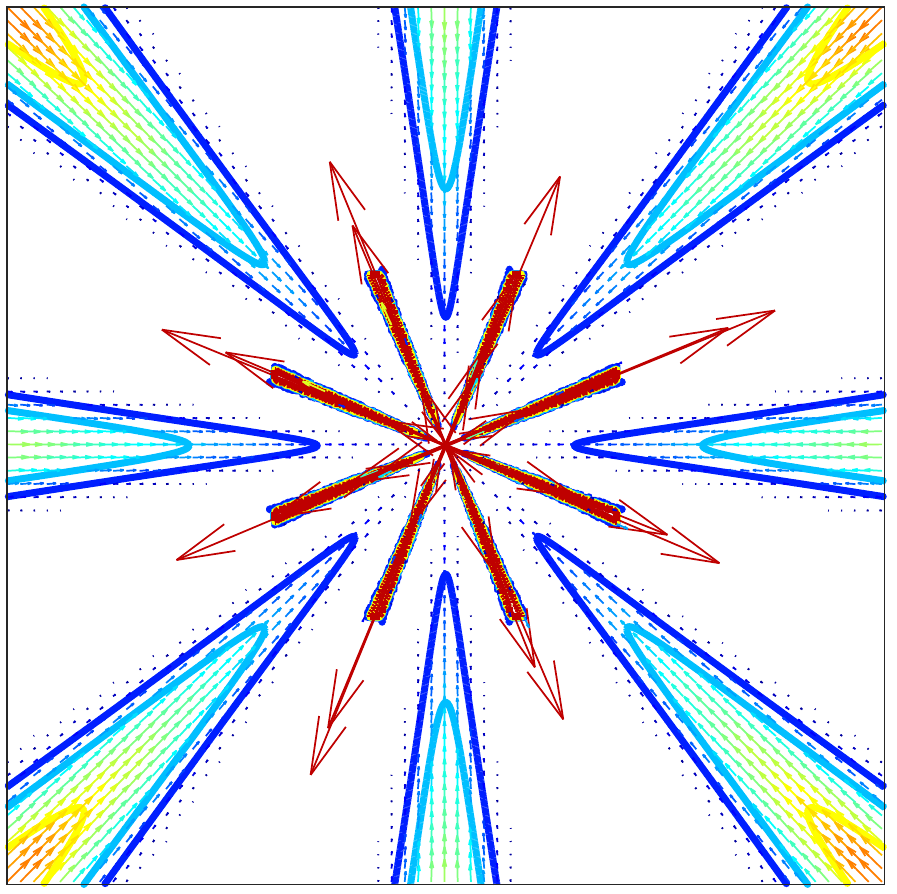}}%
      \hfill
      \subfloat{%
        \includegraphics[width=\thirdonecol]{%
        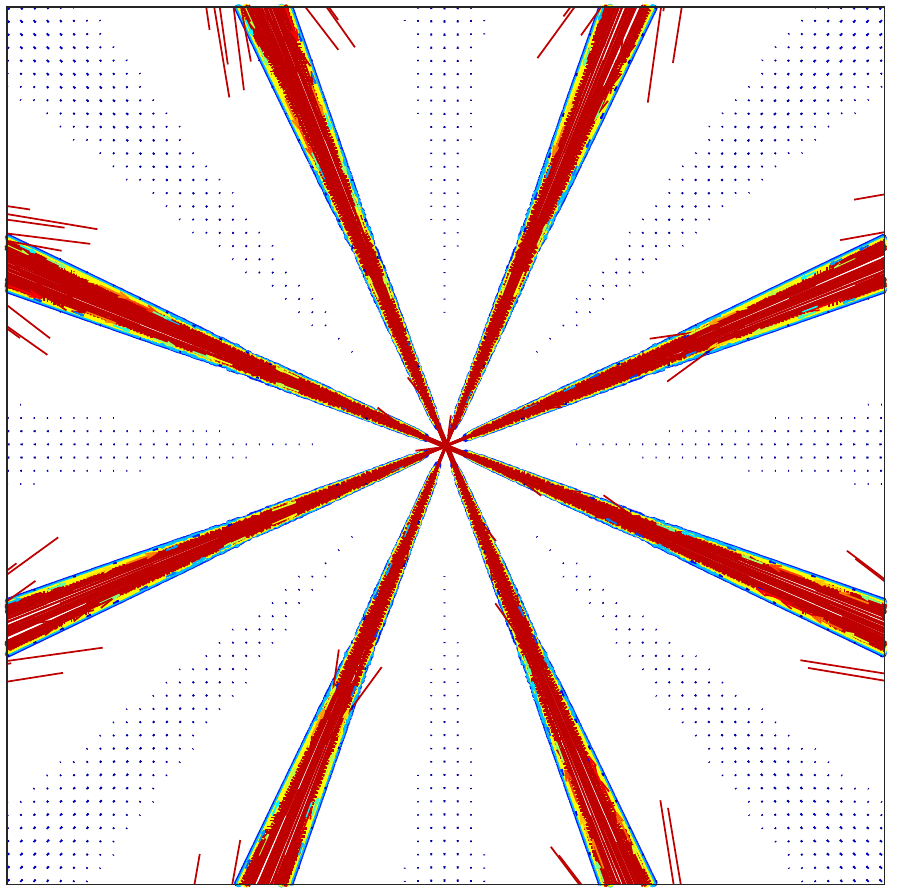}}%
    \end{minipage}
    \\
\vspace{-0.5em}%
    % 
    % ==================== error MAPS (MU alternating)
    %
      \begin{minipage}{\ictextmarginleft}
        \centering
        \begin{turn}{90}
            {\normalsize $\mu_{\text{oe}}$}
        \end{turn}
      \end{minipage}%
      \begin{minipage}{\icwidthImgleft}
      \subfloat{%
        \includegraphics[width=\thirdonecol]{%
        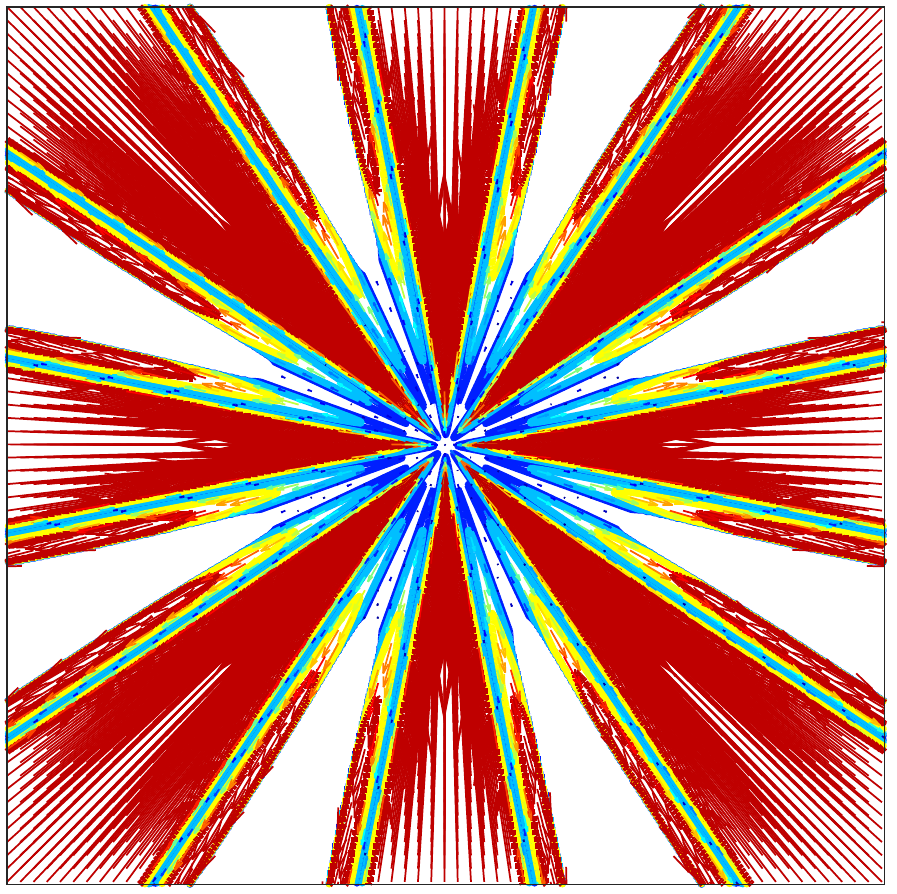}}%
      \hfill
       \subfloat{%
        \includegraphics[width=\thirdonecol]{%
        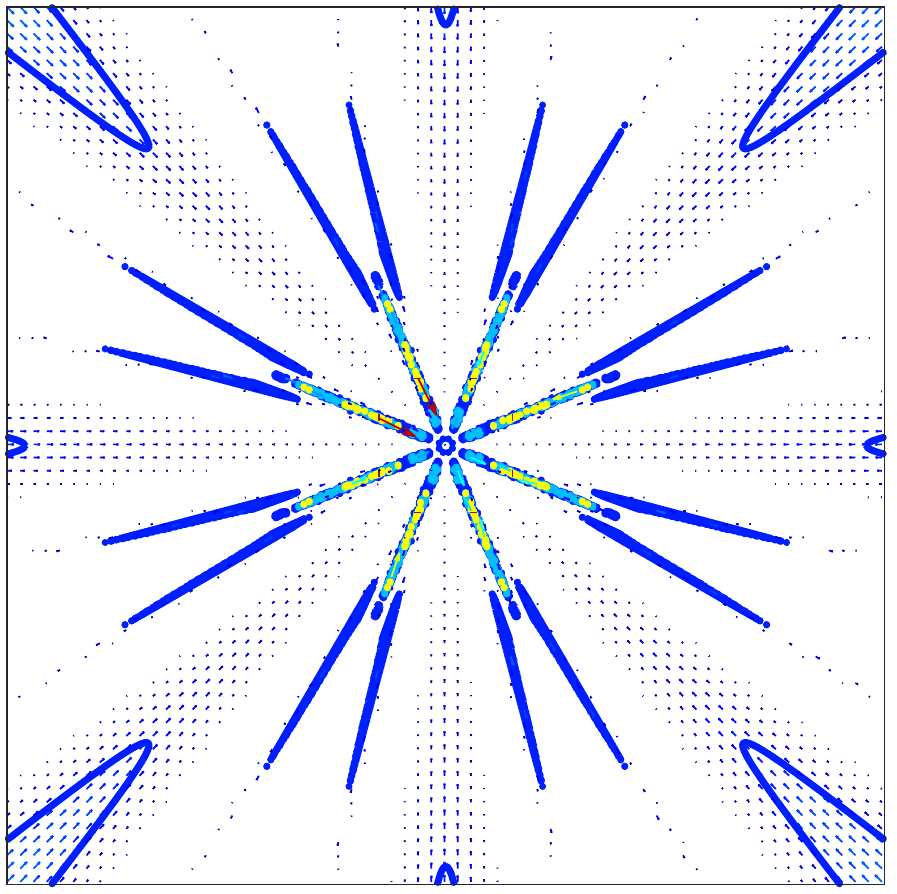}}%
      \hfill
      \subfloat{%
        \includegraphics[width=\thirdonecol]{%
        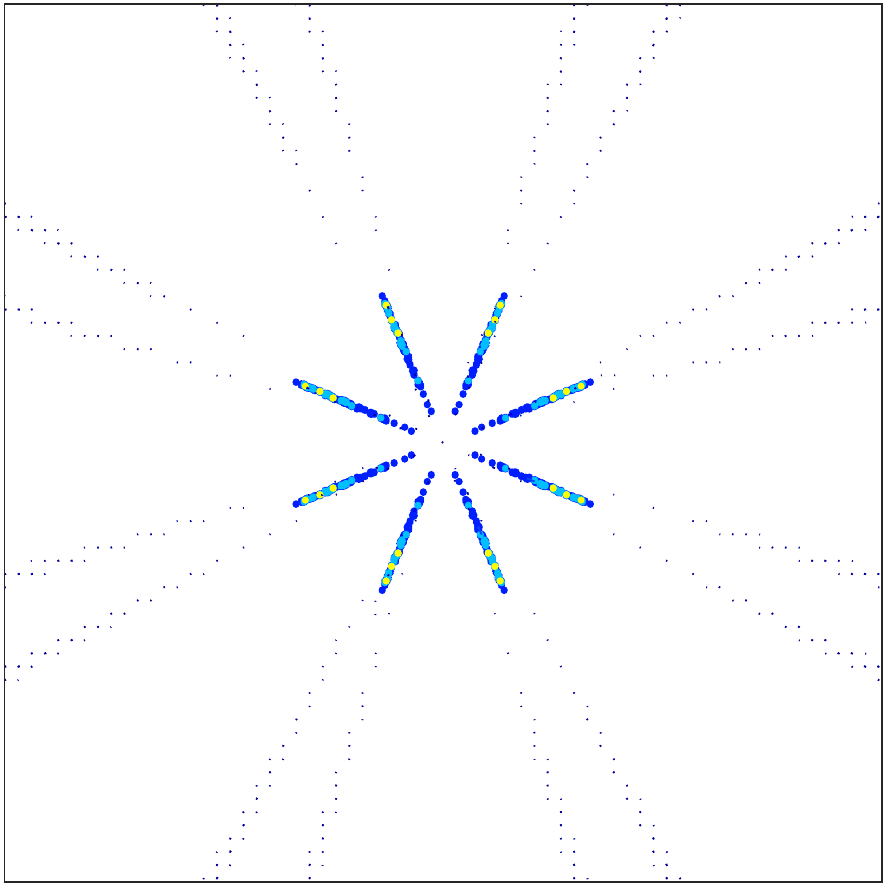}}%
      \end{minipage}%
    \\
\vspace{-0.5em}%
    % 
    % 
    % ==================== error MAPS (MU map)
    %
    % ----- images
    % 
    \begin{minipage}{\ictextmarginleft}
       \centering
        \begin{turn}{90}
            {\normalsize $\mu_{*}(\x)$}
        \end{turn}
      \end{minipage}%
      \begin{minipage}{\icwidthImgleft}
      \subfloat{%
        \includegraphics[width=\thirdonecol]{%
        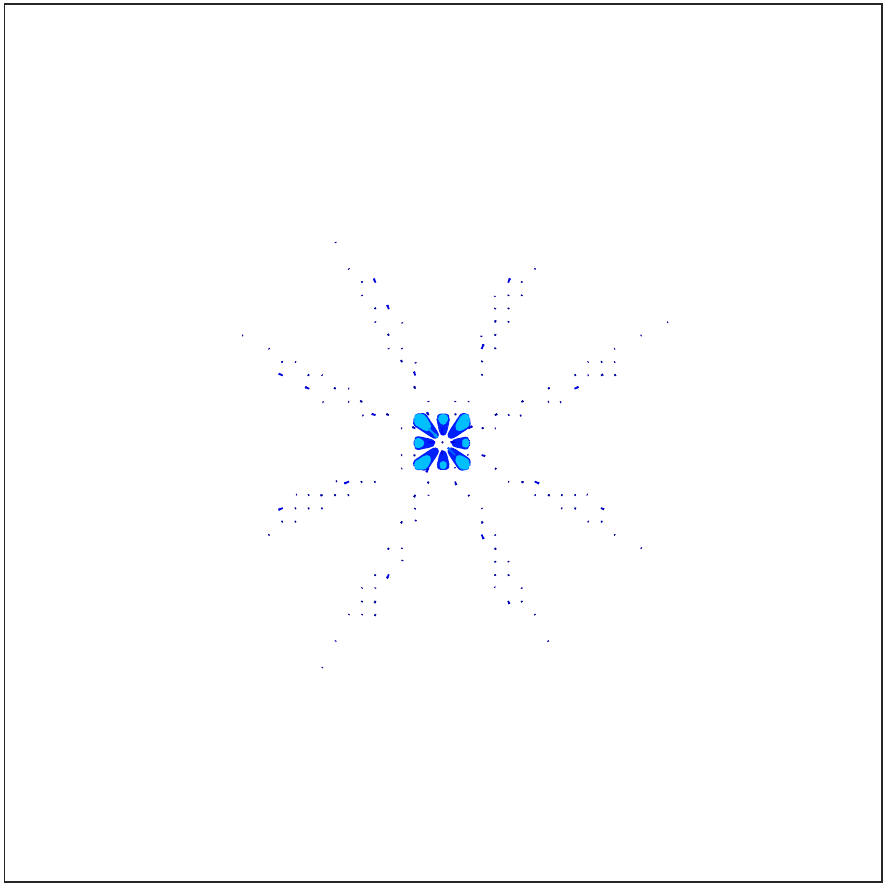}}%
   %   % 
      \hfill
      \subfloat{%
        \includegraphics[width=\thirdonecol]{%
        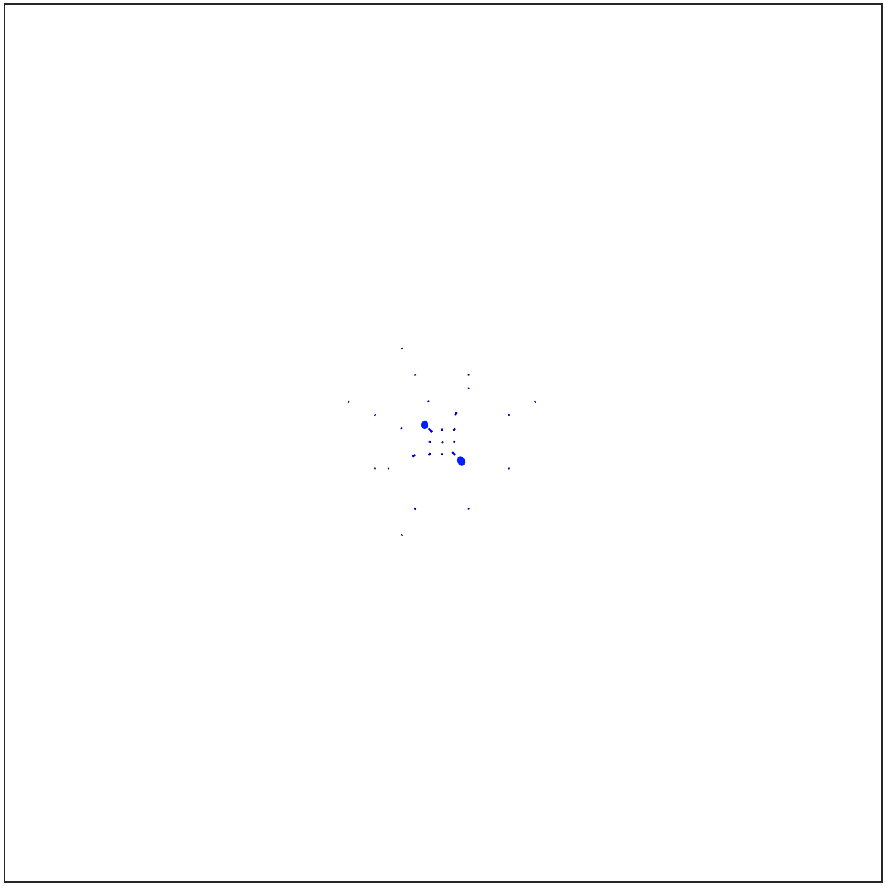}}%
      \hfill
       \subfloat{%
        \includegraphics[width=\thirdonecol]{%
        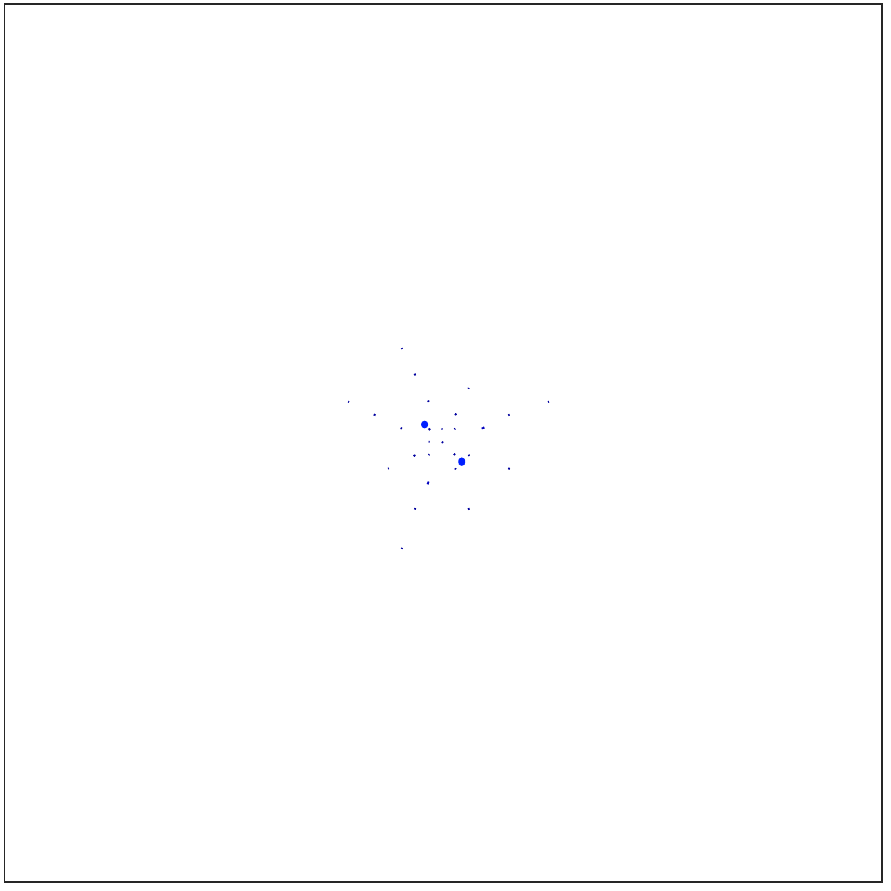}}%
   \end{minipage}
   \\
\vspace{-0.5em}%
   % ===================== subcaption
   \begin{minipage}{\icwidthImgright}
    \centering
    \vspace{1em}
    \text{(a) initial guess $\G^{[\text{a}]}_{0}(\x)$}
    \end{minipage}
   \end{minipage} % close ichalfcolleft
   \hfill
   %
   % ========================= error different initialization
   %
   %
   %
   \begin{minipage}[t]{\ichalftwocolright}%
    \hfill
    \begin{minipage}{\icwidthImgright}
      \begin{minipage}{\thirdonecol}
        \centering
        1 step
      \end{minipage}%
      \hfill
      \begin{minipage}{\thirdonecol}
        \centering
        8 steps
      \end{minipage}%
      \hfill
      \begin{minipage}{\thirdonecol}
        \centering
        15 steps
      \end{minipage}%
    \end{minipage}
    \\
    % 
    % ==================== RHO MAPS (MU = 0)
    %
    % ----- images
    % 
      %
    \begin{minipage}{\icwidthImgright}
    \vspace*{-\valignImgSkip}
    \centering
      \subfloat{%
       \includegraphics[width=\thirdonecol]{%
          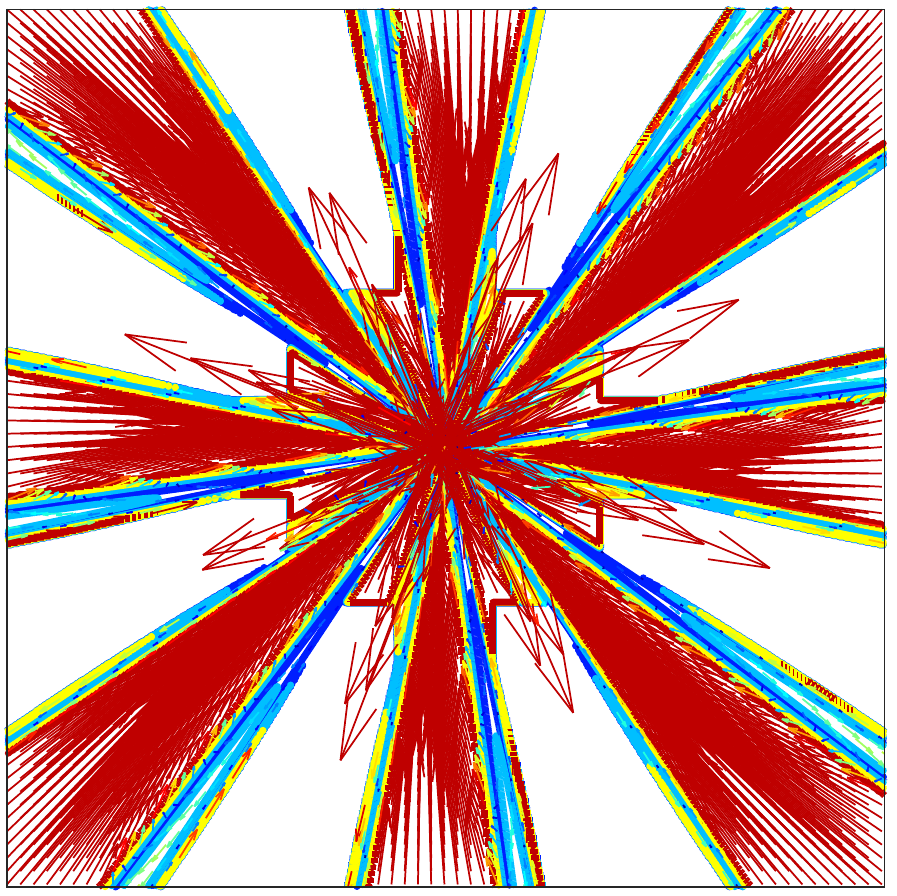}}%
      \hfill
      \subfloat{%
       \includegraphics[width=\thirdonecol]{%
          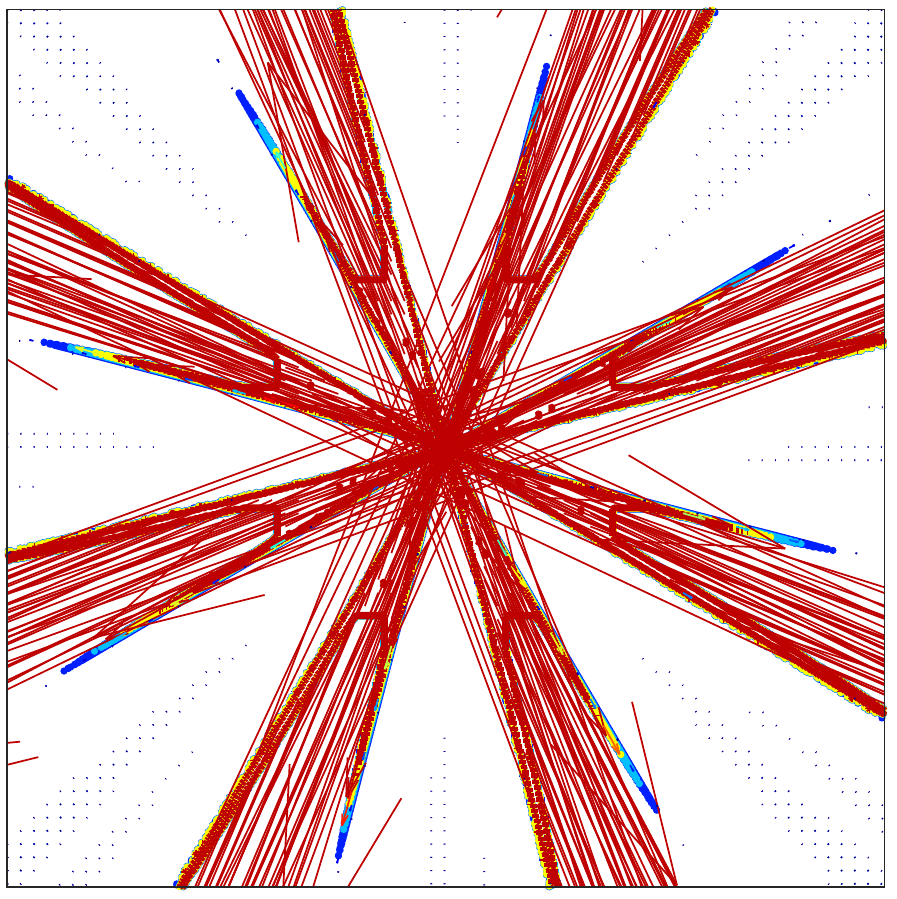}}%
      \hfill
      \subfloat{%
       \includegraphics[width=\thirdonecol]{%
          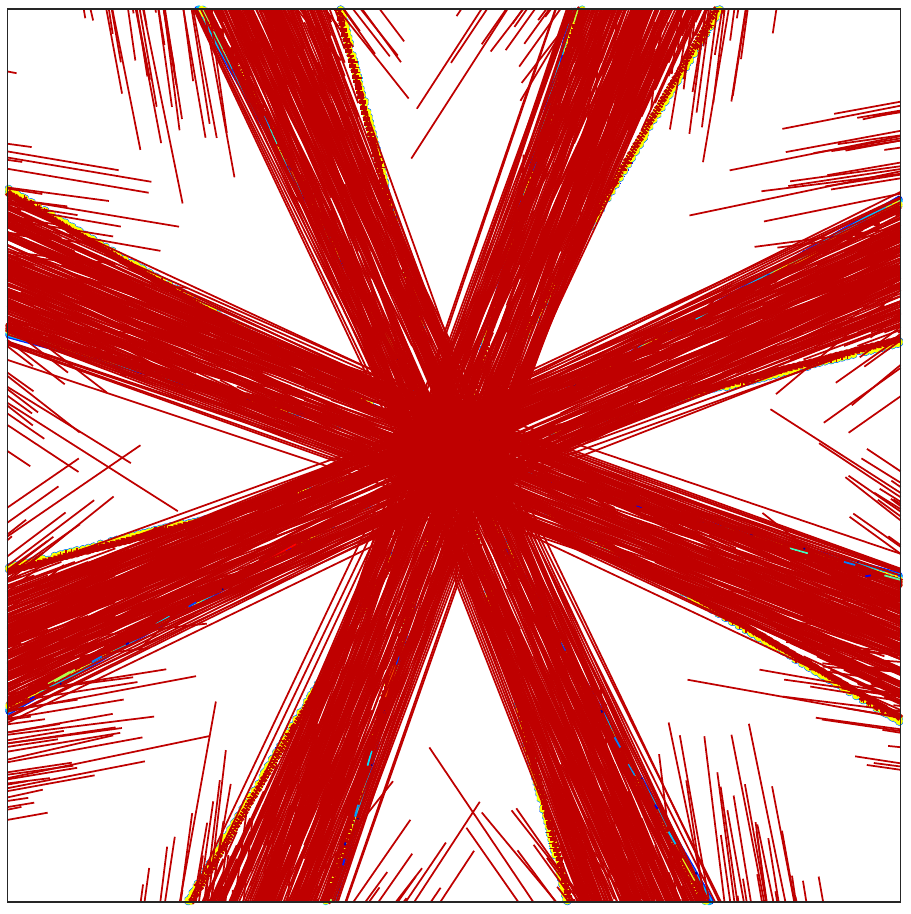}}%
  \end{minipage}%
   \\
\vspace{-0.5em}%
    % 
    % ==================== RHO MAPS (MU = 0.5)
    %
    % ----- images
    % 
      % 
      \begin{minipage}{\icwidthImgright} 
      \subfloat{%
        \includegraphics[width=\thirdonecol]{%
        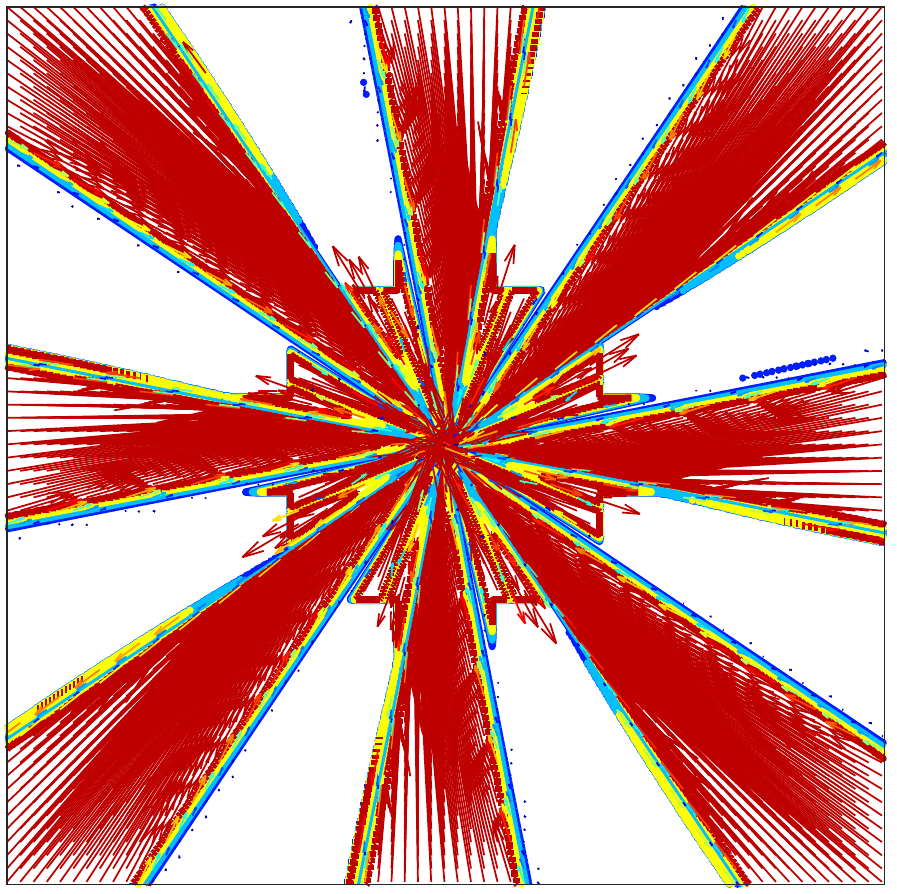}}%
      \hfill
      \subfloat{%
        \includegraphics[width=\thirdonecol]{%
        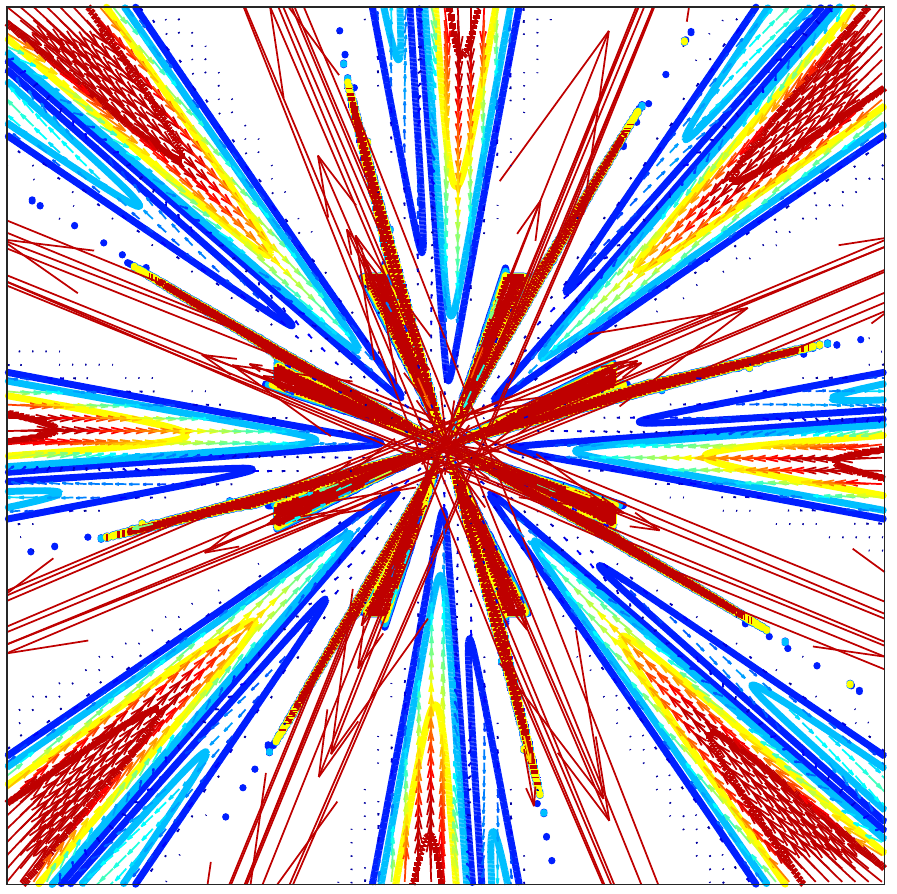}}%
      \hfill
      \subfloat{%
        \includegraphics[width=\thirdonecol]{%
        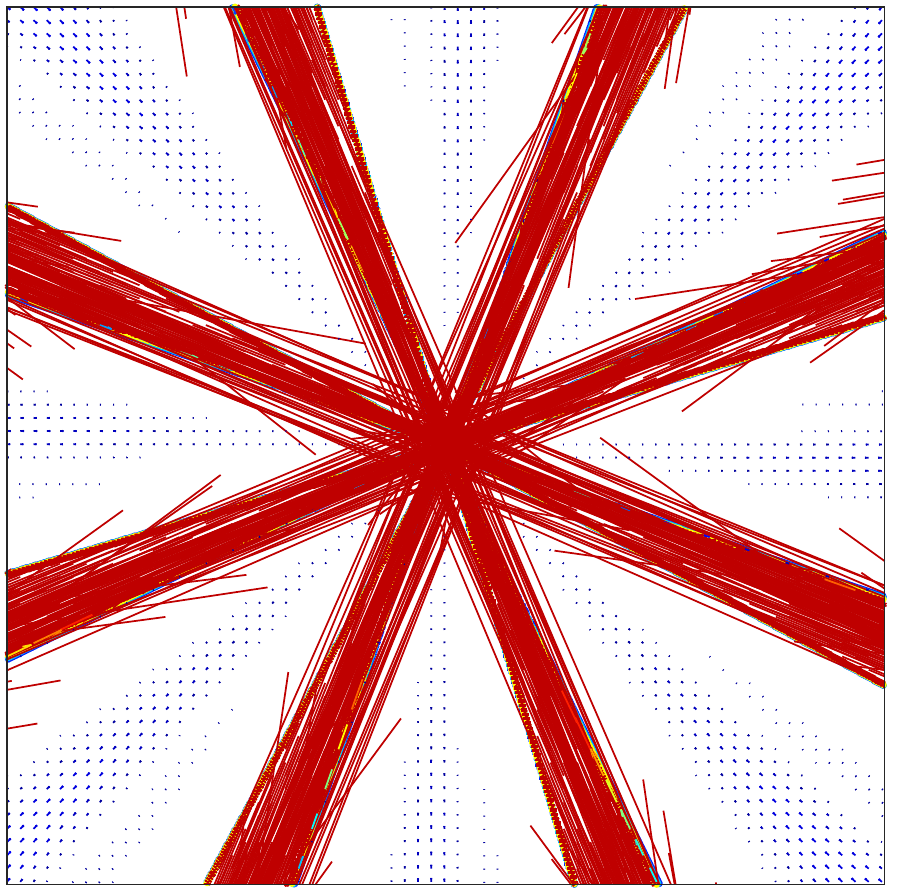}}%
    \end{minipage}
    \\
\vspace{-0.5em}%
    % 
    % ==================== error MAPS (MU alternating)
    %
      \begin{minipage}{\icwidthImgright}
      \subfloat{%
        \includegraphics[width=\thirdonecol]{%
        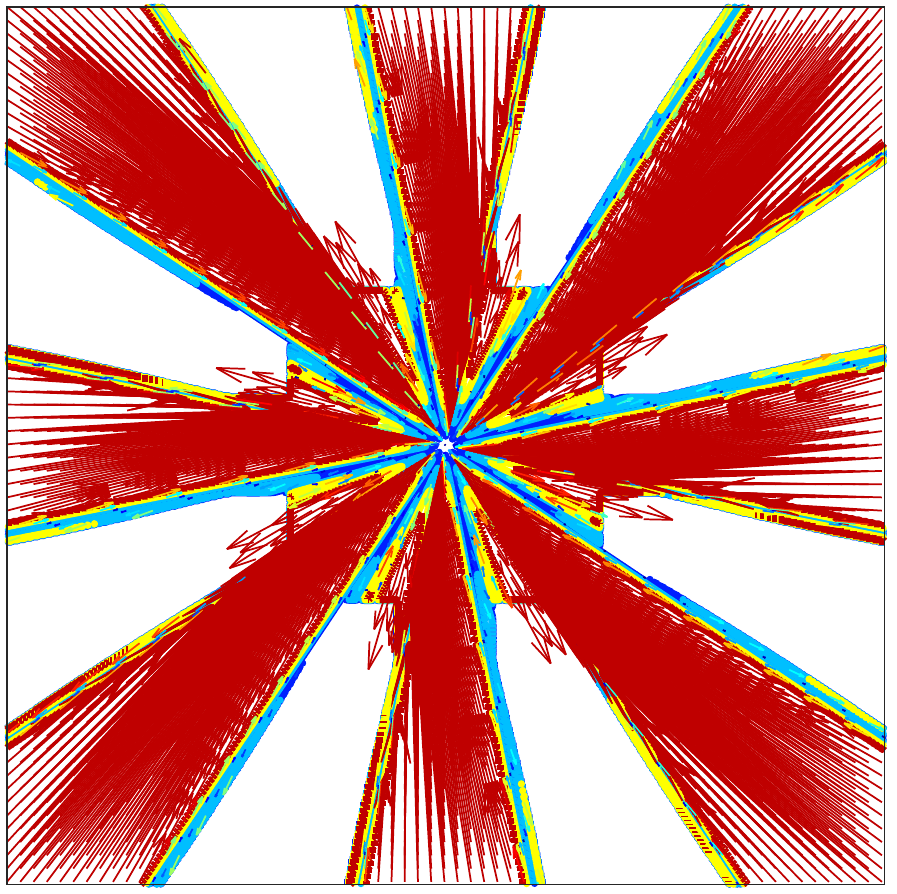}}%
      \hfill
       \subfloat{%
        \includegraphics[width=\thirdonecol]{%
        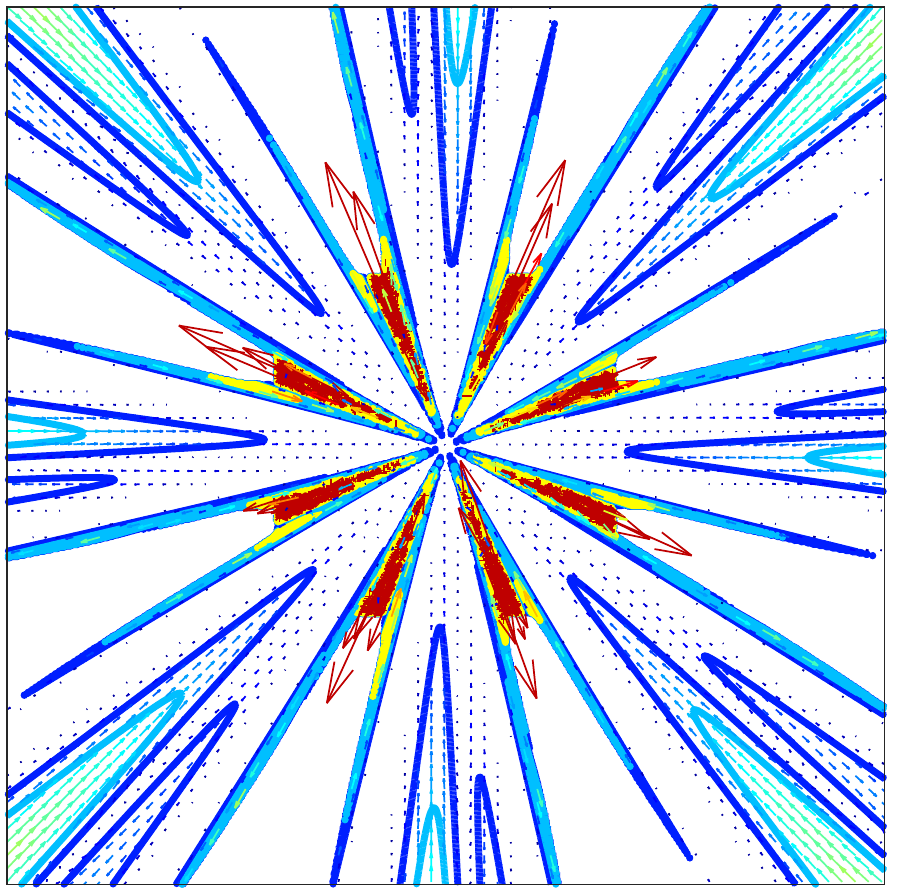}}%
      \hfill
      \subfloat{%
        \includegraphics[width=\thirdonecol]{%
        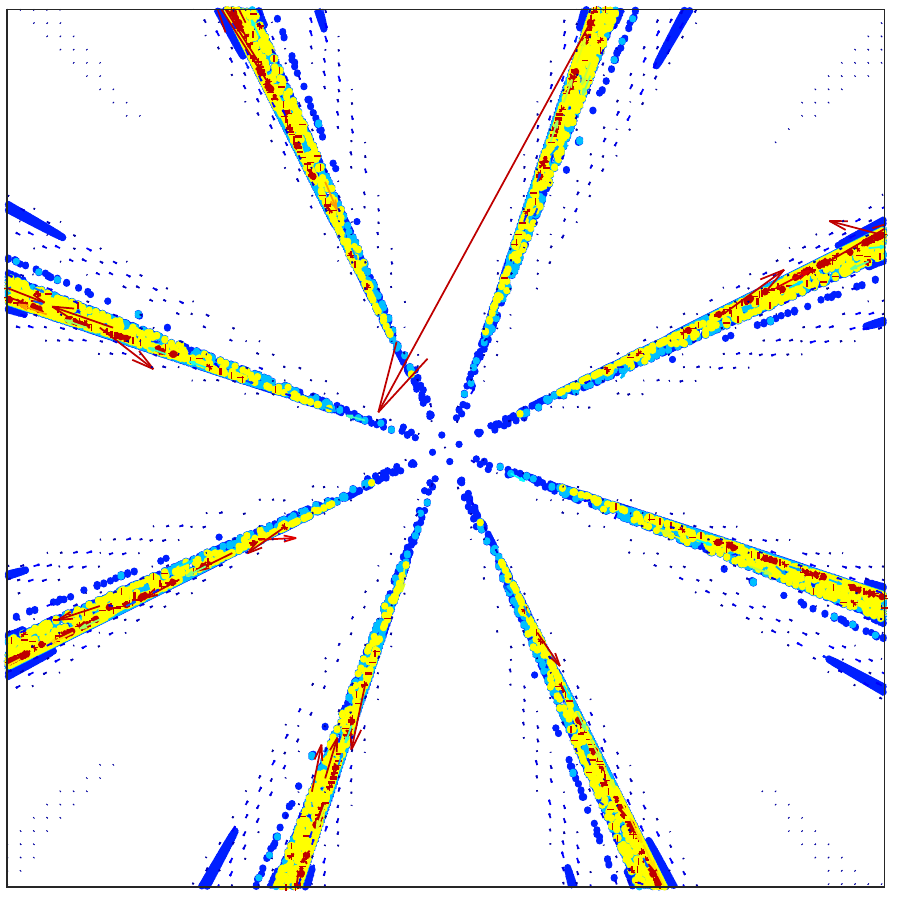}}%
      \end{minipage}%
    \\
\vspace{-0.5em}%
    % 
    % 
    % ==================== error MAPS (MU map)
    %
    % ----- images
    % 
      \begin{minipage}{\icwidthImgright}
      \subfloat{%
        \includegraphics[width=\thirdonecol]{%
        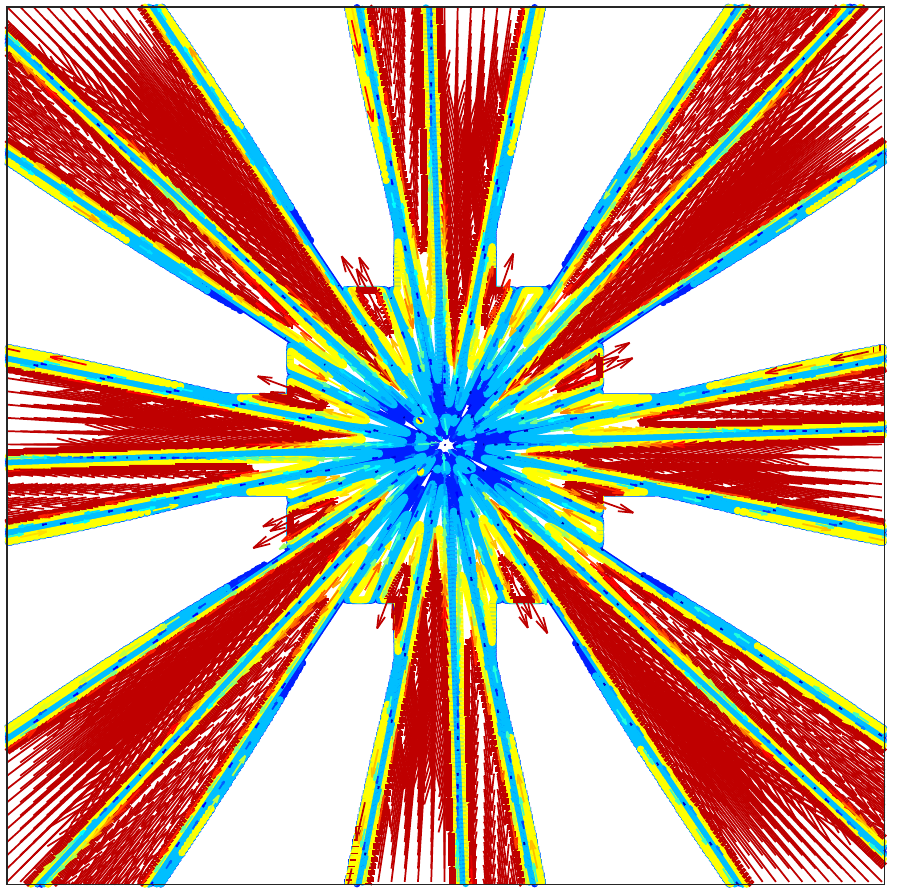}}%
   %   % 
      \hfill
      \subfloat{%
        \includegraphics[width=\thirdonecol]{%
        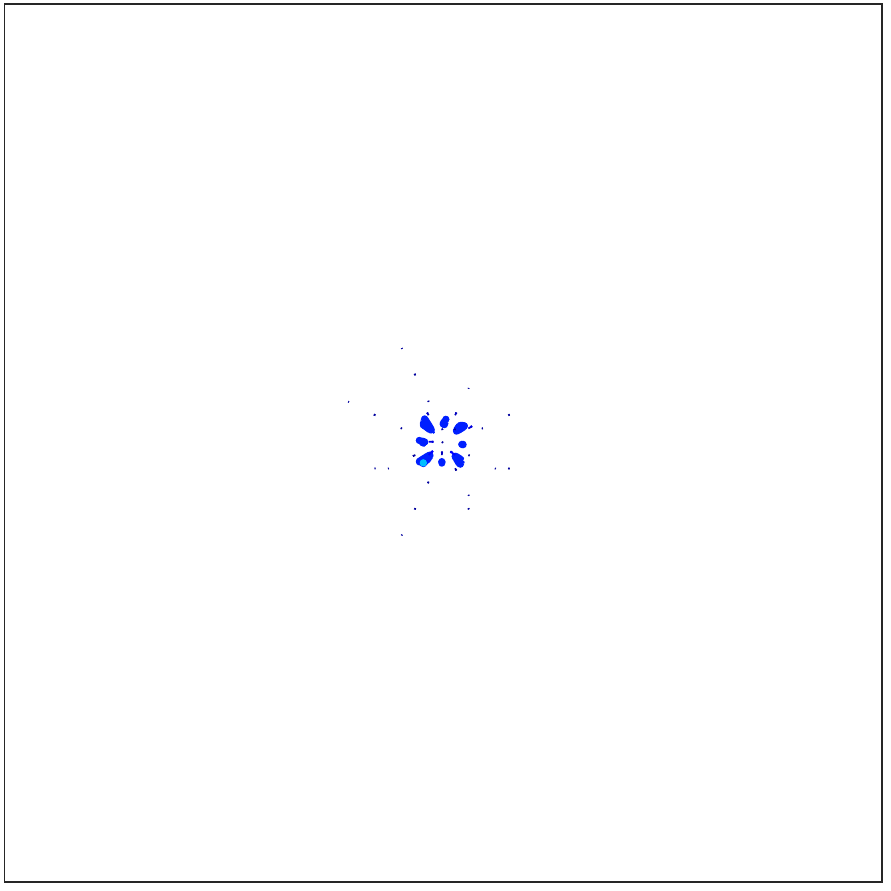}}%
      \hfill
       \subfloat{%
        \includegraphics[width=\thirdonecol]{%
        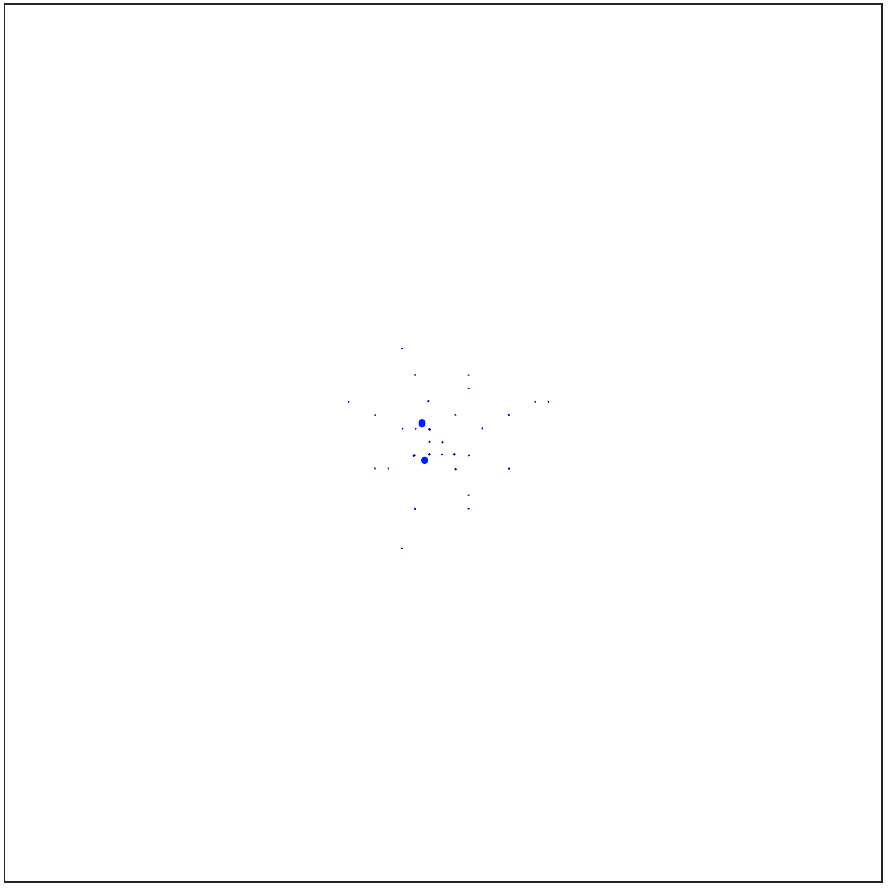}}%
   \end{minipage}
   \\
\vspace{-0.5em}%
   %
   %
   % ===================== subcaption
   \begin{minipage}{\icwidthImgright}
    \centering
    \vspace{1em}
    \text{(b) initial guess $\G^{[\text{b}]}_{0}(\x)$}
    \end{minipage}
   \end{minipage} % close ichalfcolright
   \end{minipage}
   %
   %
  % ==================== CAPTION
  % 
   \caption{%
     Inversion error map snapshots at three steps ($k=1,8,15$) during
     iterative inversion of the discretized DVF, with initial guess
     $\G_{0}^{\text{[a]}}$ (left) and  $\G_{0}^{\text{[b]}}$ (right).
     Each row contains error maps with one of the following four control
     schemes (from top to bottom): constant $\mu = 0$, constant
     $\mu = 0.5$, alternating $\mu_{\text{oe}}$ with $\mu_{\text{o}} = 0$
     and $\mu_{\text{e}} = 0.77$, and spatially variant $\mu_{\ast}(\x)$.
     Errors within a pixel are shown in white, and errors beyond $30$
     pixels are shown in red.
     The first two schemes fail in error suppression around the 8 radial
     ridge lines; the next two successfully suppress the errors over the
     entire domain.  The scheme with $\mu_{\ast}(\x)$ is robust to the
     change in the initial guess.
   }
  \label{fig:inversion-error-analytical-dvf}
\end{figure*}

Inversion errors are calculated pixel-wise by
$\E_k(\x) = \G_k(\x) - \G(\x)$, with the estimate at the $k$-th iteration
step and $\x \in \Omega$; see \cref{eq:iteratate-error}.
They are presented in two complementary views. 
In the summary view (\cref{fig:chendvf-error-vs-iter}), we provide the
sequence of inversion error percentiles \cref{eq:prctile} over each
iteration process up to the 15th step.
We show in \cref{fig:inversion-error-analytical-dvf} the spatial
distribution of errors via snapshots of inversion error maps at three
iteration steps, $k = 1, 8, 15$.
In order to take into consideration also the response of each algorithm
to the initial guess, we present the error-map snapshots with two
different initial guesses: $\G_{0}^{\text{[a]}}(\x) = \mathbf{0}$ and
$\G_{0}^{\text{[b]}}(\x) = 0.08 \cos(8 \theta) \tpose{[-x_{2} \enskip
  x_{1}]}$.

The two non-adaptive iterations fail in reducing the errors over the 8
radial ridge regions with $b = 0.8$, as shown
in~\cref{fig:inversion-error-analytical-dvf}.  The divergence regions
correspond to, and can be predicted by, those with non-small NTDCs and high
control index values in \cref{fig:chendvf-datacharacterization}.
The two adaptive iterations successfully suppress and annihilate the
inverse errors.
With the spatially variant adaptive control, inversion errors are
reduced to sub-pixel length in no more than 3 steps.  If the analytical
values of $\mu_{\ast}(\x)$ are used, the algorithm renders the inverse
in a single step.

%%%%%%%%%%%%%%%%%%%%%%%%%%%%%%%%%%%%%%%%%%%%%%%%%%
%%% DOCUMENT END

\end{document}